\documentclass{article} % For LaTeX2e

\newcommand{\iclr}{}

\usepackage{hyperref}
\usepackage{amsmath, amssymb}
\usepackage{cleveref}
\usepackage{url}
\usepackage{iclr2026_conference,times}
\usepackage{lineno}

\usepackage{booktabs} % For formal tables
\usepackage[utf8]{inputenc}
\usepackage{float}
\usepackage{afterpage}
\usepackage{color}
\usepackage[sectionbib]{bibunits}
\usepackage{multicol}
\usepackage{makecell}
\usepackage{multirow}
\usepackage{array}
\usepackage{stfloats}
\usepackage{graphicx}
\usepackage{tikz}
\usepackage{enumitem}
\usepackage{fontawesome5}
\usepackage{manfnt}
\usepackage{wrapfig}
\usepackage[export]{adjustbox}
\usepackage[font=small,skip=3pt]{caption}
\usepackage[noend]{algpseudocode}

\usepackage{layouts}
\usepackage[T1]{fontenc}
\usepackage{graphicx}
\usepackage{booktabs}
\usepackage{subcaption}
\usepackage{adjustbox}
\usepackage{booktabs}
\usepackage{multirow}
\usepackage{wrapfig}
\usepackage{layouts}
\usepackage{float}
\usepackage{array}
\usepackage{tabularx}
\usepackage[table]{xcolor}
\usepackage{tikz}
\usepackage{tcolorbox}
\usepackage{color}
\usepackage{amsthm}
\usepackage{tikz}
\usepackage{colortbl}
\usepackage{soul}
\usepackage{multicol}
\usepackage{multirow}
\usepackage{titletoc}
\usepackage{csquotes}
\usepackage{tikz}
\usepackage{makecell}
\usepackage{pifont}
\usepackage{svg}
\usepackage[flushleft]{threeparttable}
\usepackage{algorithm}
\usepackage{algpseudocode}

\catcode`\%=12
\newcommand\pcnt{%}
\catcode`\%=14

\definecolor{nvidiagreen}{HTML}{76B900}
\definecolor{nvidiagreen_light}{HTML}{E1FFAD}
\definecolor{my_green}{RGB}{51,102,0}
\definecolor{my_red}{RGB}{204, 0, 0}
\newcommand{\cmark}{\textcolor{my_green}{\ding{51}}} % ✔
\newcommand{\xmark}{\textcolor{my_red}{\ding{55}}} % ✘

\newcommand{\acronym}{VoMP} %{FRANKENSTEIN}
\usepackage{amsmath,amsfonts,bm}
\usepackage{mathtools}

\newcommand{\figleft}{{\em (Left)}}
\newcommand{\figcenter}{{\em (Center)}}
\newcommand{\figright}{{\em (Right)}}
\newcommand{\figtop}{{\em (Top)}}
\newcommand{\figbottom}{{\em (Bottom)}}
\newcommand{\captiona}{{\em (a)}}
\newcommand{\captionb}{{\em (b)}}
\newcommand{\captionc}{{\em (c)}}
\newcommand{\captiond}{{\em (d)}}

\newcommand{\newterm}[1]{{\bf #1}}

\def\figref#1{figure~\ref{#1}}
\def\Figref#1{Figure~\ref{#1}}
\def\twofigref#1#2{figures \ref{#1} and \ref{#2}}
\def\quadfigref#1#2#3#4{figures \ref{#1}, \ref{#2}, \ref{#3} and \ref{#4}}
\def\secref#1{section~\ref{#1}}
\def\Secref#1{Section~\ref{#1}}
\def\twosecrefs#1#2{sections \ref{#1} and \ref{#2}}
\def\secrefs#1#2#3{sections \ref{#1}, \ref{#2} and \ref{#3}}
\def\eqref#1{equation~\ref{#1}}
\def\Eqref#1{Equation~\ref{#1}}
\def\plaineqref#1{\ref{#1}}
\def\chapref#1{chapter~\ref{#1}}
\def\Chapref#1{Chapter~\ref{#1}}
\def\rangechapref#1#2{chapters\ref{#1}--\ref{#2}}
\def\algref#1{algorithm~\ref{#1}}
\def\Algref#1{Algorithm~\ref{#1}}
\def\twoalgref#1#2{algorithms \ref{#1} and \ref{#2}}
\def\Twoalgref#1#2{Algorithms \ref{#1} and \ref{#2}}
\def\partref#1{part~\ref{#1}}
\def\Partref#1{Part~\ref{#1}}
\def\twopartref#1#2{parts \ref{#1} and \ref{#2}}

\def\ceil#1{\lceil #1 \rceil}
\def\floor#1{\lfloor #1 \rfloor}
\def\1{\bm{1}}
\newcommand{\train}{\mathcal{D}}
\newcommand{\valid}{\mathcal{D_{\mathrm{valid}}}}
\newcommand{\test}{\mathcal{D_{\mathrm{test}}}}

\def\eps{{\epsilon}}

\def\reta{{\textnormal{$\eta$}}}
\def\ra{{\textnormal{a}}}
\def\rb{{\textnormal{b}}}
\def\rc{{\textnormal{c}}}
\def\rd{{\textnormal{d}}}
\def\re{{\textnormal{e}}}
\def\rf{{\textnormal{f}}}
\def\rg{{\textnormal{g}}}
\def\rh{{\textnormal{h}}}
\def\ri{{\textnormal{i}}}
\def\rj{{\textnormal{j}}}
\def\rk{{\textnormal{k}}}
\def\rl{{\textnormal{l}}}
\def\rn{{\textnormal{n}}}
\def\ro{{\textnormal{o}}}
\def\rp{{\textnormal{p}}}
\def\rq{{\textnormal{q}}}
\def\rr{{\textnormal{r}}}
\def\rs{{\textnormal{s}}}
\def\rt{{\textnormal{t}}}
\def\ru{{\textnormal{u}}}
\def\rv{{\textnormal{v}}}
\def\rw{{\textnormal{w}}}
\def\rx{{\textnormal{x}}}
\def\ry{{\textnormal{y}}}
\def\rz{{\textnormal{z}}}

\def\rvepsilon{{\mathbf{\epsilon}}}
\def\rvtheta{{\mathbf{\theta}}}
\def\rva{{\mathbf{a}}}
\def\rvb{{\mathbf{b}}}
\def\rvc{{\mathbf{c}}}
\def\rvd{{\mathbf{d}}}
\def\rve{{\mathbf{e}}}
\def\rvf{{\mathbf{f}}}
\def\rvg{{\mathbf{g}}}
\def\rvh{{\mathbf{h}}}
\def\rvu{{\mathbf{i}}}
\def\rvj{{\mathbf{j}}}
\def\rvk{{\mathbf{k}}}
\def\rvl{{\mathbf{l}}}
\def\rvm{{\mathbf{m}}}
\def\rvn{{\mathbf{n}}}
\def\rvo{{\mathbf{o}}}
\def\rvp{{\mathbf{p}}}
\def\rvq{{\mathbf{q}}}
\def\rvr{{\mathbf{r}}}
\def\rvs{{\mathbf{s}}}
\def\rvt{{\mathbf{t}}}
\def\rvu{{\mathbf{u}}}
\def\rvv{{\mathbf{v}}}
\def\rvw{{\mathbf{w}}}
\def\rvx{{\mathbf{x}}}
\def\rvy{{\mathbf{y}}}
\def\rvz{{\mathbf{z}}}

\def\erva{{\textnormal{a}}}
\def\ervb{{\textnormal{b}}}
\def\ervc{{\textnormal{c}}}
\def\ervd{{\textnormal{d}}}
\def\erve{{\textnormal{e}}}
\def\ervf{{\textnormal{f}}}
\def\ervg{{\textnormal{g}}}
\def\ervh{{\textnormal{h}}}
\def\ervi{{\textnormal{i}}}
\def\ervj{{\textnormal{j}}}
\def\ervk{{\textnormal{k}}}
\def\ervl{{\textnormal{l}}}
\def\ervm{{\textnormal{m}}}
\def\ervn{{\textnormal{n}}}
\def\ervo{{\textnormal{o}}}
\def\ervp{{\textnormal{p}}}
\def\ervq{{\textnormal{q}}}
\def\ervr{{\textnormal{r}}}
\def\ervs{{\textnormal{s}}}
\def\ervt{{\textnormal{t}}}
\def\ervu{{\textnormal{u}}}
\def\ervv{{\textnormal{v}}}
\def\ervw{{\textnormal{w}}}
\def\ervx{{\textnormal{x}}}
\def\ervy{{\textnormal{y}}}
\def\ervz{{\textnormal{z}}}

\def\rmA{{\mathbf{A}}}
\def\rmB{{\mathbf{B}}}
\def\rmC{{\mathbf{C}}}
\def\rmD{{\mathbf{D}}}
\def\rmE{{\mathbf{E}}}
\def\rmF{{\mathbf{F}}}
\def\rmG{{\mathbf{G}}}
\def\rmH{{\mathbf{H}}}
\def\rmI{{\mathbf{I}}}
\def\rmJ{{\mathbf{J}}}
\def\rmK{{\mathbf{K}}}
\def\rmL{{\mathbf{L}}}
\def\rmM{{\mathbf{M}}}
\def\rmN{{\mathbf{N}}}
\def\rmO{{\mathbf{O}}}
\def\rmP{{\mathbf{P}}}
\def\rmQ{{\mathbf{Q}}}
\def\rmR{{\mathbf{R}}}
\def\rmS{{\mathbf{S}}}
\def\rmT{{\mathbf{T}}}
\def\rmU{{\mathbf{U}}}
\def\rmV{{\mathbf{V}}}
\def\rmW{{\mathbf{W}}}
\def\rmX{{\mathbf{X}}}
\def\rmY{{\mathbf{Y}}}
\def\rmZ{{\mathbf{Z}}}

\def\ermA{{\textnormal{A}}}
\def\ermB{{\textnormal{B}}}
\def\ermC{{\textnormal{C}}}
\def\ermD{{\textnormal{D}}}
\def\ermE{{\textnormal{E}}}
\def\ermF{{\textnormal{F}}}
\def\ermG{{\textnormal{G}}}
\def\ermH{{\textnormal{H}}}
\def\ermI{{\textnormal{I}}}
\def\ermJ{{\textnormal{J}}}
\def\ermK{{\textnormal{K}}}
\def\ermL{{\textnormal{L}}}
\def\ermM{{\textnormal{M}}}
\def\ermN{{\textnormal{N}}}
\def\ermO{{\textnormal{O}}}
\def\ermP{{\textnormal{P}}}
\def\ermQ{{\textnormal{Q}}}
\def\ermR{{\textnormal{R}}}
\def\ermS{{\textnormal{S}}}
\def\ermT{{\textnormal{T}}}
\def\ermU{{\textnormal{U}}}
\def\ermV{{\textnormal{V}}}
\def\ermW{{\textnormal{W}}}
\def\ermX{{\textnormal{X}}}
\def\ermY{{\textnormal{Y}}}
\def\ermZ{{\textnormal{Z}}}

\def\vzero{{\bm{0}}}
\def\vone{{\bm{1}}}
\def\vmu{{\bm{\mu}}}
\def\vtheta{{\bm{\theta}}}
\def\va{{\bm{a}}}
\def\vb{{\bm{b}}}
\def\vc{{\bm{c}}}
\def\vd{{\bm{d}}}
\def\ve{{\bm{e}}}
\def\vf{{\bm{f}}}
\def\vg{{\bm{g}}}
\def\vh{{\bm{h}}}
\def\vi{{\bm{i}}}
\def\vj{{\bm{j}}}
\def\vk{{\bm{k}}}
\def\vl{{\bm{l}}}
\def\vm{{\bm{m}}}
\def\vn{{\bm{n}}}
\def\vo{{\bm{o}}}
\def\vp{{\bm{p}}}
\def\vq{{\bm{q}}}
\def\vr{{\bm{r}}}
\def\vs{{\bm{s}}}
\def\vt{{\bm{t}}}
\def\vu{{\bm{u}}}
\def\vv{{\bm{v}}}
\def\vw{{\bm{w}}}
\def\vx{{\bm{x}}}
\def\vy{{\bm{y}}}
\def\vz{{\bm{z}}}

\def\evalpha{{\alpha}}
\def\evbeta{{\beta}}
\def\evepsilon{{\epsilon}}
\def\evlambda{{\lambda}}
\def\evomega{{\omega}}
\def\evmu{{\mu}}
\def\evpsi{{\psi}}
\def\evsigma{{\sigma}}
\def\evtheta{{\theta}}
\def\eva{{a}}
\def\evb{{b}}
\def\evc{{c}}
\def\evd{{d}}
\def\eve{{e}}
\def\evf{{f}}
\def\evg{{g}}
\def\evh{{h}}
\def\evi{{i}}
\def\evj{{j}}
\def\evk{{k}}
\def\evl{{l}}
\def\evm{{m}}
\def\evn{{n}}
\def\evo{{o}}
\def\evp{{p}}
\def\evq{{q}}
\def\evr{{r}}
\def\evs{{s}}
\def\evt{{t}}
\def\evu{{u}}
\def\evv{{v}}
\def\evw{{w}}
\def\evx{{x}}
\def\evy{{y}}
\def\evz{{z}}

\def\mA{{\bm{A}}}
\def\mB{{\bm{B}}}
\def\mC{{\bm{C}}}
\def\mD{{\bm{D}}}
\def\mE{{\bm{E}}}
\def\mF{{\bm{F}}}
\def\mG{{\bm{G}}}
\def\mH{{\bm{H}}}
\def\mI{{\bm{I}}}
\def\mJ{{\bm{J}}}
\def\mK{{\bm{K}}}
\def\mL{{\bm{L}}}
\def\mM{{\bm{M}}}
\def\mN{{\bm{N}}}
\def\mO{{\bm{O}}}
\def\mP{{\bm{P}}}
\def\mQ{{\bm{Q}}}
\def\mR{{\bm{R}}}
\def\mS{{\bm{S}}}
\def\mT{{\bm{T}}}
\def\mU{{\bm{U}}}
\def\mV{{\bm{V}}}
\def\mW{{\bm{W}}}
\def\mX{{\bm{X}}}
\def\mY{{\bm{Y}}}
\def\mZ{{\bm{Z}}}
\def\mBeta{{\bm{\beta}}}
\def\mPhi{{\bm{\Phi}}}
\def\mLambda{{\bm{\Lambda}}}
\def\mSigma{{\bm{\Sigma}}}

\DeclareMathAlphabet{\mathsfit}{\encodingdefault}{\sfdefault}{m}{sl}
\SetMathAlphabet{\mathsfit}{bold}{\encodingdefault}{\sfdefault}{bx}{n}
\newcommand{\tens}[1]{\bm{\mathsfit{#1}}}
\def\tA{{\tens{A}}}
\def\tB{{\tens{B}}}
\def\tC{{\tens{C}}}
\def\tD{{\tens{D}}}
\def\tE{{\tens{E}}}
\def\tF{{\tens{F}}}
\def\tG{{\tens{G}}}
\def\tH{{\tens{H}}}
\def\tI{{\tens{I}}}
\def\tJ{{\tens{J}}}
\def\tK{{\tens{K}}}
\def\tL{{\tens{L}}}
\def\tM{{\tens{M}}}
\def\tN{{\tens{N}}}
\def\tO{{\tens{O}}}
\def\tP{{\tens{P}}}
\def\tQ{{\tens{Q}}}
\def\tR{{\tens{R}}}
\def\tS{{\tens{S}}}
\def\tT{{\tens{T}}}
\def\tU{{\tens{U}}}
\def\tV{{\tens{V}}}
\def\tW{{\tens{W}}}
\def\tX{{\tens{X}}}
\def\tY{{\tens{Y}}}
\def\tZ{{\tens{Z}}}

\def\gA{{\mathcal{A}}}
\def\gB{{\mathcal{B}}}
\def\gC{{\mathcal{C}}}
\def\gD{{\mathcal{D}}}
\def\gE{{\mathcal{E}}}
\def\gF{{\mathcal{F}}}
\def\gG{{\mathcal{G}}}
\def\gH{{\mathcal{H}}}
\def\gI{{\mathcal{I}}}
\def\gJ{{\mathcal{J}}}
\def\gK{{\mathcal{K}}}
\def\gL{{\mathcal{L}}}
\def\gM{{\mathcal{M}}}
\def\gN{{\mathcal{N}}}
\def\gO{{\mathcal{O}}}
\def\gP{{\mathcal{P}}}
\def\gQ{{\mathcal{Q}}}
\def\gR{{\mathcal{R}}}
\def\gS{{\mathcal{S}}}
\def\gT{{\mathcal{T}}}
\def\gU{{\mathcal{U}}}
\def\gV{{\mathcal{V}}}
\def\gW{{\mathcal{W}}}
\def\gX{{\mathcal{X}}}
\def\gY{{\mathcal{Y}}}
\def\gZ{{\mathcal{Z}}}

\def\sA{{\mathbb{A}}}
\def\sB{{\mathbb{B}}}
\def\sC{{\mathbb{C}}}
\def\sD{{\mathbb{D}}}
\def\sF{{\mathbb{F}}}
\def\sG{{\mathbb{G}}}
\def\sH{{\mathbb{H}}}
\def\sI{{\mathbb{I}}}
\def\sJ{{\mathbb{J}}}
\def\sK{{\mathbb{K}}}
\def\sL{{\mathbb{L}}}
\def\sM{{\mathbb{M}}}
\def\sN{{\mathbb{N}}}
\def\sO{{\mathbb{O}}}
\def\sP{{\mathbb{P}}}
\def\sQ{{\mathbb{Q}}}
\def\sR{{\mathbb{R}}}
\def\sS{{\mathbb{S}}}
\def\sT{{\mathbb{T}}}
\def\sU{{\mathbb{U}}}
\def\sV{{\mathbb{V}}}
\def\sW{{\mathbb{W}}}
\def\sX{{\mathbb{X}}}
\def\sY{{\mathbb{Y}}}
\def\sZ{{\mathbb{Z}}}

\def\emLambda{{\Lambda}}
\def\emA{{A}}
\def\emB{{B}}
\def\emC{{C}}
\def\emD{{D}}
\def\emE{{E}}
\def\emF{{F}}
\def\emG{{G}}
\def\emH{{H}}
\def\emI{{I}}
\def\emJ{{J}}
\def\emK{{K}}
\def\emL{{L}}
\def\emM{{M}}
\def\emN{{N}}
\def\emO{{O}}
\def\emP{{P}}
\def\emQ{{Q}}
\def\emR{{R}}
\def\emS{{S}}
\def\emT{{T}}
\def\emU{{U}}
\def\emV{{V}}
\def\emW{{W}}
\def\emX{{X}}
\def\emY{{Y}}
\def\emZ{{Z}}
\def\emSigma{{\Sigma}}

\newcommand{\etens}[1]{\mathsfit{#1}}
\def\etLambda{{\etens{\Lambda}}}
\def\etA{{\etens{A}}}
\def\etB{{\etens{B}}}
\def\etC{{\etens{C}}}
\def\etD{{\etens{D}}}
\def\etE{{\etens{E}}}
\def\etF{{\etens{F}}}
\def\etG{{\etens{G}}}
\def\etH{{\etens{H}}}
\def\etI{{\etens{I}}}
\def\etJ{{\etens{J}}}
\def\etK{{\etens{K}}}
\def\etL{{\etens{L}}}
\def\etM{{\etens{M}}}
\def\etN{{\etens{N}}}
\def\etO{{\etens{O}}}
\def\etP{{\etens{P}}}
\def\etQ{{\etens{Q}}}
\def\etR{{\etens{R}}}
\def\etS{{\etens{S}}}
\def\etT{{\etens{T}}}
\def\etU{{\etens{U}}}
\def\etV{{\etens{V}}}
\def\etW{{\etens{W}}}
\def\etX{{\etens{X}}}
\def\etY{{\etens{Y}}}
\def\etZ{{\etens{Z}}}

\newcommand{\pdata}{p_{\rm{data}}}
\newcommand{\ptrain}{\hat{p}_{\rm{data}}}
\newcommand{\Ptrain}{\hat{P}_{\rm{data}}}
\newcommand{\pmodel}{p_{\rm{model}}}
\newcommand{\Pmodel}{P_{\rm{model}}}
\newcommand{\ptildemodel}{\tilde{p}_{\rm{model}}}
\newcommand{\pencode}{p_{\rm{encoder}}}
\newcommand{\pdecode}{p_{\rm{decoder}}}
\newcommand{\precons}{p_{\rm{reconstruct}}}

\newcommand{\laplace}{\mathrm{Laplace}} % Laplace distribution

\newcommand{\E}{\mathbb{E}}
\newcommand{\Ls}{\mathcal{L}}
\newcommand{\RR}{\mathbb{R}}
\newcommand{\emp}{\tilde{p}}
\newcommand{\lr}{\alpha}
\newcommand{\reg}{\lambda}
\newcommand{\rect}{\mathrm{rectifier}}
\newcommand{\softmax}{\mathrm{softmax}}
\newcommand{\sigmoid}{\sigma}
\newcommand{\softplus}{\zeta}
\newcommand{\KL}{D_{\mathrm{KL}}}
\newcommand{\Var}{\mathrm{Var}}
\newcommand{\standarderror}{\mathrm{SE}}
\newcommand{\Cov}{\mathrm{Cov}}
\newcommand{\normlzero}{L^0}
\newcommand{\normlone}{L^1}
\newcommand{\normltwo}{L^2}
\newcommand{\normlp}{L^p}
\newcommand{\normmax}{L^\infty}

\newcommand{\parents}{Pa} % See usage in notation.tex. Chosen to match Daphne's book.

\DeclareMathOperator*{\argmax}{arg\,max}

\DeclareMathOperator{\sign}{sign}
\DeclareMathOperator{\Tr}{Tr}
\let\ab\allowbreak

\newcommand{\defeq}{\coloneqq}
\newcommand{\grad}{\nabla}
\newcommand{\Ea}[1]{\E\left[#1\right]}
\newcommand{\Eb}[2]{\E_{#1}\!\left[#2\right]}
\newcommand{\Vara}[1]{\Var\left[#1\right]}
\newcommand{\Varb}[2]{\Var_{#1}\left[#2\right]}
\newcommand{\kl}[2]{D_{\mathrm{KL}}\!\left(#1 ~ \| ~ #2\right)}
\newcommand{\bA}{\mathbf{A}}
\newcommand{\bI}{\mathbf{I}}
\newcommand{\bJ}{\mathbf{J}}
\newcommand{\bH}{\mathbf{H}}
\newcommand{\bL}{\mathbf{L}}
\newcommand{\bM}{\mathbf{M}}
\newcommand{\bQ}{\mathbf{Q}}
\newcommand{\bR}{\mathbf{R}}
\newcommand{\bW}{\mathbf{W}}
\newcommand{\bzero}{\mathbf{0}}
\newcommand{\bone}{\mathbf{1}}
\newcommand{\bb}{\mathbf{b}}
\newcommand{\bc}{\mathbf{c}}
\newcommand{\bd}{\mathbf{d}}
\newcommand{\be}{\mathbf{e}}
\newcommand{\bh}{\mathbf{h}}
\newcommand{\bu}{\mathbf{u}}
\newcommand{\bv}{\mathbf{v}}
\newcommand{\bw}{\mathbf{w}}
\newcommand{\bx}{\mathbf{x}}
\newcommand{\by}{\mathbf{y}}
\newcommand{\bz}{\mathbf{z}}
\newcommand{\bxh}{\hat{\mathbf{x}}}
\newcommand{\btheta}{{\boldsymbol{\theta}}}
\newcommand{\bphi}{{\boldsymbol{\phi}}}
\newcommand{\bepsilon}{{\boldsymbol{\epsilon}}}
\newcommand{\bmu}{{\boldsymbol{\mu}}}
\newcommand{\bnu}{{\boldsymbol{\nu}}}
\newcommand{\bSigma}{{\boldsymbol{\Sigma}}}
\DeclareMathOperator{\snr}{SNR}
\DeclareMathOperator{\expm1}{expm1}

\newcommand{\bP}{\mathbf{P}}
\newcommand{\bPh}{\hat{\mathbf{P}}}
\newcommand{\bV}{\mathbf{V}}
\newcommand{\bVh}{\hat{\mathbf{V}}}

\definecolor{mainblue}{HTML}{1f77b4}  % Matplotlib's C0 blue
\definecolor{lightblue}{HTML}{CCE5FF}  % Light blue for bands
\definecolor{mainorange}{HTML}{ff7f0e}  % Matplotlib's C1 orange
\definecolor{lightorange}{HTML}{FFE5CC}  % Light orange for bands

\newcommand{\bigarrow}[1]{%
  \noindent
  \begin{tikzpicture}[baseline=(txt.base)]
    \node (txt) at (.5\linewidth,0) {#1};
    \draw            (0,0) -- (txt.west);   % plain line (no head)
    \draw[->] (txt.east) -- (\linewidth,0); % arrowhead on the right
  \end{tikzpicture}%
}
\newcommand{\bigarrowleftright}[1]{%
  \noindent
  \begin{tikzpicture}[baseline=(txt.base)]
    \node (txt) at (.5\linewidth,0) {#1};
    \draw[<->] (0,0) -- (\linewidth,0); % arrowhead on both sides
    \node[fill=white] at (.5\linewidth,0) {#1};
  \end{tikzpicture}%
}
\newcommand{\video}[1]{%
  \includegraphics[height=0.6\baselineskip]{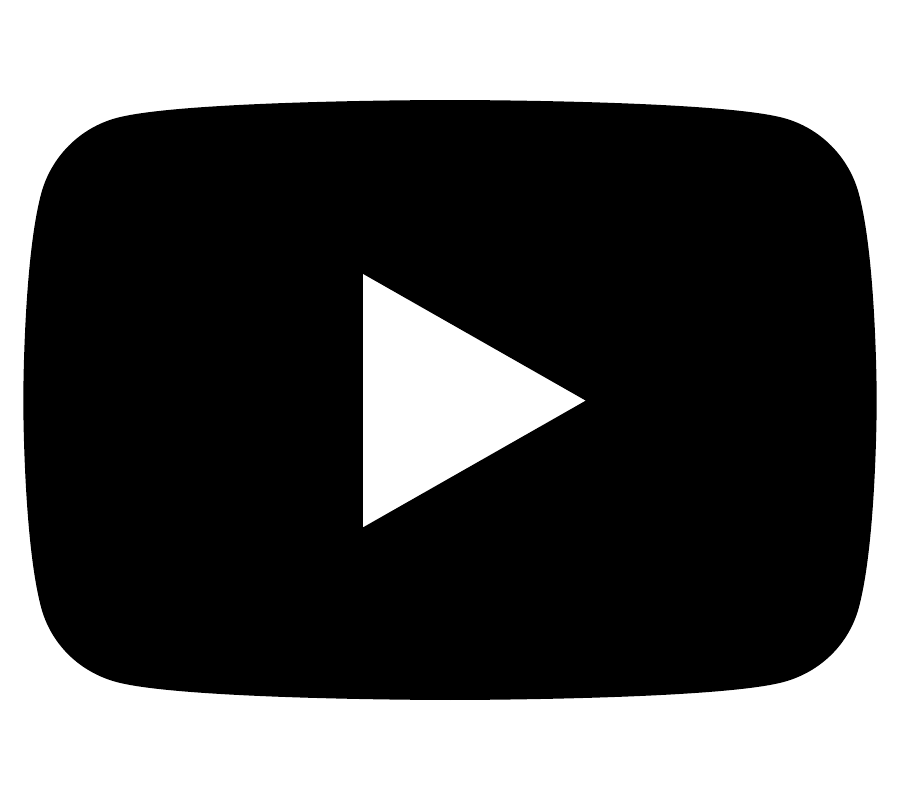}%
  {\color[HTML]{76B900}: #1}%
}
\newcommand{\hlc}[2]{\begingroup\sethlcolor{#2}\hl{#1}\endgroup}
\newcommand{\bp}{\mathbf{p}}

\newcommand{\cA}{\mathcal{A}}
\newcommand{\cB}{\mathcal{B}}
\newcommand{\cC}{\mathcal{C}}
\newcommand{\cD}{\mathcal{D}}
\newcommand{\cE}{\mathcal{E}}
\newcommand{\cF}{\mathcal{F}}
\newcommand{\cG}{\mathcal{G}}
\newcommand{\cH}{\mathcal{H}}
\newcommand{\cI}{\mathcal{I}}
\newcommand{\cJ}{\mathcal{J}}
\newcommand{\cK}{\mathcal{K}}
\newcommand{\cL}{\mathcal{L}}
\newcommand{\cM}{\mathcal{M}}
\newcommand{\cN}{\mathcal{N}}
\newcommand{\cO}{\mathcal{O}}
\newcommand{\co}{\mathcal{o}}
\newcommand{\cP}{\mathcal{P}}
\newcommand{\cQ}{\mathcal{Q}}
\newcommand{\cR}{\mathcal{R}}
\newcommand{\cS}{\mathcal{S}}
\newcommand{\cT}{\mathcal{T}}
\newcommand{\cU}{\mathcal{U}}
\newcommand{\cV}{\mathcal{V}}
\newcommand{\cW}{\mathcal{W}}
\newcommand{\cX}{\mathcal{X}}
\newcommand{\cY}{\mathcal{Y}}
\newcommand{\cZ}{\mathcal{Z}}

\crefname{section}{\S}{\S}
\Crefname{section}{\S}{\S}
\crefformat{section}{\S#2#1#3}
\Crefname{table}{Tb.}{Tb.}
\crefname{table}{Tb.}{Tb.}
\Crefname{figure}{Fig.}{Fig.}
\crefname{figure}{Fig.}{Fig.}

\newcommand{\smalltimes}{%
  {\ooalign{$\phantom{0}$\cr\hidewidth$\scriptstyle\times$\cr}}%
}

\newcommand{\mattriplet}{($E$, $\nu$, $\rho$)}
\definecolor{cvprblue}{rgb}{0.21,0.49,0.74}

\let\oldcite\cite
\renewcommand{\cite}{\citep}
\let\oldtable\table
\let\endoldtable\endtable
\renewenvironment{table*}
  {\begin{table}}
  {\end{table}}

\let\oldfigure\figure
\let\endoldfigure\endfigure
\renewenvironment{figure*}
  {\begin{figure}}
  {\end{figure}}

\title{\acronym: Predicting {\color{nvidiagreen}Vo}lumetric {\color{nvidiagreen}M}echanical {\color{nvidiagreen}P}roperty Fields}

\author{%
Rishit Dagli$^{1,2}$ \; Donglai Xiang$^{1}$ \; Vismay Modi$^{1}$ \; Charles Loop$^{1}$ \; Clement Fuji Tsang$^{1}$ \\[2pt] \textbf{Anka He Chen}$^{1}$ \; \textbf{Anita Hu}$^{1}$ \; \textbf{Gavriel State}$^{1}$ \; \textbf{David I.W. Levin}$^{1,2}$ \; \textbf{Maria Shugrina}$^{1}$ \\[2pt] $^{1}$NVIDIA\; $^{2}$University of Toronto\\[2pt]
\url{https://research.nvidia.com/labs/sil/projects/vomp}
}

\newcommand{\fix}{\marginpar{FIX}}
\newcommand{\new}{\marginpar{NEW}}

\iclrfinalcopy % Uncomment for camera-ready version, but NOT for submission.
\begin{document}
\maketitle

% \vspace{-0.6em}

  \begin{center}
    % \vspace{-2em}
    \includegraphics[width=\linewidth]{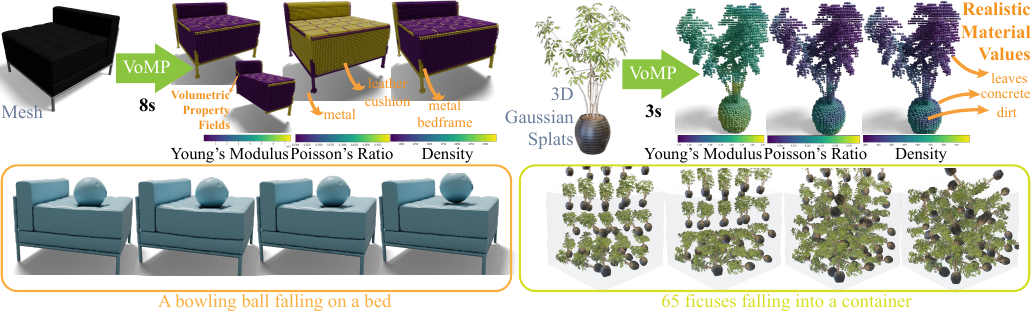}
    \captionof{figure}{\textbf{\acronym} predicts physically accurate volumetric mechanical property fields across 3D representations in just a few seconds (top), enabling their use in realistic deformable simulations (bottom).}
    \label{fig:teaser}
  \end{center}
  %\vspace{-0.8\baselineskip}
% \vspace{-0.6em}
\begin{abstract}
%Simulating objects often requires the user to handcraft spatially varying mechanical properties based on extensive experimentation and their intuition of visually pleasing simulations. 

Physical simulation relies on spatially-varying mechanical properties, often laboriously hand-crafted. \acronym\ is a feed-forward method trained to predict Young’s modulus ($E$), Poisson’s ratio ($\nu$), and density ($\rho$) throughout \emph{the volume} of 3D objects, in any representation that can be rendered and voxelized. \acronym\ aggregates per-voxel multi-view features and passes them to our trained Geometry Transformer to predict per-voxel material latent codes. These latents reside on a space of physically plausible materials, which we learn from a real-world dataset, guaranteeing the validity of decoded per-voxel materials.
To obtain object-level training data, we propose an annotation pipeline combining knowledge from segmented 3D datasets, material databases, and a vision-language model, along with a new benchmark. Experiments show that \acronym\ estimates accurate volumetric properties, far outperforming prior art in accuracy and speed.

%  and can convert 3D objects into simulation-ready assets, resulting in realistic deformable simulations and 
% \masha{It's good idea to split off separate sections into separate tex files, just as I have done here with the abstract.}
\end{abstract}

% multi-view image features and the voxelized 3D geometry to predict a latent embedding per voxel. These per-voxel latent embeddings are then passed into a VAE we train that decodes these latents into physically valid material triplets $(E, \nu,\rho)$. To obtain training data for this problem, we propose an annotation pipeline that combines the knowledge from part-segmented 3D object datasets, material databases and a vision-language model. Our experiments demonstrate that \acronym\ estimates accurate olumetric properties and can convert
% any \dave{'any' object is too strong a claim since we are data limited}
%3D objects into a simulation-ready assets without any handcrafting and produces realistic deformable simulations.

\section{Introduction}
\label{sec:intro}

Accurate physics simulation is a critical part of modern design and engineering, for example, in workflows like creating Digital Twins (virtual replicas of real systems)~\cite{Grieves2017}, Real-2-Sim (generating digital simulation from the real world)~\cite{omniverse}, and Sim-2-Real (transferring policies trained in simulation to real-world deployment)~\cite{rudin2021learning}. However, setting up reliable simulations remains labor-intensive,
partially due to the necessity to provide accurate mechanical properties \textit{throughout the volume} of every object, namely the spatially-varying 
Young's Modulus ($E$), Poisson's ratio ($\nu$), and density ($\rho$). Common 3D capture methods~\cite{kerbl3Dgaussians} and
3D repositories~\cite{objaverse} rarely contain such annotations, forcing artists and engineers to guess or copy-paste coarse material presets in a subjective, time-consuming process. We focus on automatic prediction of these parameters, addressing important limitations of prior art.

% Accurate physics simulation is a critical part of modern design and engineering, yet setting up reliable simulations remains labor-intensive. One challenge is that high-fidelity simulation of phenomena like deformation and structural failure relies on accurate mechanical properties \textit{throughout the volume} of every object, namely the spatially-varying \textit{Young's Modulus ($E$), Poisson's ratio ($\nu$), and density ($\rho$)}. Workflows like creating Digital Twins (virtual replicas of a real system)~\cite{Grieves2017}, Real-2-Sim (generating digital simulation from the real-world)~\cite{omniverse}, and Sim-2-Real (transfer policies trained in simulation to real-world deployment)~\cite{rudin2021learning} rely on simulation accuracy, yet rarely offer automated solutions to annotate mechanical properties. Most 3D repositories (ShapeNet~\cite{shapenet2015}, Thingi10K~\cite{Thingi10K}, Objaverse~\cite{objaverse}, etc.) contain geometry, but rarely volumetric materials.
% As a result, artists and engineers are forced to guess or copy-paste coarse material presets in a subjective, time-consuming process, which can result in physically incorrect values. While some works propose automatic estimation, we address many critical limitations of existing techniques.

We propose \acronym, \emph{the first feed-forward model trained to estimate simulation-ready mechanical property fields $(E, \nu, \rho)$ within the volume of 3D objects across representations}. Rather than specializing on inputs like Gaussian Splats~\cite{shuai2025pugszeroshotphysicalunderstanding, xie2024physgaussian}, our method works for any geometry that can be voxelized and rendered from turnaround views, including meshes, Gaussian Splats, NeRFs and SDFs (\Cref{fig:teaser}). Unlike virtually all prior works, \acronym\ is fully feed-forward, requiring
no per-object optimization of feature fields~\cite{zhai2024physicalpropertyunderstandinglanguageembedded, shuai2025pugszeroshotphysicalunderstanding} or run-time aggregation of Vision-Language Model (VLM) ~\cite{lin2025phys4dgenphysicscompliant4dgeneration} or Video Model~\cite{lin2025omniphysgs} supervision.  
Uniquely among others, \acronym\ outputs true mechanical properties (a.k.a.\ material parameters), like those measured in the real world.
Many existing pipelines target fast, approximate simulators, resulting in simulator-specific parameters~\cite{physdreamer, huang2024dreamphysicslearningphysicsbased3d} that may not transfer reliably across frameworks (\Cref{fig:simulators}), whereas our result is directly compatible with any accurate simulator.  Finally, unlike prior art, our method is designed to assign materials throughout the object volume, which is critical for simulation fidelity.
% \todo{Rishit: in this paragraph don't be exhaustive, select just the most representative works in each category -- make sure it's consistent with the way these are represented in related.}

%To summarize, unlike prior art in this area, our method \textit{(1)} is a trained feed-forward model with minimal preprocessing, \textit{(2)} generalizes across 3D representations, \textit{(3)} predicts physically valid properties that can be used with an accurate simulator, and \textit{(4)} captures volumetric aspects and internal structures necessary for accurate physics simulation of objects.

To enable learning physically valid mechanical properties, we first train a latent space on a database of real-world values $(E, \nu, \rho)$ using a variational auto-encoder MatVAE (\S\ref{sec:mat_latent_space}).
To predict mechanical property fields for 3D objects, our method first voxelizes the input geometry and aggregates multi-view image features
across the voxels (\S\ref{sec:aggregating}). This process accepts many representations\footnote{We describe available methods for meshes, SDFs, and NeRFs, and present a method for Splats in~\S\ref{sec:implementation}.} and is fast, unlike optimization used in concurrent work \cite{le2025pixiefastgeneralizablesupervised}. We pass the voxel features through the Geometry Transformer (\S\ref{sec:3dencdoer}), trained to output per-voxel material latents. The MatVAE latent space decouples learning material assignments for objects from learning what materials are valid, ensuring that the final volumetric properties $(E, \nu, \rho)$ decoded by MatVAE are physically valid, even in the case of interpolation. To create material property fields for training, we propose a pipeline (\S\ref{sec:data}) combining the knowledge from part-segmented 3D assets, material databases, visual textures, and a VLM. Our experiments (\S\ref{sec:exp}) show
that \acronym\ estimates simulation-ready spatially-varying mechanical properties across a range of object classes and representations, resulting in realistic elastodynamic simulations. 
We evaluate our method on an existing mass prediction benchmark and contribute a new material estimation benchmark (\S\ref{sec:estimation}), consistently outperforming prior art~\cite{shuai2025pugszeroshotphysicalunderstanding, lin2025phys4dgenphysicscompliant4dgeneration, zhai2024physicalpropertyunderstandinglanguageembedded}. In summary, our contributions are:
\begin{itemize}[leftmargin=*]
    \item The first (to our knowledge) method to estimate object mechanical material property fields that \textit{(1)} is a trained feed-forward model with minimal preprocessing, \textit{(2)} generalizes across 3D representations, \textit{(3)} predicts physically valid properties that can be used with an accurate simulator, and \textit{(4)} predicts mechanical properties \emph{within the volume} of objects (\S\ref{sec:method}).
    \item The first (to our knowledge) mechanical properties latent space (\S\ref{sec:mat_latent_space}).
    \item An automatic data annotation pipeline and a new benchmark for volumetric physics materials (\S\ref{sec:data}).
    \item Thorough evaluation through high-fidelity simulations and quantitative metrics on existing and new benchmarks, significantly outperforming the prior art (\S\ref{sec:exp}).
\end{itemize}

\ifdefined\iclr
    \begin{wrapfigure}[]{r}{0.35\textwidth}
  \includegraphics[width=0.35\textwidth]{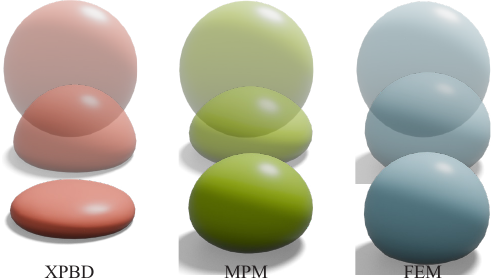}
  \caption{\textbf{Simulator differences} when dropping
    a solid sphere with \footnotesize{$(E, \nu, \rho) = (10^4 Pa, 0.3, 10^3 \operatorname{kg/m}^3)$} with XPBD~\cite{xpbd} and MPM~\cite{mpm} vs.\ more accurate FEM.}
    \label{fig:simulators}
\end{wrapfigure}

\else
    \begin{figure}[tb]
    \centering
    \includegraphics{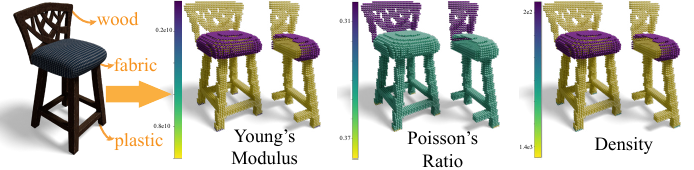}
    \caption{\textbf{\acronym}\ generates volumetric mechanical property fields from 3D objects.}
    \vspace{-2em}
    \label{fig:example}
\end{figure}
\fi

% TODO: we probably don't need to qualify in the intro
%used with an accurate physics simulator run to convergence (\Cref{sec:sim}).

% TODO: full version in related
%Predictors trained solely from surface geometry~\cite{cao2025sophylearninggeneratesimulationready}, and Language-embedded feature fields~\cite{zhai2024physicalpropertyunderstandinglanguageembedded, shuai2025pugszeroshotphysicalunderstanding, le2025pixiefastgeneralizablesupervised} learn appearance-level proxies rather than internal, spatially varying mechanics.  

% TODO: full version can go in related.
%Unlike prior works focusing exclusively on meshes~\cite{hsu2024autovfxphysicallyrealisticvideo}, NeRFs~\cite{zhai2024physicalpropertyunderstandinglanguageembedded}, Gaussian Splats~\cite{shuai2025pugszeroshotphysicalunderstanding, xie2024physgaussian, physdreamer, huang2024dreamphysicslearningphysicsbased3d, lin2025phys4dgenphysicscompliant4dgeneration}, or point clouds~\cite{cao2025sophylearninggeneratesimulationready}, our method works for any geometry that can be voxelized and rendered from turnaround views. 

\section{Related Work}
\label{sec:rw}

\subsection{Background}
\label{sec:mechbackground}
All algorithms for continuum-based simulation of solids and liquids require material models as input. The material, or constitutive, model is the function that determines the force response of a class of materials (e.g., rubbers, snow, water) to internal strains and strain rates. To produce the correct constitutive behavior for a given material, the model requires an accurate set of corresponding material parameters for every point in the simulated volume. For locally isotropic material models, Young’s modulus ($E$, in the 1D linear regime, the proportionality constant between stress and strain), Poisson’s ratio ($\nu$, the negative ratio of transverse to axial strain under uniaxial loading) and density ($\rho$, unit mass per volume) are ubiquitous. Given an accurate and valid triplet \mattriplet\ along with a reasonable material model, a consistent numerical simulation can produce accurate predictions of an object’s behavior under load. Measured, real-world parameters are portable to any consistent simulation algorithm (we use high-resolution Finite Element Methods). Further, they are portable across any material model that relies on density, Young’s modulus and Poisson’s ratio, or derived quantities, such as shear or bulk modulus (e.g., Neo-Hookean, St. Venant–Kirchhoff, As-Rigid-As-Possible, Co-Rotated Elastic, Mooney–Rivlin, and Ogden models). 
On the other hand, many physics simulation algorithms are not implemented or applied in a consistent fashion, favoring speed over accuracy~\cite{xpbd,mpm}. In these cases, material parameters must be modified to avoid inaccurate behavior (\Cref{fig:simulators}).

\subsection{Inferring Mechanical Properties of Static Objects}
\label{sec:rw:inferring}

Our goal is to predict volumetric mechanical properties given only shape and appearance, a challenging inverse problem,
which research suggests humans learn good intuition about
~\cite{10.1117/12.429489, FLEMING201462, 10.1167/13.8.9, sharan2009material}. However, progress in learning-based approaches has been hampered by limited data.
Existing datasets are small~\cite{gao2022objectfolder20multisensoryobject, downs2022googlescannedobjectshighquality, chen2025matpredictdatasetbenchmarklearning},
contain noisy labels~\cite{lin2018learning},
use simulator-specific parameters~\cite{mishra2024latticemldatadrivenapplicationpredicting, Xie__2025, belikov2015material},
provide only coarse annotations~\cite{ahmed20253dcompat200, slim20233dcompat, 10.1007/978-3-031-20074-8_7} or are biased towards rigid or man-made objects~\cite{cao2025physx-3d}. Worse, data collection is difficult, relying on rigorous physical experiments~\cite{ASTM_D638_2022, ASTM_E8_E8M_2024,pai2000robotics}, and even then lacking spatial material fields~\cite{loveday2004tensile} due to digitization and annotation challenges.
%and issues with digitizing and annotating physical assets.
% One Rare existing datasets~\cite{cao2025physx-3d}, the dataset was created from PartNet which is known to be highly biased towards rigid and man-made objects~\cite{partnet}.
% \todo{Incorporate: Creating paired training data is challenging since the material properties can not be manually annotated. For physical objects, they need to be observed by performing physical experiments with specialized machines and tests~\cite{ASTM_D638_2022, ASTM_E8_E8M_2024}. Even with these experiments, it is often challenging to come up with spatial fields of materials~\cite{loveday2004tensile} due to the difficulty in annotating 3D data and digitizing these physical assets. }
% \todo{we need to talk about physx3d}
% \todo{Rishit double check accuracy on datasets.}

As a result, works that infer physical properties from appearance often leverage knowledge from large pre-trained models.
NeRF2Physics~\cite{zhai2024physicalpropertyunderstandinglanguageembedded} and PUGS~\cite{shuai2025pugszeroshotphysicalunderstanding} optimize language-embedded feature fields for a NeRF~\cite{mildenhall2020nerf} or 3D Gaussians~\cite{kerbl3Dgaussians}, respectively, to predict coarse stiffness categories and density, but require per-object optimization and are limited in their ability to predict values inside objects due to the lack of meaningful features inside NeRFs or splats. Many approaches distill signals from a Video Generation Model and optimize physics parameters by backpropagating through fast, approximate physics simulators, resulting in a slow optimization process, yielding
materials deviating from real-world values and overfit to a specific simulation setup~\cite{physdreamer, huang2024dreamphysicslearningphysicsbased3d, Liu_2025_CVPR, cleach2023differentiablephysicssimulationdynamicsaugmented, liu2024physics3d, lin2025omniphysgs} (\S\ref{sec:mechbackground}). Many methods are also tailored
to a specific 3D representation or real-time simulation implementations, such as Splats ~\cite{xie2024physgaussian} or explicit Material Point Methods~\cite{mpm,le2025pixiefastgeneralizablesupervised}, or work with coarse material categories ~\cite{fischer2024sama, hsu2024autovfxphysicallyrealisticvideo, lin2025phys4dgenphysicscompliant4dgeneration, Xia_2025_CVPR} that must be manually mapped to simulation parameters. Instead, we aim to augment
objects across 3D representations with fine-grained spatially-varying mechanical properties that are physically accurate and compatible across accurate simulators.
Like our method, many techniques leverage vision-language (VLM) models. PhysGen~\cite{liu2024physgen} and PhysGen3D~\cite{Chen_2025_CVPR} use a VLM to infer mass, elasticity, and friction for segmented parts of a single image. Phys4DGen~\cite{lin2025phys4dgenphysicscompliant4dgeneration} uses a VLM to annotate parts of a 3D model with coarse material labels, which are then mapped to physical parameters, a baseline used in our evaluation. Most works above rely on aggregation of large model outputs for every input shape, which can be brittle and time-consuming at run-time, and can only leverage external segmentation. Instead, our method uses a VLM paired with other data sources to annotate a \emph{training dataset} for a feed-forward model leveraging 3D data to annotate and learn internal material composition.

Like our method, SOPHY~\cite{cao2025sophylearninggeneratesimulationready}, PhysX-3D~\cite{cao2025physx-3d}, PhysSplat \cite{zhao2024efficient_aka_automated,zhao2024automated} (a.k.a.\ SimAnything) and the concurrent Pixie~\cite{le2025pixiefastgeneralizablesupervised} leverage pretrained models and 3D data to annotate a \emph{training} dataset with physical materials. PhysSplat trains a network to predict spatially-varying simulator-specific material offset weights for MPM by using outputs from video distillation \cite{liu2024physics3d}, not focusing on material accuracy.
SOPHY and PhysX-3D are 3D generative models, designed to 
generate new shapes augmented with physical attributes, and cannot augment existing assets, which is our goal. Still, we detail similar aspects of these works.
Like these works, our method uses a VLM to annotate 3D objects with Young's Modulus, Poisson's ratio, and density, but
we do not rely on the human-in-the-loop and instead leverage multiple data sources, not just VLM knowledge, to ensure more accurate physical properties.
As a baseline, SOPHY does implement a material decoder, but it has not been made available, and only considers object surface, while we aim to estimate volumetric properties.
% \todo{Rishit, drill down on SOPHY baseline that is like ours, focus on surface vs. volume.}
Like our method, PhysX-3D adopts the structural latent space of TRELLIS, but trains a joint generative model over these and learned shape-aware physical properties latents in order to generate physics-augmented shapes from scratch. In contrast, 
we treat material prediction as deterministic inference for simplicity, and further adjust the TRELLIS pipeline to facilitate
accurate material prediction inside the object. Pixie~\cite{le2025pixiefastgeneralizablesupervised}, a concurrent work and the only other feed-forward approach, is trained on semantically-segmented objects and uses points from filtering NeRF densities. Thus, Pixie is trained on segments biased toward the surface, as we show in Fig.~\ref{fig:pixie-annot}, while we demonstrate being able to estimate volumetric properties with internal structures. Furthermore, unlike Pixie, we specifically focus on estimating physically plausible material properties, such as those measured in the real world. 

    \begin{figure*}[t]
        \centering
        \label{fig:pipeline}
        \includegraphics[width=\textwidth]{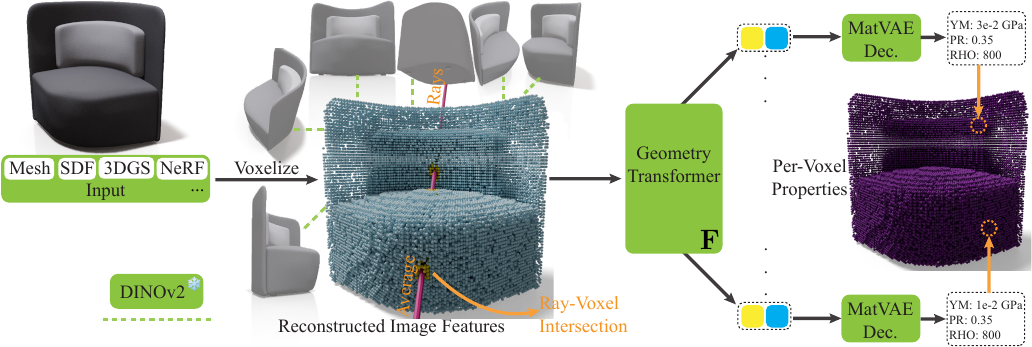}
        \caption{\textbf{\acronym\ Overview.} For any input geometry, we aggregate multi-view DINOv2 features across its volumetric voxelization (\S\ref{sec:aggregating}). A trained
        GeometryTransformer (\S\ref{sec:3dencdoer}) predicts per-voxel material latents, decoded by MatVAE (\S\ref{sec:mat_latent_space}) into mechanical properties \mattriplet.}
    \end{figure*}

\section{Mechanical Properties Latent Space}\label{sec:mat_latent_space}\label{sec:matvae}

To learn a latent space of valid Young's modulus, Poisson's ratio, and density triplets {\mattriplet} (\Cref{sec:mechbackground}), we propose MatVAE,
a variational autoencoder (VAE) trained on a dataset of real-world values $\{m_i :=(E_i,\;\nu_i,\;\rho_i)\}$ (\Cref{sec:mtd}). The model's objective is to map these triplets $m$ into a 2-dimensional latent space, $z \in \mathbb{R}^2$, from which they can be accurately reconstructed. While this offers only minor compression ($\mathbb{R}^3 \to \mathbb{R}^2$), this latent 2D space of material 
properties is now easy to visualize, sample, and interpolate within, and
results in consistent distances between material triplets with
disparate units (\Cref{fig:matvaelatentspace},\Cref{sec:reconstructtriplet}). MatVAE acts like a continuous tokenizer that allows us to always ensure \acronym\ output properties that fall inside the range of some materials. 

We build on VAE~\cite{kingma2022autoencodingvariationalbayes},
with the reconstruction component of the loss defined as
mean-squared error between the input $(E_i,\nu_i,\rho_i)$
and reconstructed material values $(\hat{E}_i, \hat{\nu}_i,\hat{\rho}_i)$:
\begin{equation}
\mathcal{L}_{\text{Recon}}
=\frac{1}{N}\sum_{i=1}^{N}
\bigl\|
((E_i,\nu_i,\rho_i)^{N})^{\mathsf T}
-
((\hat{E}_i,\hat{\nu}_i,\hat{\rho}_i)^{N})^{\mathsf T}
\bigr\|_2^{\,2},
\label{eq:recon_loss}
\end{equation}
where ${\mathsf T}$ denotes transpose and ${N}$ per-property normalization, where $E$ and $\rho$ are first log-transformed ($\log_{10}(E)$, $\log_{10}(\rho)$), then normalized to $[0,1]$, while $\nu$ is directly normalized to $[0,1]$. We find other normalization schemes without $\log$-transform or standard $z$-score normalization induce a heavy-tailed feature distribution, which is poorly conditioned for learning (\S\ref{sec:ablations}).

We make several modifications over standard VAE. \textit{First}, to capture a complex posterior beyond a simple Gaussian, the encoder's output is transformed by a (radial) Normalizing Flow~\cite{pmlr-v37-rezende15}, giving us a more flexible variational distribution $q_\phi(z|m)$ since we observe heavy-tailed distribution for Young's Modulus and Density while Poisson's Ratio concentrates near the boundaries after normalization. \textit{Second}, we decompose the KL-divergence term of the ELBO following~\cite{chen2018isolating}. This allows us to directly penalize the total correlation $\text{TC}(z) = \text{KL}(\bar{q}_\phi(z) || \prod_j \bar{q}_\phi(z_j))$ where $\bar{q}_\phi(z)$ is the aggregated posterior, $z_j$ is the $j \in \{1,2\}$-th coordinate of the latent vector $z$. Penalizing $\text{TC}$ allowed us to reduce the high dependence between latent coordinates which caused MatVAE to encode density in both dimensions. \textit{Third,} we observe imbalanced reconstruction, \textit{i.e.} the latent space collapses to one property, giving us low reconstruction errors for one property and high reconstruction error for others (\S\ref{sec:ablations}).
% \masha{What does this mean? Can you rewrite this sub-sentence to be more precise?}
Thus, to ensure the 2 latent dimensions are actively utilized, we introduce a capacity constraint ($\delta\times z_{\text{dim}}$) based on~\cite{higgins2017beta}, resulting in the following final objective: %Our final objective for MatVAE combines all these elements,
\begin{equation}
\begin{aligned}
\mathcal{L}_{\text{MatVAE}} &= \underbrace{\mathcal{L}_{\text{Recon}}}_{\text{MSE}} + \overbrace{\underbrace{\gamma \cdot \text{MI}(z)}_{\text{Mutual Information}} + \underbrace{\beta \cdot \text{TC}(z)}_{\text{Total Correlation}}}^{\text{Latent Space Regularization}} + \alpha \cdot \sum_{j=1}^{d}  \underbrace{\max\!\big(\delta,\ \text{KL}(q_\phi(z_j)\,\|\,p(z_j))\big)}_{\text{Dimension-wise KL}},
\end{aligned}
\label{eq:final_objective}
\end{equation}
where we set $\gamma, \beta, \alpha = (1.0, 2.0, 1.0)$, with a free nats constraint $\delta = 0.1$. See \S\ref{app:designmatvae} for more details.

\section{Predicting Mechanical Property Fields}
\label{sec:method}

To predict volumetric mechanical properties across 3D representation, \acronym\ first aggregates volumetric features for the input geometry (\S\ref{sec:aggregating}), which are then processed by a trained feed-forward transformer model (\S\ref{sec:3dencdoer}) that learns in the latent space of MatVAE (\S\ref{sec:mat_latent_space}). See \Cref{fig:pipeline}.

\subsection{Aggregating Features}
\label{sec:aggregating}

Our method accepts any 3D representation that can be voxelized and rendered from multiple views. Following recent works~\cite{wang2023autorecon, Dutt_2024_CVPR, xiang2025structured3dlatentsscalable}, we compute rich DINOv2~\cite{oquab2024dinov2learningrobustvisual} image features across 3D views and lift them to 3D by projecting each voxel center into every view using the camera parameters to retrieve the corresponding image features. The retreived image features are then averaged to obtain a feature for every voxel.
A critical difference with these prior works is that we also voxelize and process the interior of the objects and not just their surface, which allows us to learn and predict material properties \emph{inside} the objects (See \S\ref{sec:implementation} for voxelization schemes and see \S\ref{app:training} for details on voxelization for training).
Let's denote all active voxel center positions in a 3D grid of size $N^3$ as $\{\mathbf{p}_i\}_{i=1}^L$ where $L$ denotes the number of voxels, $\mathbf p_i \in \mathbb{R}^3$ denotes the voxel center, and \(\Pi_j:\mathbb{R}^3\to[-1,1]^2\) the camera projection for view $j \in J$ where $J$ is the set of rendered views. Let the DINOv2 patch-token map be \(T_j\in\mathbb{R}^{1024\times n\times n}\) which is bilinearly sampled to get a feature map $\mathcal{F}_j:[-1,1]^2\to\mathbb{R}^{1024}$. Then for each voxel $i \in \{1,2,\ldots, L\}$, we obtain a feature $\mathbf f_i$:
\begin{equation}
    \mathbf{f}_i = \operatorname{Average}(\mathcal{C}_i = \left\{\mathcal{F}_j\!\big(\Pi_j(\mathbf{p}_i)\big) \;\big|\; j\in J\right\}) \in \mathbb{R}^{1024}
\end{equation}
% We perform a tomographic reconstruction of features using backprojection. Let $\pi_j^{-1}(\mathbf{u}, d)$ be the function that maps a pixel at coordinate $\mathbf{u}$ in rendered view $j$ with depth $d$ back to its 3D world coordinate. \todo{(why depth? you have camera.) Using the depth map $\mathcal{D}_j$ rendered for each view, we can cast a ray from each pixel $\mathbf{u}$ with its pre-computed DINOv2 feature, $\mathcal{F}_j(\mathbf{u})$, back into the voxel grid.} The set of all features that backproject into the volume of a given voxel $\mathbf{p}_i$ can be denoted as $\mathcal{C}_i$,
% \begin{equation}
% \mathcal{C}_i = \{ \mathcal{F}_j(\mathbf{u}) \mid \text{voxel}(\pi_j^{-1}(\mathbf{u}, \mathcal{D}_j(\mathbf{u}))) = \mathbf{p}_i, \forall j, \mathbf{u} \},
% \end{equation}
% where $\text{voxel}(\cdot)$ identifies the voxel corresponding to a 3D point. The final feature vector $\mathbf{f}_i$ for the voxel is then the element-wise average of this set, $\mathbf{f}_i = \text{Average}(\mathcal{C}_i)$, giving us one image feature per voxel $\{\mathbf{f}_i\}_{i=1}^L$
This propagates multi-view information to the voxels in the interior of the object, encoding useful information that our model learns to process to predict internal material composition.
% \masha{\todo{reference a fig.}} \dave{presumably you can think of this the other way around, we are backprojecting the dino features into 3D space, effectively doing tomographic reconstruction. Is there a reason we don't do this ? Or maybe these things are equivalent, in which case the back projection explanation sounds more technically meaty.}

\subsection{Geometry Transformer}
\label{sec:3dencdoer}

% \masha{Please find a better name for this component. "Geometry Transformer" does not communicte well what this part is doing.}
The main component of \acronym\ is a Transformer $\mathbf{F}$
% (\masha{How about $\mathcal{F}$ instead, for "Frankenstein"?})
that maps voxelized image features to our trained material latent representation. The backbone of our model follows TRELLIS~\cite{xiang2025structured3dlatentsscalable} encoder/decoder, and the backbone layers of our model are initialized with TRELLIS weights. The encoder processes a variable-length set of active voxels, represented by their positions and features $\mathbf{X} = \{(\mathbf{p}_i, \mathbf{f}_i)\}_{i=1}^L$. To make this data suitable for a Transformer, we first serialize the voxel features into a sequence and then inject spatial awareness by adding sinusoidal positional encodings derived from each voxel's 3D coordinates. Similar to TRELLIS and state-of-the-art 3D Transformers, we adopt a 3D shifted window attention mechanism~\cite{liu2021swin, yang2025swin3d}.
% \masha{You have not mentioned deviating from the setup proposed by Trellis. Have you made changes? If this part closely follows Trellis, you should say so at the start of this section. If you have made changes, please make clear which.} \donglai{Does this encoder agrees with the ``encoder'' in the transformer sense? Is it a transformer encoder without autoregressive decoder? Consider making this clear to the audience, too?}
Contrary to TRELLIS~\cite{xiang2025structured3dlatentsscalable}, to handle assets of various sizes, we define a maximum sequence length of $L_N$. For assets with fewer voxels $L\leq L_N$, we use the complete set. However, for larger assets where $L>L_N$, we use a stochastic sampling strategy, selecting a random subset of $L_N$ voxels at the start of each training epoch. This dynamic resampling ensures the model is exposed to different parts of the asset over epochs and have a larger number of "effective" max voxels.
% \masha{Same here; is this yours or follows Trellis?}

% To train this model, we require a dataset of voxelized 3D geometries with ground truth material 
% triplets per voxel (See~\Cref{sec:data} for our data generation pipeline). For each training shape, we aggregate voxel features (\Cref{sec:aggregating}),
% pass in the aggregated features to $\mathbf{F}$, and then pass the structured latent output to the MatVAE, which is run $L$ times \textit{i.e.} once per voxel, which gives us material triplets $(E, \nu, \rho)$ \textit{for each voxel}. When training the Geometry Transformer, we keep the MatVAE and DINOv2~\cite{oquab2024dinov2learningrobustvisual} frozen.
For each training asset, we first define $\mathcal{S}$ as the set of voxel indices to be processed in the current iteration. The corresponding sequence of image features $\mathbf{X}_{\mathcal{S}}$ obtained from voxel features (\Cref{sec:aggregating}), is passed to $\mathbf{F}$. The resulting latent representation is then fed into the frozen decoder of pre-trained MatVAE to predict material properties. The MatVAE is run $L$ times \textit{i.e.} once per voxel, which gives us material triplets $(E, \nu, \rho)$ \textit{for each voxel}. We train this transformer with the mean squared error between the predicted materials and the ground truth materials, averaged over all voxels in the set $\mathcal{S}$,
% \masha{Oh, so the loss function is computed after the decoding? I thought you would compute MSE in the latent space. }
\begin{equation}
    \mathcal{L}_{\mathbf{F}} = \frac{1}{|\mathcal{S}|} \sum_{i \in \mathcal{S}} \| \mu_\theta(\mathbf{F}(\mathbf{X}_{\mathcal{S}})_i) - ((E_i, \nu_i, \rho_i)^N)^{\mathsf T} \|_2^2,
\label{eq:assignment_loss}
\end{equation}
where $\mu_\theta(\cdot)$ denotes the output of the frozen MatVAE decoder, $((E_i, \nu_i, \rho_i)^N)^{\mathsf T}$ is the ground truth material vector for voxel $i$, and $\mathbf{F}(\mathbf{X}_{\mathcal{S}})_i$ is the latent representation for voxel $i$.
% \masha{Define $\mathcal{S}$, or remind us: "shape $\mathcal{S}$"; -- though I think you defined it as $\mathcal{O}$, so make sure $\mathcal{S}$ is clear.}

To transfer voxel materials back to the original representation (\textit{i.e.} splat means, tets for FEM simulation, quadrature points for simulation, etc.), we use nearest neighbour interpolation as outlined in~\S\ref{app:interpolation}. The per-voxel latents are passed into the decoder model of MatVAE (\Cref{sec:matvae}), which yields per-voxel material triplets, as shown in~\Cref{fig:pipeline}.
% \todo{To transfer voxel materials back to the original representation, we...The per-voxel latents are passed into the decoder model of MatVAE (\Cref{sec:matvae}), which yields per-voxel material triplets, as shown in~\Cref{fig:pipeline}.}
% \subsection{Training \acronym}
% \label{sec:training}
% To train \acronym, we first train MatVAE in isolation and then train on the Geometry Transformer, keeping other components frozen.
% \masha{As it is, this paragraph seems redundant; you basically already said all this above. What about finetuning Trellis? You need to be a lot more transparent about building your work on that.}
% Next, for training the Geometry Transformer, we start with some voxelized input geometry, aggregate image features (\Cref{sec:aggregating}) keeping DINOv2 frozen, then pass these aggregated image features through the Geometry Transformer (\Cref{sec:3dencdoer}), and finally pass the latents for every voxel through the decoder of MatVAE (\Cref{sec:matvae}) keeping the decoder frozen. We describe the process to create the training data for the Geometry Transformer in~\Cref{sec:data}.
\section{Training Data Generation}
\label{sec:data}

\subsection {Material Triplets Dataset (MTD)}
\label{sec:mtd}

% \masha{The dataset construction can go in the dataset section; just make a subsection for the MatVAE.}
To train MatVAE (\Cref{sec:mat_latent_space}), we collect Material Triplet Dataset(MTD), containing 100,562
triplets $(E,\nu,\rho)$ for real-world materials. We first collect a dataset of measured material properties
from multiple online databases~\cite{matweb, wikipedia_density, wikipedia_poissons_ratio, wikipedia_youngs_modulus, engineeringtoolbox_materials, cambridge_materials_databook}, containing values obtained experimentally, typically
with valid \emph{ranges} for all three properties for all materials. We sample numeric triplets from
each material, with the number of samples proportional to the range size. Finally, we filter out duplicates resulting
from overlapping ranges for some materials.

%We then sample triplets from within these ranges proportional to the size of the range for each mechanical property. Finally, we filter this dataset to remove duplicate triplets resulting from overlapping property ranges for different materials.

%We first train MatVAE on the Material Triplet Dataset (MTD), a dataset of 100,562 plausible material triplet values $(E,\nu,\rho)$. To create MTD, we first collect a dataset of mechanical property ranges for many different materials from multiple online databases~\cite{matweb, wikipedia_density, wikipedia_poissons_ratio, wikipedia_youngs_modulus, engineeringtoolbox_materials, cambridge_materials_databook}, which include values obtained experimentally. We then sample triplets from within these ranges proportional to the size of the range for each mechanical property. Finally, we filter this dataset to remove duplicate triplets resulting from overlapping property ranges for different materials.
% \mashaedit{Move MTD description here; keep it brief.}

\subsection{Geometry with Volumetric Materials (GVM) Dataset}

\ifdefined\iclr
    \begin{wrapfigure}[]{r}{0.45\textwidth}
  \includegraphics[width=0.45\textwidth]{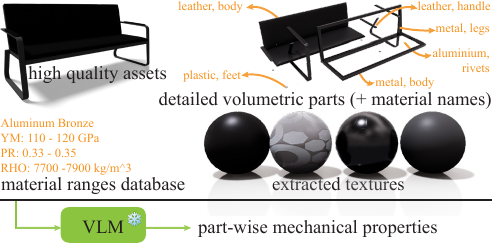}
  \caption{\textbf{Training Data} annotation leverages accurate 3D data labels together with a VLM.}
  \label{fig:datacreation}
\end{wrapfigure}

\else
    \begin{figure}
    \centering
    \includegraphics{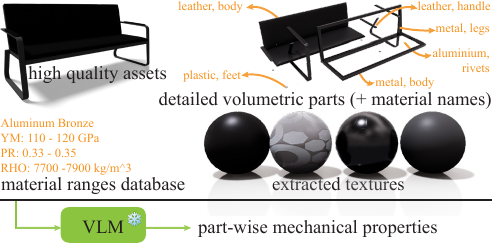}
    \caption{\textbf{Training Data Creation.} We use a VLM coupled with the information inside multiple exisitng datasets to reliably create training data.}
    \label{fig:datacreation}
\end{figure}
\fi

% \mashaedit{subsection Geometry Volumetric Materials Dataset}
% \todo{Come up with a good name for this dataset.}
To train Geometry Transformer (\Cref{sec:method}), we develop an automatic annotation pipeline to overcome the
limited availability of detailed volumetric material datasets (\S\ref{sec:rw:inferring}). Like
prior works~\cite{lin2025phys4dgenphysicscompliant4dgeneration, cao2025sophylearninggeneratesimulationready, le2025pixiefastgeneralizablesupervised}, we leverage a pre-trained VLM, but overcome
its limitations by introducing \emph{additional sources of knowledge} present in our 3D dataset and the MTD(\Cref{sec:mtd}).
We collect high-quality 3D meshes from~\cite{nvidia_omniverse_commercial_assets, nvidia_omniverse_residential_assets, nvidia_omniverse_simready, nvidia_omniverse_vegetation_assets}, containing 1624 part-segmentated 3D models, with a total of 8089 parts, and treat each part as having isotropic material. Each part contains an English material name and its own realistic PBR texture, which can be used
as additional cues to the VLM. For each part in each object, we pass the following information to the VLM: rendering of the full object, detail rendering of the part's visual material mapped onto a sphere (showing visual aspects that tend to correlate with material composition), the material names, and the ranges of three closest real-world materials in the MTD (\Cref{sec:mtd}) based on the material names (See Fig.\ref{fig:datacreation}, detailed prompt in Fig.~\ref{fig:userprompt1}). The vision-language model then outputs material triplets for each part, and we map to all volumetric voxels within it, resulting in a total of 37M voxels annotated with \mattriplet. By guiding VLM with real-world material values and extra clues, we avoid inaccuracies and implausible material values. See additional details in \S\ref{sec:implementation}, \S\ref{app:datasetdetails}.

\section{Experiments and Results}
\label{sec:exp}

We evaluate \acronym\ end-to-end, showing diverse realistic simulations in \S\ref{sec:eval:qual}. Quantitative results are presented in \S\ref{sec:estimation}, with MatVAE evaluated separately in \S\ref{sec:reconstructtriplet}. See \includegraphics[height=0.6\baselineskip]{assets/youtube-brands_svg-tex.pdf} video and \S\ref{app:more_results} for many additional results, \S\ref{app:concurrent} for extra comparisons with concurrent work and \S\ref{sec:ablations} for ablations.
% \dave{do we compare on per-object properties, mass, simulation results or all ?}
% We choose a few meshes and 3D Gaussian Splats to run simulations using mechanical property fields estimated by \acronym. We demonstrate these simulations in~\Cref{fig:simulations,fig:plinko,fig:ficus_container}. We simulate the meshes with a FEM simulator and use sparse Simplicits~\cite{10.1145/3658184, KaolinLibrary} for our large scale splat + mesh simulations involving 18 and 65 splats.

\subsection{Implementation Details}
\label{sec:implementation}

\textbf{Voxelization:}
For voxelizing 3D Gaussian splats~\cite{kerbl3Dgaussians}, we present \textit{a new voxelizer}, that works in three phases:
(1) 3D Gaussians are voxelized over a 3D grid as solid ellipsoids defined by the 99th percentile iso-surface, (2) this set of voxels is rendered from several dozen viewpoints sampled over a sphere to form depth maps, (3) these depthmaps are used to carve away empty space around the exterior of the object, but leaving unseen \emph{interior} voxels to form a solid approximation of the object. We then sample this solid at jittered sample points on a regular grid. We employ octrees as acceleration structures and GPU implementations for both phases. Our test objects can be voxelized in 31 ms (see ~\Cref{tab:wall-clock}).
To voxelize meshes and SDFs we use standard methods (see~\S\ref{app:training}).
% \todo{Rishit: to transfer materials from voxels back to splats or meshes, we...}

\textbf{Data and Training:} For material annotation, we experimentally choose Qwen 2.5 VL-72B VLM~\cite{bai2023qwenvlversatilevisionlanguagemodel, bai2025qwen25vltechnicalreport}. We partition our MTD and GVM datasets (\S\ref{sec:data}) into 80-10-10 train, validation, and test sets. See \S\ref{app:datasetdetails} for data details. For rendering we use Omniverse~\cite{nvidia_omniverse_replicator} and Blender~\cite{blender}, and for DINOv2 we use an optimized implementation~\cite{nvidia_nv_dinov2_2025}. During training and testing, we set the maximum number of non-empty voxels per object $L_N=32768$ (sampled stochastically, \Cref{sec:3dencdoer}), and sparse data structures for efficiency. See \S\ref{app:implementation} for more details. All experiments were performed on a machine with four 80GB A100 GPUs, where training took about $12$ hours for MatVAE and 5 days for the Transformer. 

\textbf{Simulation:} We used FEM simulator for meshes and sparse Simplicits~\cite{10.1145/3658184, KaolinLibrary} for our large-scale simulations combining splats and meshes. Details in \S\ref{app:sec:simulators}.

\begin{figure*}[tb]
    \begin{minipage}[t]{\textwidth}
    \includegraphics[width=\textwidth]{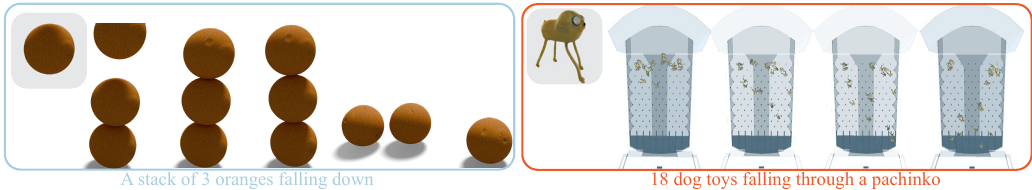}
    \end{minipage}
    \caption{\textbf{Simulation-ready physics materials of \acronym}\ enable realistic simulations for meshes and splats.}
    \label{fig:qualitative}\label{fig:simulations}
\end{figure*}

\subsection{End-to-End Qualitative Evaluation}\label{sec:eval:qual}

We qualitatively evaluate \acronym\ by using it to annotate volumetric mechanical fields for several meshes and 3D Gaussian Splats, and running
physics simulation with these exact spatially varying \mattriplet\ values, resulting in realistic simulations without any hand-tweaks (\Cref{fig:simulations}, \Cref{fig:app:qualitative}, \video{0:36}). 
We also show that our approach can work across more representations, including meshes, 3D Gaussian Splats, SDF, and NeRFs (~\Cref{fig:reps}, with additional results in \S\ref{app:sec:more_material_results}.
% Quantitative evaluation is detailed below, and ablations are included in the supplementary (\S\ref{sec:ablations}).

\subsection{Quantitative Evaluation}\label{sec:estimation}\label{sec:wall-clock-time}\label{sec:mass}

\textbf{Datasets and Metrics:} The 10\% hold-out test set of GVM (\S\ref{sec:data}) consists of 166 high-quality 3D objects with per-voxel mechanical properties for a total of 4.9 million point annotations, significantly larger than previous works, e.g.\ 31 points across 11 objects ~\cite{zhai2024physicalpropertyunderstandinglanguageembedded}.
% \dave{can you give numbers}
%Our test set also includes detailed, higher-quality objects with dense annotations.
We contribute this as \emph{a new benchmark} and use it for evaluation against baselines.
We measure standard metrics, Average Log Displacement Error (ALDE), Average Displacement Error (ADE), Average Log Relative Error (ALRE), and Average Relative Error (ARE) for each mechanical property, further detailed in~\S\ref{app:metricsfield}. We provide additional intuition for interpreting these errors through targeted simulations in ~\S\ref{app:interpret}.

\begin{wrapfigure}[]{l}{0.45\textwidth}
  \captionof{table}{\textbf{Wall-clock} comparisons and breakdown.}
    \resizebox{0.45\textwidth}{!}{%
  \begin{tabular}{lr}
  \toprule
  \rowcolor{nvidiagreen!15} Method & Time (s)\\
  \midrule NeRF2Physics & 1454.55 {\scriptsize{($\pm$1118)}}\\
  PUGS & 1058.33 {\scriptsize{($\pm$6.94)}}\\
  Pixie & 201.63 {\scriptsize{($\pm$27.74)}}\\
  Phys4DGen$^*$ & \underline{51.65} {\scriptsize{($\pm$4.07)}}\\
  Ours & \textbf{3.59} {\scriptsize{($\pm$1.36)}}\\
  \midrule  \quad Rendering & 2.11 {\scriptsize{($\pm$0.0540)}}\\
  \quad Voxelization & 0.03 {\scriptsize{($\pm$0.0016)}}\\
  \quad DINO-v2 Computation & 0.86 {\scriptsize{($\pm$0.0020)}}\\
  \quad DINO-v2 Reconstruction & 0.58 {\scriptsize{($\pm$0.0053)}}\\
  \quad Geometry Transformer & 0.0082 {\scriptsize{($\pm$0.0063)}}\\
  \quad MatVAE & 0.00032 {\scriptsize{($\pm$0.00026)}}\\
  \bottomrule
  \end{tabular}
  % \vspace{-2em}
  }
  \label{tab:wall-clock}
\end{wrapfigure}

\textbf{Baselines:} 
We compare against prior art NeRF2Physics~\cite{zhai2024physicalpropertyunderstandinglanguageembedded} and PUGS~\cite{shuai2025pugszeroshotphysicalunderstanding}, where we look up 
material properties at the voxel locations (with proper scaling) using 
their optimized representations. Note that these techniques do not output Poisson's ratio. Phys4DGen ~\cite{lin2025phys4dgenphysicscompliant4dgeneration} is an important
baseline, aggregating VLM prediction directly,
but does not provide code. We used our best effort
to replicate their method and used prompts provided by the authors,
designating this implementation Phys4DGen$^\star$. More baseline details in \S\ref{app:impl:baselines}. We also include early comparisons against
concurrent (and as yet unpublished) Pixie~\cite{le2025pixiefastgeneralizablesupervised}, with additional explorations in \S\ref{app:concurrent}.

\begin{figure*}[t!]
\vspace{-1em}
\begin{minipage}[t]{\textwidth}
 \includegraphics[width=\textwidth]{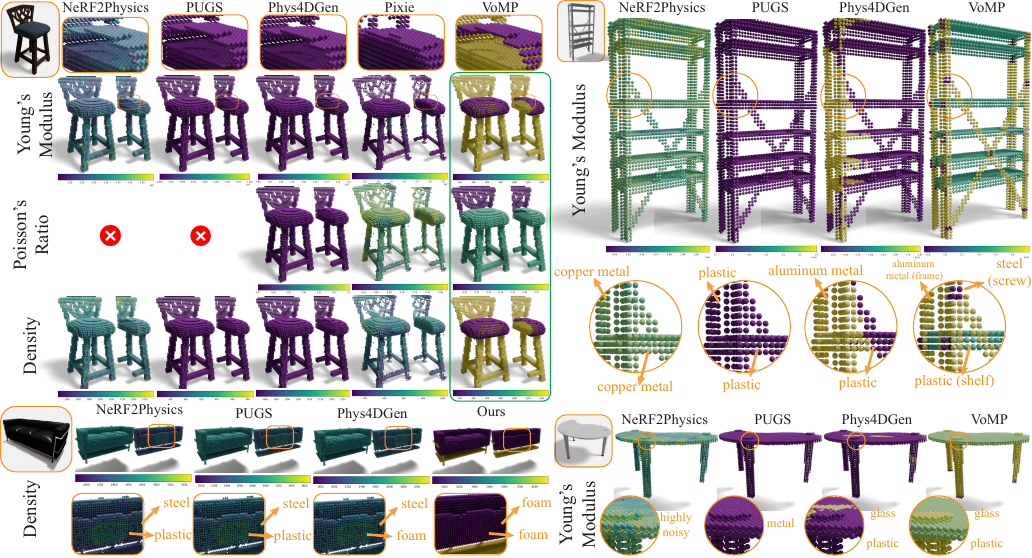}
    \subcaption{\textbf{Qualitative Comparison:} We show that qualitiative \acronym\ tends to provide less noisy volumetric? values compared to the baselines. We show the color coded fields and slice planes through the fields.}
    \label{fig:comparefields}
\end{minipage}
\begin{minipage}[t]{\textwidth}
\newcommand{\voxelproprow}{
\textbf{0.3765} {\textbf{\scriptsize\textcolor{gray}{($\pm$0.39)}}} & 
\textbf{0.0421} {\textbf{\scriptsize\textcolor{gray}{($\pm$0.05)}}} & 
\textbf{0.0250} {\textbf{\scriptsize\textcolor{gray}{($\pm$0.01)}}} & 
\textbf{0.0837} {\textbf{\scriptsize\textcolor{gray}{($\pm$0.03)}}} & 
\textbf{113.3807} {\textbf{\scriptsize\textcolor{gray}{($\pm$301.90)}}} & 
\textbf{0.0908} {\textbf{\scriptsize\textcolor{gray}{($\pm$0.14)}}}
}
\resizebox{\textwidth}{!}{%
\begin{tabular}{lrrrrrr}
\toprule
\rowcolor{nvidiagreen!15}Method & \multicolumn{2}{c}{Young's Modulus Pa ($E$)} & \multicolumn{2}{c}{Poisson's Ratio ($\nu$)} & \multicolumn{2}{c}{Density $\frac{kg}{m^3}$ ($\rho$)} \\
\cmidrule(r){2-3} \cmidrule(r){4-5} \cmidrule(r){6-7}
\rowcolor{nvidiagreen!15}& ALDE ($\downarrow$) & ALRE ($\downarrow$) & ADE ($\downarrow$) & ARE ($\downarrow$) & ADE ($\downarrow$) & ARE ($\downarrow$) \\
\midrule
NeRF2Physics & \underline{2.8000} {\scriptsize{($\pm$1.05)}} & \underline{0.1346} {\scriptsize{($\pm$0.05)}} & - & - & \underline{1432.0343} {\scriptsize{($\pm$964.88)}} & \underline{1.0365} {\scriptsize{($\pm$0.63)}} \\
PUGS & 3.3942 {\scriptsize{($\pm$1.72)}} & 0.1688 {\scriptsize{($\pm$0.10)}} & - & - & 3568.2150 {\scriptsize{($\pm$2839.13)}} & 3.2429 {\scriptsize{($\pm$3.56)}} \\
Phys4DGen$^\star$ & 4.8967 {\scriptsize{($\pm$3.17)}} & 0.2227 {\scriptsize{($\pm$0.14)}} & \underline{0.0407} {\scriptsize{($\pm$0.04)}} & \underline{0.1467} {\scriptsize{($\pm$0.18)}} & 1865.5673 {\scriptsize{($\pm$2176.90)}} & 1.4394 {\scriptsize{($\pm$2.35)}} \\
\midrule
Ours & 
\textbf{0.3793} {\textbf{\scriptsize\textcolor{gray}{($\pm$0.29)}}} & 
\textbf{0.0409} {\textbf{\scriptsize\textcolor{gray}{($\pm$0.04)}}} & 
\textbf{0.0241} {\textbf{\scriptsize\textcolor{gray}{($\pm$0.01)}}} & 
\textbf{0.0818} {\textbf{\scriptsize\textcolor{gray}{($\pm$0.03)}}} & 
\textbf{142.6949} {\textbf{\scriptsize\textcolor{gray}{($\pm$166.90)}}} & 
\textbf{0.0921} {\textbf{\scriptsize\textcolor{gray}{($\pm$0.07)}}} \\
\bottomrule
\end{tabular}
}
\subcaption{\textbf{Mechanical Property Estimates} of our method significantly outperform the baselines on all metrics. Per-voxel error rate is first computed per object, then averaged across all objects in the test set to avoid weighing some objects more. Global voxel-level normalization yields similar results, see Supplement \Cref{tab:material-properties}.}
\label{tab:material-properties-object}
\end{minipage}
\begin{minipage}[t]{0.48\textwidth}
\resizebox{0.98\textwidth}{!}{
\begin{tabular}{lrrrr}
\toprule
\rowcolor{nvidiagreen!15}Method & ALDE ($\downarrow$) & ADE ($\downarrow$) & ARE ($\downarrow$) & MnRE \textbf{($\uparrow$)}\\
\midrule
NeRF2Physics & 0.736 & 12.725 & 1.040 & 0.564\\
PUGS & \underline{0.661} & \underline{9.461} & \textbf{0.767} & \textbf{0.576}\\
Phys4DGen$^\star$ & 0.664 & 9.961 & 0.825 & 0.566\\
\midrule Ours & \textbf{0.631} & \textbf{8.433} & \underline{0.887} & \textbf{0.576}\\
\bottomrule
\end{tabular}
}
    \subcaption{\textbf{Mass Estimate:} We show the errors for estimating mass of objects on the ABO-500~\cite{Collins_2022_CVPR} dataset, the only existing benchmark, approximating the accuracy of our $\rho$ estimates.}
\label{tab:mass}
\end{minipage}
\hfill
\begin{minipage}[t]{0.48\textwidth}
\resizebox{0.98\textwidth}{!}{
\begin{tabular}{lrrr}
\toprule
\rowcolor{nvidiagreen!15}
Method & $\log(E) (\downarrow)$ & $\nu (\downarrow)$ & $\rho (\downarrow)$ \\
\midrule
NeRF2Physics 
& \underline{1.62} {\textbf{\scriptsize\textcolor{gray}{($\pm$4.96)}}} 
& -- 
& 19.75 {\textbf{\scriptsize\textcolor{gray}{($\pm$46.60)}}} \\
PUGS         
& 1.87 {\textbf{\scriptsize\textcolor{gray}{($\pm$4.50)}}} 
& -- 
& \underline{13.24} {\textbf{\scriptsize\textcolor{gray}{($\pm$12.63)}}} \\
Phys4DGen$^\star$    
& 1.77 {\textbf{\scriptsize\textcolor{gray}{($\pm$8.53)}}} 
& \underline{0.85} {\textbf{\scriptsize\textcolor{gray}{($\pm$3.01)}}} 
& 39.49 {\textbf{\scriptsize\textcolor{gray}{($\pm$35.47)}}} \\
Pixie        
& 11.90 {\textbf{\scriptsize\textcolor{gray}{($\pm$17.41)}}} 
& 3.46 {\textbf{\scriptsize\textcolor{gray}{($\pm$4.42)}}} 
& 46.58 {\textbf{\scriptsize\textcolor{gray}{($\pm$36.35)}}} \\
\midrule Ours         
&\textbf{0.29} {\textbf{\scriptsize\textcolor{gray}{($\pm$1.23)}}} 
&\textbf{0.00} {\textbf{\scriptsize\textcolor{gray}{($\pm$0.00)}}} 
&\textbf{11.75} {\textbf{\scriptsize\textcolor{gray}{($\pm$4.02)}}} \\
\bottomrule
\end{tabular}
}
\subcaption{\textbf{Material Validity:} We report mean values and relative errors (in \%) with the closest physically measured material range in MTD (\S\ref{sec:mtd}).}
\label{tab:validity}
\end{minipage}
\caption{\textbf{Quantitative Results and Comparisons:} We compare our method against prior art NeRF2Physics~\cite{zhai2024physicalpropertyunderstandinglanguageembedded}, PUGS~\cite{shuai2025pugszeroshotphysicalunderstanding} and Phys4DGen~\cite{lin2025phys4dgenphysicscompliant4dgeneration}, and include limited early results comparing with concurrent method Pixie~\cite{le2025pixiefastgeneralizablesupervised}.}
\end{figure*}

 \textbf{Estimating Mechanical Properties:} 
 % \todo{Nerf2Physics - noisy, Phys4DGen - inconsistent part labeling, interior not isolated well.}
 Quantitative evaluation of material estimates \mattriplet\ of our
 method against prior art on our new detailed benchmark shows a \emph{dramatic quality boost across all properties and metrics} (\Cref{tab:material-properties-object}). According to our explorations (\S\ref{app:interpret}), ALRE under $0.05$ for $E$ and ARE under $0.15$ for other properties result in similar simulations, suggesting that our materials will lead to more faithful simulations than competitors when using an accurate simulator.
 Qualitatively (\Cref{fig:comparefields}), we observe that this performance difference may be due to baselines occasionally mislabeling segments (e.g.\ by Phys4DGen), due to noisy estimates (e.g.\ NeRF2Physics and PUGS), and less accurate values in the objects' interior due to the baselines' design. 
%From over 1900 FEM simulations (see details in , we find that realtive errors of upto $15\%$ in $\rho,\nu$ or upto $5\%$ in $\log E$ lead to similar-looking deformable simulations.
% \todo{provide a little intuition about ALRE within 5 percent ok. }

We are unable to make the vegetation subset of our dataset publicly available. Thus, we compute the mechanical property estimations on the public version of the dataset in~\Cref{tab:public-material-properties-object,tab:public-material-properties-voxel}. We find that our results averaged over the public dataset are highly similar to the full dataset.

\textbf{Run-Time:} To show approximate speed difference, we report average material estimate speeds across 100 runs on objects with an average of 53.9K Gaussians for our method and the baselines in \Cref{tab:wall-clock}. To ensure fair compute between CPU and GPU heavy methods, we ran this experiment on a machine with only one A100 GPU and 64 CPUs. While we do not provide timing breakdown of the other methods, this result suggests a speed
up of 5-100x achieved by our method, which is not surprising given that it is the only feed-forward model among previous work. Concurrent Pixie, which is also feed-forward, involves a heavier pre-processing step, including per-object optimization, affecting its end-to-end time. 
In the timing breakdown of our method, rendering and pre-processing take the most time, and could be further optimized.

\textbf{Mass Estimation:} Following NeRF2Physics~\cite{zhai2024physicalpropertyunderstandinglanguageembedded} and PUGS~\cite{shuai2025pugszeroshotphysicalunderstanding}, we also evaluate our dataset on the ABO-500~\cite{Collins_2022_CVPR} object mass estimation 
benchmark, following the evaluation protocol of PUGS. We run our model to estimate density $\rho$ for upto 32678 voxels per object, then average these values and multiply by the known object volume to obtain mass. While this is only an imperfect proxy for measuring the accuracy
of volumetric density $\rho$, it is a benchmark used by prior works, and we include it for completeness. We achieve
better or on-par performance across most metrics (\Cref{tab:mass}), with qualitative results in ~\S\ref{app:sec:more_mass_results}.

\textbf{Validity:} To gauge how well different methods are at predicting realistic materials, such as those measured in the real world, which is our goal, we leverage our MTD dataset of real materials. First, we run all methods on GVM test set objects, and for each test voxel compute relative errors to the nearest possible material range from MTD (error is $0$ for estimates within an existing material range). These errors are averaged across all the voxels and reported in ~\Cref{tab:validity}. We observe that our method, on average, outputs much more realistic materials, as it was explicitly designed to do so.

\begin{figure*}[t!]
    \centering
    % \vspace{-1em}
    \begin{minipage}[t]{\textwidth}
    \newcommand{\matvaemetricsrow}{
0.0034 & 0.0426 & 0.0330 & 
0.0054 & 0.0036 & 0.0036 & 
0.0131 & 0.4439 & 0.0411
}

\resizebox{\textwidth}{!}{%
\begin{tabular}{rrrrrrrrrr}
\toprule
\rowcolor{nvidiagreen!15}$\log(E)$ ($\downarrow$) & $\nu$ ($\downarrow$) & $\rho$ ($\downarrow$) & $\log(E/\rho)$ ($\downarrow$) & $\log(G)$ ($\downarrow$) & $\log(K)$ ($\downarrow$) & L.S. ($\downarrow$) & E.A. ($\downarrow$) & Bray–Curtis ($\downarrow$) \\
\midrule
\matvaemetricsrow\\
\bottomrule
\end{tabular}
}

    \subcaption{\textbf{MatVAE shows excellent reconstruction errors} on the MTD test set across all metrics.}
    \label{tab:matvaemetrics}
    \end{minipage}
    \begin{minipage}[t]{0.43\textwidth}
     \vspace*{0pt}
        \centering
        \includegraphics[width=\textwidth]{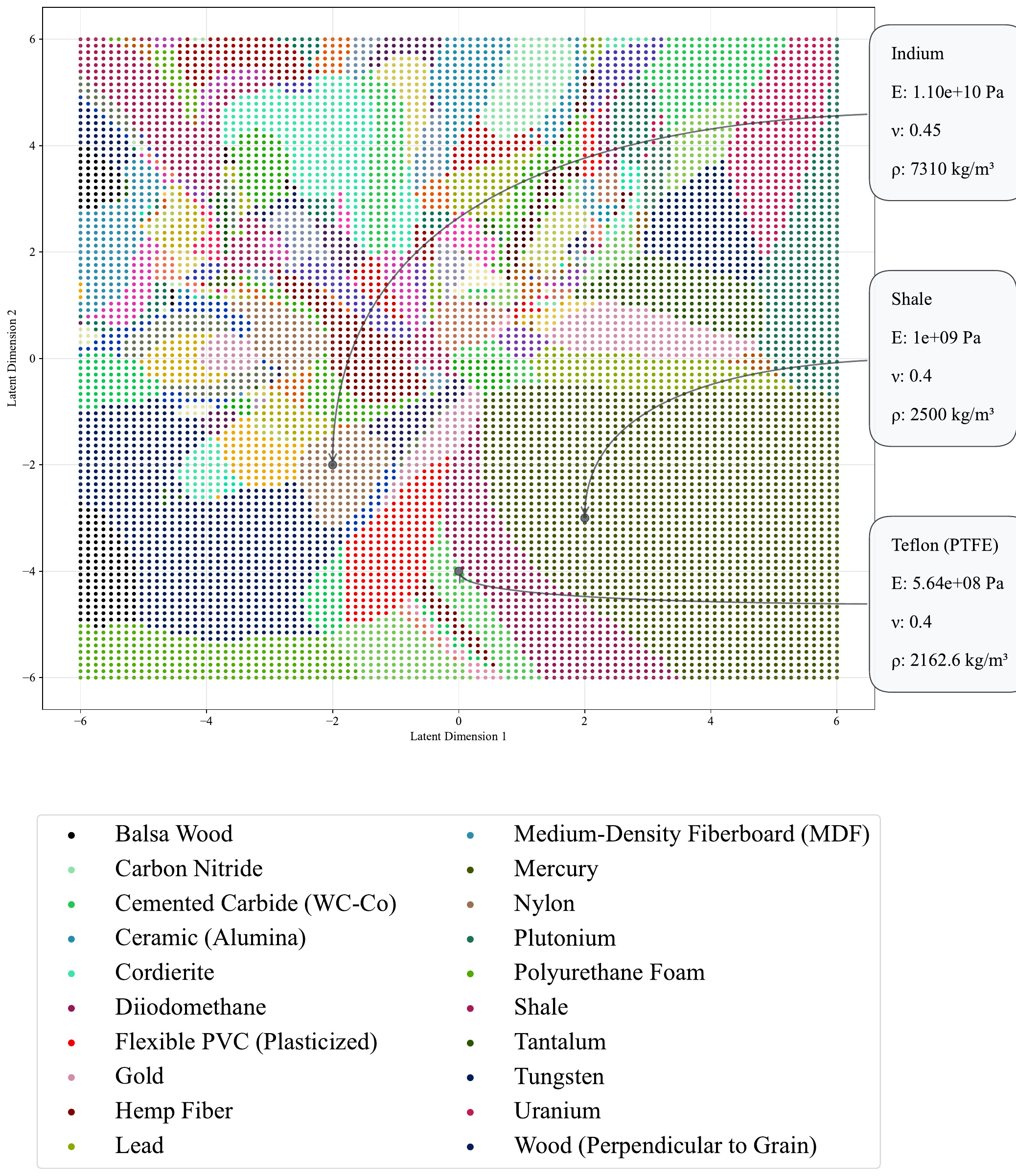}
        \subcaption{\textbf{Decoding latent samples} leads to plausible {\mattriplet} values within real-world materials.}
        \label{fig:genfrommatvae}
    \end{minipage}%
    \hfill
    \begin{minipage}[t]{0.56\textwidth}
     \vspace*{0pt}
        \centering
        \includegraphics[width=\textwidth]{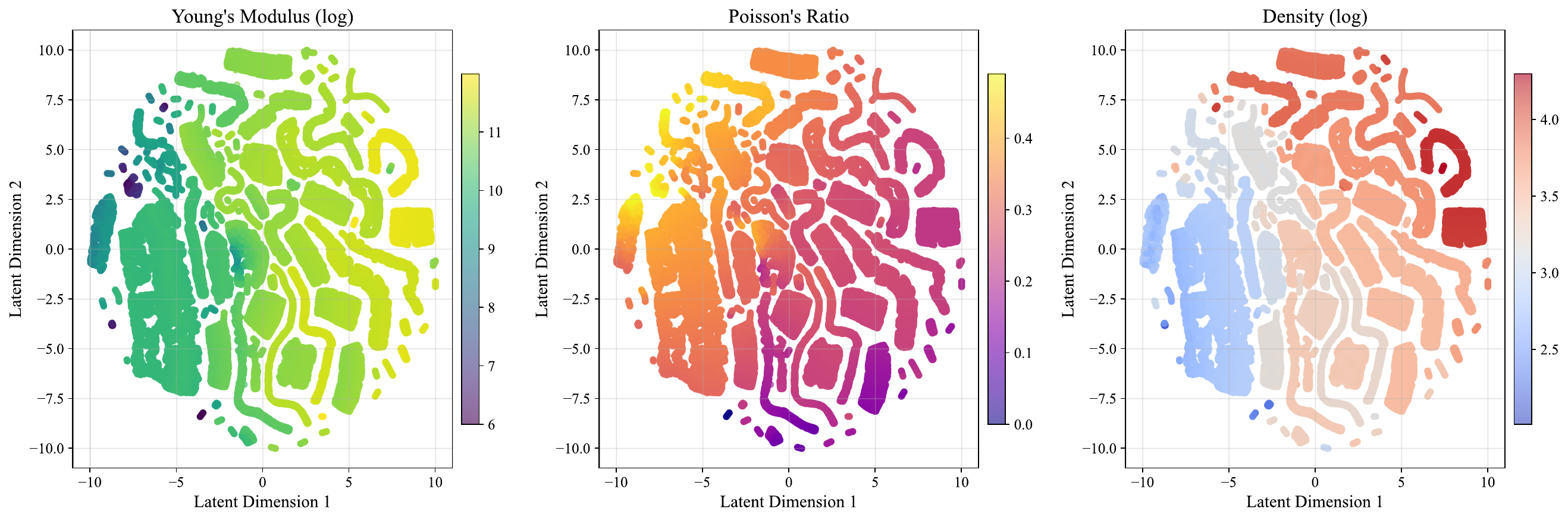} %{assets/real_materials_latent_space.pdf}
        \subcaption{\textbf{Encoding real materials} results in smoothly varying $E$, $\nu$, $\rho$ values throughout the 2D latent space.}
        \label{fig:encodematvae}

        \vspace{0.2cm}  % Add some vertical space
        {\scriptsize 
        \setlength{\tabcolsep}{2pt}
        \begin{tabular}{l>{\columncolor{nvidiagreen_light}}c|ccc|c}
        \rowcolor{nvidiagreen}
        & \textbf{Start Value} & \multicolumn{3}{c|}{\textbf{MatVAE Interpolation}}  & \textbf{End Value}\\
        & Aerographite &
        \makecell[c]{Carbon Fiber \\ (high strength)} &
        \makecell[c]{Carbon Fiber \\ (high modulus)} &
        Carbon Nitride &
        Diamond\\
        $E$ & $0.001$ GPa & $235.0$ GPa & $400.0$ GPa &
        $822.0$ GPa & $1220.0$ GPa\\
        $\nu$ & $0.25$ & $0.25$ & $0.25$ &
        $0.28$ & $0.20$\\
        $\rho$ & $0.2$ kg/m$^3$ & $1775$ kg/m$^3$ &
        $1800$ kg/m$^3$ & $3200$ kg/m$^3$ &
        $3500$ kg/m$^3$\\
        \Xhline{2pt}
        \rowcolor{nvidiagreen}
        & \textbf{Start Value} & \multicolumn{3}{c|}{\textbf{Naive Interpolation}} & \textbf{End Value} \\
        & Aerographite &
        {\color{red}Invalid (\xmark)} &
        {\color{red}Invalid (\xmark)} &
        {\color{red}Invalid (\xmark)} &
        Diamond\\
        $E$ & $0.001$ GPa & $305.0$ GPa & $610.0$ GPa &
        $915.0$ GPa & $1220.0$ GPa\\
        $\nu$ & $0.25$ & $0.24$ & $0.23$ &
        $0.21$ & $0.20$\\
        $\rho$ & $0.2$ kg/m$^3$ & $875$ kg/m$^3$ &
        $1750$ kg/m$^3$ & $2625$ kg/m$^3$ &
        $3500$ kg/m$^3$\\
        \end{tabular}%
        }
        \subcaption{\textbf{Interpolating in latent space} results in valid intermediate materials, unlike naive {\mattriplet} interpolation.}\label{fig:interp_pair1}
    \end{minipage}
    \caption{\textbf{Material Latent Space} learned by MatVAE (\S\ref{sec:mat_latent_space}) ensures faithful (a), valid (b), smoothly varying (c), and interpolatable (d) materials. "Invalid" values (c) fall outside all material ranges in MTD (\Cref{sec:mtd}).}\label{fig:matvaelatentspace}
\end{figure*}

\subsection{Reconstructing and Generating Materials with MatVAE}
\label{sec:reconstructtriplet}

Given no prior works exploring a latent space of material triplets \mattriplet, we evaluate MatVAE on the MTD test
set (\S\ref{sec:implementation}), achieving low reconstruction errors in ~\Cref{tab:matvaemetrics} (See \Cref{app:metricsfield}, \Cref{app:interpret}) for metrics). Further, in \Cref{fig:matvaelatentspace} we show the desirable properties of this learned latent space. In (a), samples throughout the 2D latent space map to real-world material ranges in MTD. In (b), we show that \mattriplet values of real materials encoded to the latent space vary smoothly. Further, 
the latent space ensures valid interpolation points between materials (c), facilitating valid assignment from predicted voxel materials back to the original geometry. We include detailed ablations of MatVAE design (\S\ref{sec:ablations}), and additional latent space explorations (\S\ref{app:sec:matvae_results}).  

\begin{table}[tb]
\centering
\caption{\textbf{Mechanical Property Estimates} of our method on the \textit{publicly released dataset} are very close to the full dataset. Per-voxel error rate is first computed per object, then averaged across all objects in the test set to avoid weighing some objects more. Global voxel-level normalization yields similar results, see Supplement \Cref{tab:public-material-properties-voxel}.}
\label{tab:public-material-properties-object}
\resizebox{\textwidth}{!}{%
\begin{tabular}{lrrrrrr}
\toprule
\rowcolor{nvidiagreen!15}Method & \multicolumn{2}{c}{Young's Modulus Pa ($E$)} & \multicolumn{2}{c}{Poisson's Ratio ($\nu$)} & \multicolumn{2}{c}{Density $\frac{kg}{m^3}$ ($\rho$)} \\
\cmidrule(r){2-3} \cmidrule(r){4-5} \cmidrule(r){6-7}
\rowcolor{nvidiagreen!15}& ALDE ($\downarrow$) & ALRE ($\downarrow$) & ADE ($\downarrow$) & ARE ($\downarrow$) & ADE ($\downarrow$) & ARE ($\downarrow$) \\
\midrule
NeRF2Physics &
\underline{2.8000} {\scriptsize{($\pm$1.05)}} &
\underline{0.1346} {\scriptsize{($\pm$0.05)}} &
- & - &
\underline{1432.0343} {\scriptsize{($\pm$964.88)}} &
\underline{1.0365} {\scriptsize{($\pm$0.63)}} \\
PUGS &
3.3942 {\scriptsize{($\pm$1.72)}} &
0.1688 {\scriptsize{($\pm$0.10)}} &
- & - &
3568.2150 {\scriptsize{($\pm$2839.13)}} &
3.2429 {\scriptsize{($\pm$3.56)}} \\
Phys4DGen$^\star$ &
4.8967 {\scriptsize{($\pm$3.17)}} &
0.2227 {\scriptsize{($\pm$0.14)}} &
\underline{0.0407} {\scriptsize{($\pm$0.04)}} &
\underline{0.1467} {\scriptsize{($\pm$0.18)}} &
1865.5673 {\scriptsize{($\pm$2176.90)}} &
1.4394 {\scriptsize{($\pm$2.35)}} \\
\midrule
Ours &
\textbf{0.3794} {\textbf{\scriptsize\textcolor{gray}{($\pm$0.29)}}} &
\textbf{0.0409} {\textbf{\scriptsize\textcolor{gray}{($\pm$0.04)}}} &
\textbf{0.0241} {\textbf{\scriptsize\textcolor{gray}{($\pm$0.01)}}} &
\textbf{0.0818} {\textbf{\scriptsize\textcolor{gray}{($\pm$0.03)}}} &
\textbf{142.7017} {\textbf{\scriptsize\textcolor{gray}{($\pm$166.92)}}} &
\textbf{0.0921} {\textbf{\scriptsize\textcolor{gray}{($\pm$0.07)}}} \\
\bottomrule
\end{tabular}
}
\end{table}
\section{Discussion}\label{sec:discussion}\label{sec:conclusion}

We introduce a representation-agnostic method that maps any 3D asset (mesh, SDF, Gaussian splat, or voxel grid) to a volumetric field of physically valid mechanical properties $(E,\nu,\rho)$. We show that our method significantly outperforms prior art in accuracy and speed, lowering the barrier for integrating accurate physics into digital workflows across 3D representations, with potential impact across digital twins, robotics, and beyond.

While we show important advances over existing works, our method is not without limitations, which we hope will open exciting avenues of future research. Due to fixed-grid voxelization, our output resolution is limited, causing oversmoothing in highly heterogeneous regions, and may result in approximation errors when transferring results to more detailed input geometry. 
During annotation, we assume part-level materials are isotropic, which is not a true assumption for some common materials like wood.
Further, future work could extend our method to predict additional properties like yield strength, shear modulus and thermal expansion,
or to adapt true material properties output by our method to simulator-specific scales required for faster algorithms or implementations. We hope to support future directions in this area by releasing our material estimation benchmark, and trained models. 

\subsubsection*{Acknowledgments}

We thank Gilles Daviet for help in setting up some of the simulations. We thank Jean-Francois Lafleche for help with rendering. We thank Beau Perschall for help in using the datasets.

\newpage
\nocite{polyscope}
\nocite{playcanvas_supersplat_2025}
\nocite{brussee_brush_2025}
\bibliography{main}
\bibliographystyle{iclr2026_conference}

\newpage
\clearpage
\appendix

\begin{center}
    \begin{center}
        {\LARGE \bfseries Supplementary Material for \acronym: Predicting {\color{nvidiagreen}Vo}lumetric {\color{nvidiagreen}M}echanical {\color{nvidiagreen}P}roperty Fields}
    \end{center}
    \vspace{2.5em}
\end{center}

\section*{Supplementary Contents}
% \tableofcontents
\addcontentsline{toc}{section}{Supplementary Contents}
\begingroup
\setcounter{tocdepth}{2}
\startcontents[appendix]
\printcontents[appendix]{l}{1}{\setcounter{tocdepth}{2}}
\endgroup

\clearpage

\section{Additional Results}
\label{app:qual}\label{app:more_results}

\subsection{End-to-end Examples with Simulation}\label{app:sed:end2end}

\begin{figure}
    \centering
    \setlength{\tabcolsep}{0pt}
    \begin{tabular}{lll}
        \includegraphics[width=0.33\textwidth]{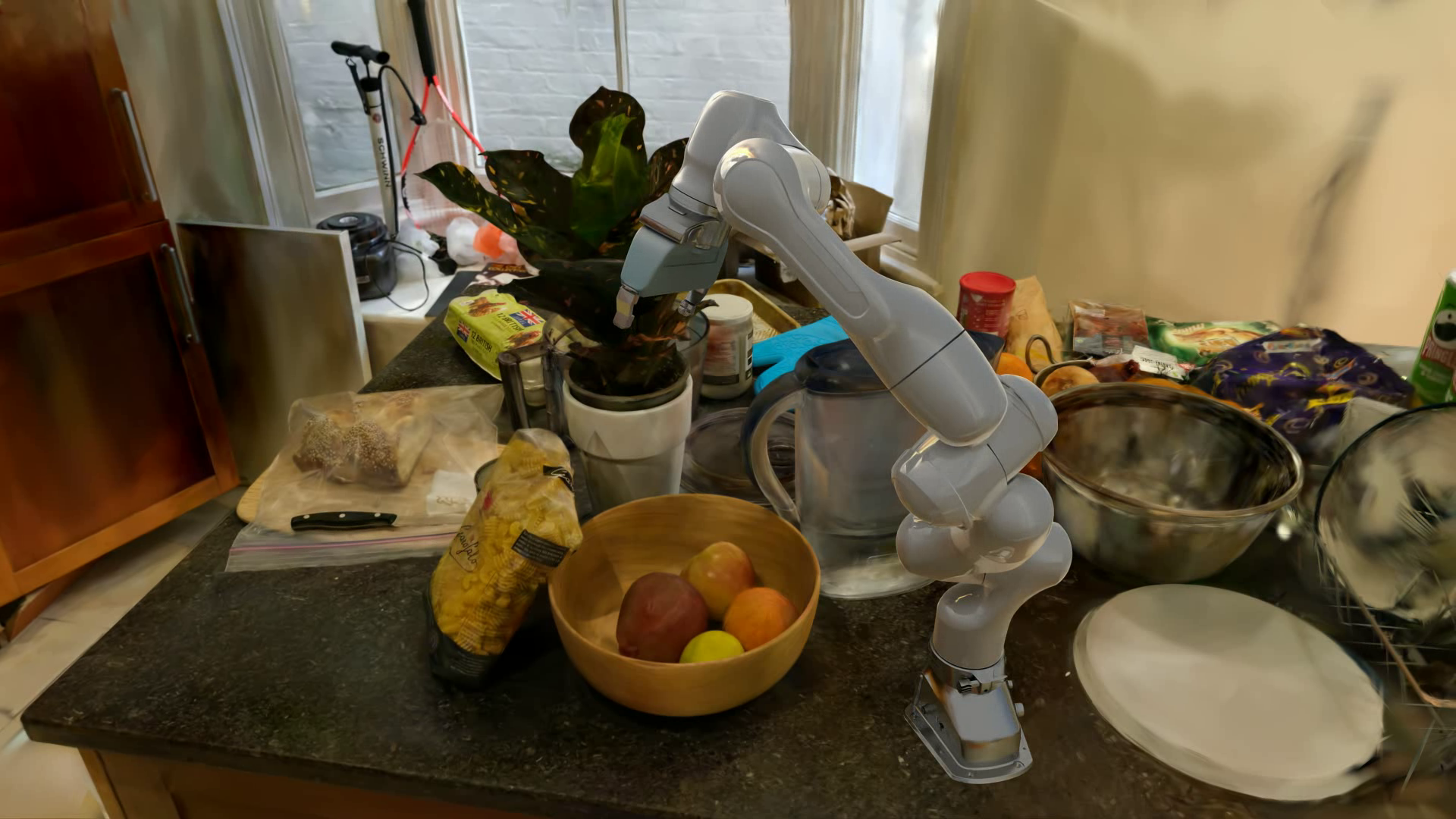} & \includegraphics[width=0.33\textwidth]{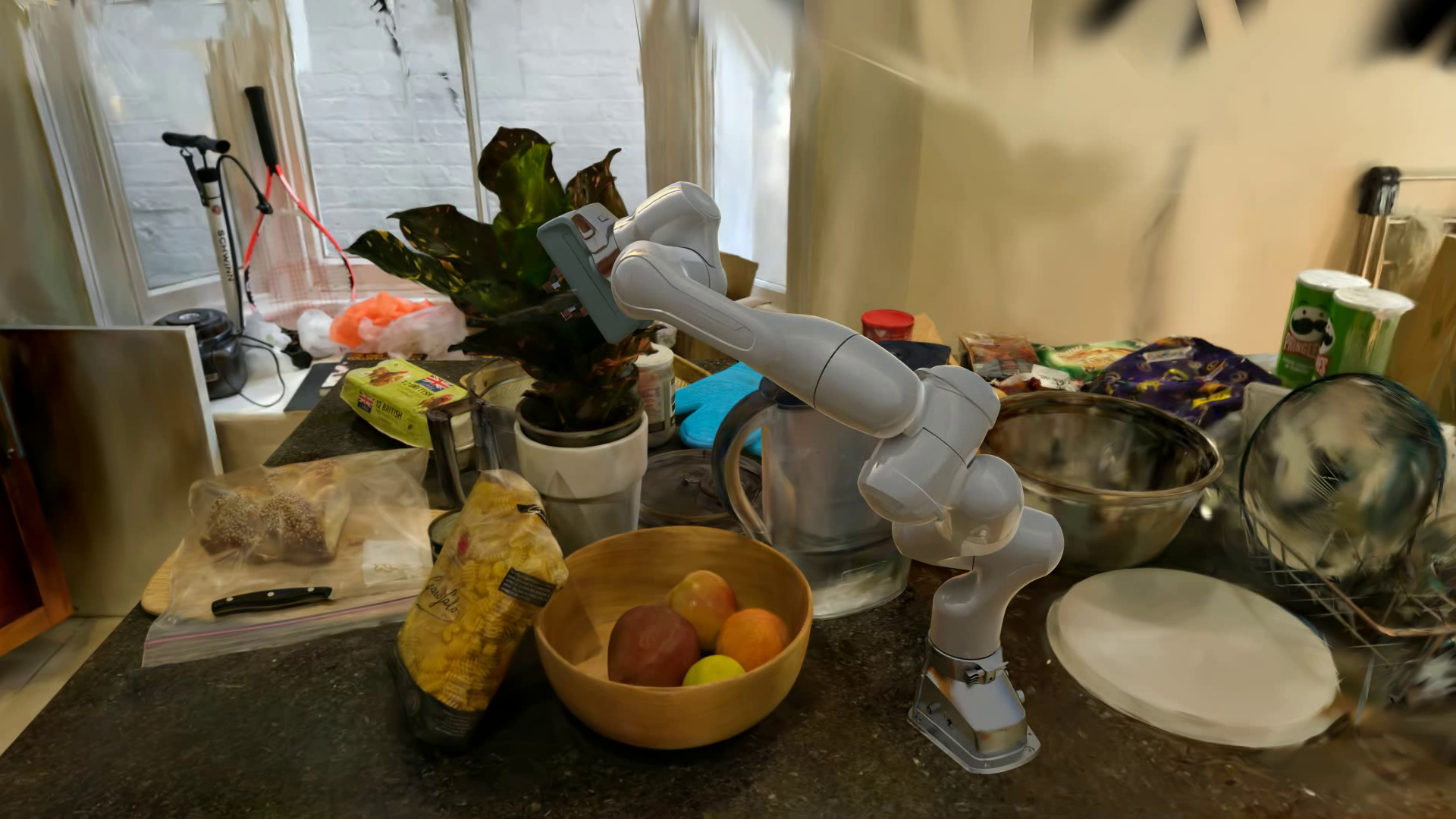} & \includegraphics[width=0.33\textwidth]{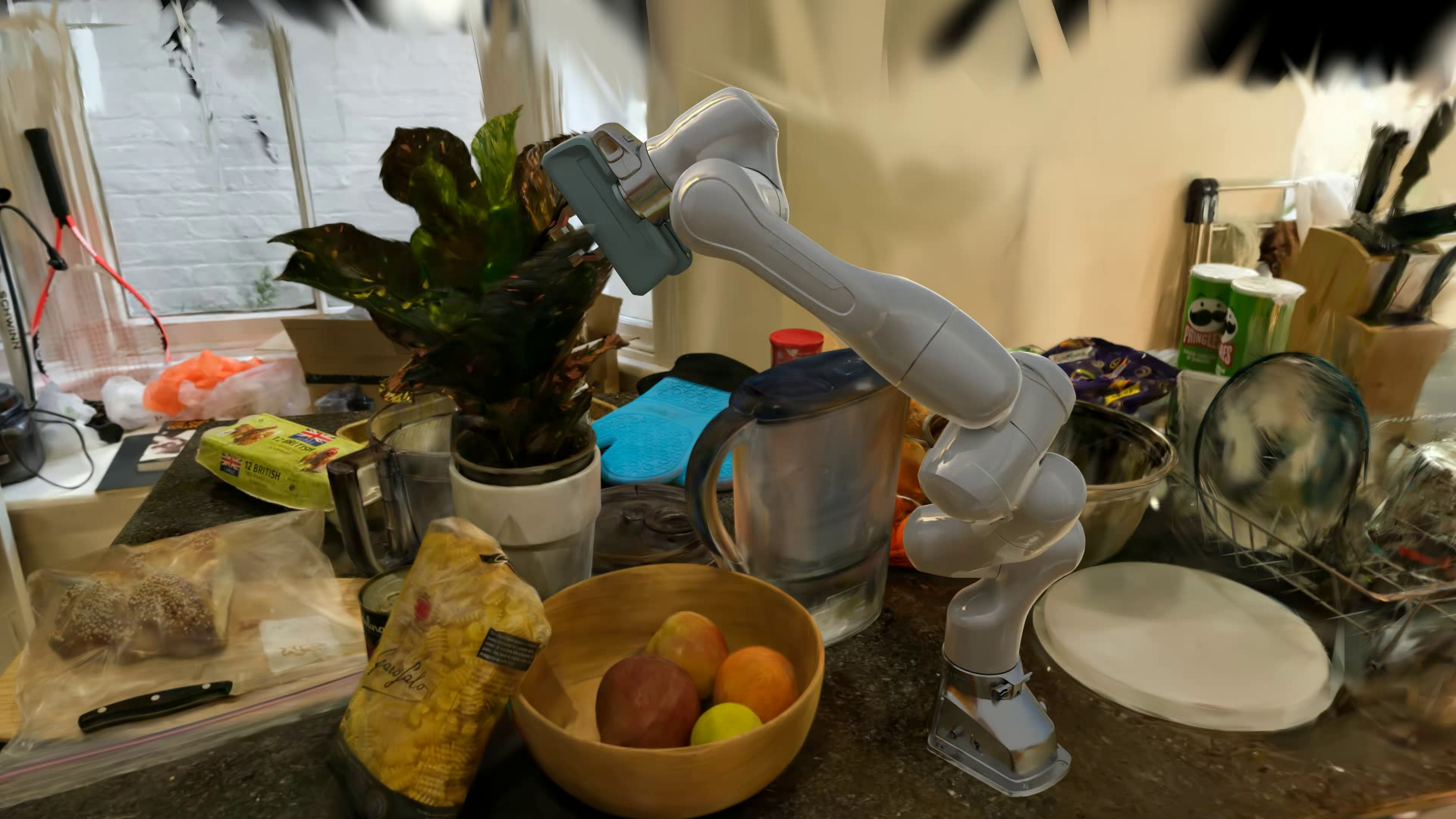}\\
     \multicolumn{3}{c}{\bigarrow{time}}\\
        \includegraphics[width=0.33\textwidth]{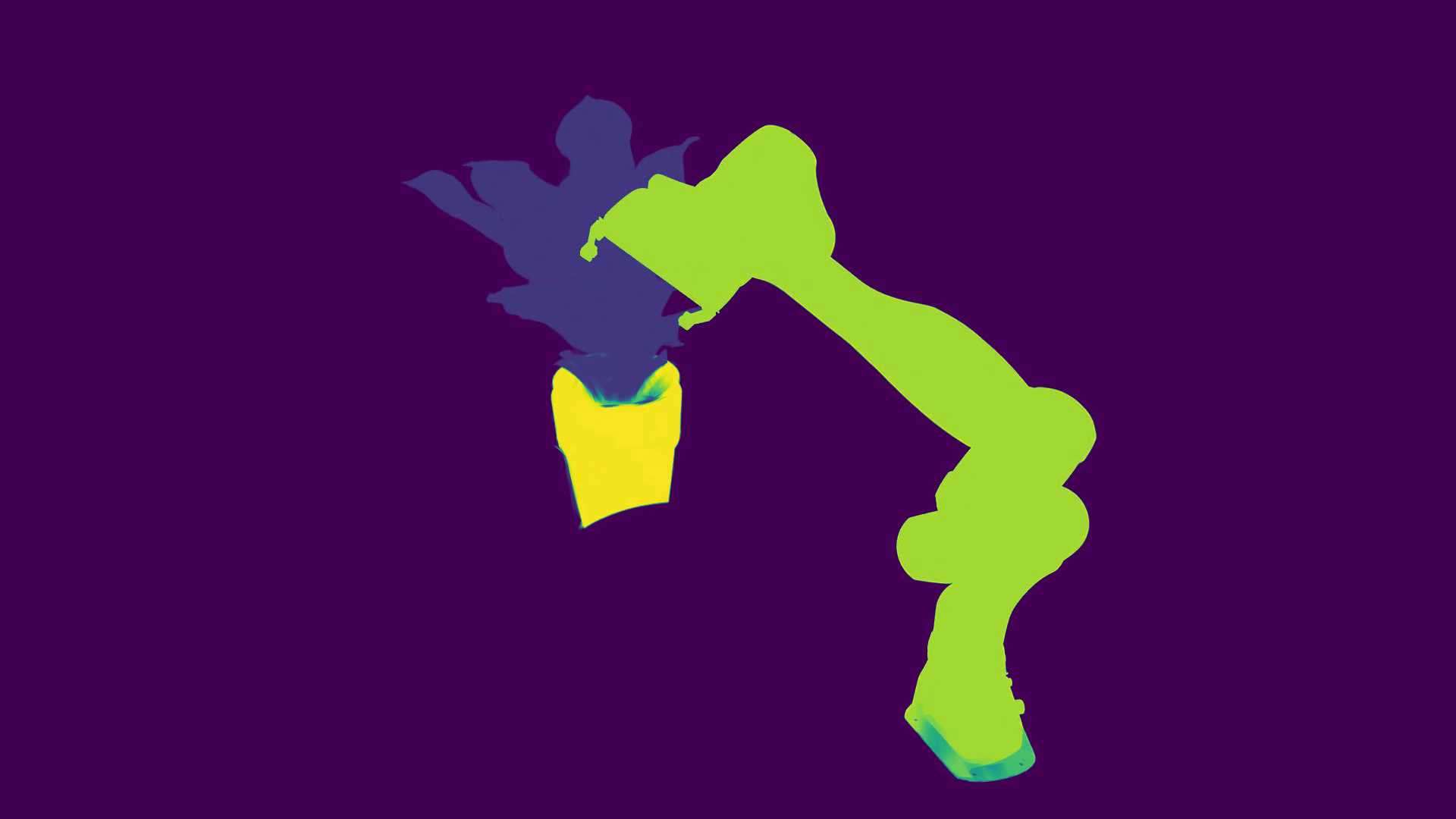} & \includegraphics[width=0.33\textwidth]{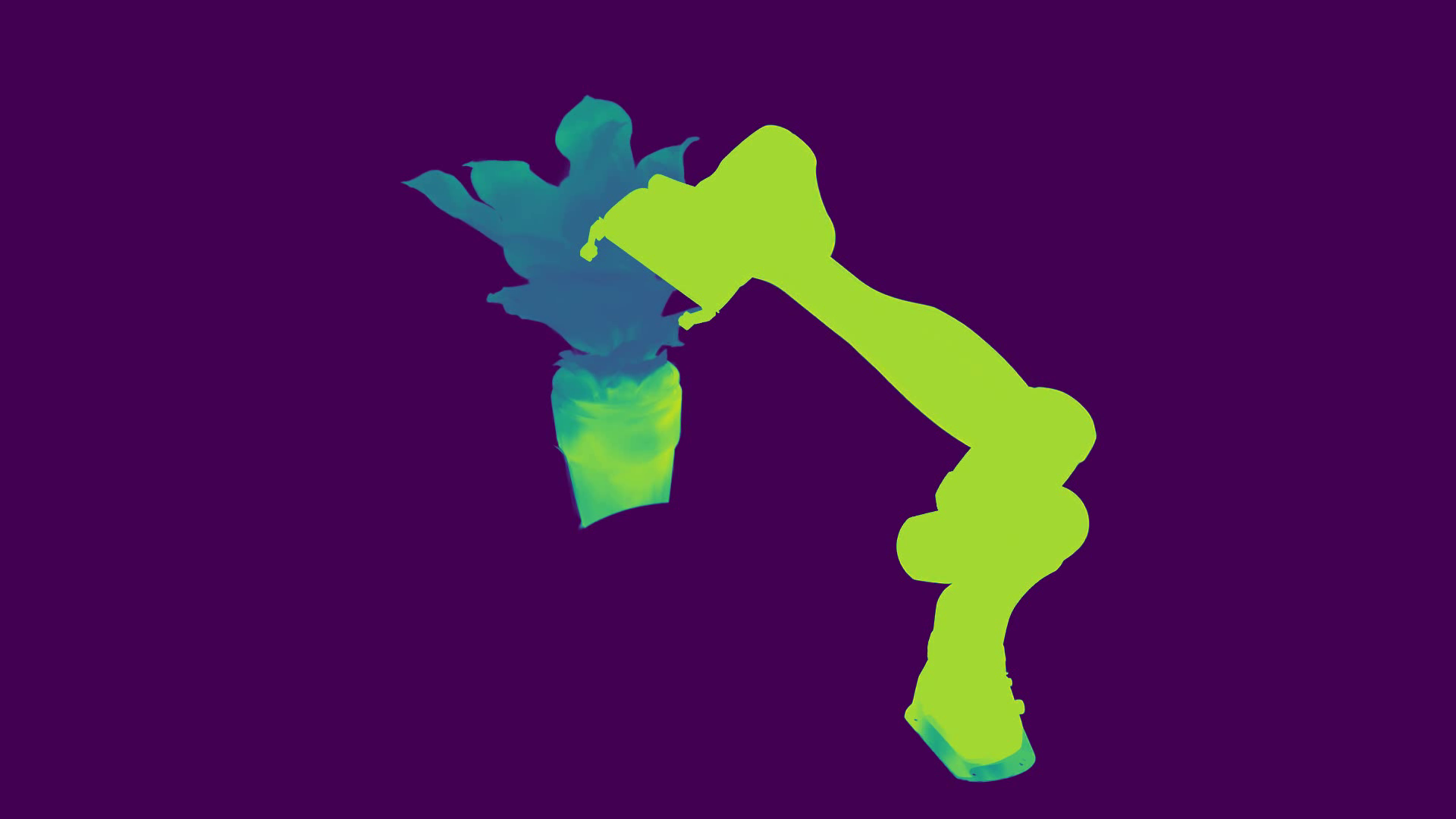} & \includegraphics[width=0.33\textwidth]{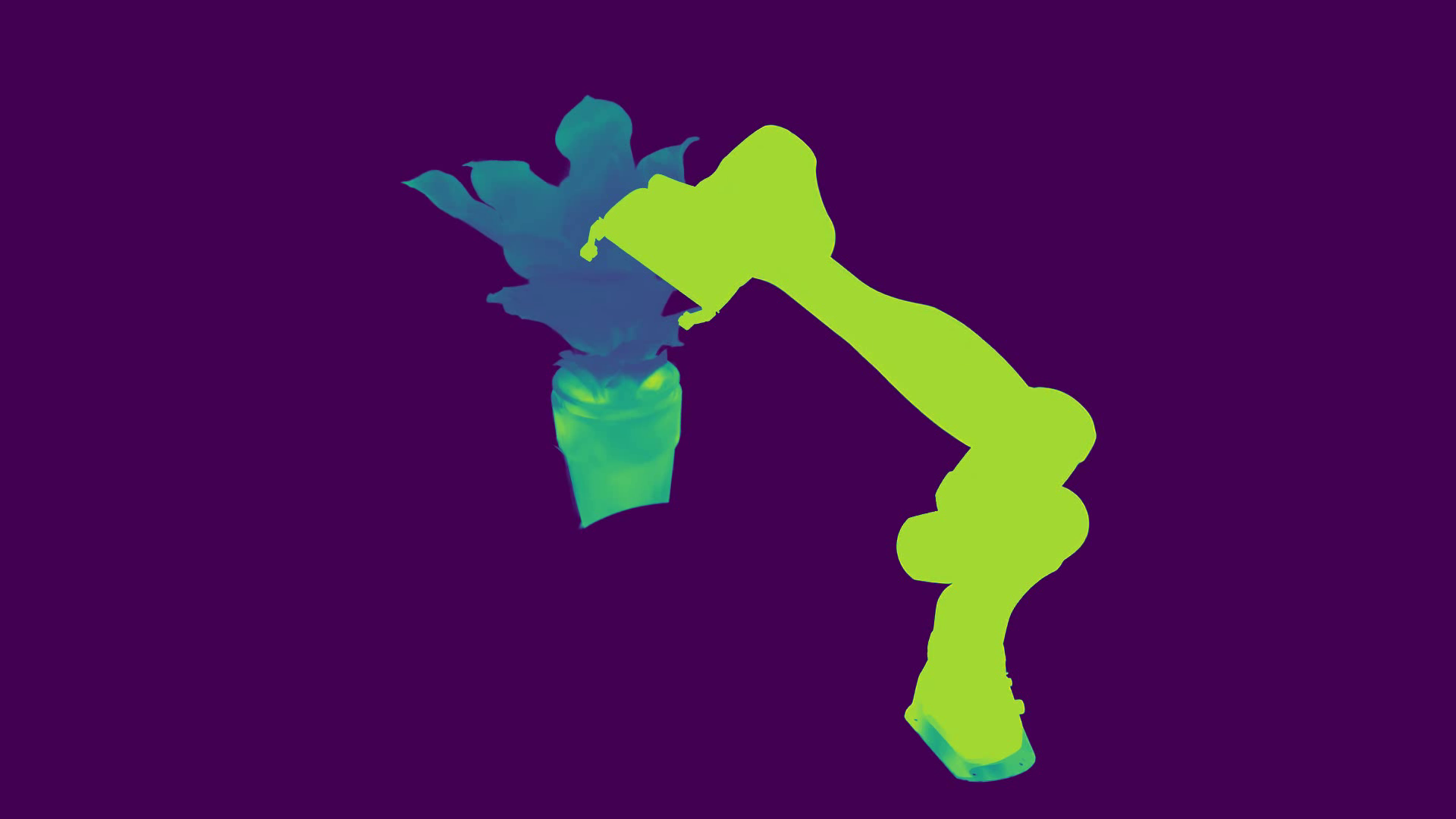}\\
        \hfill Young's Modulus & \hfill Density & \hfill Poisson's Ratio\\
        \includegraphics[width=0.162\textwidth]{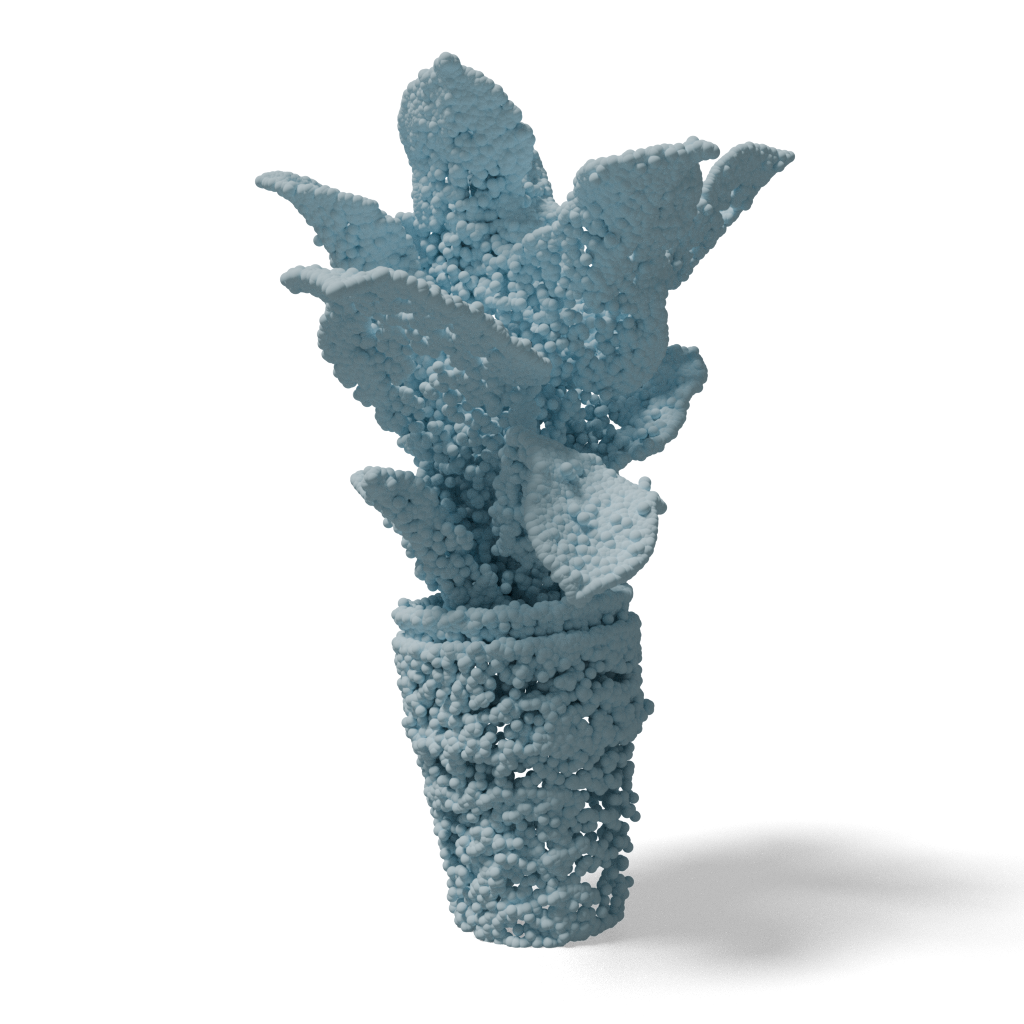}\includegraphics[width=0.162\textwidth]{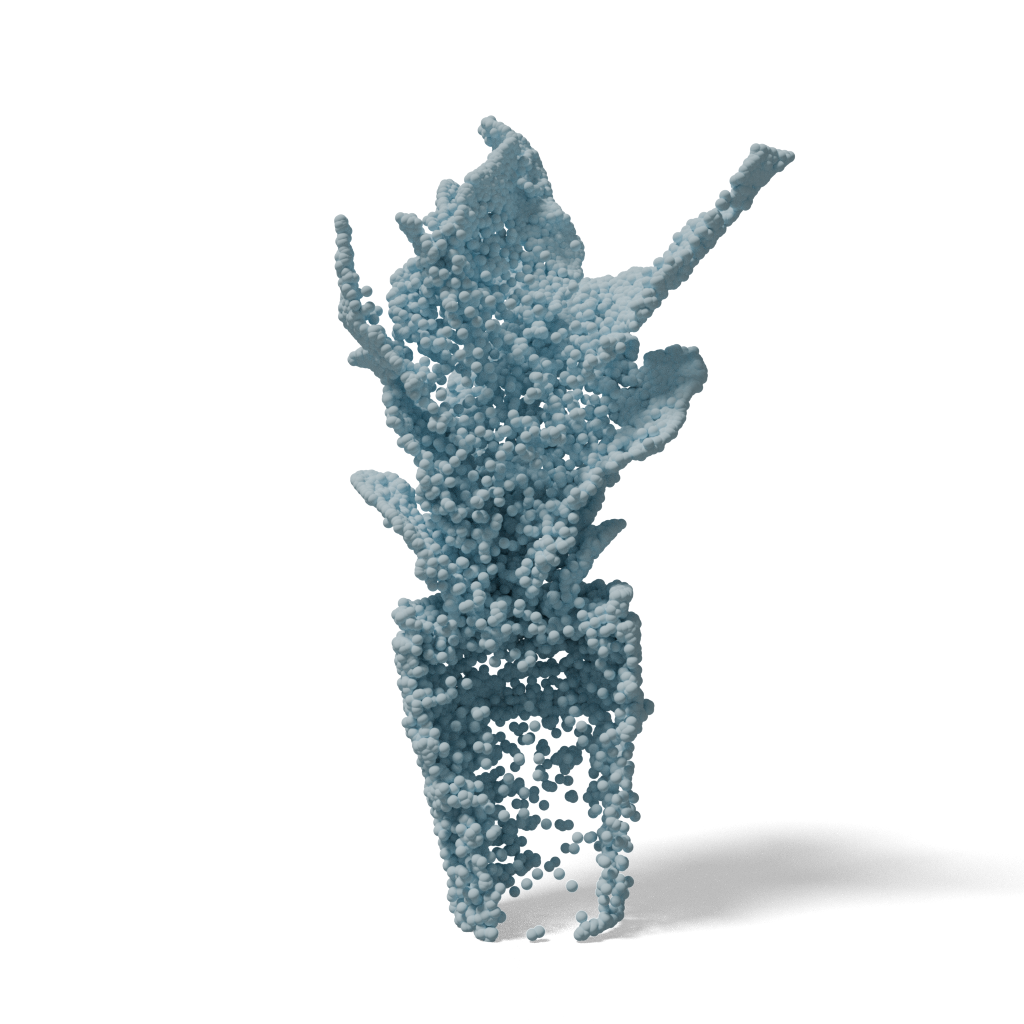} &         \includegraphics[width=0.162\textwidth]{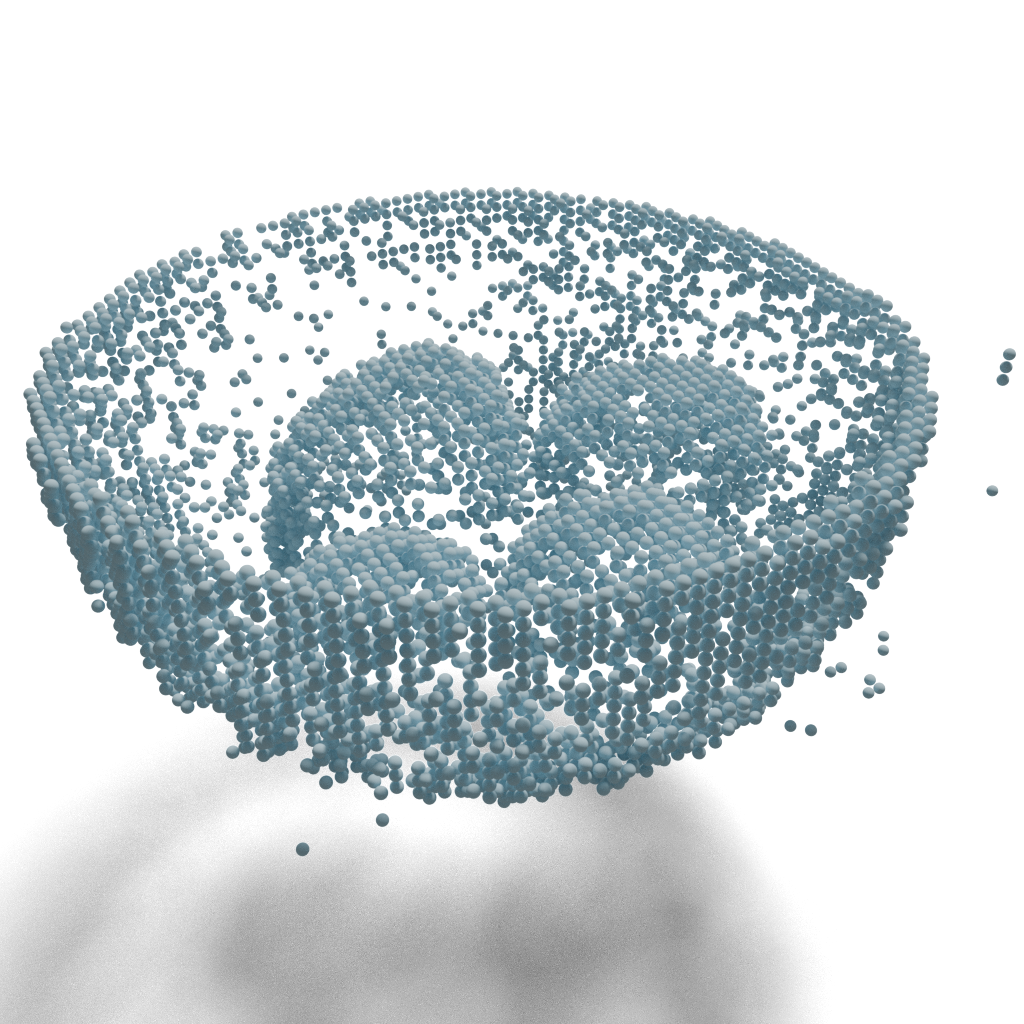}\includegraphics[width=0.162\textwidth]{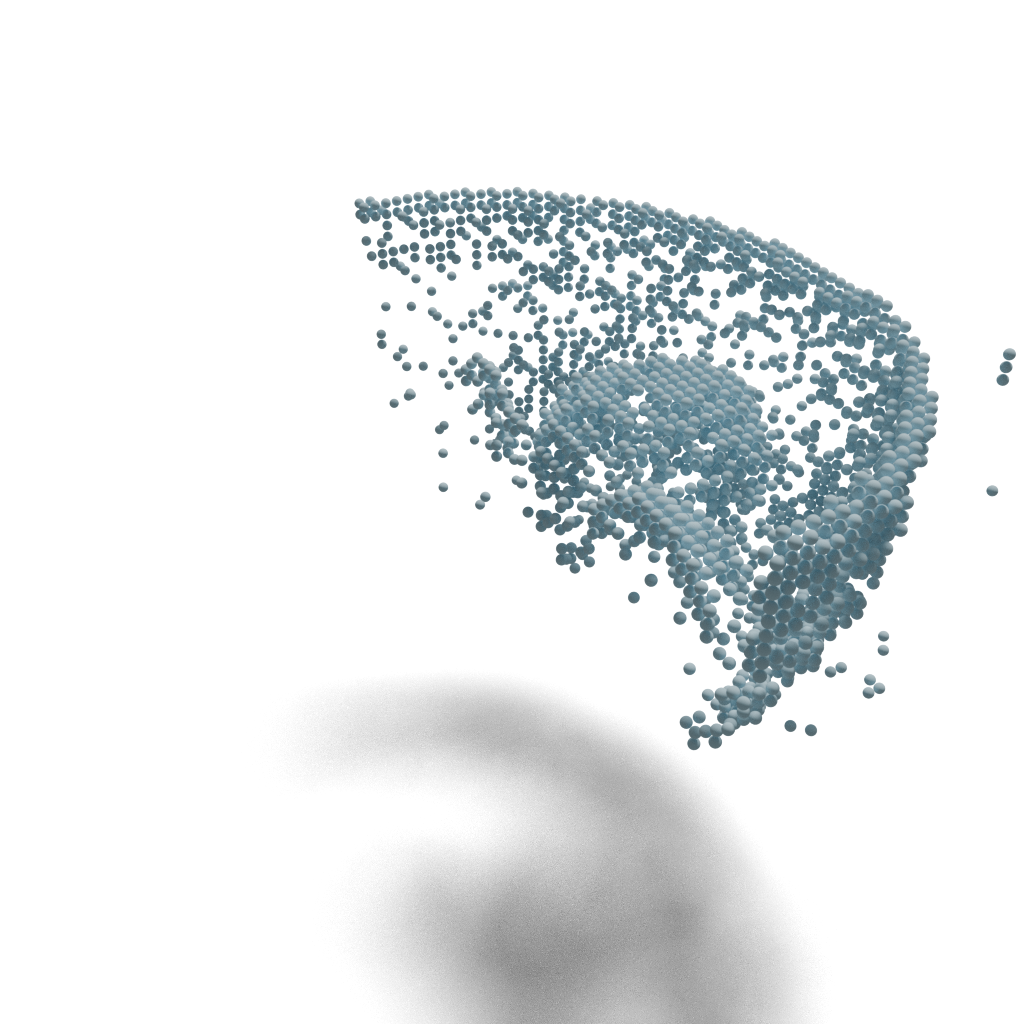} &         \includegraphics[width=0.162\textwidth]{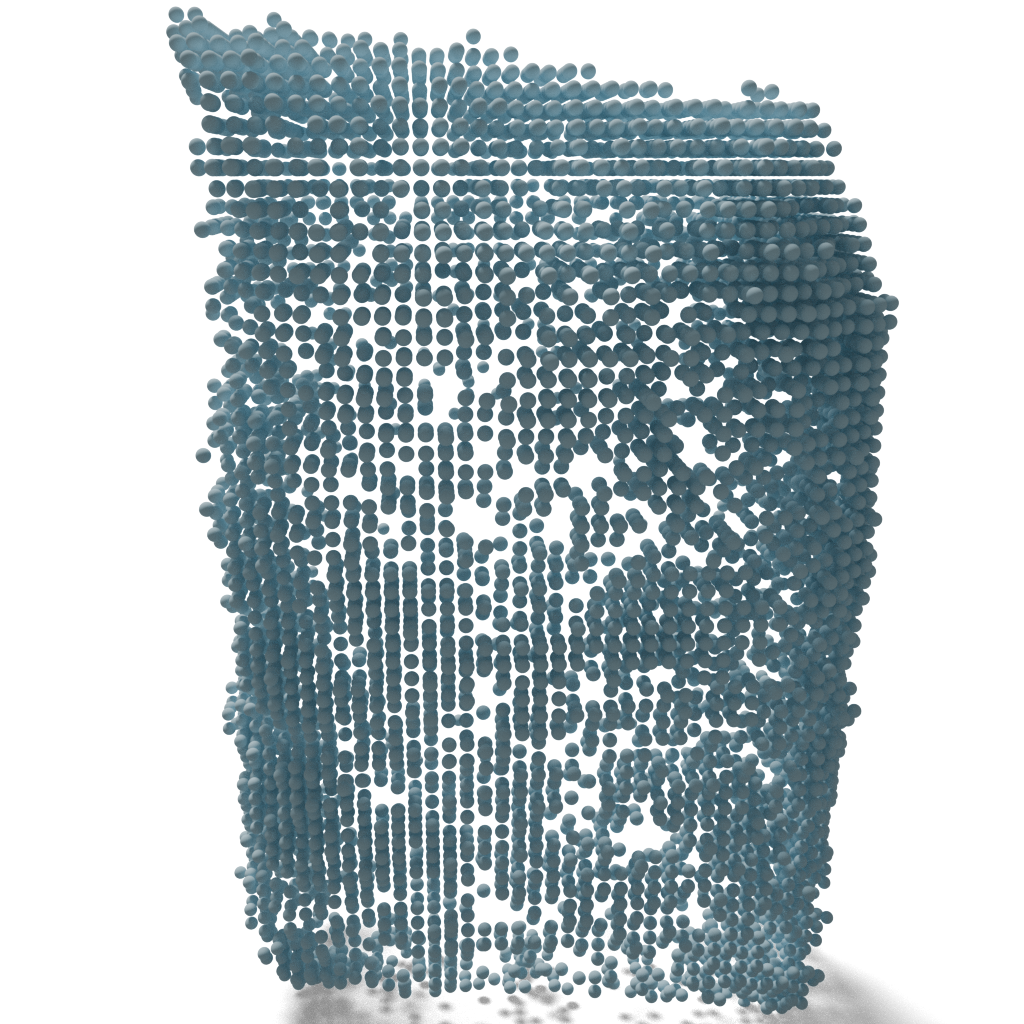}\includegraphics[width=0.162\textwidth]{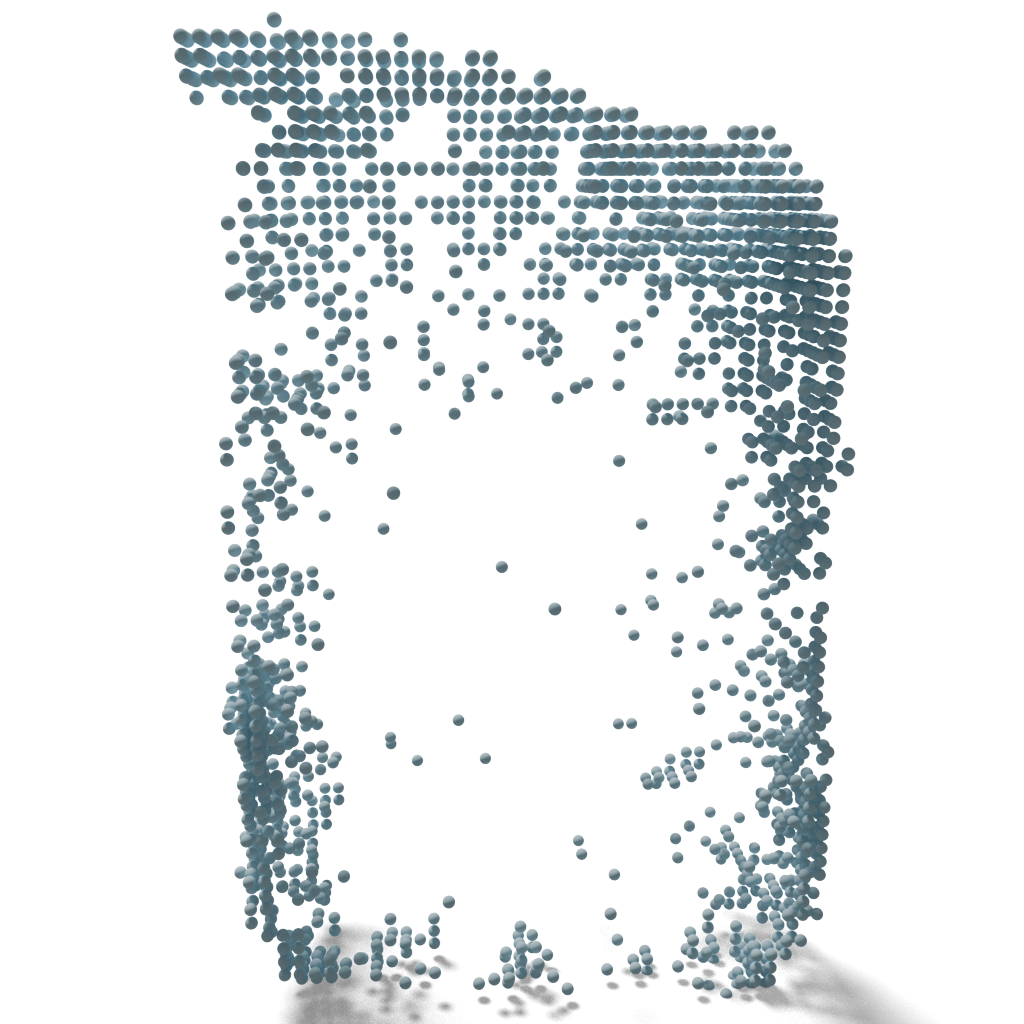}\\
         \multicolumn{3}{c}{Voxelized Splats}\\
    \end{tabular}
    \caption{\textbf{Simulating a Large Gaussian Splat Scene.} We demonstrate an elastodynamic simulation of a large Gaussian Splat scene with multiple objects segmented out and being assigned properties with \acronym.}
    \label{fig:siggraph}
\end{figure}

We demonstrate the process that enables creating simulation-ready, realistic assets from Gaussian Splats and meshes in \Cref{fig:app:qualitative}, demonstrating convincing simulations without any fine-tuning. For example, in~\Cref{fig:grapes}, we first capture a video of an object, and then train a 3D Gaussian Splatting model. Then, we pass it to \acronym\ which estimates the mechanical properties in a couple of seconds. We then use these properties in a simulator to produce a realistic  (see \video{0:36}), greatly reducing the barrier toward constructing realistic interactive digital worlds directly 
from our physical reality.
We demonstrate a Gaussian Splat scene with multiple simulated objects each of which has properties estimated with \acronym, we then place a robot in the scene interacting with the splats in~\Cref{fig:siggraph}.
Through our experiments on Gaussian Splats we find the Gaussian Splat voxelization scheme (\S\ref{sec:implementation}) we introduce empirically qualitatively accurately voxelizes complex, noisy real-world splat objects.

\subsection{More Mechanical Property Prediction Results}\label{app:sec:more_material_results}

\begin{figure*}[ht]
    \centering
    \setlength{\tabcolsep}{0pt}
    \begin{tabular}{p{0.125\textwidth}p{0.125\textwidth}p{0.125\textwidth}p{0.125\textwidth}|p{0.125\textwidth}p{0.125\textwidth}p{0.125\textwidth}p{0.125\textwidth}}
    \centering Object & \centering Young's Modulus ($E$, Pa) & \centering Poisson's Ratio ($\nu$) & \centering Density ($\rho, \frac{kg}{m^3}$) & \centering Object & \centering Young's Modulus ($E$, Pa) & \centering Poisson's Ratio ($\nu$) & \centering Density ($\rho, \frac{kg}{m^3}$) \\
    \end{tabular}
    \begin{tabular}{c|c}
    \includegraphics[width=0.5\textwidth]{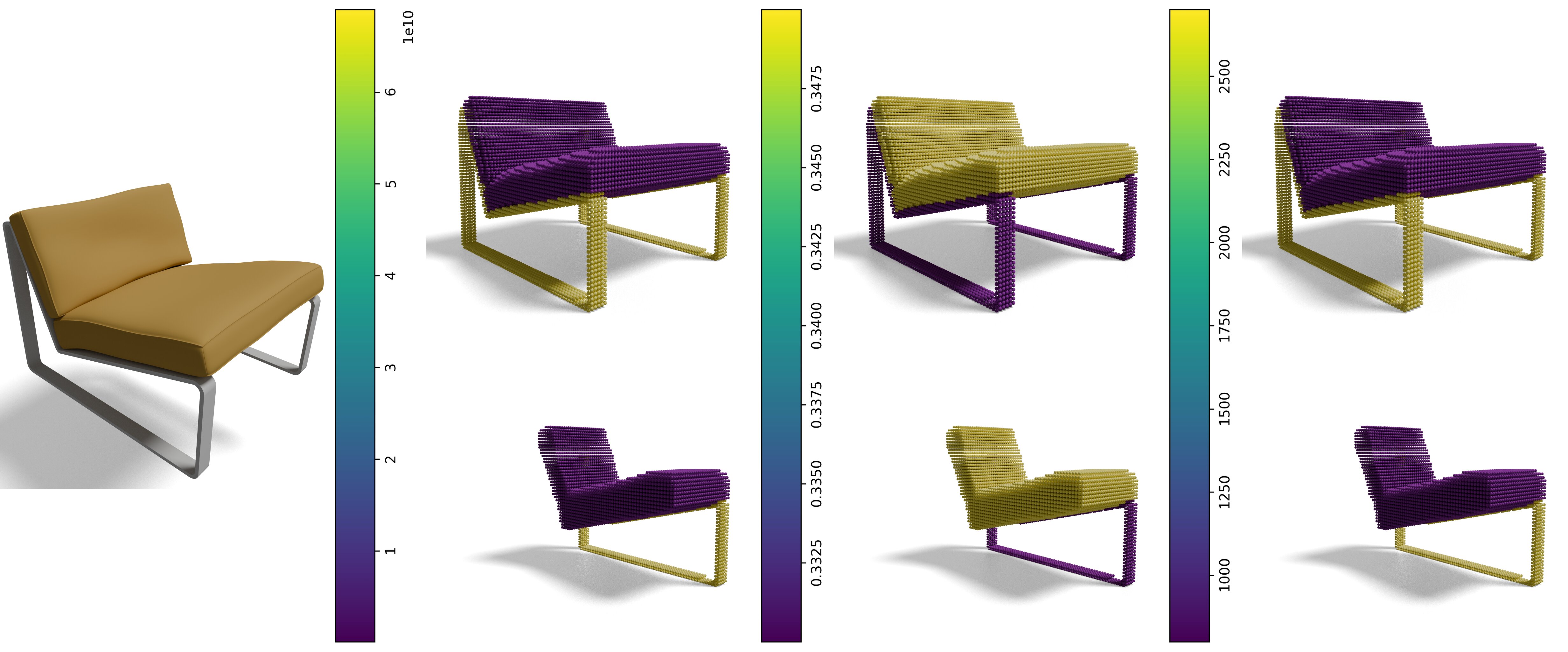} & \includegraphics[width=0.5\textwidth]{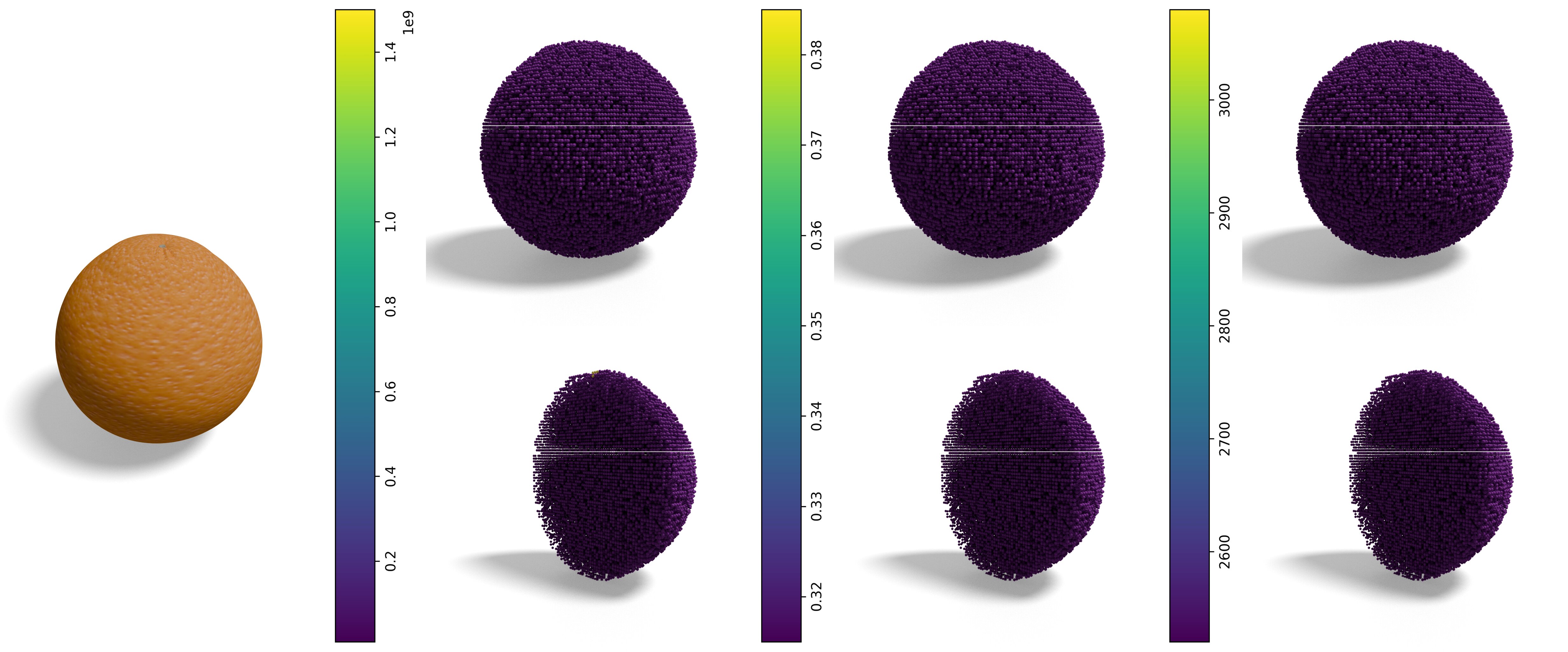} \\
    \includegraphics[width=0.5\textwidth]{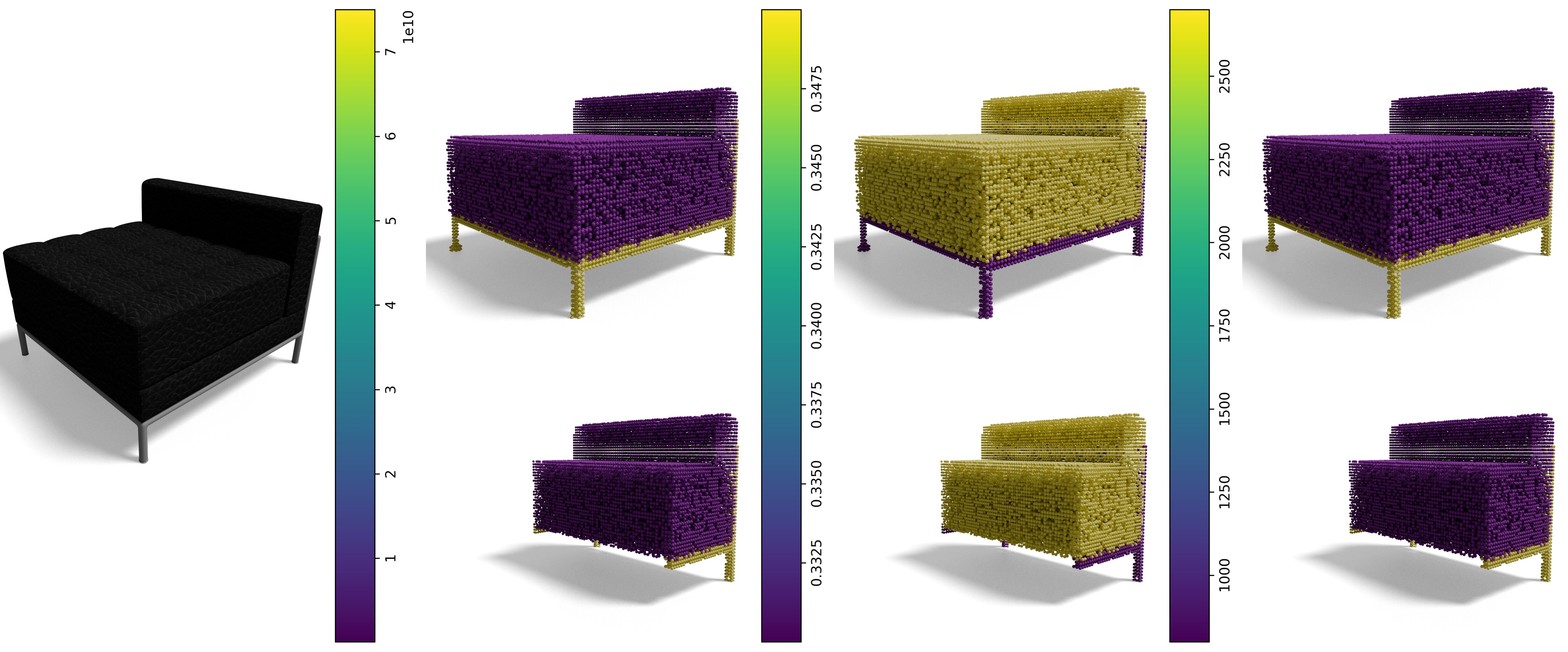} & \includegraphics[width=0.5\textwidth]{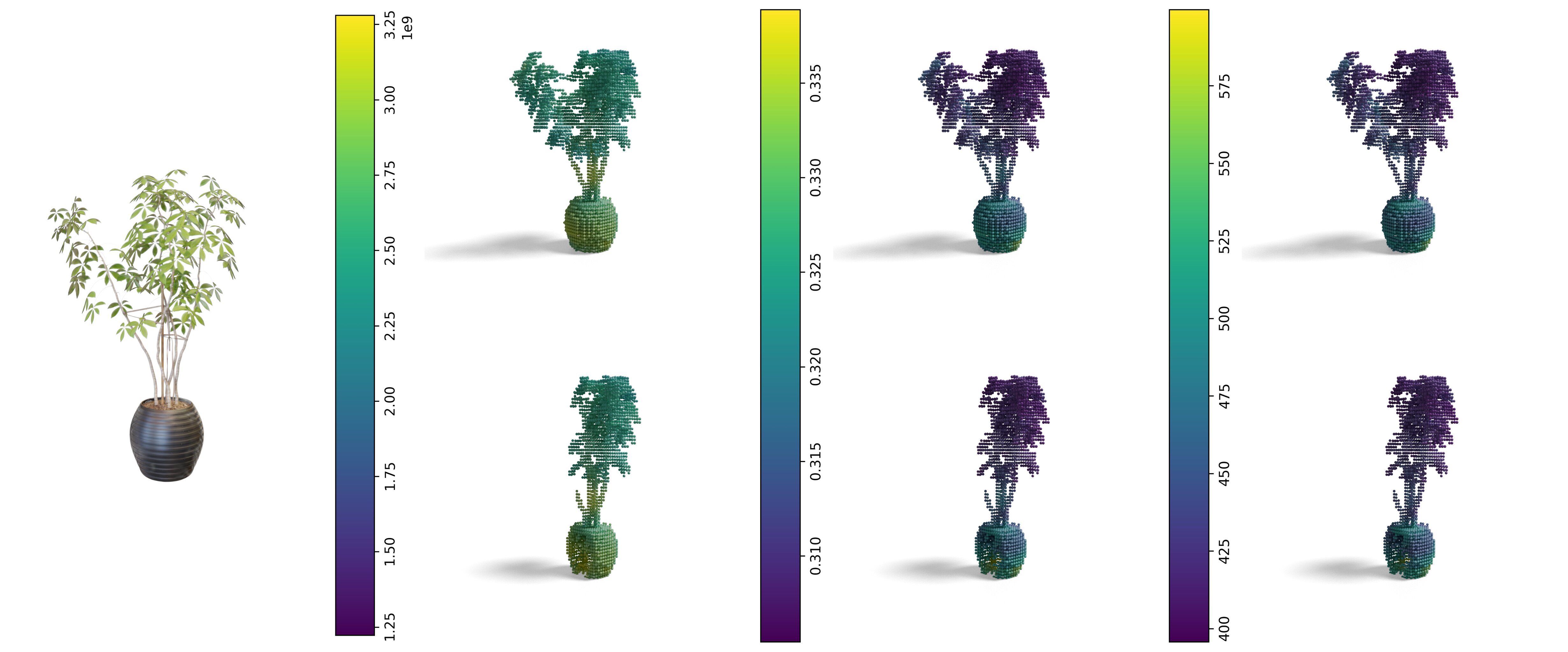} \\
    \includegraphics[width=0.5\textwidth]{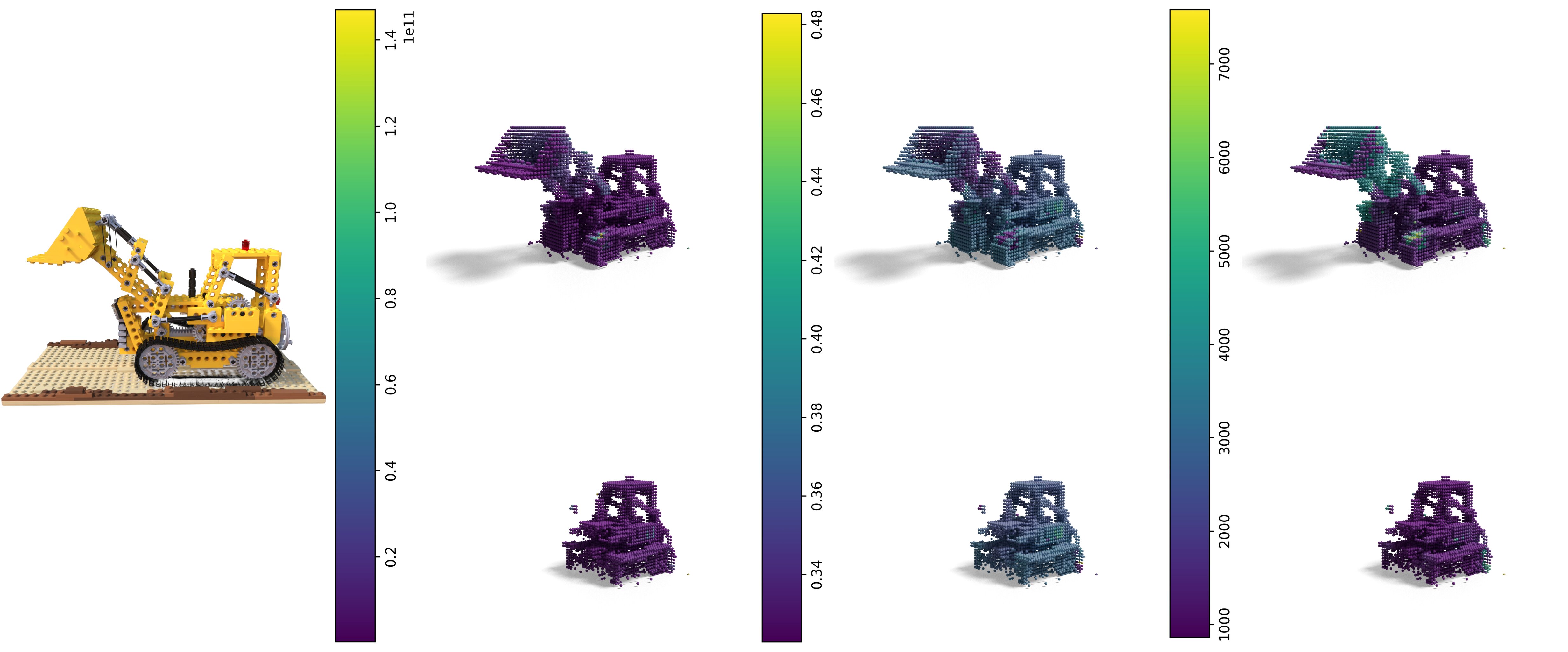} & \includegraphics[width=0.5\textwidth]{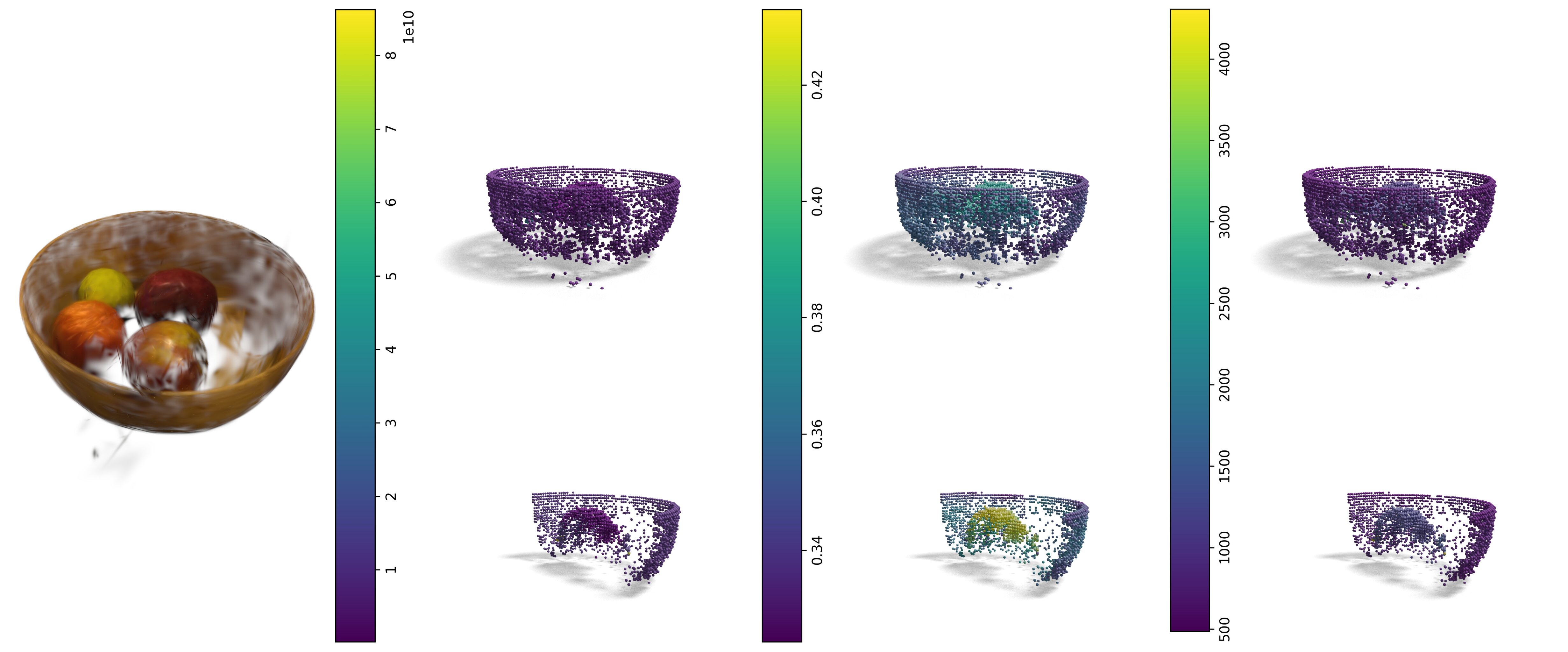} \\
    \includegraphics[width=0.5\textwidth]{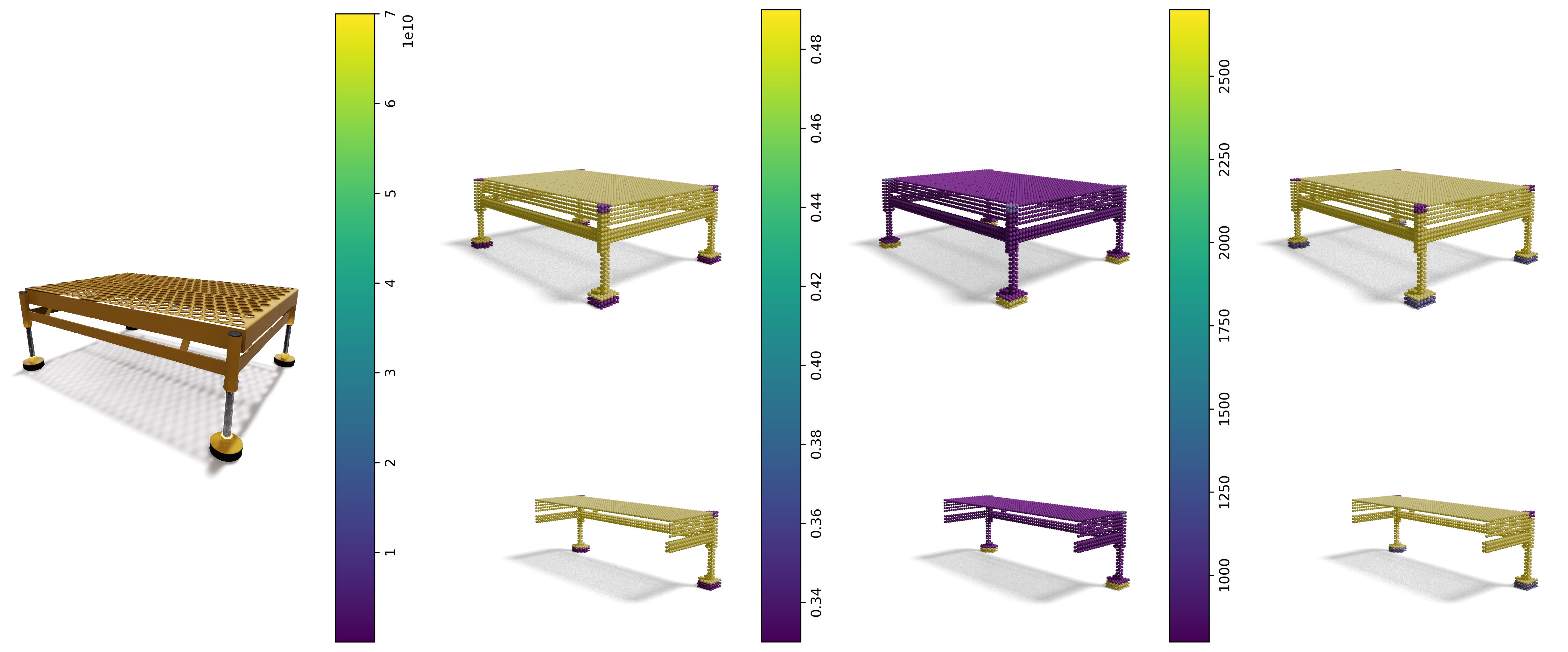} & \includegraphics[width=0.5\textwidth]{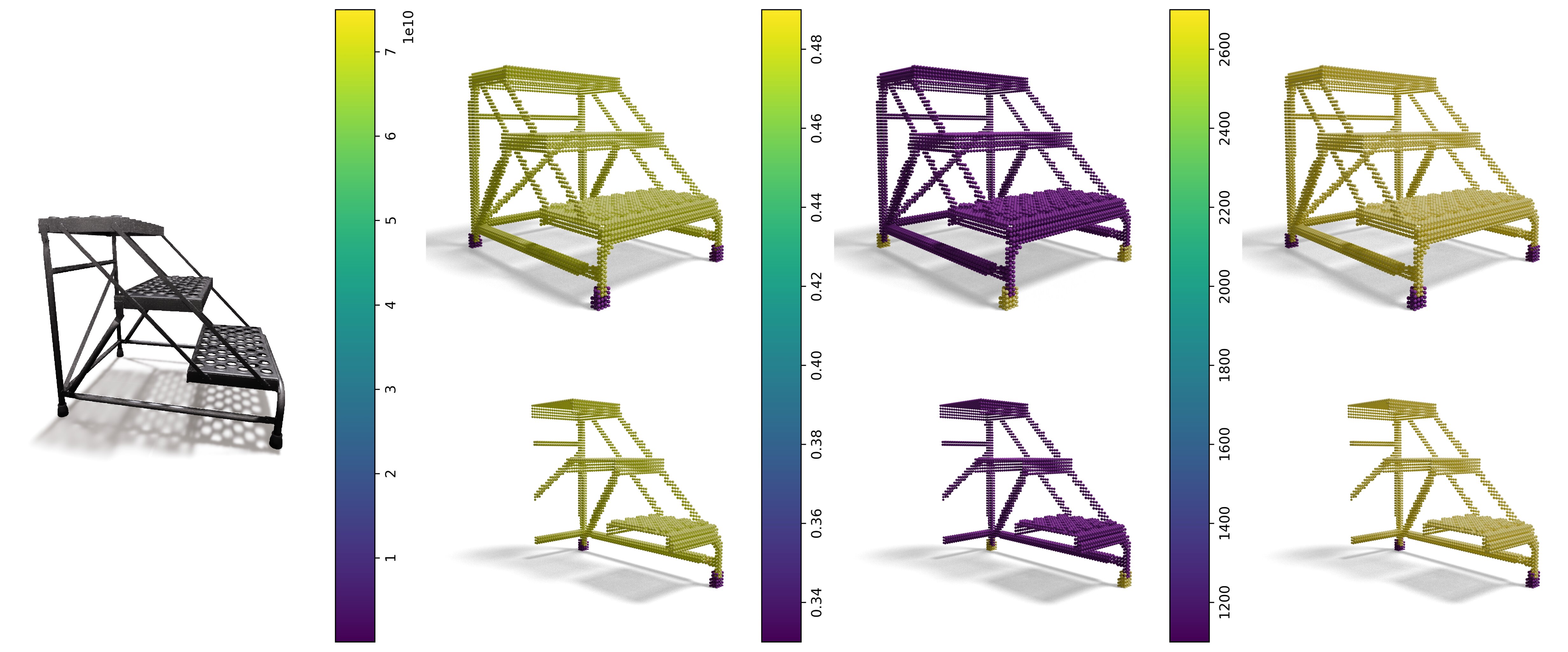} \\
    \includegraphics[width=0.5\textwidth]{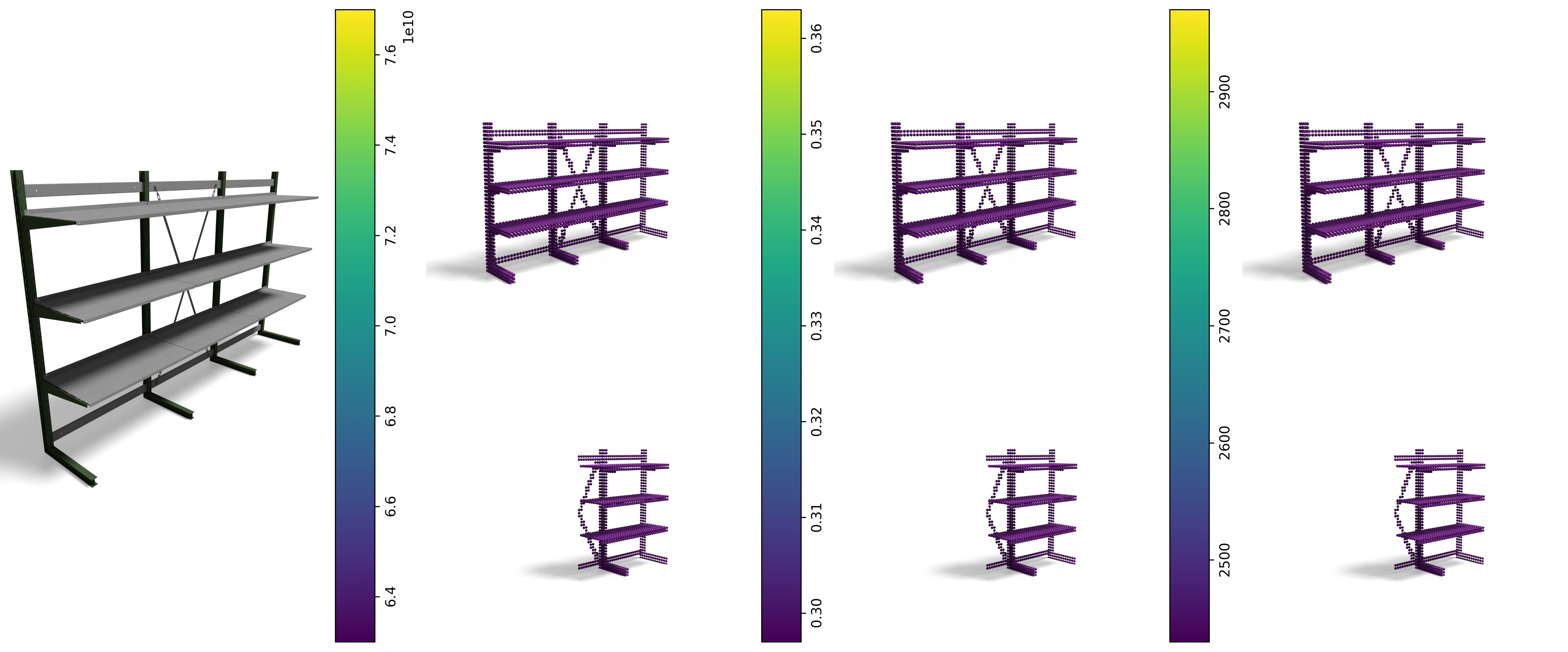} & \includegraphics[width=0.5\textwidth]{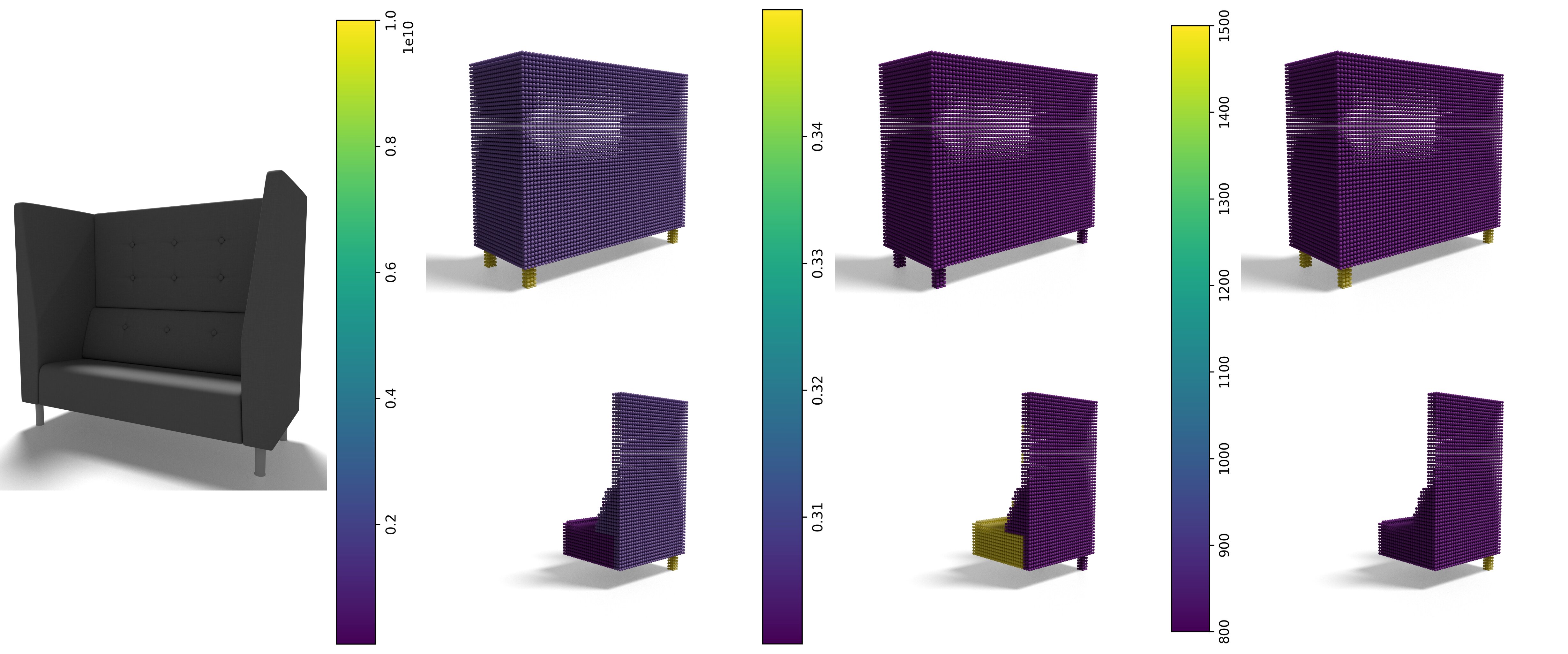} \\
    \end{tabular}
    \caption{\textbf{Inferred Mechanical Property Fields.} We show additional mechnical property fields and slice planes through mechanical property fields estimated by \acronym\ (\video{1:40}).}
    \label{fig:extrafields1}
\end{figure*}

\begin{figure*}[tb]
    \centering
    \setlength{\tabcolsep}{0pt}
    \begin{tabular}{p{0.125\textwidth}p{0.125\textwidth}p{0.125\textwidth}p{0.125\textwidth}|p{0.125\textwidth}p{0.125\textwidth}p{0.125\textwidth}p{0.125\textwidth}}
    \centering Object & \centering Young's Modulus ($E$, Pa) & \centering Poisson's Ratio ($\nu$) & \centering Density ($\rho, \frac{kg}{m^3}$) & \centering Object & \centering Young's Modulus ($E$, Pa) & \centering Poisson's Ratio ($\nu$) & \centering Density ($\rho, \frac{kg}{m^3}$) \\
    \end{tabular}
    \begin{tabular}{c|c}
    \includegraphics[width=0.5\textwidth]{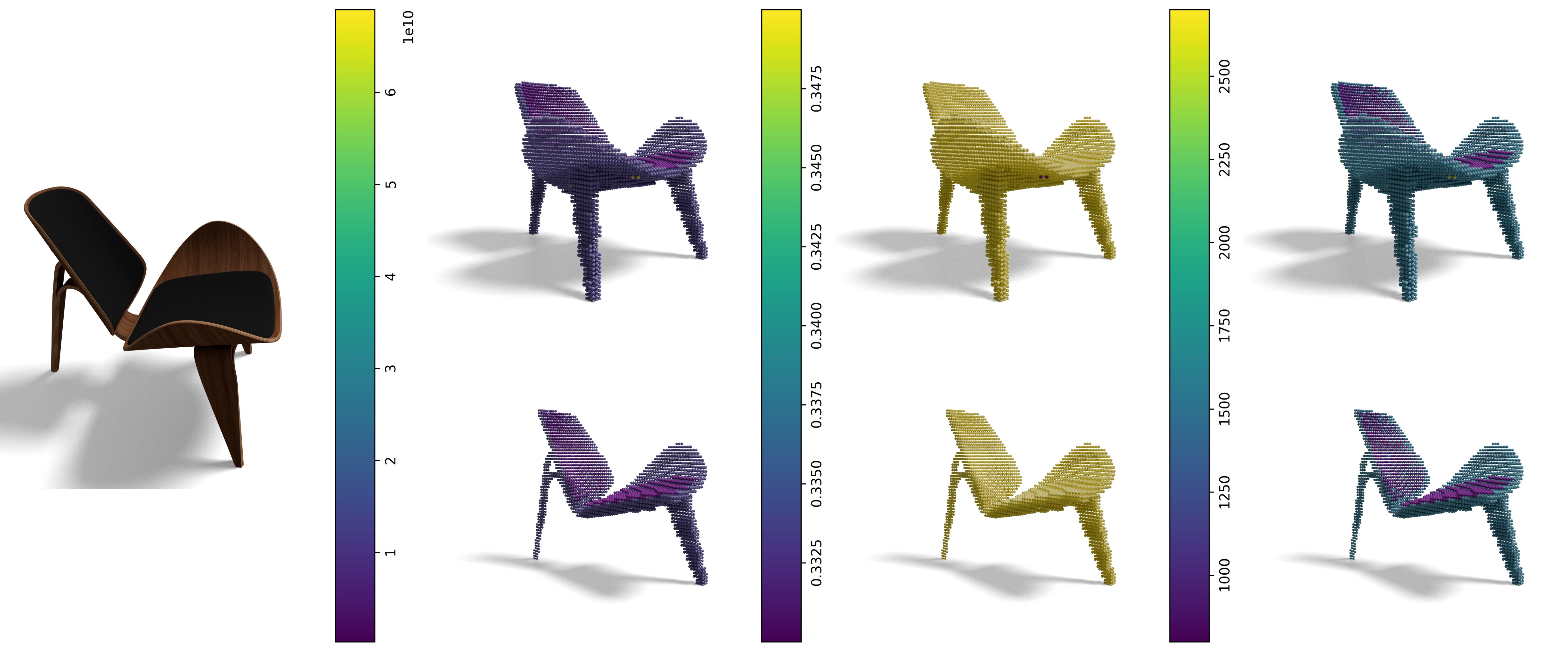} & \includegraphics[width=0.5\textwidth]{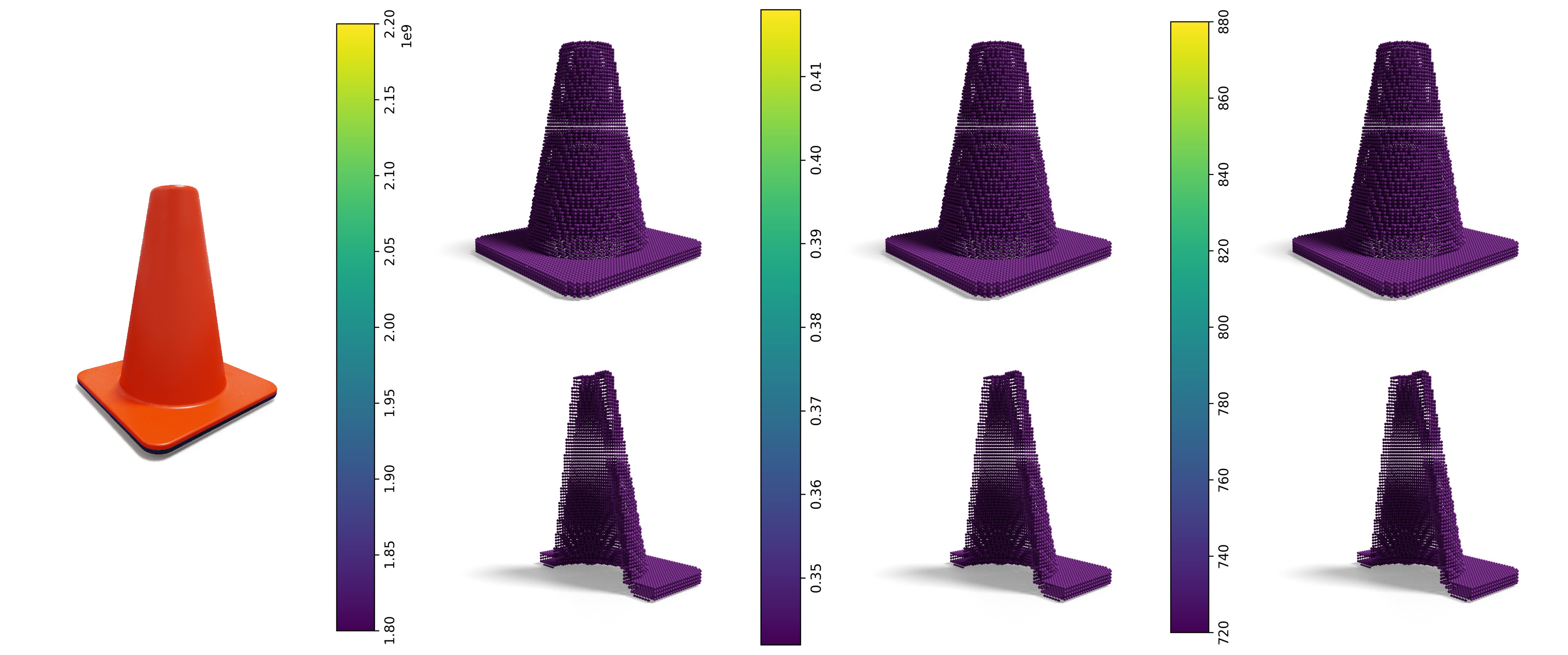} \\
    \includegraphics[width=0.5\textwidth]{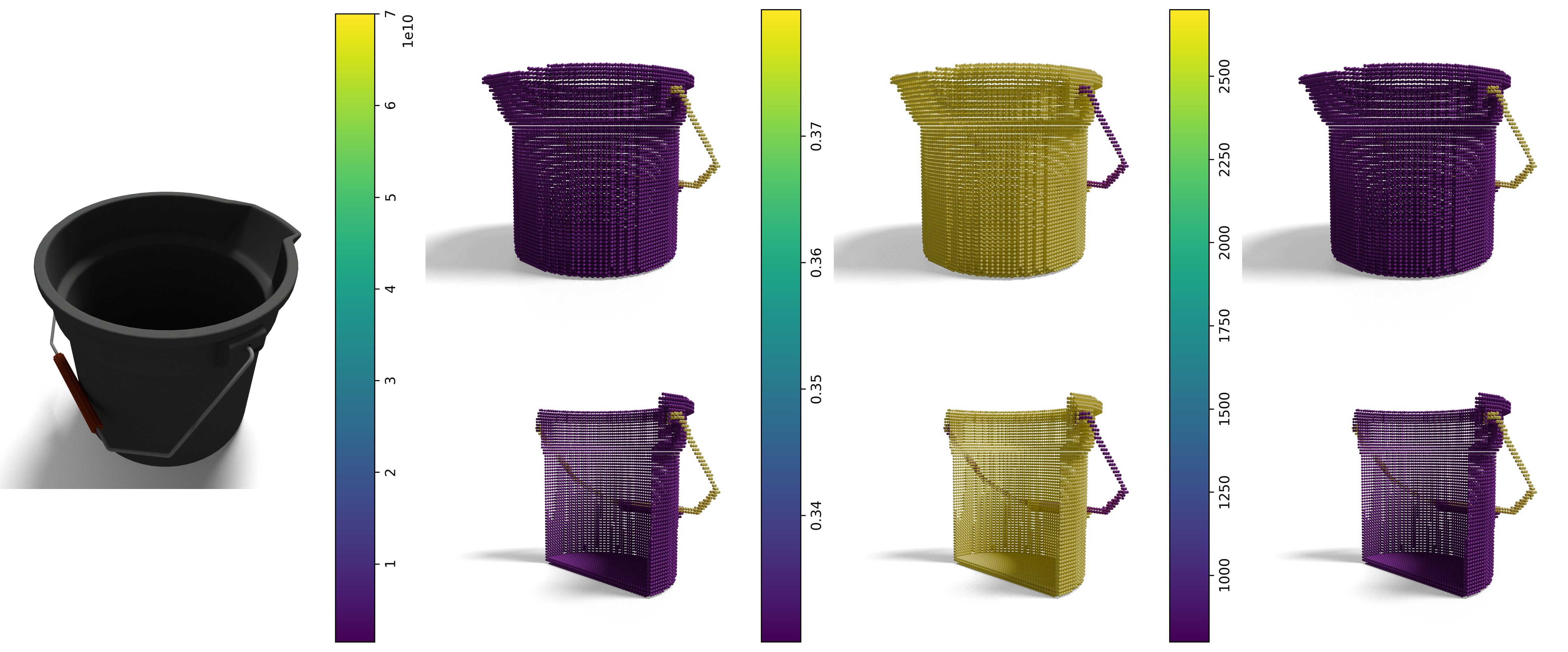} & \includegraphics[width=0.5\textwidth]{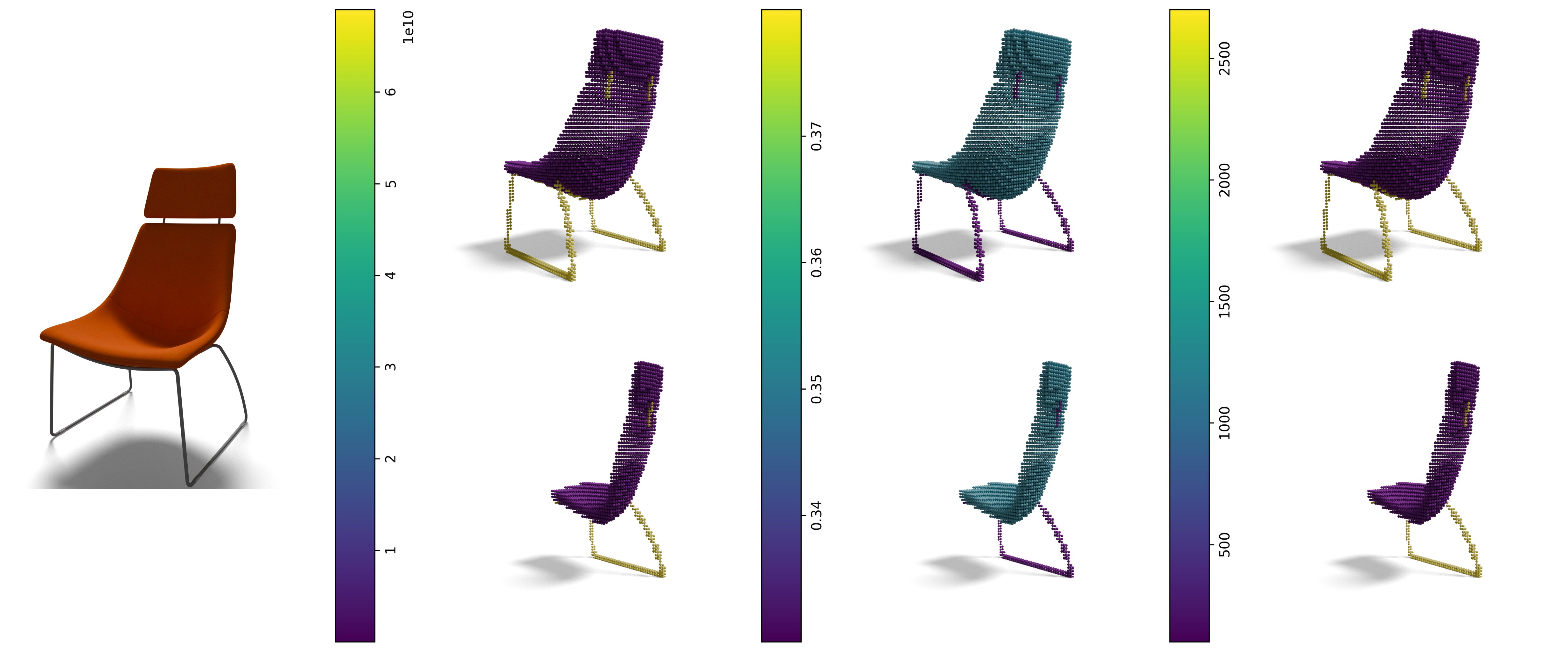} \\
    \includegraphics[width=0.5\textwidth]{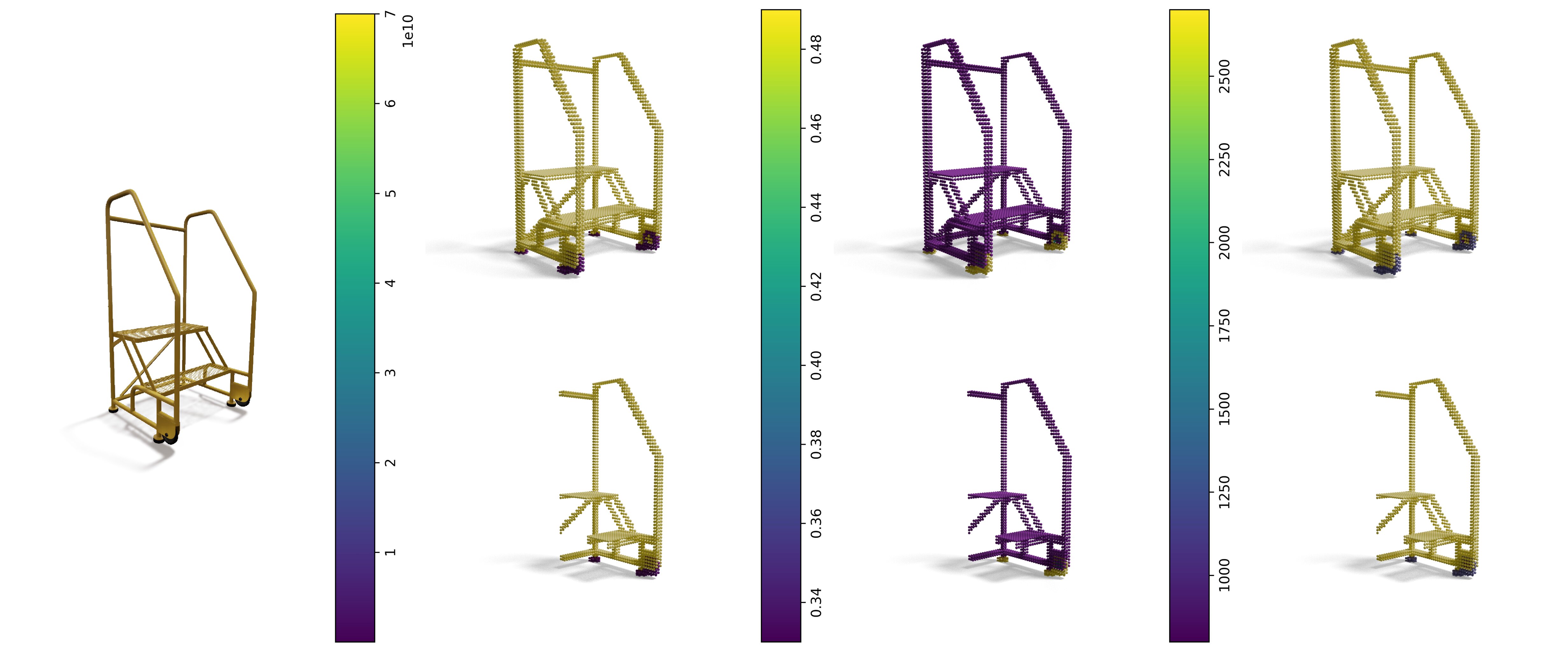} & \includegraphics[width=0.5\textwidth]{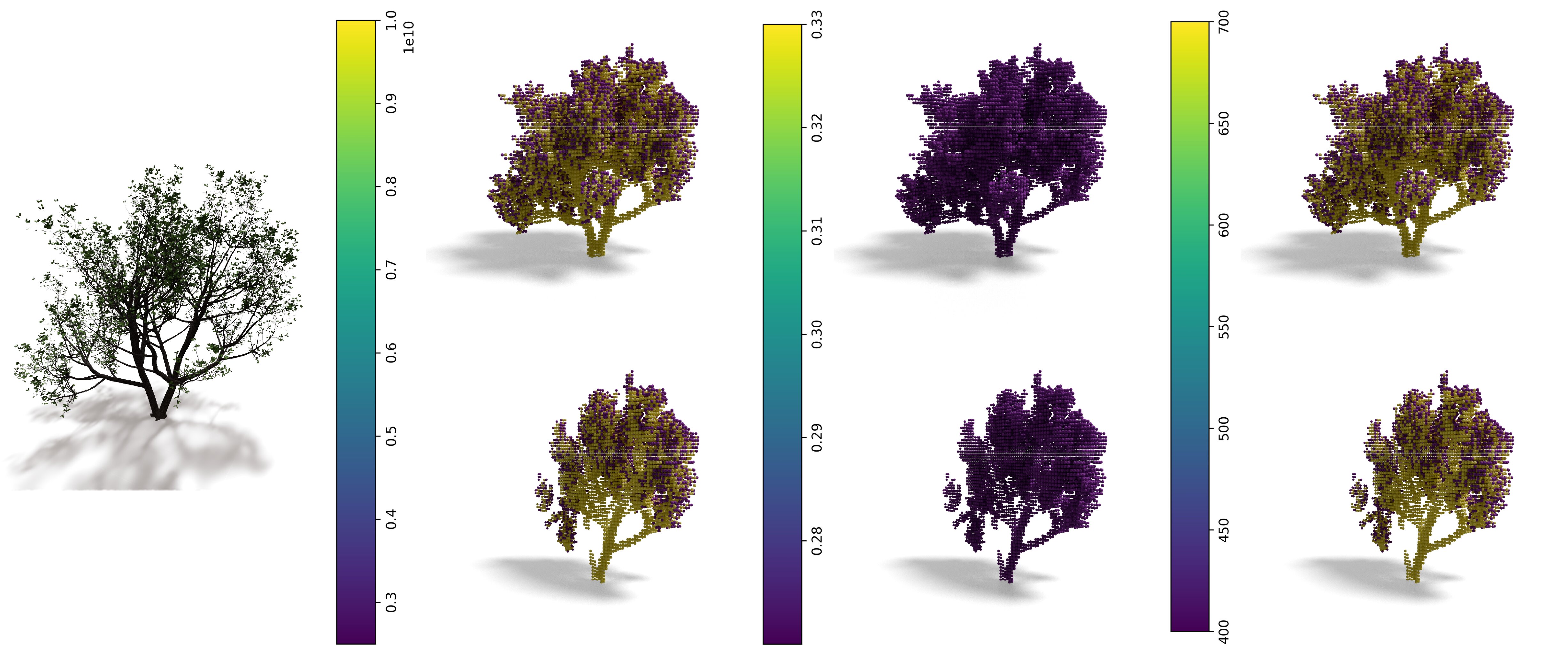} \\
    \includegraphics[width=0.5\textwidth]{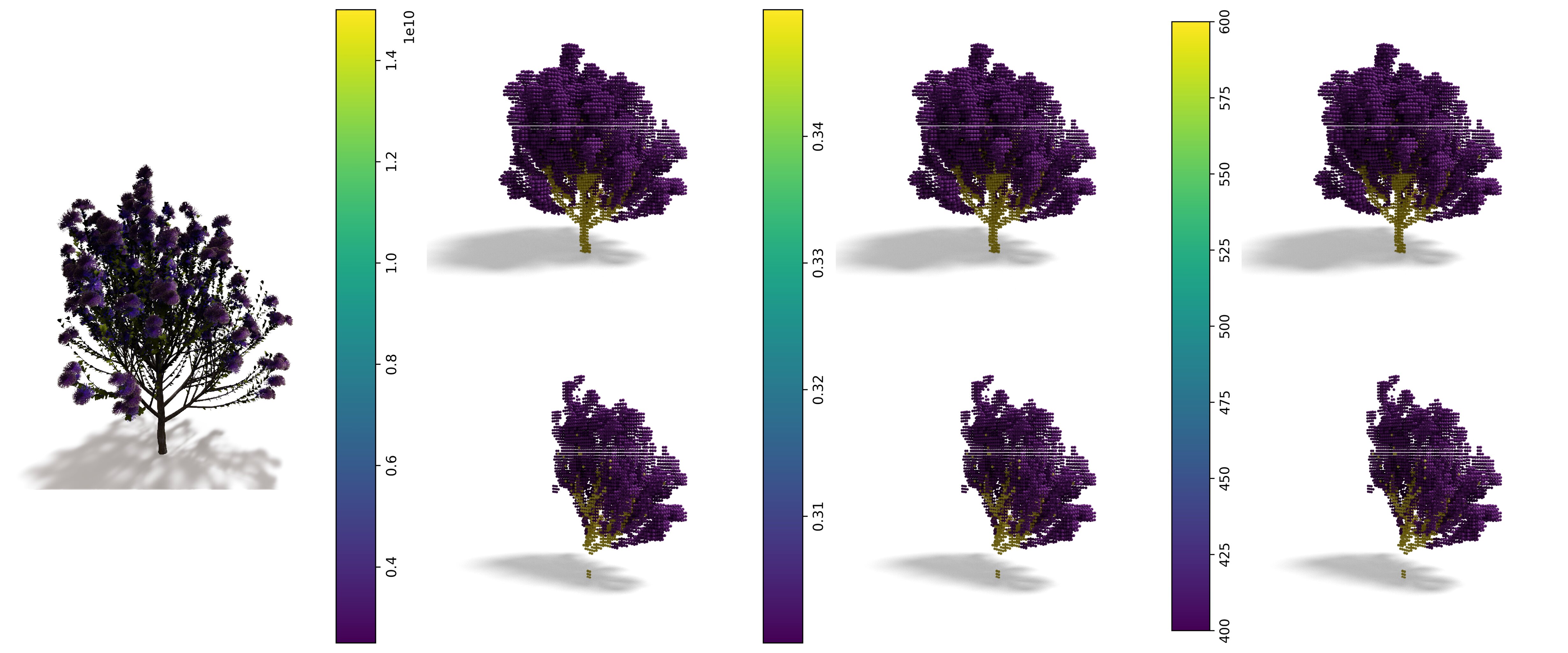} & \includegraphics[width=0.5\textwidth]{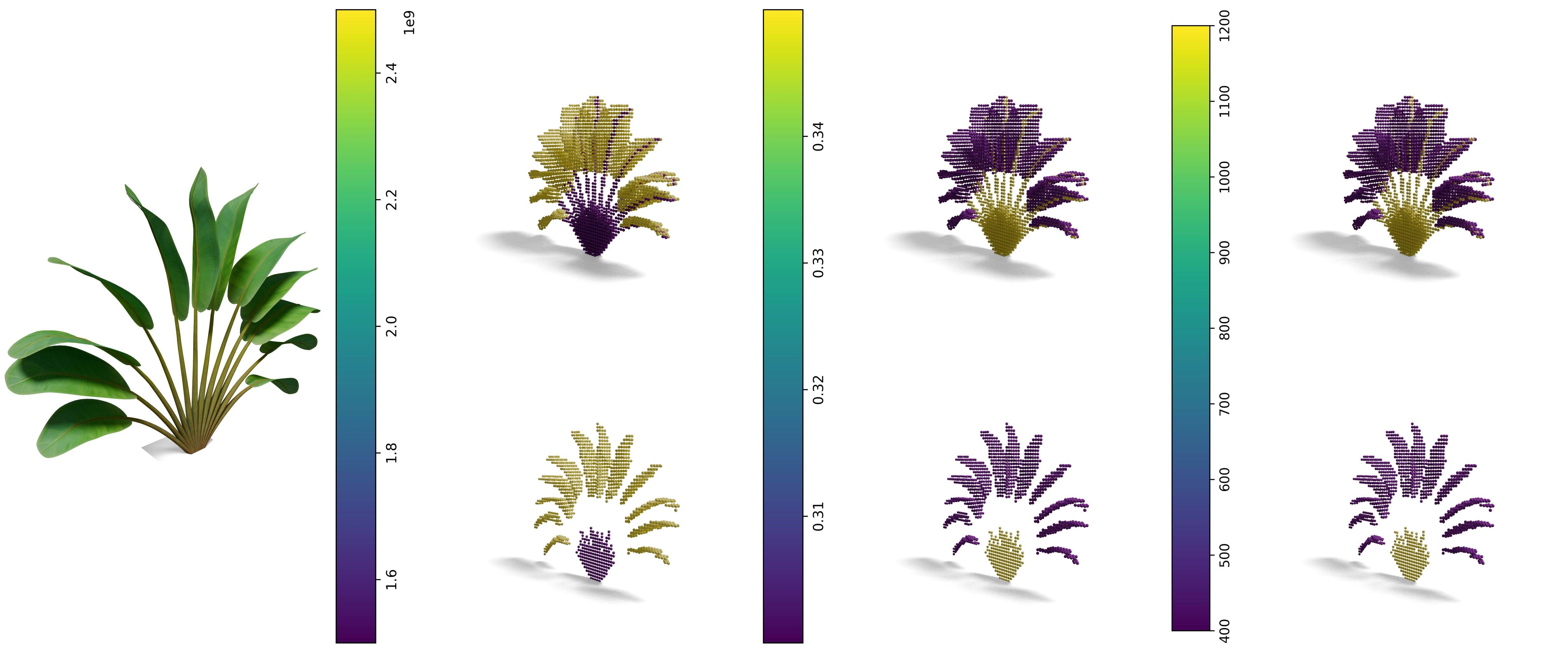} \\
    \includegraphics[width=0.5\textwidth]{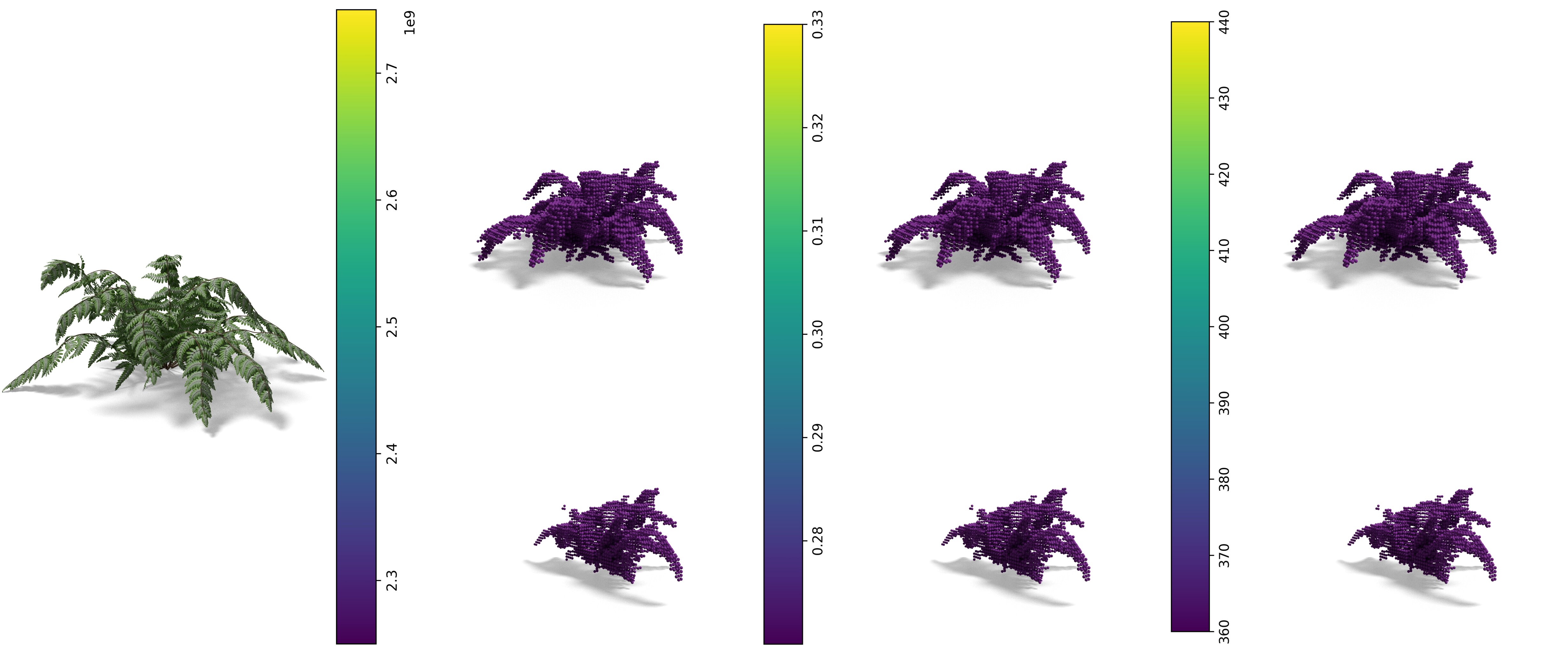} & \includegraphics[width=0.5\textwidth]{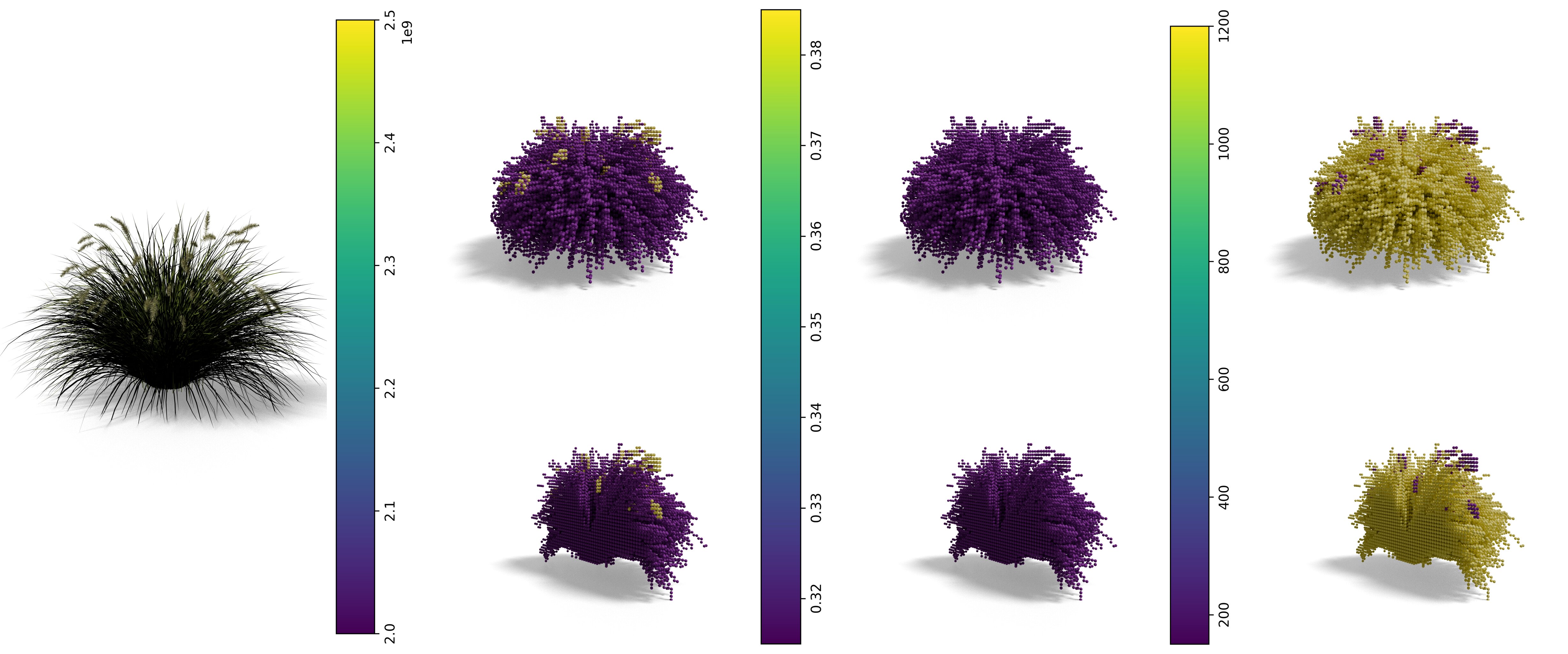} \\
    \end{tabular}
    \caption{\textbf{Inferred Mechanical Property Fields.} We show additional mechnical property fields and slice planes through mechanical property fields estimated by \acronym\ (\video{2:10}).}
    \label{fig:extrafields2}
\end{figure*}

We show qualitative results for inferring mechanical property fields in~\Cref{fig:extrafields1,fig:extrafields2}.
We notice from~\Cref{fig:extrafields1} row 1, column 2 that our model can pick up small details like the stem of the orange at the top of the object, which is given a different Young's modulus, though it only spans a few voxels (see \video{1:38}). We notice from~\Cref{fig:extrafields1} row 2, column 2 that our model finds that the space inside the pot should be made up of properties that fall in the range of dirt, even though the inside of the pot was not observed through external renders (see \video{1:50}).  We notice from~\Cref{fig:extrafields1} row 3, column 2 that our model can tolerate some noise in assets such as the Gaussian splat of a bowl with fruits segmented from a larger Gaussian splat (see \video{2:01}). We notice from~\Cref{fig:extrafields2} row 1, column 1 that our model can accurately predict thin segments and thin boundaries, like for the seat of the chair (see \video{2:10}). We notice from~\Cref{fig:extrafields1} row 3, column 2 that our models can handle complex assets and complex materials like trees and accurately handle fine details, such as understanding where all the leaves lie and giving them different material properties than wood (see \video{2:16}). We notice from~\Cref{fig:extrafields1} row 4, column 1 that our models can handle complex volumetric materials, such as annotating the properties of wood under the flowers (see \video{2:28}). We notice from~\Cref{fig:extrafields1} row 4, column 2 that our models can handle thin materials and identify the vein of the leaves (see \video{2:35}). For completeness, we also include our metrics normalized by total test set voxels in \Cref{tab:material-properties} and observe the
same performance boost compared to prior art as with the per-object normalization as presented in the main paper \Cref{tab:material-properties-object}.

We present dataset voxel-averaged results on the publicly-released dataset in \Cref{tab:public-material-properties-voxel}. The publicly-released dataset does not include the vegetation subset.

\begin{table}[tb]
\centering
\caption{\textbf{Voxel Mechanical Property Estimation.} Errors for predicting mechanical properties on the publicly released dataset, which does not include the vegetation subset, are very close to the full dataset. These metrics are computed by averaging across all voxels across all 3D objects in the public test set.}
\label{tab:public-material-properties-voxel}
\resizebox{\textwidth}{!}{%
\begin{tabular}{lrrrrrr}
\toprule
\rowcolor{nvidiagreen!15}Method & \multicolumn{2}{c}{Young's Modulus Pa ($E$)} & \multicolumn{2}{c}{Poisson's Ratio ($\nu$)} & \multicolumn{2}{c}{Density $\frac{kg}{m^3}$ ($\rho$)} \\
\cmidrule(r){2-3} \cmidrule(r){4-5} \cmidrule(r){6-7}
\rowcolor{nvidiagreen!15}& ALDE ($\downarrow$) & ALRE ($\downarrow$) & ADE ($\downarrow$) & ARE ($\downarrow$) & ADE ($\downarrow$) & ARE ($\downarrow$) \\
\midrule
NeRF2Physics &
\underline{2.8000} {\scriptsize{($\pm$1.05)}} &
\underline{0.1346} {\scriptsize{($\pm$0.05)}} &
- & - &
\underline{1432.0343} {\scriptsize{($\pm$964.88)}} &
\underline{1.0365} {\scriptsize{($\pm$0.63)}} \\
PUGS &
3.3942 {\scriptsize{($\pm$1.72)}} &
0.1688 {\scriptsize{($\pm$0.10)}} &
- & - &
3568.2150 {\scriptsize{($\pm$2839.13)}} &
3.2429 {\scriptsize{($\pm$3.56)}} \\
Phys4DGen$^\star$ &
4.8967 {\scriptsize{($\pm$3.17)}} &
0.2227 {\scriptsize{($\pm$0.14)}} &
\underline{0.0407} {\scriptsize{($\pm$0.04)}} &
\underline{0.1467} {\scriptsize{($\pm$0.18)}} &
1865.5673 {\scriptsize{($\pm$2176.90)}} &
1.4394 {\scriptsize{($\pm$2.35)}} \\
\midrule
Ours &
\textbf{0.3766} {\textbf{\scriptsize\textcolor{gray}{($\pm$0.39)}}} &
\textbf{0.0421} {\textbf{\scriptsize\textcolor{gray}{($\pm$0.05)}}} &
\textbf{0.0250} {\textbf{\scriptsize\textcolor{gray}{($\pm$0.01)}}} &
\textbf{0.0837} {\textbf{\scriptsize\textcolor{gray}{($\pm$0.03)}}} &
\textbf{113.4683} {\textbf{\scriptsize\textcolor{gray}{($\pm$302.19)}}} &
\textbf{0.0909} {\textbf{\scriptsize\textcolor{gray}{($\pm$0.14)}}} \\
\bottomrule
\end{tabular}
}
\end{table}

\subsection{Mass Estimation Example Results}\label{app:sec:more_mass_results}

We show qualitative results from our model on mass estimation on the ABO-500~\cite{Collins_2022_CVPR} dataset in~\Cref{fig:abo}.

\clearpage

\begin{figure*}[ht!]
\begin{minipage}[t]{\textwidth}
     %\vspace*{0pt}
        \centering
    \includegraphics[width=\textwidth]{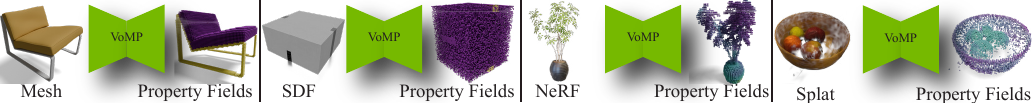}
    \subcaption{\textbf{Inference across representations:} We show that \acronym\ can operate on any geometry that can be voxelized and rendered, including meshes, SDF, NeRFs, and Gaussian Splats (\video{1:30}).}
    \label{fig:reps}
    \end{minipage}
     \begin{minipage}[t]{\textwidth}
     \centering
    \includegraphics[width=\textwidth]{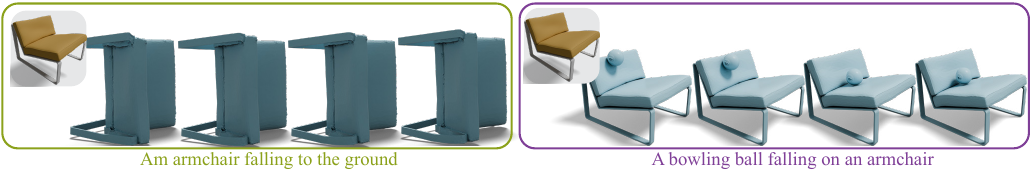}
    \subcaption{\textbf{Simulating meshes:} We show realistic simulations for meshes using predicted material values (\video{3:40}).}\label{fig:app:simulations2}
    \end{minipage}
    \begin{minipage}[t]{\textwidth}
    \centering
    \includegraphics[width=\textwidth]{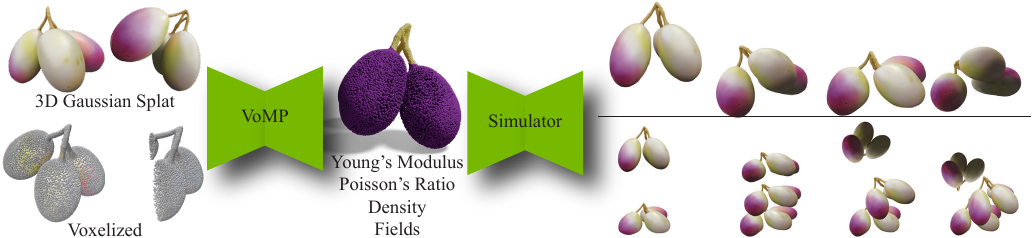}
    \subcaption{\textbf{Simulating captured Gaussian Splat model}: In this example, we apply \acronym\ to this Gaussian Splat model that we captured using a commercial app. Our method converts this model into a simulation-ready asset (\video{0:36}).}\label{fig:grapes}
    \end{minipage}
     \begin{minipage}[t]{\textwidth}
     \centering
    \includegraphics[width=\textwidth]{assets/simulations.pdf}
    \subcaption{\textbf{Simulating meshes and splats:} We show realistic simulations for meshes (left) and splats (right) using predicted material values (\video{3:20}).}\label{fig:app:simulations}
    \end{minipage}
\begin{minipage}[t]{\textwidth}
 \centering
    \includegraphics[width=\textwidth]{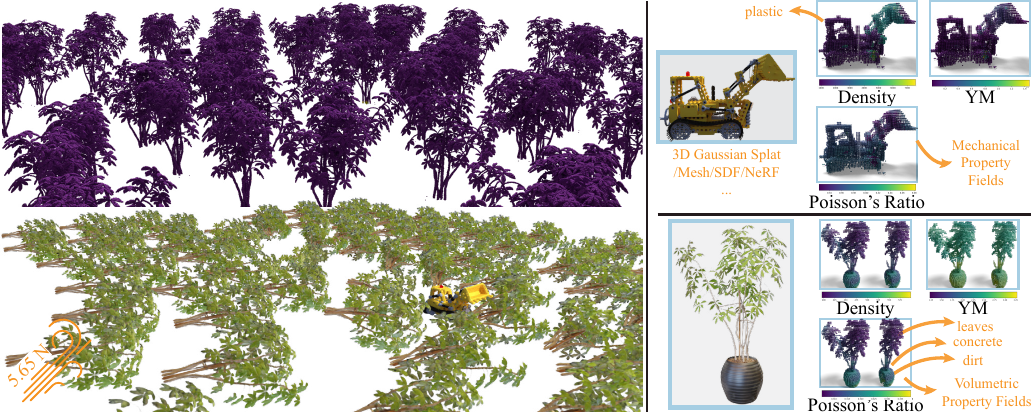}
    \subcaption{\textbf{Simulating Gaussian Splats at scale:} an elastodynamic simulation of a Gaussian Splat bulldozer going through a forest of 100 Gaussian splat ficuses in the presence of wind; all materials predicted by \acronym (\video{3:00}).}\label{teaser_old}
\end{minipage}
\caption{\textbf{End To End Results:} We test \acronym\ material estimates on a variety of input representations (a), and show realistic simulations without any hand-tuning for meshes and splats across diverse scenarios. No hand-tuning of our predicted material parameters was performed, showing that \acronym\ directly
predicts simulation-ready parameters. See our \includegraphics[height=0.6\baselineskip]{assets/youtube-brands_svg-tex.pdf} video for the simulation of these examples.}
\label{fig:app:qualitative}
\end{figure*}

\clearpage

\newcommand{\voxelproprow}{
\textbf{0.3765} {\textbf{\scriptsize\textcolor{gray}{($\pm$0.39)}}} & 
\textbf{0.0421} {\textbf{\scriptsize\textcolor{gray}{($\pm$0.05)}}} & 
\textbf{0.0250} {\textbf{\scriptsize\textcolor{gray}{($\pm$0.01)}}} & 
\textbf{0.0837} {\textbf{\scriptsize\textcolor{gray}{($\pm$0.03)}}} & 
\textbf{113.3807} {\textbf{\scriptsize\textcolor{gray}{($\pm$301.90)}}} & 
\textbf{0.0908} {\textbf{\scriptsize\textcolor{gray}{($\pm$0.14)}}}
}
\vspace{2em}
\begin{table*}[hb]
\centering
\caption{\textbf{Voxel Mechanical Property Estimation.} Errors for predicting mechanical properties from 3D objects averaged across all voxels in the test set.}
\resizebox{\textwidth}{!}{%
\begin{tabular}{lrrrrrr}
\toprule
\rowcolor{nvidiagreen!15}Method & \multicolumn{2}{c}{Young's Modulus Pa ($E$)} & \multicolumn{2}{c}{Poisson's Ratio ($\nu$)} & \multicolumn{2}{c}{Density $\frac{kg}{m^3}$ ($\rho$)} \\
\cmidrule(r){2-3} \cmidrule(r){4-5} \cmidrule(r){6-7}
\rowcolor{nvidiagreen!15}& ALDE ($\downarrow$) & ALRE ($\downarrow$) & ADE ($\downarrow$) & ARE ($\downarrow$) & ADE ($\downarrow$) & ARE ($\downarrow$) \\
\midrule
NeRF2Physics~\cite{zhai2024physicalpropertyunderstandinglanguageembedded} & \underline{2.5719} {\scriptsize{($\pm$1.15)}} & \underline{0.4122} {\scriptsize{($\pm$0.08)}} & - & - & 1354.9458 {\scriptsize{($\pm$1315.71)}} & 1.1496 {\scriptsize{($\pm$0.67)}} \\
PUGS~\cite{shuai2025pugszeroshotphysicalunderstanding} & 3.8619 {\scriptsize{($\pm$2.01)}} & 0.4512 {\scriptsize{($\pm$0.11)}} & - & - & 3641.0715 {\scriptsize{($\pm$3320.78)}} & 4.0413 {\scriptsize{($\pm$4.16)}} \\
Phys4DGen$^\star$~\cite{lin2025phys4dgenphysicscompliant4dgeneration} & 5.2977 {\scriptsize{($\pm$3.36)}} & 0.4825 {\scriptsize{($\pm$0.14)}} & \underline{0.0394} {\scriptsize{($\pm$0.05)}} & \underline{0.1425} {\scriptsize{($\pm$0.21)}} & \underline{1285.9489} {\scriptsize{($\pm$1981.11)}} & \underline{1.0445} {\scriptsize{($\pm$2.53)}} \\
\midrule
Ours &  \voxelproprow\\
\bottomrule
\end{tabular}
}
\label{tab:material-properties}
\end{table*}
\clearpage

\ifdefined\iclr
    \begin{figure*}[ht]
    \centering
    \setlength{\tabcolsep}{0pt}%
    \begin{tabular}{cccc}
    \includegraphics[width=0.24\textwidth]{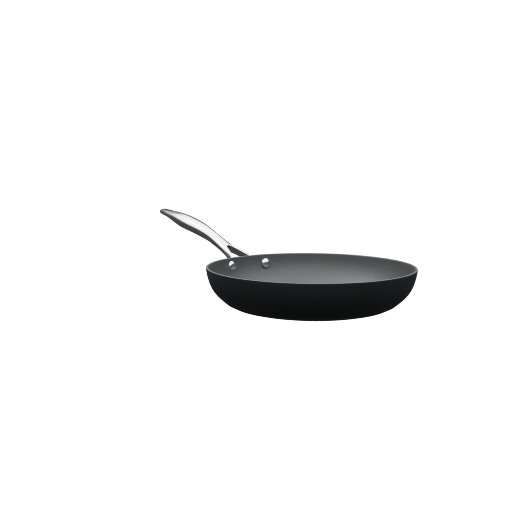} & \includegraphics[width=0.24\textwidth]{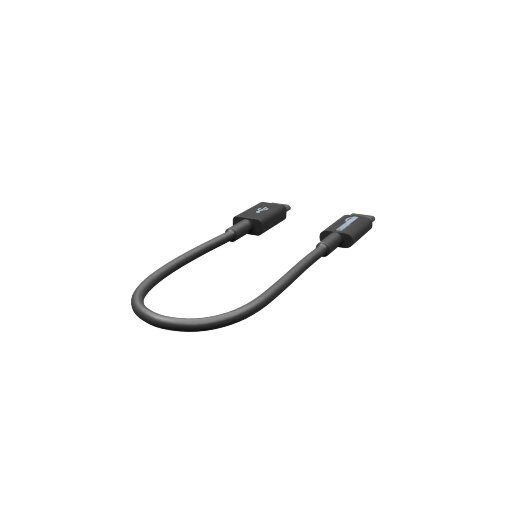} & \includegraphics[width=0.24\textwidth]{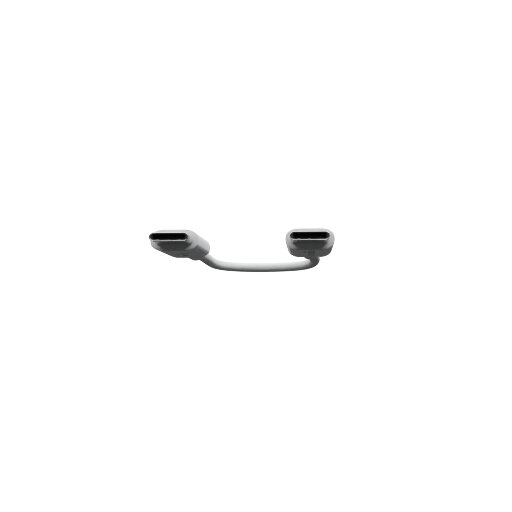} & \includegraphics[width=0.24\textwidth]{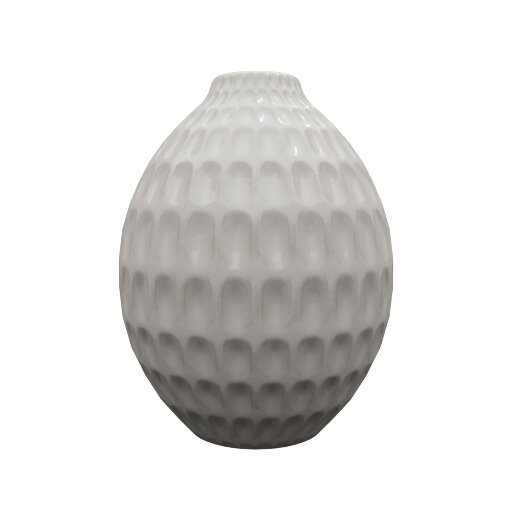} \\
    \begin{tabular}{@{}r@{: }l@{}} Predicted & 1.58 kg \\ Ground Truth & 1.58 kg \end{tabular} & \begin{tabular}{@{}r@{: }l@{}} Predicted & 0.01 kg \\ Ground Truth & 0.01 kg \end{tabular} & \begin{tabular}{@{}r@{: }l@{}} Predicted & 0.02 kg \\ Ground Truth & 0.01 kg \end{tabular} & \begin{tabular}{@{}r@{: }l@{}} Predicted & 0.71 kg \\ Ground Truth & 0.70 kg \end{tabular} \\
    \\
    \includegraphics[width=0.24\textwidth]{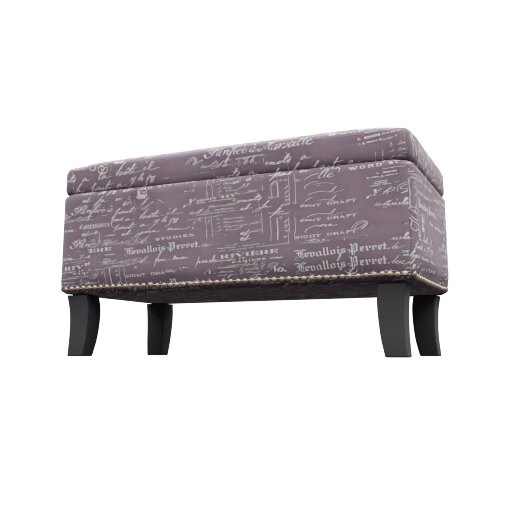} & \includegraphics[width=0.24\textwidth]{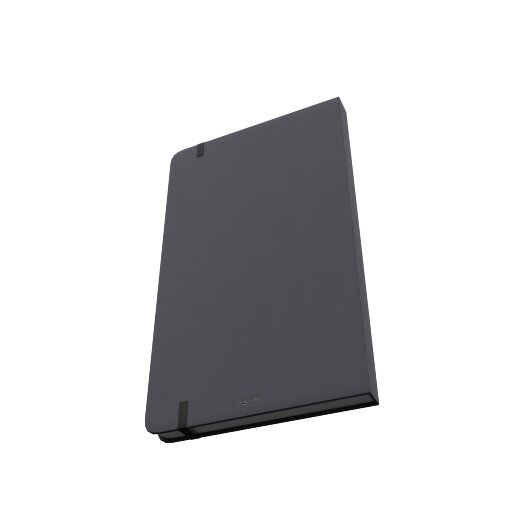} & \includegraphics[width=0.24\textwidth]{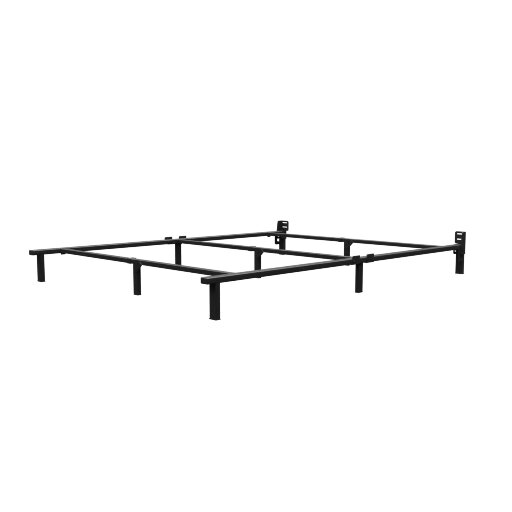} & \includegraphics[width=0.24\textwidth]{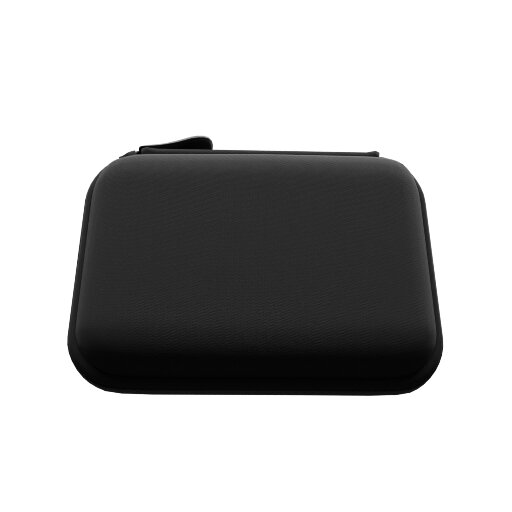} \\
    \begin{tabular}{@{}r@{: }l@{}} Predicted & 9.09 kg \\ Ground Truth & 9.07 kg \end{tabular} & \begin{tabular}{@{}r@{: }l@{}} Predicted & 0.37 kg \\ Ground Truth & 0.34 kg \end{tabular} & \begin{tabular}{@{}r@{: }l@{}} Predicted & 7.52 kg \\ Ground Truth & 7.54 kg \end{tabular} & \begin{tabular}{@{}r@{: }l@{}} Predicted & 0.14 kg \\ Ground Truth & 0.10 kg \end{tabular} \\
    \\
    \includegraphics[width=0.24\textwidth]{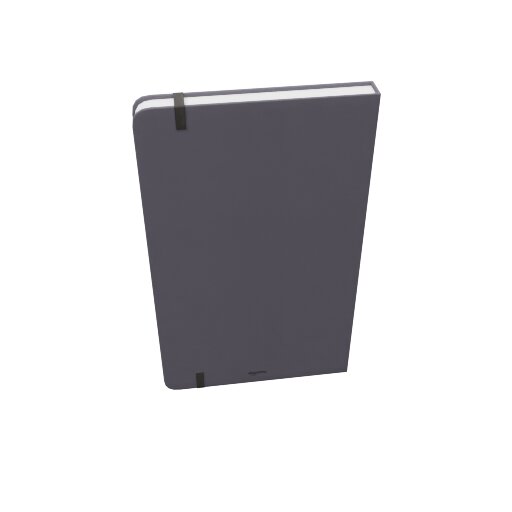} & \includegraphics[width=0.24\textwidth]{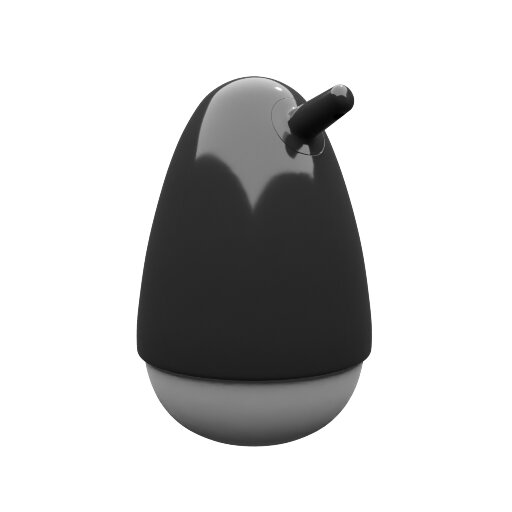} & \includegraphics[width=0.24\textwidth]{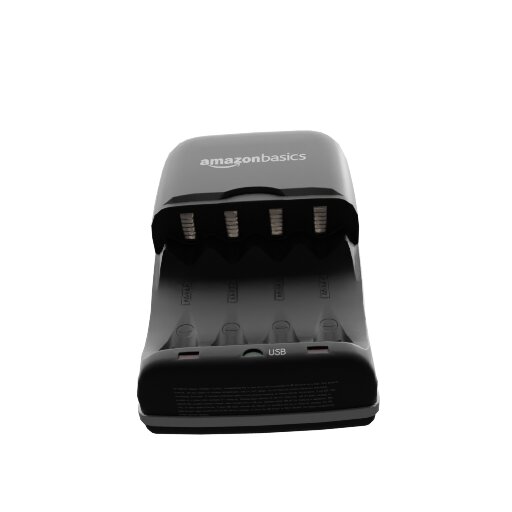} & \includegraphics[width=0.24\textwidth]{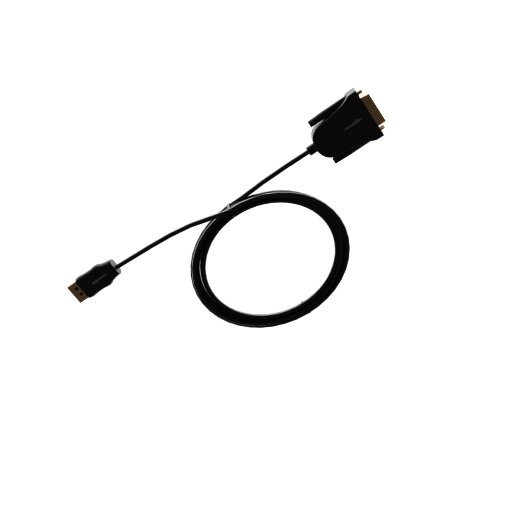} \\
    \begin{tabular}{@{}r@{: }l@{}} Predicted & 0.38 kg \\ Ground Truth & 0.32 kg \end{tabular} & \begin{tabular}{@{}r@{: }l@{}} Predicted & 0.25 kg \\ Ground Truth & 0.19 kg \end{tabular} & \begin{tabular}{@{}r@{: }l@{}} Predicted & 0.05 kg \\ Ground Truth & 0.11 kg \end{tabular} & \begin{tabular}{@{}r@{: }l@{}} Predicted & 0.19 kg \\ Ground Truth & 0.12 kg \end{tabular} \\
    \\
    \includegraphics[width=0.24\textwidth]{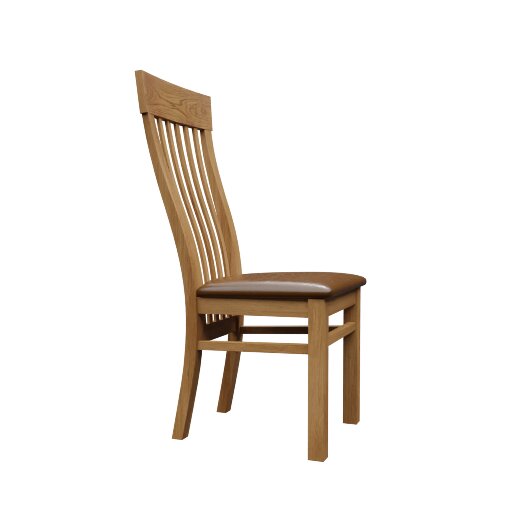} & \includegraphics[width=0.24\textwidth]{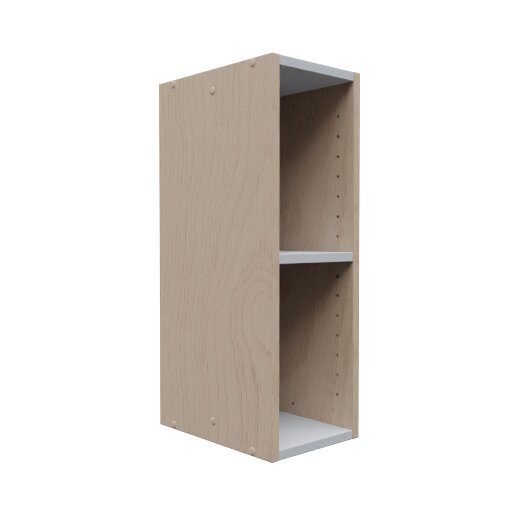} & \includegraphics[width=0.24\textwidth]{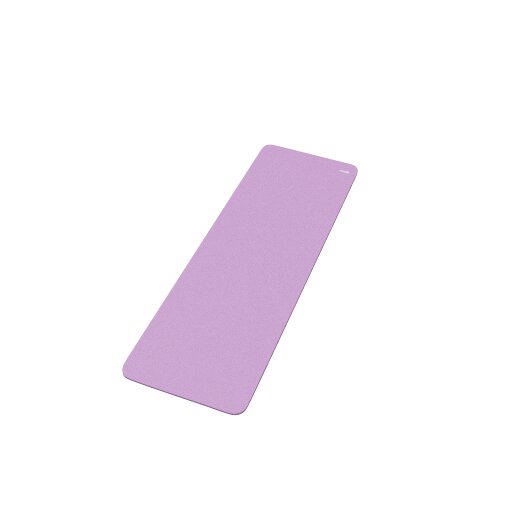} & \includegraphics[width=0.24\textwidth]{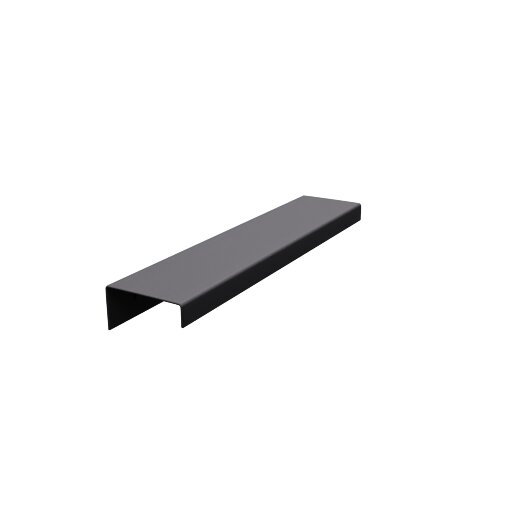} \\
    \begin{tabular}{@{}r@{: }l@{}} Predicted & 7.68 kg \\ Ground Truth & 7.60 kg \end{tabular} & \begin{tabular}{@{}r@{: }l@{}} Predicted & 4.56 kg \\ Ground Truth & 4.65 kg \end{tabular} & \begin{tabular}{@{}r@{: }l@{}} Predicted & 1.23 kg \\ Ground Truth & 1.12 kg \end{tabular} & \begin{tabular}{@{}r@{: }l@{}} Predicted & 0.80 kg \\ Ground Truth & 0.91 kg \end{tabular} \\
    \\
    \includegraphics[width=0.24\textwidth]{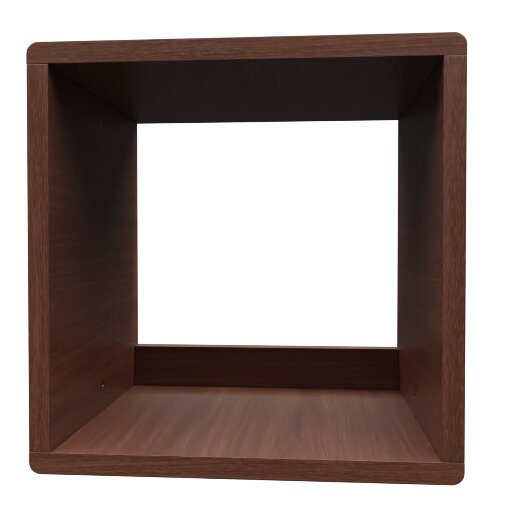} & \includegraphics[width=0.24\textwidth]{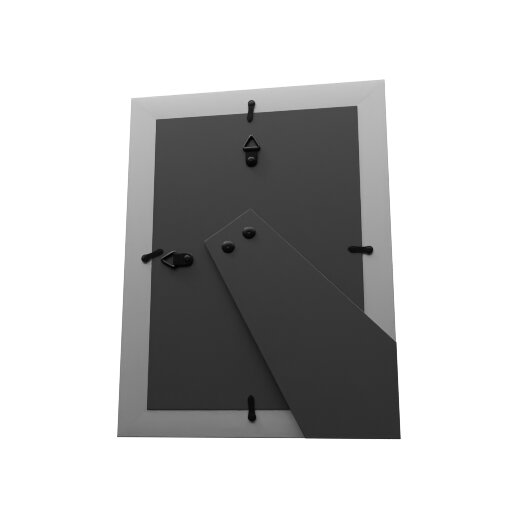} & \includegraphics[width=0.24\textwidth]{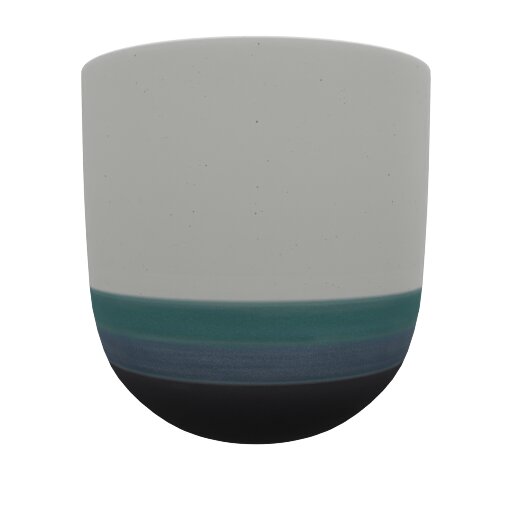} & \includegraphics[width=0.24\textwidth]{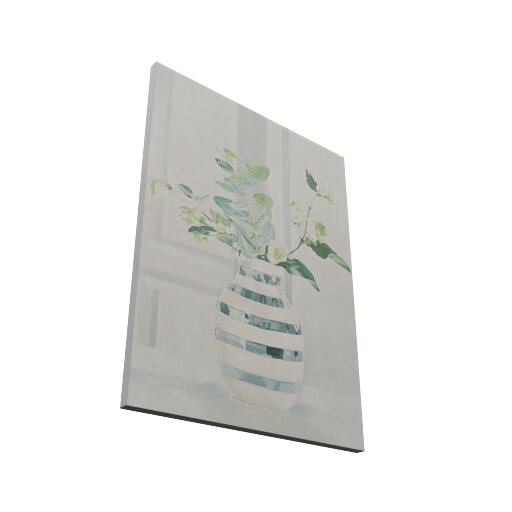} \\
    \begin{tabular}{@{}r@{: }l@{}} Predicted & 4.47 kg \\ Ground Truth & 4.36 kg \end{tabular} & \begin{tabular}{@{}r@{: }l@{}} Predicted & 0.42 kg \\ Ground Truth & 0.29 kg \end{tabular} & \begin{tabular}{@{}r@{: }l@{}} Predicted & 2.40 kg \\ Ground Truth & 2.27 kg \end{tabular} & \begin{tabular}{@{}r@{: }l@{}} Predicted & 2.85 kg \\ Ground Truth & 2.72 kg \end{tabular} \\
    \end{tabular}
    \caption{\textbf{Mass Estimation.} We show qualitative results of estimating mass from the ABO-500~\cite{Collins_2022_CVPR} dataset.}
    \label{fig:abo}
\end{figure*}

\else
    \begin{figure*}[tb]
    \centering
    \setlength{\tabcolsep}{0pt}%
    \begin{tabular}{ccccc}
    \includegraphics[width=0.2\textwidth]{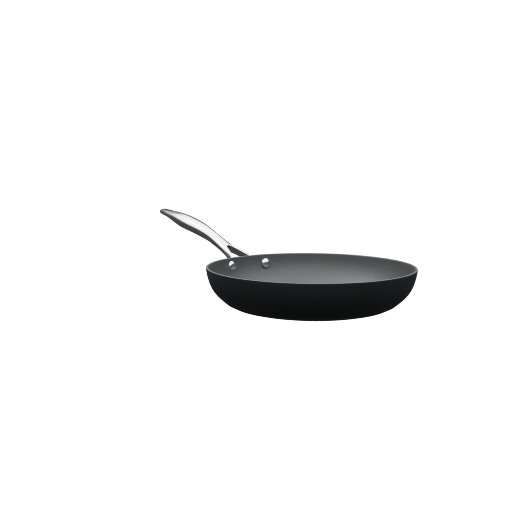} & \includegraphics[width=0.2\textwidth]{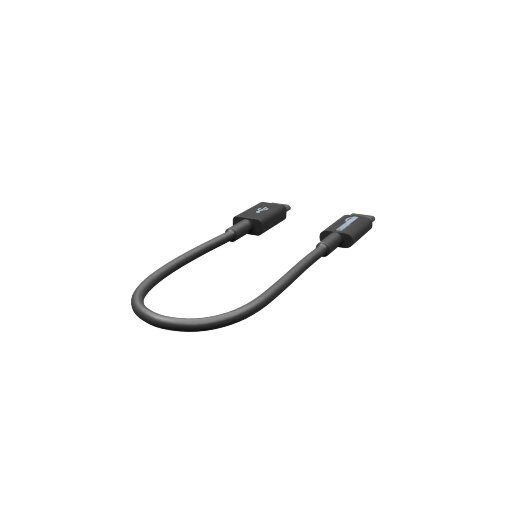} & \includegraphics[width=0.2\textwidth]{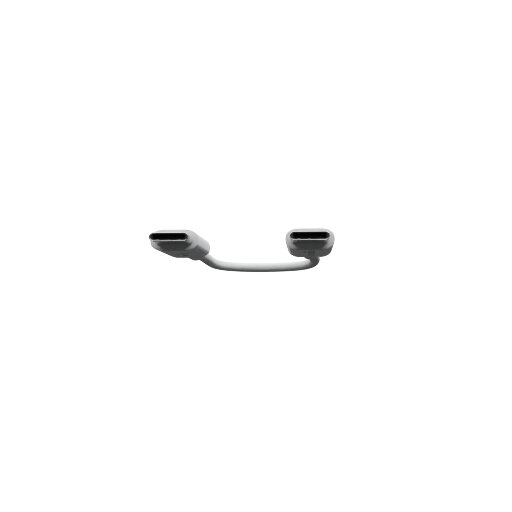} & \includegraphics[width=0.2\textwidth]{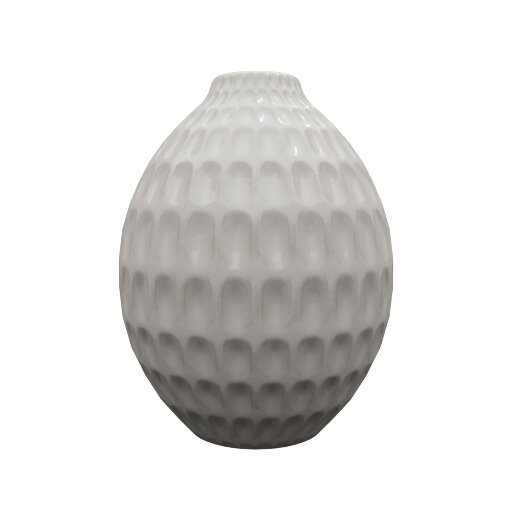} & \includegraphics[width=0.2\textwidth]{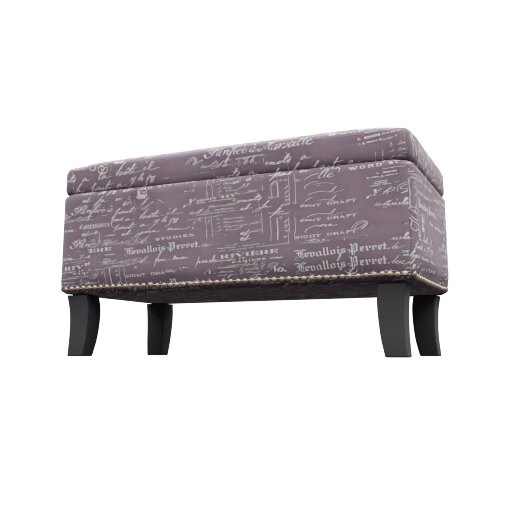} \\
    \begin{tabular}{@{}r@{: }l@{}} Predicted & 1.58 kg \\ Ground Truth & 1.58 kg \end{tabular} & \begin{tabular}{@{}r@{: }l@{}} Predicted & 0.01 kg \\ Ground Truth & 0.01 kg \end{tabular} & \begin{tabular}{@{}r@{: }l@{}} Predicted & 0.02 kg \\ Ground Truth & 0.01 kg \end{tabular} & \begin{tabular}{@{}r@{: }l@{}} Predicted & 0.71 kg \\ Ground Truth & 0.70 kg \end{tabular} & \begin{tabular}{@{}r@{: }l@{}} Predicted & 9.09 kg \\ Ground Truth & 9.07 kg \end{tabular} \\
    \\
    \includegraphics[width=0.2\textwidth]{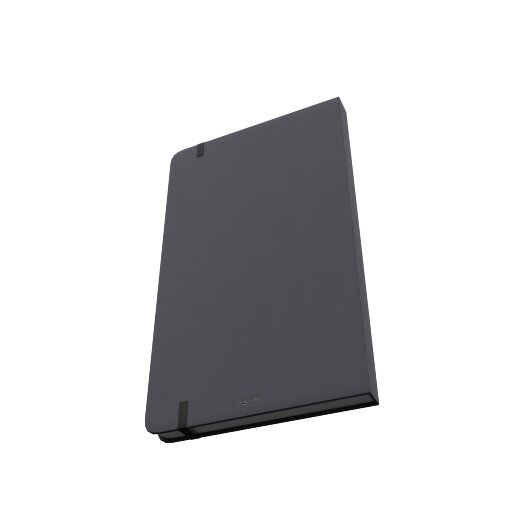} & \includegraphics[width=0.2\textwidth]{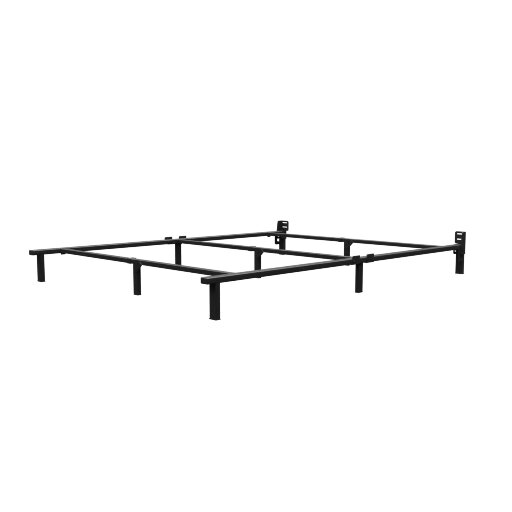} & \includegraphics[width=0.2\textwidth]{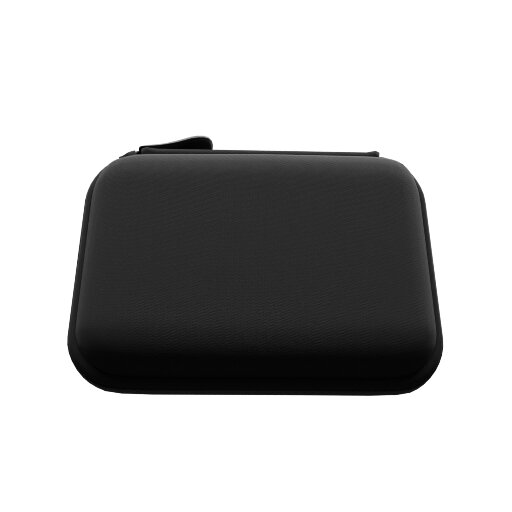} & \includegraphics[width=0.2\textwidth]{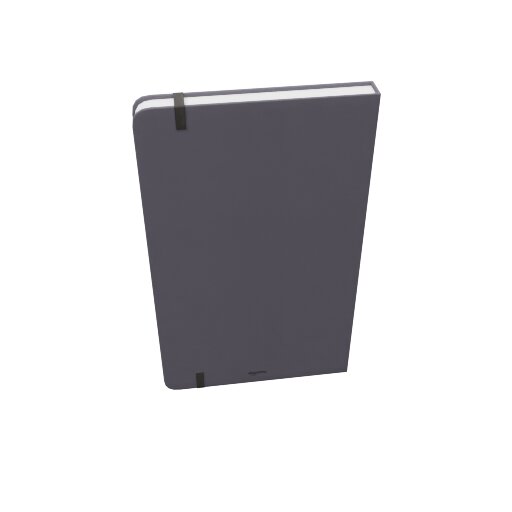} & \includegraphics[width=0.2\textwidth]{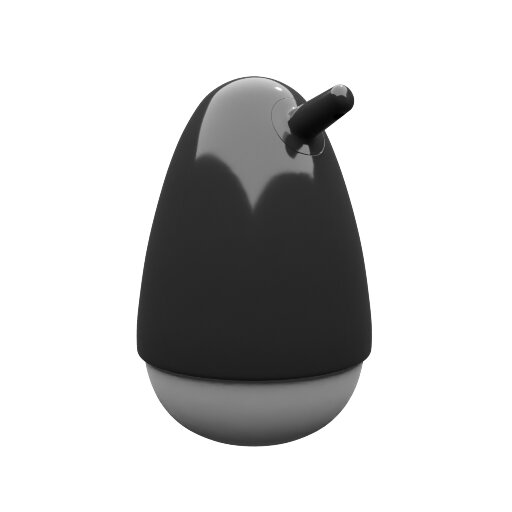} \\
    \begin{tabular}{@{}r@{: }l@{}} Predicted & 0.37 kg \\ Ground Truth & 0.34 kg \end{tabular} & \begin{tabular}{@{}r@{: }l@{}} Predicted & 7.52 kg \\ Ground Truth & 7.54 kg \end{tabular} & \begin{tabular}{@{}r@{: }l@{}} Predicted & 0.14 kg \\ Ground Truth & 0.10 kg \end{tabular} & \begin{tabular}{@{}r@{: }l@{}} Predicted & 0.38 kg \\ Ground Truth & 0.32 kg \end{tabular} & \begin{tabular}{@{}r@{: }l@{}} Predicted & 0.25 kg \\ Ground Truth & 0.19 kg \end{tabular} \\
    \\
    \includegraphics[width=0.2\textwidth]{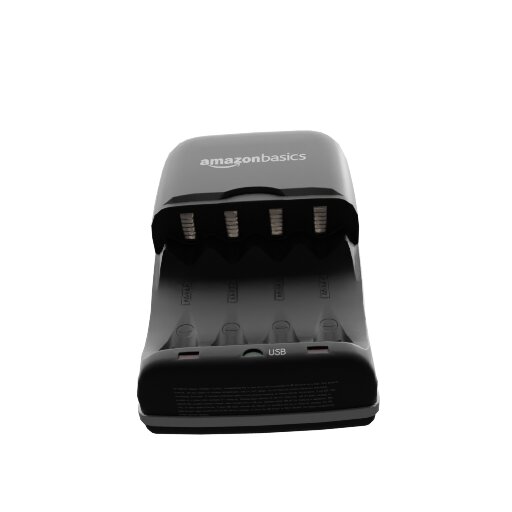} & \includegraphics[width=0.2\textwidth]{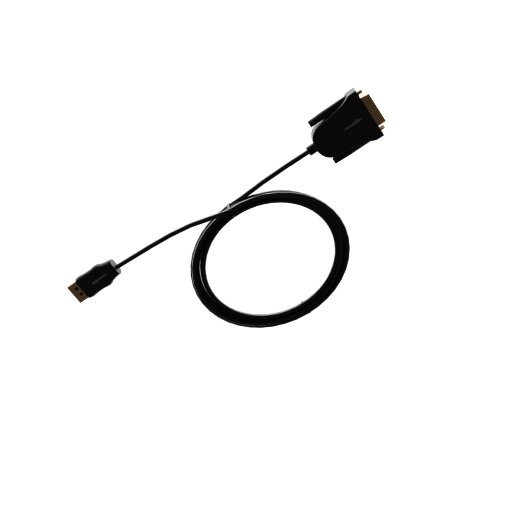} & \includegraphics[width=0.2\textwidth]{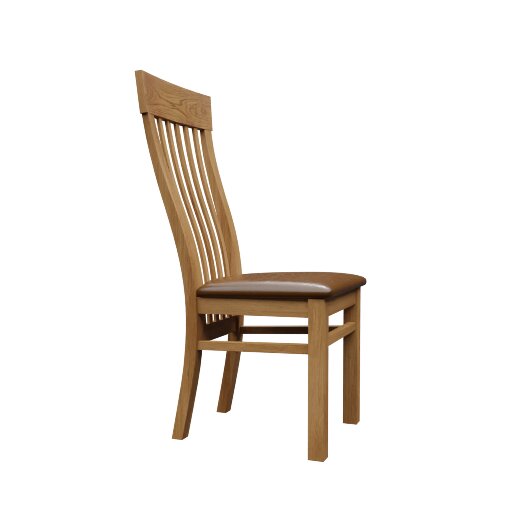} & \includegraphics[width=0.2\textwidth]{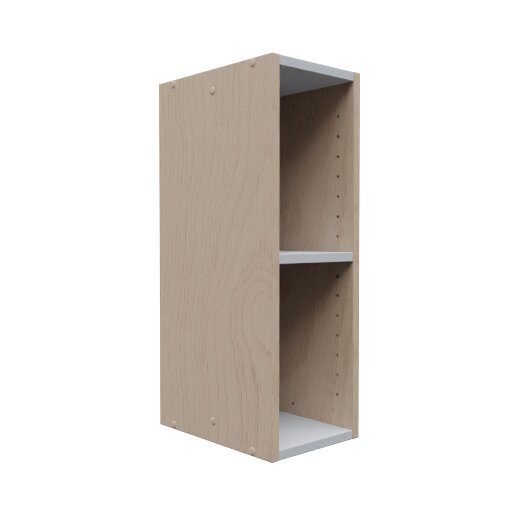} & \includegraphics[width=0.2\textwidth]{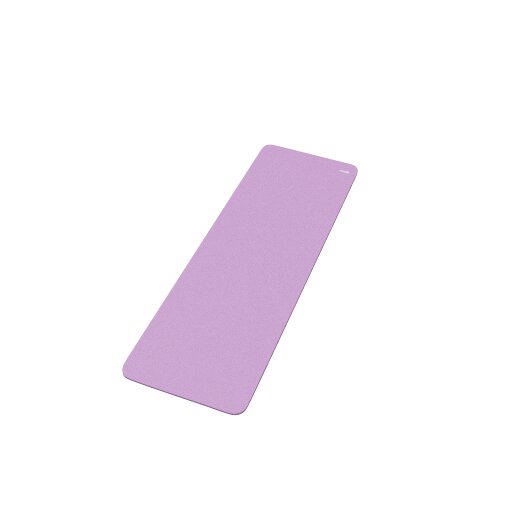} \\
    \begin{tabular}{@{}r@{: }l@{}} Predicted & 0.05 kg \\ Ground Truth & 0.11 kg \end{tabular} & \begin{tabular}{@{}r@{: }l@{}} Predicted & 0.19 kg \\ Ground Truth & 0.12 kg \end{tabular} & \begin{tabular}{@{}r@{: }l@{}} Predicted & 7.68 kg \\ Ground Truth & 7.60 kg \end{tabular} & \begin{tabular}{@{}r@{: }l@{}} Predicted & 4.56 kg \\ Ground Truth & 4.65 kg \end{tabular} & \begin{tabular}{@{}r@{: }l@{}} Predicted & 1.23 kg \\ Ground Truth & 1.12 kg \end{tabular} \\
    \\
    \includegraphics[width=0.2\textwidth]{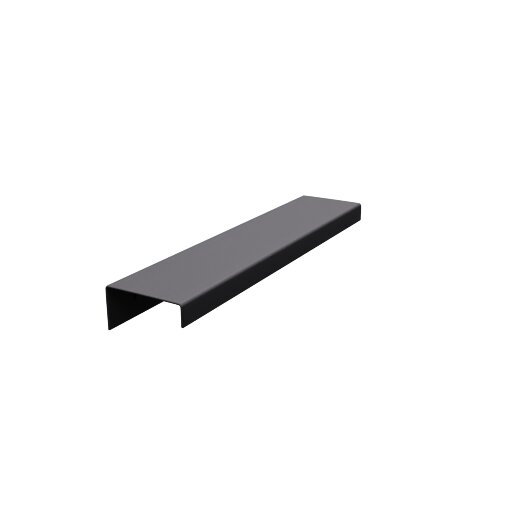} & \includegraphics[width=0.2\textwidth]{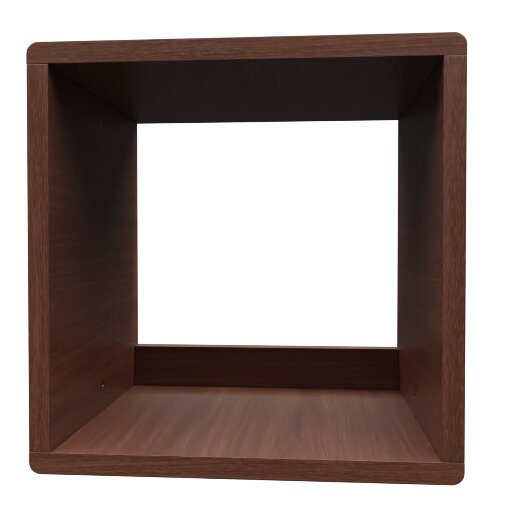} & \includegraphics[width=0.2\textwidth]{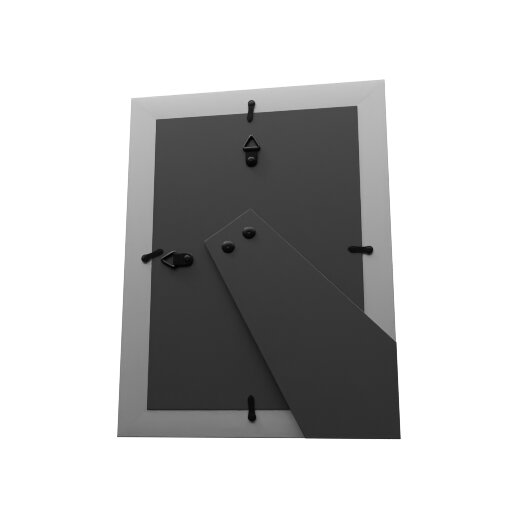} & \includegraphics[width=0.2\textwidth]{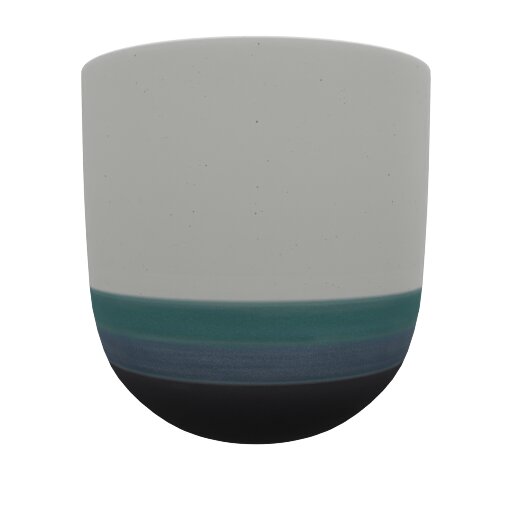} & \includegraphics[width=0.2\textwidth]{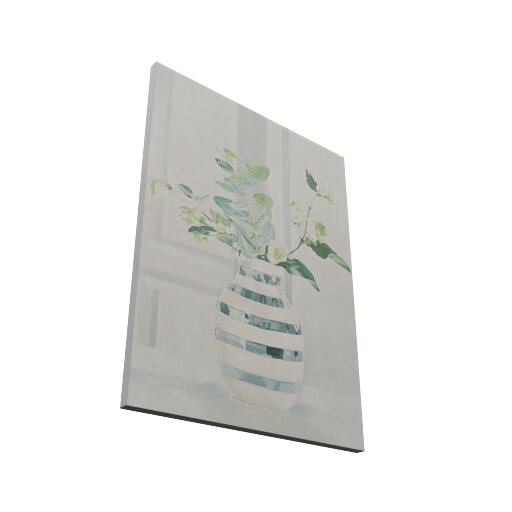} \\
    \begin{tabular}{@{}r@{: }l@{}} Predicted & 0.80 kg \\ Ground Truth & 0.91 kg \end{tabular} & \begin{tabular}{@{}r@{: }l@{}} Predicted & 4.47 kg \\ Ground Truth & 4.36 kg \end{tabular} & \begin{tabular}{@{}r@{: }l@{}} Predicted & 0.42 kg \\ Ground Truth & 0.29 kg \end{tabular} & \begin{tabular}{@{}r@{: }l@{}} Predicted & 2.40 kg \\ Ground Truth & 2.27 kg \end{tabular} & \begin{tabular}{@{}r@{: }l@{}} Predicted & 2.85 kg \\ Ground Truth & 2.72 kg \end{tabular} \\
    \end{tabular}
    \caption{\textbf{Mass Estimation.} We show qualitative results of estimating mass from the ABO-500~\cite{Collins_2022_CVPR} dataset.}
    \label{fig:abo}
\end{figure*}
\fi

\clearpage

\subsection{Additional MatVAE Results}\label{app:sec:matvae_results}

\subsubsection{Distribution learned by MatVAE}

We measure standard metrics used for measuring the difference in the distribution learned by MatVAE and the distribution of the test set in Tb.~\ref{tab:matvaedistmetrics}, observing small errors which suggest that MatVAE learned a good approximation of the true material distribution.

\newcommand{\matvaedistrow}{
0.0405 & 0.0798 & 0.1379 & 
0.0317 & 0.0437 & 0.0342 & 
0.0132 & 0.0172 & 0.0260
}
\vspace{1em}
\begin{table*}[tb]
\centering
\caption{\textbf{Distribution learned by MatVAE} compared to the distribution of MTD test set.}
\resizebox{0.8\textwidth}{!}{%
\begin{tabular}{rrrrrrrrr}
\toprule
\rowcolor{nvidiagreen!15}
\multicolumn{3}{c}{Young's Modulus ($E$)}
& \multicolumn{3}{c}{Poisson's Ratio ($\nu$)}
& \multicolumn{3}{c}{Density ($\rho$)}\\ 
\cmidrule(lr){1-3}\cmidrule(lr){4-6}\cmidrule(lr){7-9}
\rowcolor{nvidiagreen!15}W$_1$ ($\downarrow$) & W$_2$ ($\downarrow$) & $D_{\text{KL}}$ ($\downarrow$) & W$_1$ ($\downarrow$) & W$_2$ ($\downarrow$) & $D_{\text{KL}}$ ($\downarrow$) & W$_1$ ($\downarrow$) & W$_2$ ($\downarrow$) & $D_{\text{KL}}$ ($\downarrow$) \\
\midrule
\matvaedistrow\\
\bottomrule
\end{tabular}
}
\label{tab:matvaedistmetrics}
\end{table*}

\subsubsection{Moving Across MatVAE Latent Space}

We show an example of moving across the latent space in \Cref{fig:app:matvae_traverse}. For each setting, we take a point in our latent space and move across both of its dimensions to obtain multiple smoothly varying material properties. We apply these properties to a bunny and simulate dropping it to the ground with a FEM simulator and these various material properties. The color plots show the average displacements of the mesh from its rest state across simulation steps (e.g.~\Cref{fig:grid_cubes_fall_setting1}), demonstrating that even the actual physical behavior correlates with the dimensions of the latent space. Thus, we find that MatVAE learns a rich, meaningful latent space with smooth interpolation and ensures generating physically valid material triplets.

\subsubsection{Interpolation with MatVAE}

We show additional examples of interpolating in the MatVAE latent space (\Cref{sec:reconstructtriplet}) in~\Cref{fig:interp_pair2}.

\vspace{1em}
\begin{figure*}[hb!]
    \centering
    \resizebox{0.5\textwidth}{!}{%
    % \scriptsize
    % \resizebox{0.75\textwidth}{!}{%
    \setlength{\tabcolsep}{2pt}
    \begin{tabular}{l>{\columncolor{nvidiagreen_light}}c|ccc|c}
    \rowcolor{nvidiagreen}
    & \textbf{Start Value} & \multicolumn{3}{c|}{\textbf{MatVAE Interpolation}}  & \textbf{End Value}\\
    & Styrofoam &
    Vanadium &
    Rhodium &
    Tungsten &
    Osmium\\
    $E$ & $0.002$ GPa & $128.0$ GPa & $275.0$ GPa &
    $411.0$ GPa & $550.0$ GPa\\
    $\nu$ & $0.33$ & $0.37$ & $0.30$ &
    $0.28$ & $0.25$\\
    $\rho$ & $25$ kg/m$^3$ & $6110$ kg/m$^3$ &
    $12450$ kg/m$^3$ & $19250$ kg/m$^3$ &
    $22570$ kg/m$^3$\\
    \Xhline{2pt}
    \rowcolor{nvidiagreen}
    & \textbf{Start Value} & \multicolumn{3}{c|}{\textbf{Naive Interpolation}} & \textbf{End Value} \\
    & Styrofoam &
    {\color{red}Invalid (\xmark)} &
    {\color{red}Invalid (\xmark)} &
    {\color{red}Invalid (\xmark)} &
    Osmium\\
    $E$ & $0.002$ GPa & $137.5$ GPa & $275.0$ GPa &
    $412.5$ GPa & $550.0$ GPa\\
    $\nu$ & $0.33$ & $0.31$ & $0.29$ &
    $0.27$ & $0.25$\\
    $\rho$ & $25$ kg/m$^3$ & $5661$ kg/m$^3$ &
    $11298$ kg/m$^3$ & $16934$ kg/m$^3$ &
    $22570$ kg/m$^3$\\
    \end{tabular}}
    \caption{\textbf{Interpolating in MatVAE latent space:} an additional example of interpolation, complementary to \Cref{fig:interp_pair1}.}
    \label{fig:interp_pair2}
\end{figure*}

\ifdefined\iclr
    \begin{figure*}[t!]
	\centering

\begin{minipage}[t]{0.98\textwidth}
     \vspace*{0pt}
        \centering
        \resizebox{\textwidth}{!}{
        \begin{tabular}{lllrrrrrr}
\toprule
\rowcolor{nvidiagreen!15}Setting & Position & Material & \multicolumn{2}{c}{Young's Modulus (Pa)} & \multicolumn{2}{c}{Poisson's Ratio} & \multicolumn{2}{c}{Density (kg/m$^3$)} \\
\cmidrule(lr){4-5} \cmidrule(lr){6-7} \cmidrule(lr){8-9}
 \rowcolor{nvidiagreen!15}& & & Interpolated & True Range & Interpolated & True Range & Interpolated & True Range \\
\midrule
\Cref{fig:grid_cubes_fall_setting1} & Top-left ($\nwarrow$) & Aerographite & $4.4 \times 10^5$ & $1.0 \times 10^5$ -- $1.0 \times 10^6$ & 0.241 & 0.2--0.3 & 0.2 & 0.2--0.2 \\
 & Top-right ($\nearrow$) & Polyurethane Foam & $4.8 \times 10^6$ & $1.0 \times 10^5$ -- $5.0 \times 10^6$ & 0.304 & 0.30--0.30 & 298.2.0 & 50--300 \\
 & Bottom-left ($\swarrow$) & Rubber (soft) & $3.1 \times 10^6$ & $3.0 \times 10^6$ -- $5.0 \times 10^6$ & 0.488 & 0.48--0.50 & 952.0 & 950--950 \\
 & Bottom-right ($\searrow$) & Styrofoam & $1.6 \times 10^6$ & $1.0 \times 10^6$ -- $3.0 \times 10^6$ & 0.322 & 0.3--0.35 & 22.6 & 15--35 \\
\midrule
\Cref{fig:grid_cubes_fall_setting2} & Top-left ($\nwarrow$) & Aerogel & $4.4 \times 10^6$ & $1.0 \times 10^6$ -- $1.0 \times 10^7$ & 0.257 & 0.2--0.3 & 1.0 & 1.0--1.0 \\
 & Top-right ($\nearrow$) & Neoprene & $1.0 \times 10^7$ & $1.0 \times 10^6$ -- $1.0 \times 10^7$ & 0.494 & 0.45--0.5 & 1232.0 & 1230--1250 \\
 & Bottom-left ($\swarrow$) & EPDM Rubber & $6.6 \times 10^6$ & $5.0 \times 10^6$ -- $1.0 \times 10^7$ & 0.488 & 0.49--0.49 & 1100.9 & 1100--1100 \\
 & Bottom-right ($\searrow$) & Flexible PVC (Plasticized) & $4.8 \times 10^7$ & $2.0 \times 10^7$ -- $1.0 \times 10^8$ & 0.450 & 0.45--0.45 & 1209.5 & 1200--1400 \\
\midrule
\Cref{fig:grid_cubes_fall_setting3} & Top-left ($\nwarrow$) & Polystyrene Foam (EPS) & $2.6 \times 10^6$ & $1.0 \times 10^6$ -- $5.0 \times 10^6$ & 0.104 & 0.10--0.10 & 59.1 & 30--100 \\
 & Top-right ($\nearrow$) & Chloroprene Rubber (Neoprene) & $5.0 \times 10^6$ & $5.0 \times 10^6$ -- $5.0 \times 10^6$ & 0.490 & 0.49--0.49 & 1200.8 & 1200--1200 \\
 & Bottom-left ($\swarrow$) & Polystyrene (Foam) & $5.8 \times 10^6$ & $2.5 \times 10^6$ -- $7.0 \times 10^6$ & 0.371 & 0.34--0.4 & 34.8 & 15--35 \\
 & Bottom-right ($\searrow$) & Polybutylene (PB) & $2.5 \times 10^8$ & $2.5 \times 10^8$ -- $3.0 \times 10^8$ & 0.400 & 0.4--0.42 & 932.0 & 930--950 \\
\bottomrule
\end{tabular}
}
        \subcaption{Corner Materials for our experiments on moving across the latent space (\Cref{fig:grid_cubes_fall_setting1,fig:grid_cubes_fall_setting2,fig:grid_cubes_fall_setting3}).}
    \end{minipage} \\
\begin{minipage}[t]{0.31\textwidth}

\setlength{\tabcolsep}{0pt}%
    \renewcommand{\arraystretch}{0}%
    \newlength{\arrowcol}
    \setlength{\arrowcol}{0.08\columnwidth} % width of the arrow column
    \newlength{\cellsize}
    \setlength{\cellsize}{\dimexpr(\columnwidth-\arrowcol)/5\relax}
    \newcommand{\squarecell}[2]{%
      \cellcolor{#1}{%
        \color{white}\parbox[c][\cellsize][c]{\cellsize}{\centering #2}%
      }%
    }
    \resizebox{\linewidth}{!}{%
    \begin{tabular}{c c c c c c}
    % Top header row
    & \multicolumn{5}{c}{\bigarrowleftright{Latent Dimension 1}} \\
    % Left multirow rotated label + data rows
    \multirow{5}{*}{\rotatebox{90}{\parbox{6.cm}{\centering\bigarrowleftright{Latent Dimension 2}}}}
    & \cellcolor[rgb]{0.993,0.906,0.144} \raisebox{0pt}[0.7cm][0.7cm]{\centering{$2.23 \times 10^{-1}$}} 
    & \cellcolor[rgb]{0.148,0.512,0.557} \raisebox{0pt}[0.7cm][0.7cm]{\centering\textcolor{white}{$1.37 \times 10^{-1}$}} 
    & \cellcolor[rgb]{0.207,0.372,0.553} \raisebox{0pt}[0.7cm][0.7cm]{\centering\textcolor{white}{$1.14 \times 10^{-1}$}} 
    & \cellcolor[rgb]{0.252,0.270,0.532} \raisebox{0pt}[0.7cm][0.7cm]{\centering\textcolor{white}{$1.00 \times 10^{-1}$}} 
    & \cellcolor[rgb]{0.267,0.005,0.329} \raisebox{0pt}[0.7cm][0.7cm]{\centering\textcolor{white}{$6.85 \times 10^{-2}$}} \\
    & \cellcolor[rgb]{0.158,0.684,0.502} \raisebox{0pt}[0.7cm][0.7cm]{\centering\textcolor{white}{$1.65 \times 10^{-1}$}} 
    & \cellcolor[rgb]{0.720,0.870,0.163} \raisebox{0pt}[0.7cm][0.7cm]{\centering\textcolor{white}{$2.06 \times 10^{-1}$}} 
    & \cellcolor[rgb]{0.168,0.460,0.558} \raisebox{0pt}[0.7cm][0.7cm]{\centering\textcolor{white}{$1.29 \times 10^{-1}$}} 
    & \cellcolor[rgb]{0.226,0.331,0.547} \raisebox{0pt}[0.7cm][0.7cm]{\centering\textcolor{white}{$1.08 \times 10^{-1}$}} 
    & \cellcolor[rgb]{0.256,0.261,0.528} \raisebox{0pt}[0.7cm][0.7cm]{\centering\textcolor{white}{$9.86 \times 10^{-2}$}} \\
    & \cellcolor[rgb]{0.167,0.464,0.558} \raisebox{0pt}[0.7cm][0.7cm]{\centering\textcolor{white}{$1.29 \times 10^{-1}$}} 
    & \cellcolor[rgb]{0.142,0.526,0.556} \raisebox{0pt}[0.7cm][0.7cm]{\centering\textcolor{white}{$1.39 \times 10^{-1}$}} 
    & \cellcolor[rgb]{0.369,0.789,0.383} \raisebox{0pt}[0.7cm][0.7cm]{\centering\textcolor{white}{$1.84 \times 10^{-1}$}} 
    & \cellcolor[rgb]{0.184,0.422,0.557} \raisebox{0pt}[0.7cm][0.7cm]{\centering\textcolor{white}{$1.22 \times 10^{-1}$}} 
    & \cellcolor[rgb]{0.241,0.296,0.540} \raisebox{0pt}[0.7cm][0.7cm]{\centering\textcolor{white}{$1.04 \times 10^{-1}$}} \\
    & \cellcolor[rgb]{0.205,0.376,0.554} \raisebox{0pt}[0.7cm][0.7cm]{\centering\textcolor{white}{$1.15 \times 10^{-1}$}} 
    & \cellcolor[rgb]{0.184,0.422,0.557} \raisebox{0pt}[0.7cm][0.7cm]{\centering\textcolor{white}{$1.23 \times 10^{-1}$}} 
    & \cellcolor[rgb]{0.156,0.490,0.558} \raisebox{0pt}[0.7cm][0.7cm]{\centering\textcolor{white}{$1.33 \times 10^{-1}$}} 
    & \cellcolor[rgb]{0.123,0.585,0.547} \raisebox{0pt}[0.7cm][0.7cm]{\centering\textcolor{white}{$1.49 \times 10^{-1}$}} 
    & \cellcolor[rgb]{0.201,0.384,0.554} \raisebox{0pt}[0.7cm][0.7cm]{\centering\textcolor{white}{$1.17 \times 10^{-1}$}} \\
    & \cellcolor[rgb]{0.236,0.310,0.543} \raisebox{0pt}[0.7cm][0.7cm]{\centering\textcolor{white}{$1.05 \times 10^{-1}$}} 
    & \cellcolor[rgb]{0.218,0.347,0.550} \raisebox{0pt}[0.7cm][0.7cm]{\centering\textcolor{white}{$1.11 \times 10^{-1}$}} 
    & \cellcolor[rgb]{0.199,0.388,0.555} \raisebox{0pt}[0.7cm][0.7cm]{\centering\textcolor{white}{$1.17 \times 10^{-1}$}} 
    & \cellcolor[rgb]{0.173,0.449,0.558} \raisebox{0pt}[0.7cm][0.7cm]{\centering\textcolor{white}{$1.26 \times 10^{-1}$}} 
    & \cellcolor[rgb]{0.148,0.512,0.557} \raisebox{0pt}[0.7cm][0.7cm]{\centering\textcolor{white}{$1.37 \times 10^{-1}$}} \\
    \end{tabular}}
    \subcaption{Setting 1.}
    \label{fig:grid_cubes_fall_setting1}
\end{minipage}
\begin{minipage}[t]{0.31\textwidth}
        \setlength{\tabcolsep}{0pt}%
        \setlength{\arrowcol}{0.08\columnwidth} % width of the arrow column
        \setlength{\cellsize}{\dimexpr(\columnwidth-\arrowcol)/5\relax}
        \resizebox{\linewidth}{!}{%
        \begin{tabular}{c c c c c c}
        % Top header row
        & \multicolumn{5}{c}{\bigarrowleftright{Latent Dimension 1}} \\
        % Left multirow rotated label + data rows
        \multirow{5}{*}{\rotatebox{90}{\parbox{6.cm}{\centering\bigarrowleftright{Latent Dimension 2}}}}
&
    \cellcolor[rgb]{0.993,0.906,0.144} \raisebox{0pt}[0.7cm][0.7cm]{\centering{$1.66 \times 10^{-1}$}} &
    \cellcolor[rgb]{0.147,0.673,0.509} \raisebox{0pt}[0.7cm][0.7cm]{\centering\textcolor{white}{$1.11 \times 10^{-1}$}} &
    \cellcolor[rgb]{0.143,0.669,0.511} \raisebox{0pt}[0.7cm][0.7cm]{\centering\textcolor{white}{$1.10 \times 10^{-1}$}} &
    \cellcolor[rgb]{0.140,0.666,0.513} \raisebox{0pt}[0.7cm][0.7cm]{\centering\textcolor{white}{$1.10 \times 10^{-1}$}} &
    \cellcolor[rgb]{0.130,0.651,0.522} \raisebox{0pt}[0.7cm][0.7cm]{\centering\textcolor{white}{$1.08 \times 10^{-1}$}} \\
    &
    \cellcolor[rgb]{0.158,0.684,0.502} \raisebox{0pt}[0.7cm][0.7cm]{\centering\textcolor{white}{$1.13 \times 10^{-1}$}} &
    \cellcolor[rgb]{0.128,0.648,0.523} \raisebox{0pt}[0.7cm][0.7cm]{\centering\textcolor{white}{$1.07 \times 10^{-1}$}} &
    \cellcolor[rgb]{0.121,0.593,0.545} \raisebox{0pt}[0.7cm][0.7cm]{\centering\textcolor{white}{$9.85 \times 10^{-2}$}} &
    \cellcolor[rgb]{0.123,0.585,0.547} \raisebox{0pt}[0.7cm][0.7cm]{\centering\textcolor{white}{$9.75 \times 10^{-2}$}} &
    \cellcolor[rgb]{0.124,0.578,0.548} \raisebox{0pt}[0.7cm][0.7cm]{\centering\textcolor{white}{$9.65 \times 10^{-2}$}} \\
    &
    \cellcolor[rgb]{0.158,0.684,0.502} \raisebox{0pt}[0.7cm][0.7cm]{\centering\textcolor{white}{$1.13 \times 10^{-1}$}} &
    \cellcolor[rgb]{0.126,0.644,0.525} \raisebox{0pt}[0.7cm][0.7cm]{\centering\textcolor{white}{$1.06 \times 10^{-1}$}} &
    \cellcolor[rgb]{0.125,0.574,0.549} \raisebox{0pt}[0.7cm][0.7cm]{\centering\textcolor{white}{$9.57 \times 10^{-2}$}} &
    \cellcolor[rgb]{0.143,0.523,0.556} \raisebox{0pt}[0.7cm][0.7cm]{\centering\textcolor{white}{$8.76 \times 10^{-2}$}} &
    \cellcolor[rgb]{0.158,0.486,0.558} \raisebox{0pt}[0.7cm][0.7cm]{\centering\textcolor{white}{$8.19 \times 10^{-2}$}} \\
    &
    \cellcolor[rgb]{0.147,0.673,0.509} \raisebox{0pt}[0.7cm][0.7cm]{\centering\textcolor{white}{$1.11 \times 10^{-1}$}} &
    \cellcolor[rgb]{0.121,0.626,0.533} \raisebox{0pt}[0.7cm][0.7cm]{\centering\textcolor{white}{$1.04 \times 10^{-1}$}} &
    \cellcolor[rgb]{0.134,0.549,0.554} \raisebox{0pt}[0.7cm][0.7cm]{\centering\textcolor{white}{$9.19 \times 10^{-2}$}} &
    \cellcolor[rgb]{0.164,0.471,0.558} \raisebox{0pt}[0.7cm][0.7cm]{\centering\textcolor{white}{$7.99 \times 10^{-2}$}} &
    \cellcolor[rgb]{0.280,0.171,0.480} \raisebox{0pt}[0.7cm][0.7cm]{\centering\textcolor{white}{$3.97 \times 10^{-2}$}} \\
    &
    \cellcolor[rgb]{0.143,0.669,0.511} \raisebox{0pt}[0.7cm][0.7cm]{\centering\textcolor{white}{$1.10 \times 10^{-1}$}} &
    \cellcolor[rgb]{0.122,0.589,0.546} \raisebox{0pt}[0.7cm][0.7cm]{\centering\textcolor{white}{$9.78 \times 10^{-2}$}} &
    \cellcolor[rgb]{0.135,0.545,0.554} \raisebox{0pt}[0.7cm][0.7cm]{\centering\textcolor{white}{$9.13 \times 10^{-2}$}} &
    \cellcolor[rgb]{0.274,0.200,0.499} \raisebox{0pt}[0.7cm][0.7cm]{\centering\textcolor{white}{$4.35 \times 10^{-2}$}} &
    \cellcolor[rgb]{0.267,0.005,0.329} \raisebox{0pt}[0.7cm][0.7cm]{\centering\textcolor{white}{$2.21 \times 10^{-2}$}} \\
    \end{tabular}}
    \subcaption{Setting 2.}\label{fig:grid_cubes_fall_setting2}
\end{minipage}
\begin{minipage}[t]{0.31\textwidth}
        \centering
        \setlength{\tabcolsep}{0pt}%
        \renewcommand{\arraystretch}{0}%
        \setlength{\arrowcol}{0.08\columnwidth} % width of the arrow column
        \setlength{\cellsize}{\dimexpr(\columnwidth-\arrowcol)/5\relax}
        \resizebox{\linewidth}{!}{%
        \begin{tabular}{c c c c c c}
        % Top header row
        & \multicolumn{5}{c}{\bigarrowleftright{Latent Dimension 1}} \\
        % Left multirow rotated label + data rows
        \multirow{5}{*}{\rotatebox{90}{\parbox{6.cm}{\centering\bigarrowleftright{Latent Dimension 2}}}}
&
    \cellcolor[rgb]{0.993,0.906,0.144} \raisebox{0pt}[0.7cm][0.7cm]{\centering{$2.25 \times 10^{-1}$}} &
    \cellcolor[rgb]{0.993,0.906,0.144} \raisebox{0pt}[0.7cm][0.7cm]{\centering{$2.25 \times 10^{-1}$}} &
    \cellcolor[rgb]{0.214,0.722,0.470} \raisebox{0pt}[0.7cm][0.7cm]{\centering\textcolor{white}{$1.56 \times 10^{-1}$}} &
    \cellcolor[rgb]{0.176,0.698,0.491} \raisebox{0pt}[0.7cm][0.7cm]{\centering\textcolor{white}{$1.51 \times 10^{-1}$}} &
    \cellcolor[rgb]{0.166,0.691,0.497} \raisebox{0pt}[0.7cm][0.7cm]{\centering\textcolor{white}{$1.49 \times 10^{-1}$}} \\
    &
    \cellcolor[rgb]{0.404,0.800,0.363} \raisebox{0pt}[0.7cm][0.7cm]{\centering\textcolor{white}{$1.77 \times 10^{-1}$}} &
    \cellcolor[rgb]{0.130,0.560,0.552} \raisebox{0pt}[0.7cm][0.7cm]{\centering\textcolor{white}{$1.20 \times 10^{-1}$}} &
    \cellcolor[rgb]{0.146,0.515,0.557} \raisebox{0pt}[0.7cm][0.7cm]{\centering\textcolor{white}{$1.10 \times 10^{-1}$}} &
    \cellcolor[rgb]{0.228,0.327,0.547} \raisebox{0pt}[0.7cm][0.7cm]{\centering\textcolor{white}{$7.01 \times 10^{-2}$}} &
    \cellcolor[rgb]{0.283,0.111,0.432} \raisebox{0pt}[0.7cm][0.7cm]{\centering\textcolor{white}{$3.30 \times 10^{-2}$}} \\
    &
    \cellcolor[rgb]{0.208,0.719,0.473} \raisebox{0pt}[0.7cm][0.7cm]{\centering\textcolor{white}{$1.56 \times 10^{-1}$}} &
    \cellcolor[rgb]{0.150,0.504,0.557} \raisebox{0pt}[0.7cm][0.7cm]{\centering\textcolor{white}{$1.07 \times 10^{-1}$}} &
    \cellcolor[rgb]{0.277,0.050,0.376} \raisebox{0pt}[0.7cm][0.7cm]{\centering\textcolor{white}{$2.38 \times 10^{-2}$}} &
    \cellcolor[rgb]{0.277,0.050,0.376} \raisebox{0pt}[0.7cm][0.7cm]{\centering\textcolor{white}{$2.37 \times 10^{-2}$}} &
    \cellcolor[rgb]{0.276,0.044,0.370} \raisebox{0pt}[0.7cm][0.7cm]{\centering\textcolor{white}{$2.27 \times 10^{-2}$}} \\
    &
    \cellcolor[rgb]{0.120,0.618,0.536} \raisebox{0pt}[0.7cm][0.7cm]{\centering\textcolor{white}{$1.33 \times 10^{-1}$}} &
    \cellcolor[rgb]{0.280,0.166,0.476} \raisebox{0pt}[0.7cm][0.7cm]{\centering\textcolor{white}{$4.18 \times 10^{-2}$}} &
    \cellcolor[rgb]{0.277,0.050,0.376} \raisebox{0pt}[0.7cm][0.7cm]{\centering\textcolor{white}{$2.37 \times 10^{-2}$}} &
    \cellcolor[rgb]{0.269,0.010,0.335} \raisebox{0pt}[0.7cm][0.7cm]{\centering\textcolor{white}{$1.82 \times 10^{-2}$}} &
    \cellcolor[rgb]{0.269,0.010,0.335} \raisebox{0pt}[0.7cm][0.7cm]{\centering\textcolor{white}{$1.80 \times 10^{-2}$}} \\
    &
    \cellcolor[rgb]{0.121,0.593,0.545} \raisebox{0pt}[0.7cm][0.7cm]{\centering\textcolor{white}{$1.27 \times 10^{-1}$}} &
    \cellcolor[rgb]{0.278,0.056,0.381} \raisebox{0pt}[0.7cm][0.7cm]{\centering\textcolor{white}{$2.44 \times 10^{-2}$}} &
    \cellcolor[rgb]{0.270,0.015,0.341} \raisebox{0pt}[0.7cm][0.7cm]{\centering\textcolor{white}{$1.89 \times 10^{-2}$}} &
    \cellcolor[rgb]{0.267,0.005,0.329} \raisebox{0pt}[0.7cm][0.7cm]{\centering\textcolor{white}{$1.73 \times 10^{-2}$}} &
    \cellcolor[rgb]{0.267,0.005,0.329} \raisebox{0pt}[0.7cm][0.7cm]{\centering\textcolor{white}{$1.70 \times 10^{-2}$}} \\
        \end{tabular}}
        \subcaption{Setting 3.}
        \label{fig:grid_cubes_fall_setting3}
\end{minipage}
\caption{\textbf{Moving Across MatVAE latent space.} We sample 3 different valid mechanical property triplets \mattriplet\ (Setting 1,2,3), corresponding to the middle square in the three color diagrams. We encode each of these triplets with MatVAE, and then traverse the 2D latent space to build a $5\times5$ grid of latents around the starting value, which are each decoded to actual mechanical properties. To visualize if latent space dimensions correlate with actual simulation performance, we apply each meachnical property triplet to a dropping bunny simulation and measure its mean displacement from rest, which is color coded in the graphs below. Each diagram (b, c, d) thus corresponds to 25 simulation runs with different parameters. We observe a clear correlation between latent dimensions and simulation behavior. (\video{4:40})}\label{fig:app:matvae_traverse}
\end{figure*}
\else
    \begin{table*}[tb]
\centering
\resizebox{\textwidth}{!}{%
\begin{tabular}{lllrrrrrr}
\toprule
\rowcolor{blue!15}Setting & Position & Material & \multicolumn{2}{c}{Young's Modulus (Pa)} & \multicolumn{2}{c}{Poisson's Ratio} & \multicolumn{2}{c}{Density (kg/m$^3$)} \\
\cmidrule(lr){4-5} \cmidrule(lr){6-7} \cmidrule(lr){8-9}
 \rowcolor{blue!15}& & & Interpolated & True Range & Interpolated & True Range & Interpolated & True Range \\
\midrule
\Cref{fig:grid_cubes_fall_setting1} & Top-left ($\nwarrow$) & Aerographite & $4.4 \times 10^5$ & $1.0 \times 10^5$ -- $1.0 \times 10^6$ & 0.241 & 0.2--0.3 & 0.2 & 0.2--0.2 \\
 & Top-right ($\nearrow$) & Polyurethane Foam & $4.8 \times 10^6$ & $1.0 \times 10^5$ -- $5.0 \times 10^6$ & 0.304 & 0.30--0.30 & 298.2.0 & 50--300 \\
 & Bottom-left ($\swarrow$) & Rubber (soft) & $3.1 \times 10^6$ & $3.0 \times 10^6$ -- $5.0 \times 10^6$ & 0.488 & 0.48--0.50 & 952.0 & 950--950 \\
 & Bottom-right ($\searrow$) & Styrofoam & $1.6 \times 10^6$ & $1.0 \times 10^6$ -- $3.0 \times 10^6$ & 0.322 & 0.3--0.35 & 22.6 & 15--35 \\
\midrule
\Cref{fig:grid_cubes_fall_setting2} & Top-left ($\nwarrow$) & Aerogel & $4.4 \times 10^6$ & $1.0 \times 10^6$ -- $1.0 \times 10^7$ & 0.257 & 0.2--0.3 & 1.0 & 1.0--1.0 \\
 & Top-right ($\nearrow$) & Neoprene & $1.0 \times 10^7$ & $1.0 \times 10^6$ -- $1.0 \times 10^7$ & 0.494 & 0.45--0.5 & 1232.0 & 1230--1250 \\
 & Bottom-left ($\swarrow$) & EPDM Rubber & $6.6 \times 10^6$ & $5.0 \times 10^6$ -- $1.0 \times 10^7$ & 0.488 & 0.49--0.49 & 1100.9 & 1100--1100 \\
 & Bottom-right ($\searrow$) & Flexible PVC (Plasticized) & $4.8 \times 10^7$ & $2.0 \times 10^7$ -- $1.0 \times 10^8$ & 0.450 & 0.45--0.45 & 1209.5 & 1200--1400 \\
\midrule
\Cref{fig:grid_cubes_fall_setting3} & Top-left ($\nwarrow$) & Polystyrene Foam (EPS) & $2.6 \times 10^6$ & $1.0 \times 10^6$ -- $5.0 \times 10^6$ & 0.104 & 0.10--0.10 & 59.1 & 30--100 \\
 & Top-right ($\nearrow$) & Chloroprene Rubber (Neoprene) & $5.0 \times 10^6$ & $5.0 \times 10^6$ -- $5.0 \times 10^6$ & 0.490 & 0.49--0.49 & 1200.8 & 1200--1200 \\
 & Bottom-left ($\swarrow$) & Polystyrene (Foam) & $5.8 \times 10^6$ & $2.5 \times 10^6$ -- $7.0 \times 10^6$ & 0.371 & 0.34--0.4 & 34.8 & 15--35 \\
 & Bottom-right ($\searrow$) & Polybutylene (PB) & $2.5 \times 10^8$ & $2.5 \times 10^8$ -- $3.0 \times 10^8$ & 0.400 & 0.4--0.42 & 932.0 & 930--950 \\
\bottomrule
\end{tabular}%
}
\caption{Corner Materials for our experiments on moving across the latent space (\Cref{fig:grid_cubes_fall_setting1,fig:grid_cubes_fall_setting2,fig:grid_cubes_fall_setting3}).}
\label{tab:corner_materials}
\end{table*}

\newcolumntype{P}{>{\centering\arraybackslash}m{1.5cm}}
% Figure for Setting 2
\begin{figure}[tb]
    \centering
    \setlength{\tabcolsep}{0pt}%
    \renewcommand{\arraystretch}{0}%
    % \newlength{\arrowcol}
    \setlength{\arrowcol}{0.08\columnwidth} % width of the arrow column
    % \newlength{\cellsize}
    \setlength{\cellsize}{\dimexpr(\columnwidth-\arrowcol)/5\relax}
    \resizebox{\linewidth}{!}{%
    \begin{tabular}{c c c c c c}
    % Top header row
    & \multicolumn{5}{c}{\bigarrowleftright{Latent Dimension 1}} \\
    % Left multirow rotated label + data rows
    \multirow{5}{*}{\rotatebox{90}{\parbox{6.cm}{\centering\bigarrowleftright{Latent Dimension 2}}}} &
    \cellcolor[rgb]{0.993,0.906,0.144} \raisebox{0pt}[0.7cm][0.7cm]{\centering{$1.66 \times 10^{-1}$}} &
    \cellcolor[rgb]{0.147,0.673,0.509} \raisebox{0pt}[0.7cm][0.7cm]{\centering\textcolor{white}{$1.11 \times 10^{-1}$}} &
    \cellcolor[rgb]{0.143,0.669,0.511} \raisebox{0pt}[0.7cm][0.7cm]{\centering\textcolor{white}{$1.10 \times 10^{-1}$}} &
    \cellcolor[rgb]{0.140,0.666,0.513} \raisebox{0pt}[0.7cm][0.7cm]{\centering\textcolor{white}{$1.10 \times 10^{-1}$}} &
    \cellcolor[rgb]{0.130,0.651,0.522} \raisebox{0pt}[0.7cm][0.7cm]{\centering\textcolor{white}{$1.08 \times 10^{-1}$}} \\
    &
    \cellcolor[rgb]{0.158,0.684,0.502} \raisebox{0pt}[0.7cm][0.7cm]{\centering\textcolor{white}{$1.13 \times 10^{-1}$}} &
    \cellcolor[rgb]{0.128,0.648,0.523} \raisebox{0pt}[0.7cm][0.7cm]{\centering\textcolor{white}{$1.07 \times 10^{-1}$}} &
    \cellcolor[rgb]{0.121,0.593,0.545} \raisebox{0pt}[0.7cm][0.7cm]{\centering\textcolor{white}{$9.85 \times 10^{-2}$}} &
    \cellcolor[rgb]{0.123,0.585,0.547} \raisebox{0pt}[0.7cm][0.7cm]{\centering\textcolor{white}{$9.75 \times 10^{-2}$}} &
    \cellcolor[rgb]{0.124,0.578,0.548} \raisebox{0pt}[0.7cm][0.7cm]{\centering\textcolor{white}{$9.65 \times 10^{-2}$}} \\
    &
    \cellcolor[rgb]{0.158,0.684,0.502} \raisebox{0pt}[0.7cm][0.7cm]{\centering\textcolor{white}{$1.13 \times 10^{-1}$}} &
    \cellcolor[rgb]{0.126,0.644,0.525} \raisebox{0pt}[0.7cm][0.7cm]{\centering\textcolor{white}{$1.06 \times 10^{-1}$}} &
    \cellcolor[rgb]{0.125,0.574,0.549} \raisebox{0pt}[0.7cm][0.7cm]{\centering\textcolor{white}{$9.57 \times 10^{-2}$}} &
    \cellcolor[rgb]{0.143,0.523,0.556} \raisebox{0pt}[0.7cm][0.7cm]{\centering\textcolor{white}{$8.76 \times 10^{-2}$}} &
    \cellcolor[rgb]{0.158,0.486,0.558} \raisebox{0pt}[0.7cm][0.7cm]{\centering\textcolor{white}{$8.19 \times 10^{-2}$}} \\
    &
    \cellcolor[rgb]{0.147,0.673,0.509} \raisebox{0pt}[0.7cm][0.7cm]{\centering\textcolor{white}{$1.11 \times 10^{-1}$}} &
    \cellcolor[rgb]{0.121,0.626,0.533} \raisebox{0pt}[0.7cm][0.7cm]{\centering\textcolor{white}{$1.04 \times 10^{-1}$}} &
    \cellcolor[rgb]{0.134,0.549,0.554} \raisebox{0pt}[0.7cm][0.7cm]{\centering\textcolor{white}{$9.19 \times 10^{-2}$}} &
    \cellcolor[rgb]{0.164,0.471,0.558} \raisebox{0pt}[0.7cm][0.7cm]{\centering\textcolor{white}{$7.99 \times 10^{-2}$}} &
    \cellcolor[rgb]{0.280,0.171,0.480} \raisebox{0pt}[0.7cm][0.7cm]{\centering\textcolor{white}{$3.97 \times 10^{-2}$}} \\
    &
    \cellcolor[rgb]{0.143,0.669,0.511} \raisebox{0pt}[0.7cm][0.7cm]{\centering\textcolor{white}{$1.10 \times 10^{-1}$}} &
    \cellcolor[rgb]{0.122,0.589,0.546} \raisebox{0pt}[0.7cm][0.7cm]{\centering\textcolor{white}{$9.78 \times 10^{-2}$}} &
    \cellcolor[rgb]{0.135,0.545,0.554} \raisebox{0pt}[0.7cm][0.7cm]{\centering\textcolor{white}{$9.13 \times 10^{-2}$}} &
    \cellcolor[rgb]{0.274,0.200,0.499} \raisebox{0pt}[0.7cm][0.7cm]{\centering\textcolor{white}{$4.35 \times 10^{-2}$}} &
    \cellcolor[rgb]{0.267,0.005,0.329} \raisebox{0pt}[0.7cm][0.7cm]{\centering\textcolor{white}{$2.21 \times 10^{-2}$}} \\
    \end{tabular}}
    \caption{\textbf{Moving Across MatVAE latent space II.} We manually set a triplet of valid mechanical properties (for the object in the middle) which is encoded by MatVAE. We use the produced latent and move across the two-dimensional latent space to build a $5\times5$ grid of valid mechanical property triplets. We show the mean displacement from the simulation of bunnies with these mechanical property triplets falling to the ground. (\video{00:00})}
    \label{fig:grid_cubes_fall_setting2}
\end{figure}

% Figure for Setting 3
\begin{figure}[tb]
    \centering
    \setlength{\tabcolsep}{0pt}%
    \renewcommand{\arraystretch}{0}%
    \setlength{\arrowcol}{0.08\columnwidth} % width of the arrow column
    \setlength{\cellsize}{\dimexpr(\columnwidth-\arrowcol)/5\relax}
    \resizebox{\linewidth}{!}{%
    \begin{tabular}{c c c c c c}
    % Top header row
    & \multicolumn{5}{c}{\bigarrowleftright{Latent Dimension 1}} \\
    % Left multirow rotated label + data rows
    \multirow{5}{*}{\rotatebox{90}{\parbox{6.cm}{\centering\bigarrowleftright{Latent Dimension 2}}}} &
    \cellcolor[rgb]{0.993,0.906,0.144} \raisebox{0pt}[0.7cm][0.7cm]{\centering{$2.25 \times 10^{-1}$}} &
    \cellcolor[rgb]{0.993,0.906,0.144} \raisebox{0pt}[0.7cm][0.7cm]{\centering{$2.25 \times 10^{-1}$}} &
    \cellcolor[rgb]{0.214,0.722,0.470} \raisebox{0pt}[0.7cm][0.7cm]{\centering\textcolor{white}{$1.56 \times 10^{-1}$}} &
    \cellcolor[rgb]{0.176,0.698,0.491} \raisebox{0pt}[0.7cm][0.7cm]{\centering\textcolor{white}{$1.51 \times 10^{-1}$}} &
    \cellcolor[rgb]{0.166,0.691,0.497} \raisebox{0pt}[0.7cm][0.7cm]{\centering\textcolor{white}{$1.49 \times 10^{-1}$}} \\
    &
    \cellcolor[rgb]{0.404,0.800,0.363} \raisebox{0pt}[0.7cm][0.7cm]{\centering\textcolor{white}{$1.77 \times 10^{-1}$}} &
    \cellcolor[rgb]{0.130,0.560,0.552} \raisebox{0pt}[0.7cm][0.7cm]{\centering\textcolor{white}{$1.20 \times 10^{-1}$}} &
    \cellcolor[rgb]{0.146,0.515,0.557} \raisebox{0pt}[0.7cm][0.7cm]{\centering\textcolor{white}{$1.10 \times 10^{-1}$}} &
    \cellcolor[rgb]{0.228,0.327,0.547} \raisebox{0pt}[0.7cm][0.7cm]{\centering\textcolor{white}{$7.01 \times 10^{-2}$}} &
    \cellcolor[rgb]{0.283,0.111,0.432} \raisebox{0pt}[0.7cm][0.7cm]{\centering\textcolor{white}{$3.30 \times 10^{-2}$}} \\
    &
    \cellcolor[rgb]{0.208,0.719,0.473} \raisebox{0pt}[0.7cm][0.7cm]{\centering\textcolor{white}{$1.56 \times 10^{-1}$}} &
    \cellcolor[rgb]{0.150,0.504,0.557} \raisebox{0pt}[0.7cm][0.7cm]{\centering\textcolor{white}{$1.07 \times 10^{-1}$}} &
    \cellcolor[rgb]{0.277,0.050,0.376} \raisebox{0pt}[0.7cm][0.7cm]{\centering\textcolor{white}{$2.38 \times 10^{-2}$}} &
    \cellcolor[rgb]{0.277,0.050,0.376} \raisebox{0pt}[0.7cm][0.7cm]{\centering\textcolor{white}{$2.37 \times 10^{-2}$}} &
    \cellcolor[rgb]{0.276,0.044,0.370} \raisebox{0pt}[0.7cm][0.7cm]{\centering\textcolor{white}{$2.27 \times 10^{-2}$}} \\
    &
    \cellcolor[rgb]{0.120,0.618,0.536} \raisebox{0pt}[0.7cm][0.7cm]{\centering\textcolor{white}{$1.33 \times 10^{-1}$}} &
    \cellcolor[rgb]{0.280,0.166,0.476} \raisebox{0pt}[0.7cm][0.7cm]{\centering\textcolor{white}{$4.18 \times 10^{-2}$}} &
    \cellcolor[rgb]{0.277,0.050,0.376} \raisebox{0pt}[0.7cm][0.7cm]{\centering\textcolor{white}{$2.37 \times 10^{-2}$}} &
    \cellcolor[rgb]{0.269,0.010,0.335} \raisebox{0pt}[0.7cm][0.7cm]{\centering\textcolor{white}{$1.82 \times 10^{-2}$}} &
    \cellcolor[rgb]{0.269,0.010,0.335} \raisebox{0pt}[0.7cm][0.7cm]{\centering\textcolor{white}{$1.80 \times 10^{-2}$}} \\
    &
    \cellcolor[rgb]{0.121,0.593,0.545} \raisebox{0pt}[0.7cm][0.7cm]{\centering\textcolor{white}{$1.27 \times 10^{-1}$}} &
    \cellcolor[rgb]{0.278,0.056,0.381} \raisebox{0pt}[0.7cm][0.7cm]{\centering\textcolor{white}{$2.44 \times 10^{-2}$}} &
    \cellcolor[rgb]{0.270,0.015,0.341} \raisebox{0pt}[0.7cm][0.7cm]{\centering\textcolor{white}{$1.89 \times 10^{-2}$}} &
    \cellcolor[rgb]{0.267,0.005,0.329} \raisebox{0pt}[0.7cm][0.7cm]{\centering\textcolor{white}{$1.73 \times 10^{-2}$}} &
    \cellcolor[rgb]{0.267,0.005,0.329} \raisebox{0pt}[0.7cm][0.7cm]{\centering\textcolor{white}{$1.70 \times 10^{-2}$}} \\
    \end{tabular}}
    \caption{\textbf{Moving Across MatVAE latent space III.} We manually set a triplet of valid mechanical properties (for the object in the middle) which is encoded by MatVAE. We use the produced latent and move across the two-dimensional latent space to build a $5\times5$ grid of valid mechanical property triplets. We show the mean displacement from the simulation of bunnies with these mechanical property triplets falling to the ground. (\video{00:00})}
    \label{fig:grid_cubes_fall_setting3}
\end{figure}

% figure for interp_cubes_fall pair 2
\begin{figure}[tb]
    \centering
    \scriptsize
    \resizebox{\columnwidth}{!}{%
    \setlength{\tabcolsep}{0pt}
    \begin{tabular}{c|ccc|c}
        \multicolumn{5}{c}{\normalsize\textbf{MatVAE}} \\
        Styrofoam &
        Vanadium &
        Rhodium &
        Tungsten &
        Osmium\\
        $E = 0.002$ GPa & $E = 128.0$ GPa & $E = 275.0$ GPa &
        $E = 411.0$ GPa & $E = 550.0$ GPa\\
        $\nu = 0.33$ & $\nu = 0.37$ & $\nu = 0.30$ &
        $\nu = 0.28$ & $\nu = 0.25$\\
        $\rho = 25$ kg/m$^3$ & $\rho = 6110$ kg/m$^3$ &
        $\rho = 12450$ kg/m$^3$ & $\rho = 19250$ kg/m$^3$ &
        $\rho = 22570$ kg/m$^3$\\
        \midrule
        \multicolumn{5}{c}{\normalsize\textbf{Naive Interpolation}} \\
        Styrofoam &
        {\color{red}Invalid (\xmark)} &
        {\color{red}Invalid (\xmark)} &
        {\color{red}Invalid (\xmark)} &
        Osmium\\
        $E = 0.002$ GPa & $E = 137.5$ GPa & $E = 275.0$ GPa &
        $E = 412.5$ GPa & $E = 550.0$ GPa\\
        $\nu = 0.33$ & $\nu = 0.31$ & $\nu = 0.29$ &
        $\nu = 0.27$ & $\nu = 0.25$\\
        $\rho = 25$ kg/m$^3$ & $\rho = 5661$ kg/m$^3$ &
        $\rho = 11298$ kg/m$^3$ & $\rho = 16934$ kg/m$^3$ &
        $\rho = 22570$ kg/m$^3$\\
    \end{tabular}%
    }
    \caption{\textbf{MatVAE Allows Interpolation II.} We set the left-most and right-most mechanical property triplets and use MatVAE latent space to linearly interpolate between the materials. We use "Invalid" to indicate properties that do not fall into any ranges from our vast MTD (\Cref{sec:mtd}). (\video{00:00})}
    \label{fig:interp_pair2}
\end{figure}

\fi

\clearpage

\section{Comparison with Concurrent work}
\label{app:concurrent}

Although Pixie~\cite{le2025pixiefastgeneralizablesupervised} is concurrent with us, we still compare aspects of our approach with Pixie. We discuss differences with Pixie in~\Cref{sec:rw:inferring}.

\begin{figure*}[tb]
    \centering
    \includegraphics[]{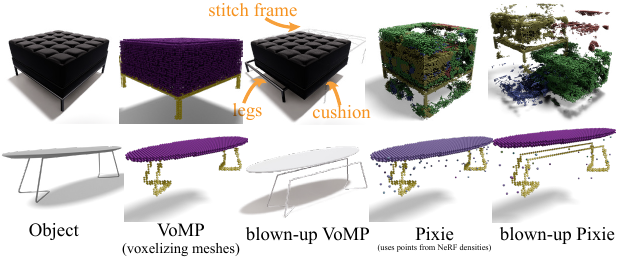}
    \caption{\textbf{Data Annotation.} We compare data annotation process of \acronym\ and Pixie.}
    \label{fig:pixie-annot}
\end{figure*}

\paragraph{Data Annotation Process.} We compare our data annotation process with the annotation process of Pixie~\cite{le2025pixiefastgeneralizablesupervised} in ~\Cref{fig:pixie-annot}. Our method performs annotation from meshes while Pixie~\cite{le2025pixiefastgeneralizablesupervised} gets points from training a NeRF, which often produces noisy points, and the segmentation is performed based on CLIP features, which often produces noisy segmentation for difficult objects.

\paragraph{Validity of Materials for Data Annotation.} Although we do not have access to the Pixie~\cite{le2025pixiefastgeneralizablesupervised} dataset, Pixie uses in-context physics examples, which include material names and ranges of mechanical property triplets in the annotation process. We analyze these in-context physics examples and compare them with real material ranges from MTD (\Cref{sec:mtd}) in~\Cref{tab:mapped-materials}. We find that some of these in-context properties might create pleasing simulations with a particular simulator but can fall outside the range of real materials.

\begin{table*}[tb]
\centering
\caption{\textbf{Comparison of Mapped Material Properties} between Pixie's in-context physics examples~\cite{le2025pixiefastgeneralizablesupervised} and the datasets of known material properties.}
\begin{adjustbox}{max width=\textwidth}
\begin{tabular}{l l l l l l l l}
\toprule
\rowcolor{nvidiagreen!15}Item & Mapped Materials & Pixie $\rho$ & Dataset $\rho$ & Pixie $E$ & Dataset $E$ & Pixie $\nu$ & Dataset $\nu$ \\
\midrule
  \multirow[c]{3}{*}{tree/pot} & Clay Brick & \multirow[c]{3}{*}{[400, 400]} & [1900, 1900] & \multirow[c]{3}{*}{[2.000e+08, 2.000e+08]} & [2.000e+09, 6.000e+09] & \multirow[c]{3}{*}{[0.40, 0.40]} & [0.20, 0.20] \\
  & Porcelain (Ceramic) & & [2400, 2400] & & [7.000e+10, 7.000e+10] & & [0.20, 0.20] \\
  & Glass Ceramic & & [2400, 2600] & & [9.000e+10, 1.100e+11] & & [0.24, 0.25] \\
  \midrule\multirow[c]{3}{*}{tree/trunk} & Wood & \multirow[c]{3}{*}{[400, 400]} & [700, 700] & \multirow[c]{3}{*}{[2.000e+06, 2.000e+06]} & [8.000e+09, 1.100e+10] & \multirow[c]{3}{*}{[0.40, 0.40]} & [0.30, 0.50] \\
  & Oak (White) & & [770, 800] & & [1.200e+10, 1.500e+10] & & [0.30, 0.40] \\
  & Maple Wood (Sugar) & & [630, 690] & & [1.000e+10, 1.300e+10] & & [0.30, 0.40] \\
  \midrule tree/leaves & — & [200, 200] & — & [2.000e+04, 2.000e+04] & — & [0.40, 0.40] & — \\
  \midrule\multirow[c]{2}{*}{flowers/vase} & Glass (Soda-Lime) & \multirow[c]{2}{*}{[500, 500]} & [2500, 2500] & \multirow[c]{2}{*}{[1.000e+06, 1.000e+06]} & [7.200e+10, 7.400e+10] & \multirow[c]{2}{*}{[0.30, 0.30]} & [0.23, 0.23] \\
  & Glass (Borosilicate) & & [2300, 2300] & & [6.200e+10, 8.100e+10] & & [0.20, 0.20] \\
  \midrule flowers/flowers & — & [100, 100] & — & [1.000e+04, 1.000e+04] & — & [0.40, 0.40] & — \\
  \midrule shrub/stems & Wood & [300, 300] & [700, 700] & [1.000e+05, 1.000e+05] & [8.000e+09, 1.100e+10] & [0.35, 0.35] & [0.30, 0.50] \\
  \midrule shrub/twigs & Wood & [250, 250] & [700, 700] & [6.000e+04, 6.000e+04] & [8.000e+09, 1.100e+10] & [0.38, 0.38] & [0.30, 0.50] \\
  \midrule shrub/foliage & — & [150, 150] & — & [2.000e+04, 2.000e+04] & — & [0.40, 0.40] & — \\
  \midrule\multirow[c]{3}{*}{grass/blades} & Rubber (soft) & \multirow[c]{3}{*}{[80, 80]} & [950, 950] & \multirow[c]{3}{*}{[1.000e+04, 1.000e+04]} & [3.000e+06, 5.000e+06] & \multirow[c]{3}{*}{[0.45, 0.45]} & [0.48, 0.50] \\
  & EPDM Rubber & & [1100, 1100] & & [1.000e+07, 1.000e+07] & & [0.49, 0.49] \\
  & Neoprene & & [1230, 1250] & & [1.000e+06, 1.000e+07] & & [0.45, 0.50] \\
  \midrule soil (if visible) & Sandy Loam & [1200, 1200] & [1600, 1800] & [5.000e+05, 5.000e+05] & [1.000e+08, 5.000e+08] & [0.30, 0.30] & [0.31, 0.31] \\
  \midrule\multirow[c]{4}{*}{rubber\_ducks\_and\_toys/toy} & Rubber (soft) & \multirow[c]{4}{*}{[80, 150]} & [950, 950] & \multirow[c]{4}{*}{[3.000e+04, 5.000e+04]} & [3.000e+06, 5.000e+06] & \multirow[c]{4}{*}{[0.40, 0.45]} & [0.48, 0.50] \\
  & EPDM Rubber & & [1100, 1100] & & [1.000e+07, 1.000e+07] & & [0.49, 0.49] \\
  & Neoprene & & [1230, 1250] & & [1.000e+06, 1.000e+07] & & [0.45, 0.50] \\
  & Flexible PVC (Plasticized) & & [1200, 1400] & & [2.000e+07, 1.000e+08] & & [0.45, 0.45] \\
  \midrule\multirow[c]{3}{*}{sport\_balls/ball} & Rubber (soft) & \multirow[c]{3}{*}{[80, 150]} & [950, 950] & \multirow[c]{3}{*}{[3.000e+04, 5.000e+04]} & [3.000e+06, 5.000e+06] & \multirow[c]{3}{*}{[0.40, 0.45]} & [0.48, 0.50] \\
  & EPDM Rubber & & [1100, 1100] & & [1.000e+07, 1.000e+07] & & [0.49, 0.49] \\
  & Neoprene & & [1230, 1250] & & [1.000e+06, 1.000e+07] & & [0.45, 0.50] \\
  \midrule\multirow[c]{3}{*}{soda\_cans/can} & Aluminium & \multirow[c]{3}{*}{[2600, 2800]} & [2700, 2700] & \multirow[c]{3}{*}{[5.000e+10, 8.000e+10]} & [7.000e+10, 7.000e+10] & \multirow[c]{3}{*}{[0.25, 0.35]} & [0.35, 0.35] \\
  & Aluminum 2024-T3 & & [2780, 2780] & & [7.240e+10, 7.240e+10] & & [0.33, 0.33] \\
  & Aluminum 7075-T6 & & [2810, 2810] & & [7.100e+10, 7.100e+10] & & [0.33, 0.33] \\
  \midrule\multirow[c]{3}{*}{metal\_crates/crate} & Steel & \multirow[c]{3}{*}{[2500, 2900]} & [7700, 7700] & \multirow[c]{3}{*}{[8.000e+07, 1.200e+08]} & [2.000e+11, 2.000e+11] & \multirow[c]{3}{*}{[0.25, 0.35]} & [0.31, 0.31] \\
  & Stainless Steel 17-7PH & & [7800, 7800] & & [2.040e+11, 2.040e+11] & & [0.30, 0.30] \\
  & Stainless Steel 440A & & [7800, 7800] & & [2.000e+11, 2.000e+11] & & [0.30, 0.30] \\
  \midrule sand/sand & Sandy Loam & [1800, 2200] & [1600, 1800] & [4.000e+07, 6.000e+07] & [1.000e+08, 5.000e+08] & [0.25, 0.35] & [0.31, 0.31] \\
  \midrule jello\_block/jello & — & [40, 60] & — & [8.000e+02, 1.200e+03] & — & [0.25, 0.35] & — \\
  \midrule snow\_and\_mud/snow\_and\_mud & Sandy Loam & [2000, 3000] & [1600, 1800] & [8.000e+04, 1.200e+05] & [1.000e+08, 5.000e+08] & [0.15, 0.25] & [0.31, 0.31] \\
\bottomrule
\end{tabular}
\end{adjustbox}
\label{tab:mapped-materials}
\end{table*}

\begin{figure*}[tb]
    \includegraphics[width=\textwidth]{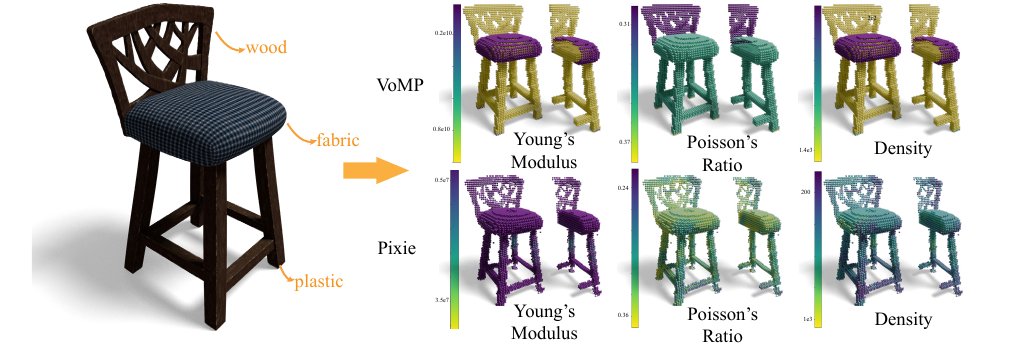}
    \caption{\textbf{Comparing Estimations.} We compare property estimates between \acronym\ and Pixie for a mesh.}
    \label{fig:pixie-compare}
\end{figure*}

\paragraph{Comparison of Outputs.} Although Pixie~\cite{le2025pixiefastgeneralizablesupervised} is concurrent with us, we still compare the outputs of Pixie with \acronym\ on one object in~\Cref{fig:pixie-compare}.

\section{Ablations}
\label{sec:ablations}

We provide an in-depth analysis motivating our MatVAE and Geometry Transformer training scheme by ablating each component in~\Cref{tab:matvaeablation,tab:ablations}. Our ablations require changing the hyperparameters for fair comparisons; thus, for each ablation, we tune our hyperparameters within an identical compute budget.

\paragraph{MatVAE vs Vanilla VAE.} Technically, MatVAE (\Cref{sec:matvae}) is built on top of the vanilla VAE~\cite{kingma2022autoencodingvariationalbayes} and we can use a vanilla VAE~\cite{kingma2022autoencodingvariationalbayes} in its place. In~\Cref{tab:matvaeablation}, we show material property reconstruction and distributional metrics between a vanilla VAE and our MatVAE. MatVAE outperforms the vanilla VAE in almost all metrics. We find that Vanilla VAE collapses to the Young's Modulus property, giving us a low reconstruction error for Young's Modulus but significantly higher errors for other properties.

\paragraph{Image Features.} For aggregating image features (\Cref{sec:aggregating}), we experiment with using DINOv2~\cite{oquab2024dinov2learningrobustvisual}, CLIP~\cite{radford2021learningtransferablevisualmodels}, and RGB colors by average pooling in the voxel. The results are shown in~\Cref{tab:ablations}. Our model had many layers in the Geometry Transformer (\Cref{sec:3dencdoer,sec:networkdesign}) initialized from a generation model~\cite{xiang2025structured3dlatentsscalable}. This set of ablations was trained starting from random weights, due to the absence of generation weights for these settings. We find that using DINOv2~\cite{oquab2024dinov2learningrobustvisual} and CLIP~\cite{radford2021learningtransferablevisualmodels} without initializing the weights from TRELLIS~\cite{xiang2025structured3dlatentsscalable} performs slightly worse, whereas simply using RGB colors performs significantly worse.

\paragraph{MatVAE.} While MatVAE acts as a continuous tokenizer, it is possible to have the Geometry Transformer directly predict a $\mathbb{R}^3$ vector \textit{i.e.} directly predict the material triplets. We find this produces significantly worse results for Young's Modulus estimation and Poisson's Ratio estimation.

\paragraph{Normalization Scheme.} We experiment with different normalization schemes (\Cref{sec:matvae}) like $Z$-score, and not using $\log$-space transform for either Young's Modulus or Density. All of these normalization schemes lead to a significant degradation in performance. Most notably, removing the logarithmic scaling for Young's Modulus (w/o $\log(E)$) or using a simple $Z$-score severely harms prediction accuracy.

\paragraph{Loss.} Our Geometry Transformer (\Cref{sec:3dencdoer}) is trained with $\ell_2$ loss for reconstruction (\Cref{eq:assignment_loss}). We test the effect of replacing this with an $\ell_1$ loss. This change results in a substantial drop in performance across all metrics, with errors increasing by a factor of 2-3x for most properties. This indicates that the squared error penalty of the $\ell_2$ loss is more effective for this material property regression task.

\newcommand{\matvaedistrowablation}{
\textbf{0.0405} & \textbf{0.0798} & \underline{0.1379} & 
\textbf{0.0317} & \textbf{0.0437} & \textbf{0.0342} & 
\textbf{0.0132} & \textbf{0.0172} & \underline{0.0260}
}
\newcommand{\matvaemetricsrowablation}{
\textbf{0.0034} & \textbf{0.0426} & \textbf{0.0330} & 
\textbf{0.0054} & \textbf{0.0036} & \textbf{0.0036} & 
\textbf{0.0131} & \textbf{0.4439} & \textbf{0.0411}
}
\newcommand{\vanillavaedistrow}{
\underline{0.0653} & \underline{0.0868} & \textbf{0.0547} & 
\underline{0.0849} & \underline{0.1057} & \underline{0.0689} & 
\underline{0.0547} & \underline{0.0744} & \textbf{0.0175}
}
\newcommand{\vanillavaemetricsrow}{
\underline{0.0512} & \underline{15366.8750} & \underline{0.8306} & 
\underline{0.0542} & \underline{0.0544} & \underline{0.0447} & 
\underline{0.2384} & \underline{2.1893} & \underline{0.4690}
}
\newcommand{\matvaedistbenefitrow}{
{\textbf{\textcolor{green}{(-0.025)}}} & 
{\textbf{\textcolor{green}{(-0.007)}}} & 
{\textbf{\textcolor{red}{(+0.083)}}} & 
{\textbf{\textcolor{green}{(-0.053)}}} & 
{\textbf{\textcolor{green}{(-0.062)}}} & 
{\textbf{\textcolor{green}{(-0.035)}}} & 
{\textbf{\textcolor{green}{(-0.042)}}} & 
{\textbf{\textcolor{green}{(-0.057)}}} & 
{\textbf{\textcolor{red}{(+0.009)}}}
}
\newcommand{\matvaemetricsbenefitrow}{
{\textbf{\textcolor{green}{(-0.048)}}} & 
{\textbf{\textcolor{green}{(-15366.8)}}} & 
{\textbf{\textcolor{green}{(-0.798)}}} & 
{\textbf{\textcolor{green}{(-0.049)}}} & 
{\textbf{\textcolor{green}{(-0.051)}}} & 
{\textbf{\textcolor{green}{(-0.041)}}} & 
{\textbf{\textcolor{green}{(-0.225)}}} & 
{\textbf{\textcolor{green}{(-1.745)}}} & 
{\textbf{\textcolor{green}{(-0.428)}}}
}
\begin{table*}[hb]
\centering
\caption{\textbf{MatVAE Ablation.} We ablate MatVAE by comparing it against a vanilla VAE and present reconstruction and distirbutional metrics.}
\resizebox{\textwidth}{!}{%
\begin{tabular}{lrrrrrrrrr}
\toprule
\rowcolor{nvidiagreen!15}
Model & 
\multicolumn{3}{c}{Young's Modulus ($E$)} & 
\multicolumn{3}{c}{Poisson's Ratio ($\nu$)} & 
\multicolumn{3}{c}{Density ($\rho$)}\\
\cmidrule(lr){2-4} \cmidrule(lr){5-7} \cmidrule(lr){8-10}
\rowcolor{nvidiagreen!15}
& W$_1$ ($\downarrow$) & W$_2$ ($\downarrow$) & $D_{\text{KL}}$ ($\downarrow$) 
& W$_1$ ($\downarrow$) & W$_2$ ($\downarrow$) & $D_{\text{KL}}$ ($\downarrow$) 
& W$_1$ ($\downarrow$) & W$_2$ ($\downarrow$) & $D_{\text{KL}}$ ($\downarrow$) \\
\midrule
Vanilla VAE~\cite{kingma2022autoencodingvariationalbayes} & \vanillavaedistrow \\
MatVAE & \matvaedistrowablation \\
& \matvaedistbenefitrow \\
w/o NF & 0.0339 & 0.0447 & 0.0441 & 0.0417 & 0.0504 & 0.0848 & 0.0599 & 0.0819 & 0.0529\\
w/o TC penalty & 0.0633 & 0.0855 & 0.0500 & 0.0844 & 0.1052 & 0.0672 & 0.0525 & 0.0715 & 0.0109 \\
w/o free nats & 0.0749 & 0.1168 & 0.1311 & 0.2014 & 0.2064 & 0.6376 & 0.0421 & 0.0507 & 0.0223 \\\midrule
\rowcolor{nvidiagreen!15} 
& $\log(E)$ ($\downarrow$) & $\nu$ ($\downarrow$) & $\rho$ ($\downarrow$) 
& $\log(E/\rho)$ ($\downarrow$) & $\log(G)$ ($\downarrow$) & $\log(K)$ ($\downarrow$) 
& L.S. ($\downarrow$) & E.A. ($\downarrow$) & Bray–Curtis ($\downarrow$) \\
\midrule
Vanilla VAE~\cite{kingma2022autoencodingvariationalbayes} & \vanillavaemetricsrow \\
MatVAE & \matvaemetricsrowablation \\
& \matvaemetricsbenefitrow \\
w/o NF & 0.0020 & \texttt{nan} & 0.0160 & 0.0033 & 0.0021 & 0.0021 & 0.0086 & 0.1729 & 0.0234\\
w/o TC penalty & 0.0499 & 15567.0986 & 0.8298 & 0.0537 & 0.0530 & 0.0437 & 0.2382 & 2.1514 & 0.4562\\
w/o free nats & 0.0036 & 16332.7829 & 0.0276 & 0.0053 & 0.0037 & 0.0041 & 0.0116 & 0.2966 & 0.0436\\
\bottomrule
\end{tabular}
}
\label{tab:matvaeablation}
\end{table*}
\ifdefined\iclr
    \begin{table*}[ht]
\centering
\caption{\textbf{Ablations.} We ablate \acronym\ with choice of image features, using MatVAE, normalization scheme, and choice of a loss function. We report the voxel-level mechanical property difference errors.}
\resizebox{\textwidth}{!}{%
\begin{tabular}{lrrrrrr}
\toprule
\rowcolor{nvidiagreen!15}Method & \multicolumn{2}{c}{Young's Modulus Pa ($E$)} & \multicolumn{2}{c}{Poisson's Ratio ($\nu$)} & \multicolumn{2}{c}{Density $\frac{kg}{m^3}$ ($\rho$)} \\
\cmidrule(r){2-3} \cmidrule(r){4-5} \cmidrule(r){6-7}
\rowcolor{nvidiagreen!15}& ALDE ($\downarrow$) & ALRE ($\downarrow$) & ADE ($\downarrow$) & ARE ($\downarrow$) & ADE ($\downarrow$) & ARE ($\downarrow$) \\
\midrule \rowcolor{gray!15} \multicolumn{7}{l}{Image Features (Geometry Transformer start from random weights,~\Cref{sec:aggregating})} \\
w/ DINOv2~\cite{oquab2024dinov2learningrobustvisual} & \textbf{0.2888} {\textbf{\scriptsize\textcolor{gray}{($\pm$0.41)}}} & 0.0536 {\textbf{\scriptsize\textcolor{gray}{($\pm$0.06)}}} & 0.0259 {\textbf{\scriptsize\textcolor{gray}{($\pm$0.02)}}} & 0.0803 {\textbf{\scriptsize\textcolor{gray}{($\pm$0.08)}}} & \underline{373.5183} {\textbf{\scriptsize\textcolor{gray}{($\pm$675.90)}}} & 0.3126 {\textbf{\scriptsize\textcolor{gray}{($\pm$0.79)}}}\\
w/ CLIP~\cite{radford2021learningtransferablevisualmodels} & \underline{0.2695} {\textbf{\scriptsize\textcolor{gray}{($\pm$0.42)}}} & \underline{0.0508} {\textbf{\scriptsize\textcolor{gray}{($\pm$0.06)}}} & \underline{0.0250} {\textbf{\scriptsize\textcolor{gray}{($\pm$0.02)}}} & \underline{0.0771} {\textbf{\scriptsize\textcolor{gray}{($\pm$0.07)}}} & 383.5844 {\textbf{\scriptsize\textcolor{gray}{($\pm$766.41)}}} & \underline{0.3110} {\textbf{\scriptsize\textcolor{gray}{($\pm$0.85)}}}\\
w/ RGB colors  & 1.2176 {\textbf{\scriptsize\textcolor{gray}{($\pm$0.88)}}} & 0.6593 {\textbf{\scriptsize\textcolor{gray}{($\pm$0.49)}}} & 0.1379 {\textbf{\scriptsize\textcolor{gray}{($\pm$0.06)}}} & 1.1642 {\textbf{\scriptsize\textcolor{gray}{($\pm$0.78)}}} & 3678.4451 {\textbf{\scriptsize\textcolor{gray}{($\pm$8421.17)}}} & 4.7430 {\textbf{\scriptsize\textcolor{gray}{($\pm$2.85)}}} \\
\midrule \rowcolor{gray!15} \multicolumn{7}{l}{MatVAE (\Cref{sec:matvae})} \\
w/o MatVAE & 1.1284 {\textbf{\scriptsize\textcolor{gray}{($\pm$0.52)}}} & 0.1289 {\textbf{\scriptsize\textcolor{gray}{($\pm$0.08)}}} & 0.0480 {\textbf{\scriptsize\textcolor{gray}{($\pm$0.02)}}} & 0.1638 {\textbf{\scriptsize\textcolor{gray}{($\pm$0.08)}}} & 917.5879 {\textbf{\scriptsize\textcolor{gray}{($\pm$428.50)}}} & 0.8637 {\textbf{\scriptsize\textcolor{gray}{($\pm$0.61)}}}\\
\midrule \rowcolor{gray!15} \multicolumn{7}{l}{Normalization Scheme (\Cref{sec:matvae})} \\
w/ $Z$-score & 0.8838 {\textbf{\scriptsize\textcolor{gray}{($\pm$0.61)}}} & 0.0996 {\textbf{\scriptsize\textcolor{gray}{($\pm$0.08)}}} & 0.0814 {\textbf{\scriptsize\textcolor{gray}{($\pm$0.04)}}} & 0.2938 {\textbf{\scriptsize\textcolor{gray}{($\pm$0.20)}}} & 5269.2900 {\textbf{\scriptsize\textcolor{gray}{($\pm$946.03)}}} & 6.4656 {\textbf{\scriptsize\textcolor{gray}{($\pm$4.00)}}}\\
w/o $\log(\rho)$ & 0.6654 {\textbf{\scriptsize\textcolor{gray}{($\pm$0.54)}}} & 0.0748 {\textbf{\scriptsize\textcolor{gray}{($\pm$0.07)}}} & 0.0542 {\textbf{\scriptsize\textcolor{gray}{($\pm$0.03)}}} & 0.1806 {\textbf{\scriptsize\textcolor{gray}{($\pm$0.11)}}} & 549.9512 {\textbf{\scriptsize\textcolor{gray}{($\pm$513.57)}}} & 0.5976 {\textbf{\scriptsize\textcolor{gray}{($\pm$0.39)}}}\\
w/o $\log(E)$ & 0.9033 {\textbf{\scriptsize\textcolor{gray}{($\pm$0.60)}}} & 0.1024 {\textbf{\scriptsize\textcolor{gray}{($\pm$0.09)}}} & 0.1182 {\textbf{\scriptsize\textcolor{gray}{($\pm$0.05)}}} & 0.4189 {\textbf{\scriptsize\textcolor{gray}{($\pm$0.25)}}} & 4051.9121 {\textbf{\scriptsize\textcolor{gray}{($\pm$838.98)}}} & 5.0428 {\textbf{\scriptsize\textcolor{gray}{($\pm$3.22)}}}\\
\midrule \rowcolor{gray!15} \multicolumn{7}{l}{Loss} \\
w/ $\ell_1$ & 0.8947 {\textbf{\scriptsize\textcolor{gray}{($\pm$0.62)}}} & 0.1038 {\textbf{\scriptsize\textcolor{gray}{($\pm$0.09)}}} & 0.0468 {\textbf{\scriptsize\textcolor{gray}{($\pm$0.04)}}} & 0.1666 {\textbf{\scriptsize\textcolor{gray}{($\pm$0.16)}}} & 568.7543 {\textbf{\scriptsize\textcolor{gray}{($\pm$734.10)}}} & 0.6337 {\textbf{\scriptsize\textcolor{gray}{($\pm$0.88)}}}\\
\midrule Ours &  0.3765 {\textbf{\scriptsize\textcolor{gray}{($\pm$0.39)}}} & 
\textbf{0.0421} {\textbf{\scriptsize\textcolor{gray}{($\pm$0.05)}}} & 
\textbf{0.0250} {\textbf{\scriptsize\textcolor{gray}{($\pm$0.01)}}} & 
\textbf{0.0837} {\textbf{\scriptsize\textcolor{gray}{($\pm$0.03)}}} & 
\textbf{113.3807} {\textbf{\scriptsize\textcolor{gray}{($\pm$301.90)}}} & 
\textbf{0.0908} {\textbf{\scriptsize\textcolor{gray}{($\pm$0.14)}}}\\
\bottomrule
\end{tabular}
}
\label{tab:ablations}
\end{table*}

\else
    \begin{table*}[tb]
\centering
\begin{tabular}{lrrrrrr}
\toprule
\rowcolor{blue!15}Method & \multicolumn{2}{c}{Young's Modulus Pa ($E$)} & \multicolumn{2}{c}{Poisson's Ratio ($\nu$)} & \multicolumn{2}{c}{Density $\frac{kg}{m^3}$ ($\rho$)} \\
\cmidrule(r){2-3} \cmidrule(r){4-5} \cmidrule(r){6-7}
\rowcolor{blue!15}& ALDE ($\downarrow$) & ALRE ($\downarrow$) & ADE ($\downarrow$) & ARE ($\downarrow$) & ADE ($\downarrow$) & ARE ($\downarrow$) \\
\midrule \rowcolor{gray!15} \multicolumn{7}{l}{Image Features (Geometry Transformer start from random weights,~\Cref{sec:aggregating})} \\
w/ DINOv2~\cite{oquab2024dinov2learningrobustvisual} & 0.2888 {\textbf{\scriptsize\textcolor{gray}{($\pm$0.41)}}} & 0.0536 {\textbf{\scriptsize\textcolor{gray}{($\pm$0.06)}}} & 0.0259 {\textbf{\scriptsize\textcolor{gray}{($\pm$0.02)}}} & 0.0803 {\textbf{\scriptsize\textcolor{gray}{($\pm$0.08)}}} & \underline{373.5183} {\textbf{\scriptsize\textcolor{gray}{($\pm$675.90)}}} & 0.3126 {\textbf{\scriptsize\textcolor{gray}{($\pm$0.79)}}}\\
w/ CLIP~\cite{radford2021learningtransferablevisualmodels} & \underline{0.2695} {\textbf{\scriptsize\textcolor{gray}{($\pm$0.42)}}} & \underline{0.0508} {\textbf{\scriptsize\textcolor{gray}{($\pm$0.06)}}} & \underline{0.0250} {\textbf{\scriptsize\textcolor{gray}{($\pm$0.02)}}} & \underline{0.0771} {\textbf{\scriptsize\textcolor{gray}{($\pm$0.07)}}} & 383.5844 {\textbf{\scriptsize\textcolor{gray}{($\pm$766.41)}}} & \underline{0.3110} {\textbf{\scriptsize\textcolor{gray}{($\pm$0.85)}}}\\
w/ RGB colors  & 1.2176 {\textbf{\scriptsize\textcolor{gray}{($\pm$0.88)}}} & 0.6593 {\textbf{\scriptsize\textcolor{gray}{($\pm$0.49)}}} & 0.1379 {\textbf{\scriptsize\textcolor{gray}{($\pm$0.06)}}} & 1.1642 {\textbf{\scriptsize\textcolor{gray}{($\pm$0.78)}}} & 3678.4451 {\textbf{\scriptsize\textcolor{gray}{($\pm$8421.17)}}} & 4.7430 {\textbf{\scriptsize\textcolor{gray}{($\pm$2.85)}}} \\
\midrule \rowcolor{gray!15} \multicolumn{7}{l}{MatVAE (\Cref{sec:matvae})} \\
w/o MatVAE & 1.1284 {\textbf{\scriptsize\textcolor{gray}{($\pm$0.52)}}} & 0.1289 {\textbf{\scriptsize\textcolor{gray}{($\pm$0.08)}}} & 0.0480 {\textbf{\scriptsize\textcolor{gray}{($\pm$0.02)}}} & 0.1638 {\textbf{\scriptsize\textcolor{gray}{($\pm$0.08)}}} & 917.5879 {\textbf{\scriptsize\textcolor{gray}{($\pm$428.50)}}} & 0.8637 {\textbf{\scriptsize\textcolor{gray}{($\pm$0.61)}}}\\
\midrule \rowcolor{gray!15} \multicolumn{7}{l}{Normalization Scheme (\Cref{sec:matvae})} \\
w/ $Z$-score & 0.8838 {\textbf{\scriptsize\textcolor{gray}{($\pm$0.61)}}} & 0.0996 {\textbf{\scriptsize\textcolor{gray}{($\pm$0.08)}}} & 0.0814 {\textbf{\scriptsize\textcolor{gray}{($\pm$0.04)}}} & 0.2938 {\textbf{\scriptsize\textcolor{gray}{($\pm$0.20)}}} & 5269.2900 {\textbf{\scriptsize\textcolor{gray}{($\pm$946.03)}}} & 6.4656 {\textbf{\scriptsize\textcolor{gray}{($\pm$4.00)}}}\\
w/o $\log(\rho)$ & 0.6654 {\textbf{\scriptsize\textcolor{gray}{($\pm$0.54)}}} & 0.0748 {\textbf{\scriptsize\textcolor{gray}{($\pm$0.07)}}} & 0.0542 {\textbf{\scriptsize\textcolor{gray}{($\pm$0.03)}}} & 0.1806 {\textbf{\scriptsize\textcolor{gray}{($\pm$0.11)}}} & 549.9512 {\textbf{\scriptsize\textcolor{gray}{($\pm$513.57)}}} & 0.5976 {\textbf{\scriptsize\textcolor{gray}{($\pm$0.39)}}}\\
w/o $\log(E)$ & 0.9033 {\textbf{\scriptsize\textcolor{gray}{($\pm$0.60)}}} & 0.1024 {\textbf{\scriptsize\textcolor{gray}{($\pm$0.09)}}} & 0.1182 {\textbf{\scriptsize\textcolor{gray}{($\pm$0.05)}}} & 0.4189 {\textbf{\scriptsize\textcolor{gray}{($\pm$0.25)}}} & 4051.9121 {\textbf{\scriptsize\textcolor{gray}{($\pm$838.98)}}} & 5.0428 {\textbf{\scriptsize\textcolor{gray}{($\pm$3.22)}}}\\
\midrule \rowcolor{gray!15} \multicolumn{7}{l}{Loss} \\
w/ $\ell_1$ & 0.8947 {\textbf{\scriptsize\textcolor{gray}{($\pm$0.62)}}} & 0.1038 {\textbf{\scriptsize\textcolor{gray}{($\pm$0.09)}}} & 0.0468 {\textbf{\scriptsize\textcolor{gray}{($\pm$0.04)}}} & 0.1666 {\textbf{\scriptsize\textcolor{gray}{($\pm$0.16)}}} & 568.7543 {\textbf{\scriptsize\textcolor{gray}{($\pm$734.10)}}} & 0.6337 {\textbf{\scriptsize\textcolor{gray}{($\pm$0.88)}}}\\
\midrule Ours &  \voxelproprow\\
\bottomrule
\end{tabular}
\caption{\textbf{Ablations.} We ablate \acronym\ with choice of image features, using MatVAE, normalization scheme, and choice of a loss function. We report the voxel-level mechanical property difference errors.}
\label{tab:ablations}
\end{table*}

\fi
\section{Metrics}
\label{app:metrics}

We present an explanation of the metrics we use, and experiments on interpreting these metrics. 

\subsection{Metrics for Mass and Field Estimation}
\label{app:metricsfield}

To evaluate the accuracy of predicted scalar quantities such as object mass, as well as continuous scalar fields like density or stiffness, we use several commonly adopted metrics. Let $y$ denote a ground-truth scalar value or voxel-wise field (e.g., density), and $\hat{y}$ its predicted counterpart.

\paragraph{Absolute Difference Error (ADE).}  
The average absolute error between predicted and ground-truth values:
\begin{equation}
    \mathrm{ADE} = \frac{1}{N} \sum_{i=1}^N |y_i - \hat{y}_i|.
\end{equation}

This metric is scale-sensitive and reports the error in physical units (e.g., $\mathrm{kg/m^3}$ for density, $\mathrm{kg}$ for mass).

\paragraph{Absolute Log Difference Error (ALDE).}  
The average absolute error in logarithmic space:
\begin{equation}
    \mathrm{ALDE} = \frac{1}{N} \sum_{i=1}^N |\log y_i - \log \hat{y}_i|.
\end{equation}

This metric captures multiplicative error and is particularly useful for quantities that vary over several orders of magnitude.

\paragraph{Average Relative Error (ARE).}  
The mean relative deviation between predictions and ground truth:
\begin{equation}
    \mathrm{ARE} = \frac{1}{N} \sum_{i=1}^N \left| \frac{y_i - \hat{y}_i}{y_i} \right|.
\end{equation}

This dimensionless metric penalizes over- and under-estimates proportionally, making it appropriate for comparing across varying scales.

\paragraph{Minimum Ratio Error (MnRE).}  
A symmetric and bounded measure of relative accuracy:
\begin{equation}
\mathrm{MnRE} = \frac{1}{N} \sum_{i=1}^N \min\left( \frac{y_i}{\hat{y}_i}, \frac{\hat{y}_i}{y_i} \right).
\end{equation}

This metric ranges from $0$ to $1$ and is maximized when predictions are perfectly accurate. As suggested in prior work~\cite{mass_estimation_visual}, MnRE avoids bias toward systematic over- or under-estimation and reduces sensitivity to outliers, making it particularly effective for evaluating physical quantity predictions across heterogeneous samples.

\subsection{Metrics to Measure Differences in Mechanical Properties}
\label{app:metricsmech}

We use multiple commonly used metrics for measuring differences between mechanical properties.

\paragraph{Relative Error in $\log(E)$.}  
Relative error between predicted and true values of the logarithm of Young’s modulus $E$ reported in units of Pa. This captures relative error in material stiffness across several orders of magnitude.

\paragraph{Relative Error in $\nu$.}  
Relative Error in linear space for Poisson’s ratio $\nu$, a dimensionless measure of lateral contraction under uniaxial loading.

\paragraph{Relative Error  in $\rho$.}  
Relative Error between predicted and true values of material density $\rho$, reported in units of $\mathrm{kg/m^3}$.

\paragraph{Relative Error in $\log(E/\rho)$.}  
Relative Error in the logarithm of specific modulus, where $E$ is Young’s modulus and $\rho$ is density. Reflects relative deviation in stiffness-to-weight efficiency.

\paragraph{Relative Error  in $\log(G)$.}  
Relative Error in the logarithm of shear modulus $G = \frac{E}{2(1 + \nu)}$, representing resistance to shear deformation.

\paragraph{Relative Error  in $\log(K)$.}  
Relative Error in the logarithm of bulk modulus $K = \frac{E}{3(1 - 2\nu)}$, characterizing resistance to uniform volumetric compression.

\paragraph{Lightweight Stiffness Ashby Index ($P = E^{1/2}/\rho$).} The Relative Error in $\log(P)$, where $P = E^{1/2}/\rho$, reflecting relative error in predicting material efficiency for maximizing stiffness per unit weight~\cite{ashby1993materials}.

\paragraph{Energy Absorption Ashby Index ($P = E^{1/3}/\rho$).} The Relative Error in $\log(P)$, where $P = E^{1/3}/\rho$, quantifying relative deviation in predicted energy absorption efficiency~\cite{ashby1993materials}.

\paragraph{Bray–Curtis dissimilarity.}  
Bray–Curtis dissimilarity~\cite{https://doi.org/10.2307/1942268} between predicted and ground-truth property vectors $\mathbf{x}$ and $\mathbf{y}$:
\begin{equation}
\mathrm{BC}(\mathbf{x}, \mathbf{y}) = \frac{\sum_i |x_i - y_i|}{\sum_i (x_i + y_i)}.
\end{equation}

A normalized, dimensionless measure in $[0, 1]$ capturing overall distributional divergence across multiple material properties.

\subsection{Metrics to Measure Differences in Distributions}
\label{app:metricsdist}

We use multiple commonly-used metrics for measuring differences between distributions.

\paragraph{Wasserstein--1 Distance ($W_1$).}
For probability measures $\mu,\nu$ on a space $\mathcal{X}$,
\begin{equation}
    W_1(\mu,\nu)=\inf_{\gamma\in\Pi(\mu,\nu)}
\int_{\mathcal{X}\times\mathcal{X}}\|x-y\|\,d\gamma(x,y),
\end{equation}
where $\Pi(\mu,\nu)$ is the set of all couplings of $\mu$ and $\nu$. $W_1$ equals the minimum average “work” to move mass from $\mu$ to $\nu$.

\paragraph{Wasserstein--2 Distance ($W_2$).}
For probability measures $\mu,\nu$ on a space $\mathcal{X}$,
\begin{equation}
    W_2(\mu,\nu)=
\Bigl(\inf_{\gamma\in\Pi(\mu,\nu)}
\int_{\mathcal{X}\times\mathcal{X}}\|x-y\|^{2}\,d\gamma(x,y)\Bigr)^{1/2},
\end{equation}
where $\Pi(\mu,\nu)$ is the set of all couplings of $\mu$ and $\nu$. The root‐mean‐square transport cost between $\mu$ and $\nu$.

\paragraph{Kullback–Leibler Divergence ($D_{\mathrm{KL}}$).}
For densities $p,q$ on $\mathcal{X}$,
\begin{equation}
    D_{\mathrm{KL}}(p\|q)=\int_{\mathcal{X}} p(x)\,\log\frac{p(x)}{q(x)}\,dx.
\end{equation}

$D_{\mathrm{KL}}$ measures the expected extra log‐likelihood of data drawn from $p$ when it is coded using $q$ instead of $p$.

\begin{figure*}[tb]
    \centering
    \setlength{\tabcolsep}{0pt}
    \begin{tabular}{@{}ccccc@{}}
        \includegraphics[width=0.2\textwidth]{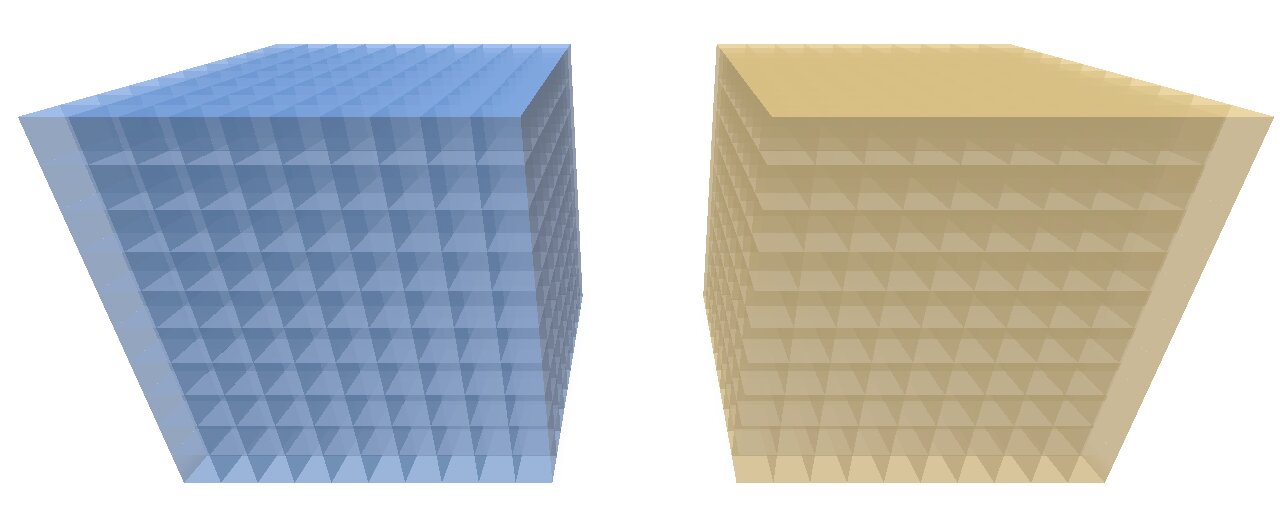}%
        \includegraphics[width=0.2\textwidth]{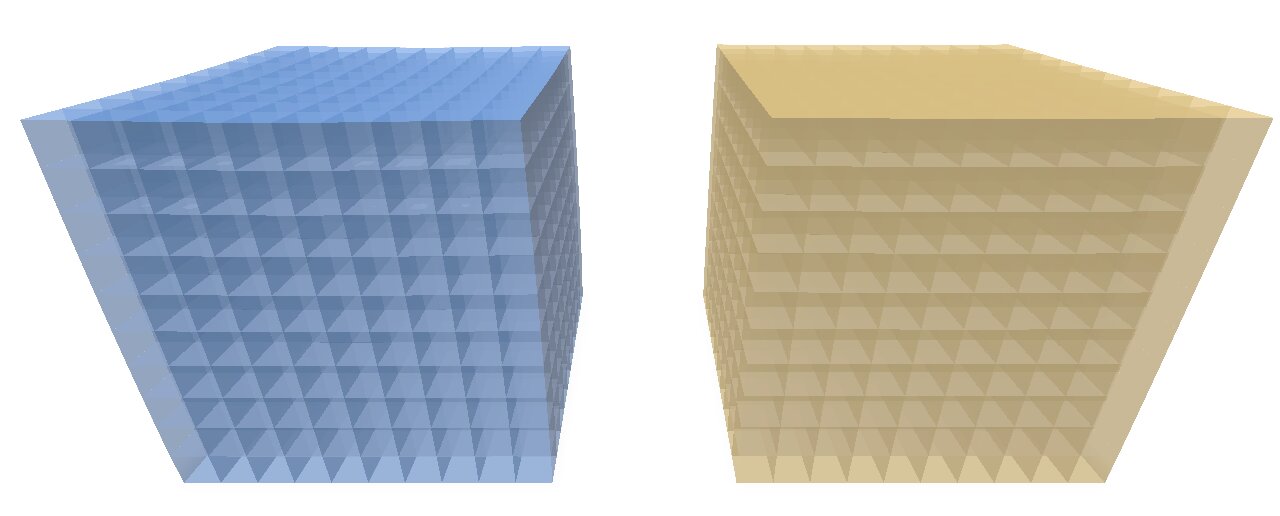}%
        \includegraphics[width=0.2\textwidth]{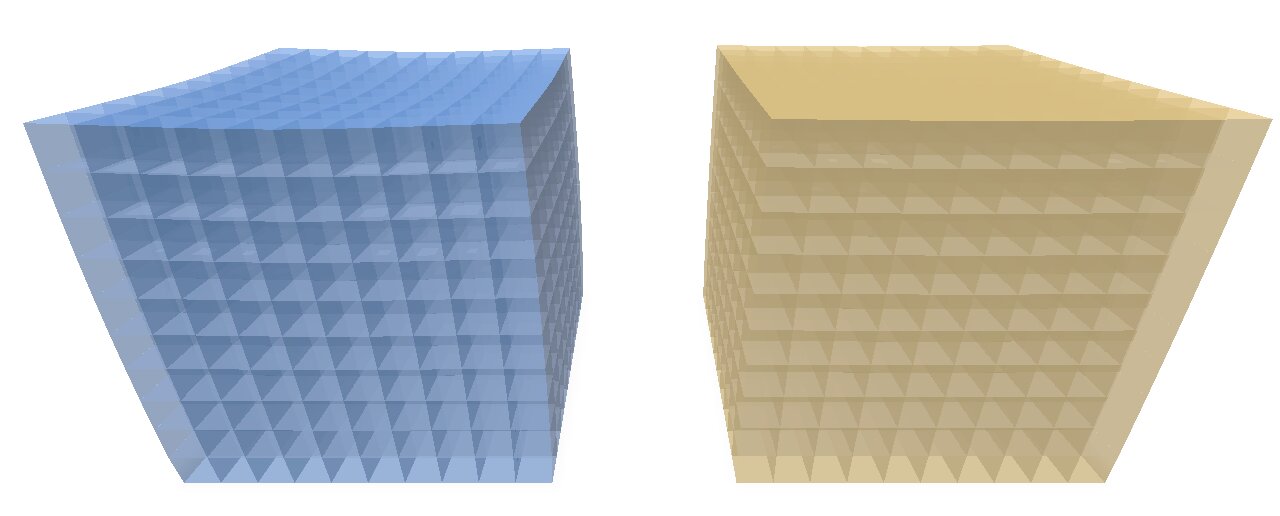}%
        \includegraphics[width=0.2\textwidth]{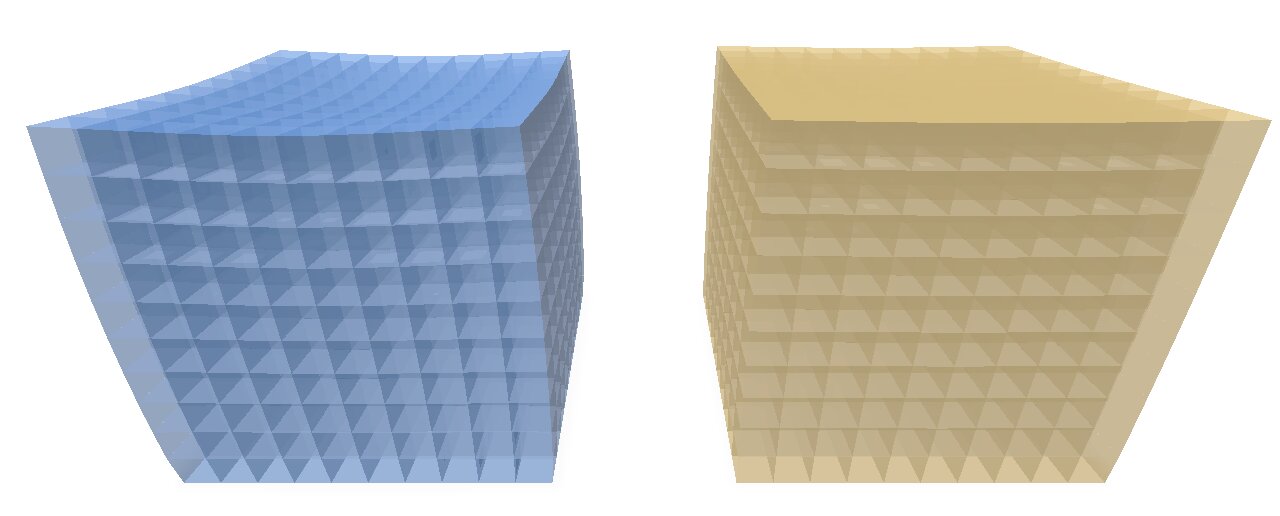}%
        \includegraphics[width=0.2\textwidth]{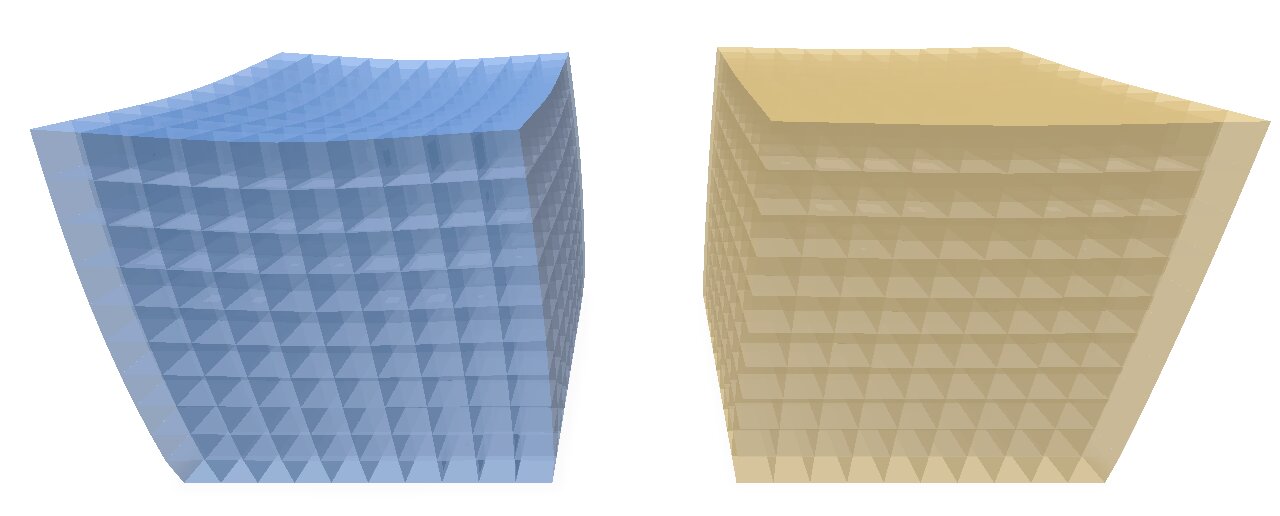}\\
        \includegraphics[width=0.2\textwidth]{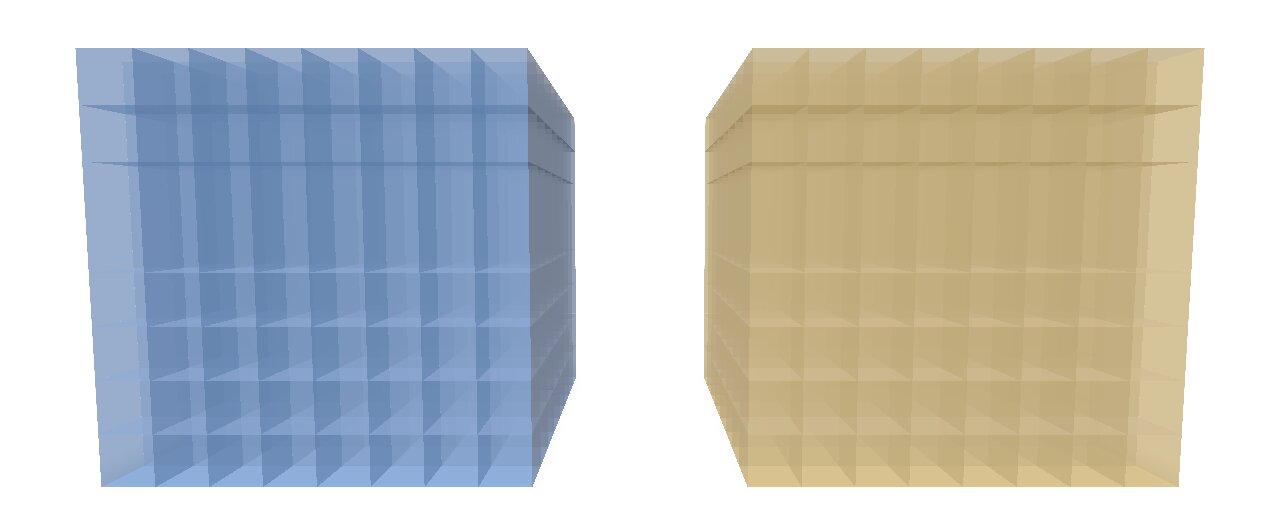}%
        \includegraphics[width=0.2\textwidth]{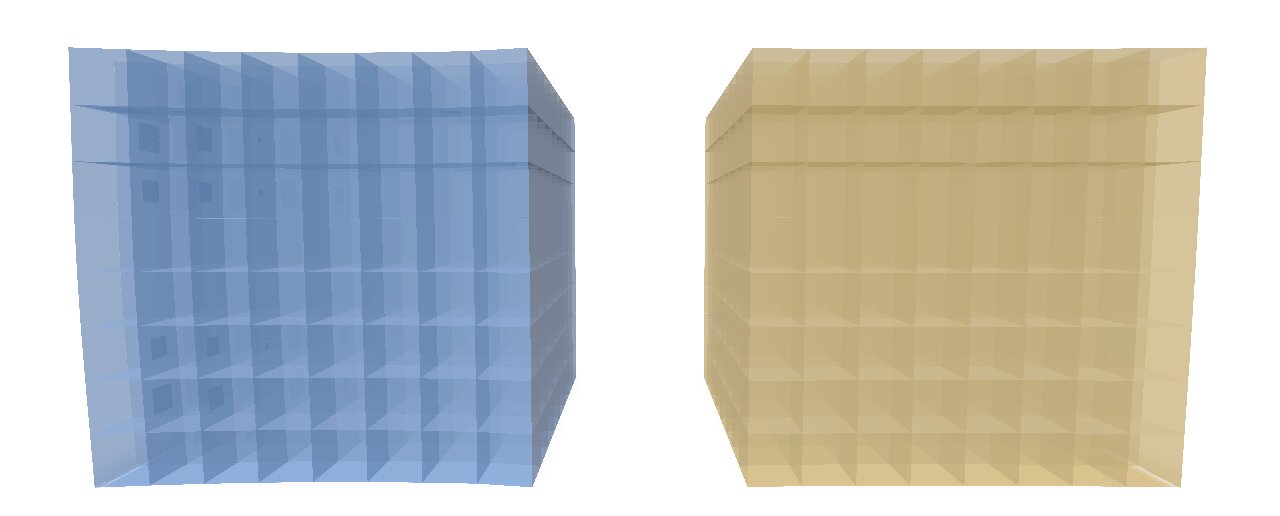}%
        \includegraphics[width=0.2\textwidth]{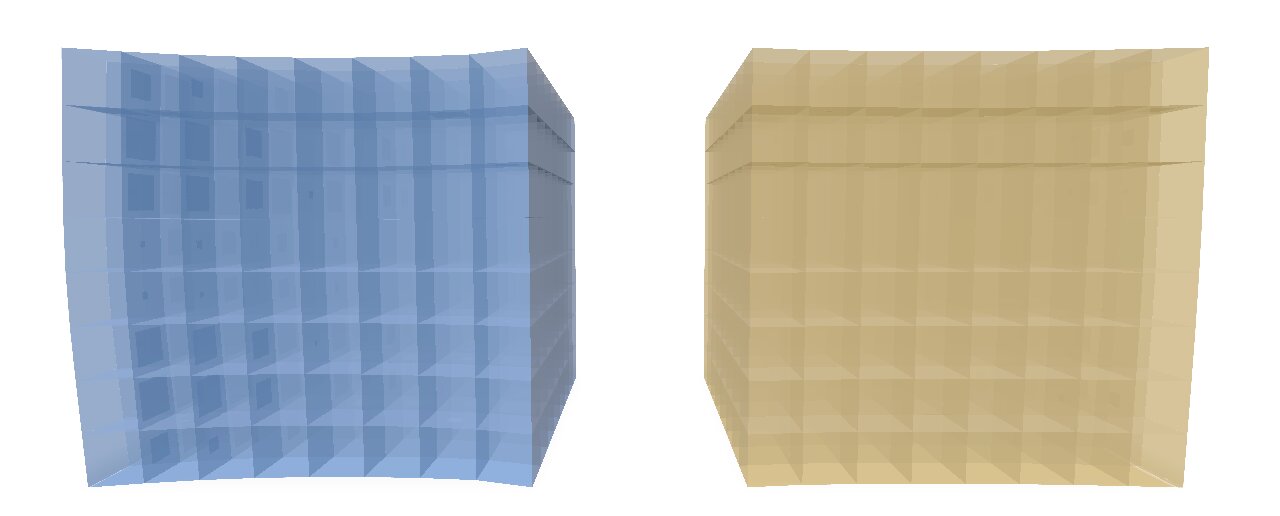}%
        \includegraphics[width=0.2\textwidth]{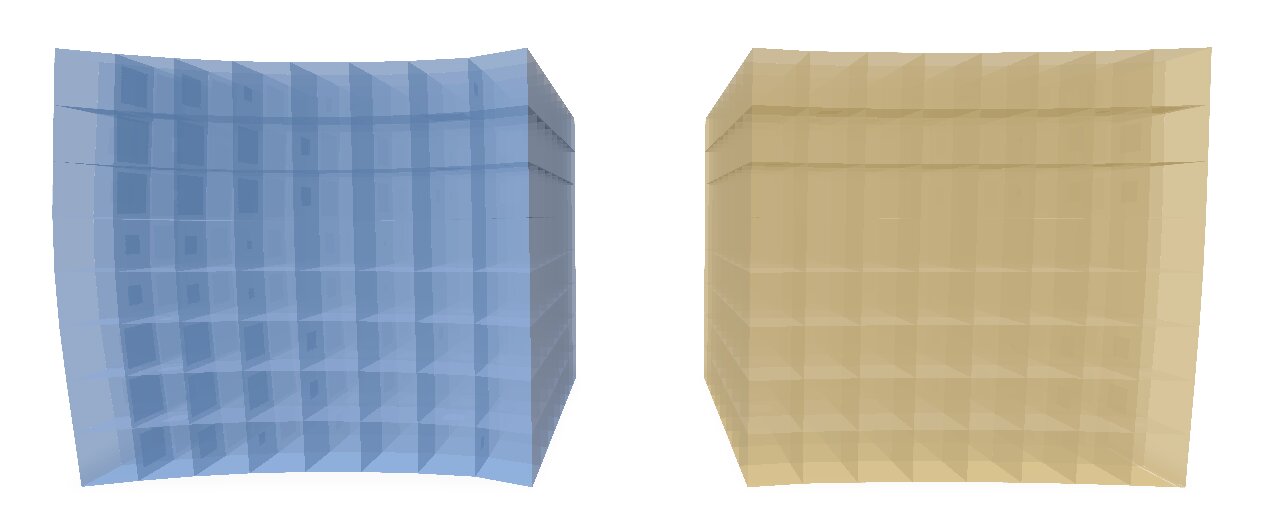}%
        \includegraphics[width=0.2\textwidth]{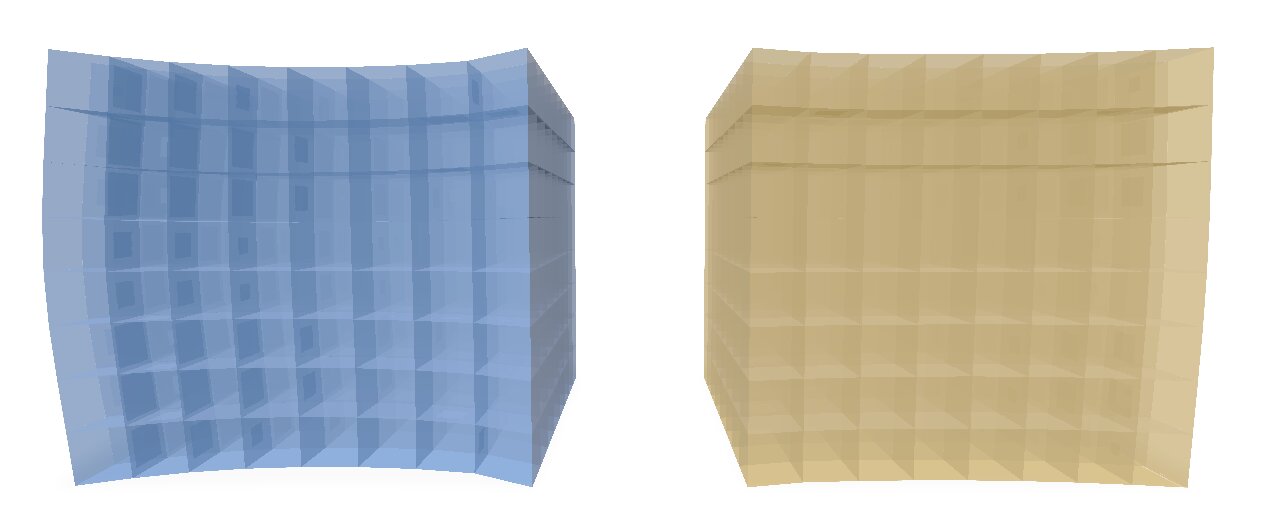}\\
        \bigarrow{time}
    \end{tabular}
    \caption{\textbf{Simulations to Interpret Errors.} We demonstrate simulations performed to show the relation between relative error and simulation error.}
    \label{fig:interpreterrorssim}
\end{figure*}

\subsection{Interpreting Errors for Material Property Estimation}
\label{app:interpret}

We experimentally demonstrate an interpretation of how relative changes in material properties affect simulations of the finite element method (FEM) solver. We do so by simulating the deformation of unit cubes under many different material properties and scenarios with an FEM solver.

For each baseline triplet $(E_0, \nu_0, \rho_0)$ representing Young's modulus, Poisson's ratio, and density, we introduce variations following the scaling laws: density variations follow linear scaling $\rho_{\text{new}} = \rho_0(1 + \Delta)$, Poisson's ratio variations use the same linear relationship $\nu_{\text{new}} = \nu_0(1 + \Delta)$, while Young's modulus variations use exponential scaling $E_{\text{new}} = E_0 e^{\Delta}$ to accommodate the wide range of stiffness values. We then apply every such unique material triplet to a unit cube and perform a simulation under some external forces.

During each of these simulations, we measure the final volume and potential energy of the cube after the Newton iterations have converged.

\paragraph{Measuring Volume.} For a body undergoing deformation, the deformation gradient $\mathbf{F} = \nabla \mathbf{u} + \mathbb{I}$ maps material points from the reference to the current configuration, where $\mathbf{u}$ represents the displacement field and $\mathbb{I}$ is the $3\times3$ identity tensor. The local volume change is quantified by the Jacobian $J = \det(\mathbf{F})$, which represents the ratio of deformed to reference volume at each material point. The total deformed volume is: $V_{\text{def}} = \int_{\Omega_0} J \, dV$, where $\Omega_0$ denotes the reference configuration. The relative volume change, defined as $\Delta V/V = (V_{\text{def}} - V_0)/V_0$, provides a dimensionless measure of volumetric deformation.
\paragraph{Measuring Potential Energy.} We compute the total potential energy by combining elastic strain energy and kinetic-potential contributions. We use corotated linear elasticity, where we calculate the deformation gradient $\mathbf{F} = \nabla \mathbf{u} + \mathbb{I}$ and symmetric strain tensor $\mathbf{S} = \frac{1}{2}(\mathbf{F} + \mathbf{F}^T) - \mathbb{I}$ to obtain the energy density $W = \mu \, \text{tr}(\mathbf{S}^2) + \frac{\lambda}{2} (\text{tr}(\mathbf{S}))^2$, with Lamé parameters $\mu$ and $\lambda$ derived from the Young's modulus and Poisson's ratio. We use three distinct contributions in the kinetic-potential term: an inertial component $E_{\text{inertia}} = \int_{\Omega} \frac{\rho}{2\Delta t^2} |\mathbf{u}^{n+1} - \mathbf{u}^n|^2 \, dV$ that captures displacement changes between iterations in our quasi-static solver, a gravitational potential $E_{\text{gravity}} = -\int_{\Omega} \rho \, \mathbf{u} \cdot \mathbf{g} \, dV$ accounting for body forces, and an external work term $E_{\text{ext}} = -\int_{\Omega} \mathbf{u} \cdot \mathbf{f}_{\text{ext}} \, dV$ representing the applied loads. We thus compute the total potential energy as $E_{\text{total}} = \int_{\Omega} W \, dV + E_{\text{inertia}} + E_{\text{gravity}} + E_{\text{ext}}$, evaluated at the converged displacement field.

\begin{figure*}[tb]
    \centering
    {\setlength{\tabcolsep}{0pt}
    \begin{tabular}{ccc}
    Young's Modulus ($E$) & Poisson's Ratio ($\nu$) & Density ($\rho$) \\
    \includegraphics[width=0.32\textwidth]{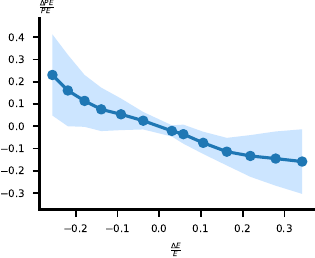} & \includegraphics[width=0.32\textwidth]{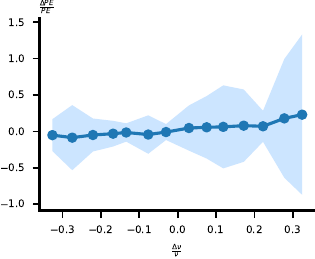} & \includegraphics[width=0.32\textwidth]{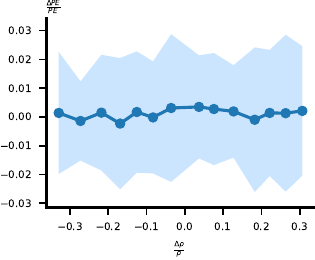} \\
    \includegraphics[width=0.32\textwidth]{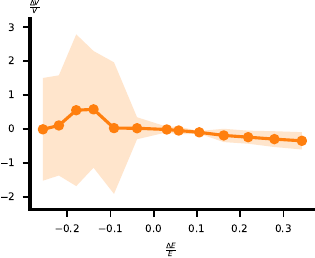} & \includegraphics[width=0.32\textwidth]{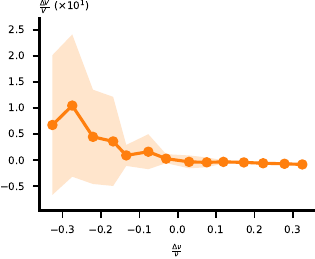} & \includegraphics[width=0.32\textwidth]{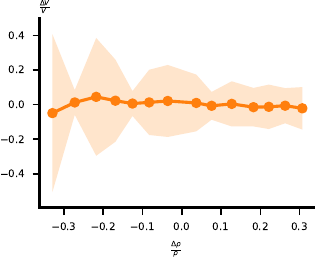}
    \end{tabular}}
    \caption{\textbf{Gripping Force by Robots.} We demonstrate the relation between relative errors in materials and relative change in {\color{mainblue}P.E. (top)} and {\color{mainorange}volume (bottom)}. We then show the {\color{lightblue}confidence} {\color{lightorange}bounds} in light shaded regions.}
    \label{fig:gripper}
\end{figure*}

We perform the simulations in the following scenarios,

\paragraph{Gripping force by robots.} We simulate a 140 N compressive force, which is common in robotic gripping applications, for example, the Franka Emika~\cite{franka_description} ``Hand'' end effector applies a maximum of 70 N per finger with a maximum clamping force of 140 N. We demonstrate the results from 486 simulations in this setting, all of which were run to convergence in~\Cref{fig:gripper}.

\begin{figure*}[tb]
    \centering
    {\setlength{\tabcolsep}{0pt}
    \begin{tabular}{ccc}
    Young's Modulus ($E$) & Poisson's Ratio ($\nu$) & Density ($\rho$) \\
    \includegraphics[width=0.32\textwidth]{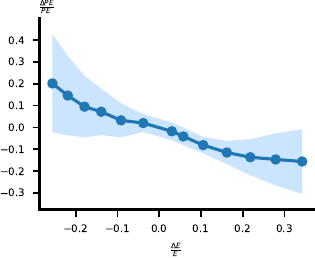} & \includegraphics[width=0.32\textwidth]{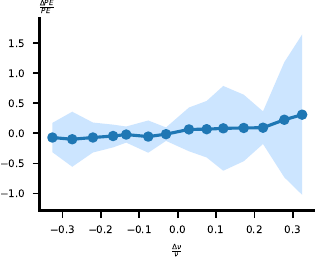} & \includegraphics[width=0.32\textwidth]{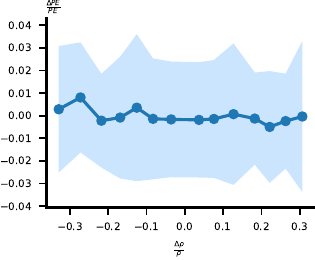} \\
    \includegraphics[width=0.32\textwidth]{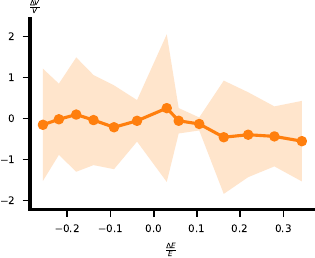} & \includegraphics[width=0.32\textwidth]{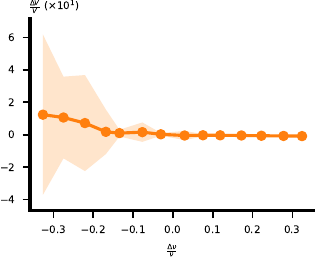} & \includegraphics[width=0.32\textwidth]{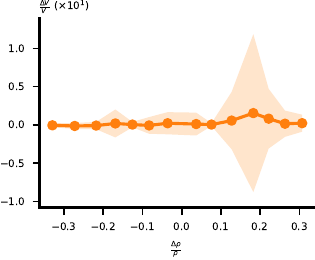}
    \end{tabular}}
    \caption{\textbf{Impact Force on Dropping Objects.} We demonstrate the relation between relative errors in materials and relative change in {\color{mainblue}P.E. (top)} and {\color{mainorange}volume (bottom)}. We show the {\color{lightblue}confidence} {\color{lightorange}bounds} in light shaded regions.}
    \label{fig:packagedrop}
\end{figure*}

\paragraph{Impact Force on Dropping Objects.} We simulate a 120 N impact force that simulates package drop scenarios, calculated from the impact dynamics of a 1 kg package dropped from 0.6 m height with a 5 cm deformation distance. We demonstrate the results from 486 simulations in this setting, all of which were run to convergence in~\Cref{fig:packagedrop}.

\begin{figure*}[tb]
    \centering
    {\setlength{\tabcolsep}{0pt}
    \begin{tabular}{ccc}
    Young's Modulus ($E$) & Poisson's Ratio ($\nu$) & Density ($\rho$) \\
    \includegraphics[width=0.32\textwidth]{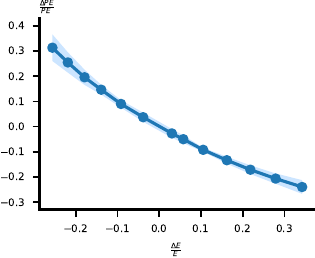} & \includegraphics[width=0.32\textwidth]{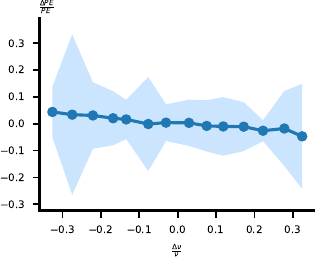} & \includegraphics[width=0.32\textwidth]{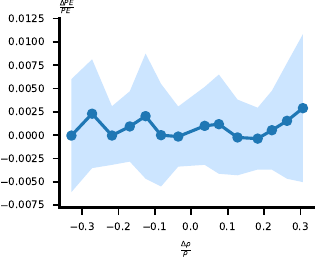} \\
    \includegraphics[width=0.32\textwidth]{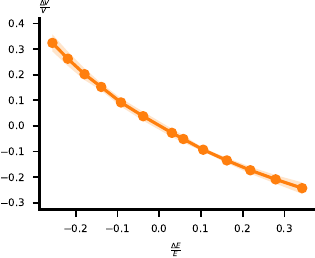} & \includegraphics[width=0.32\textwidth]{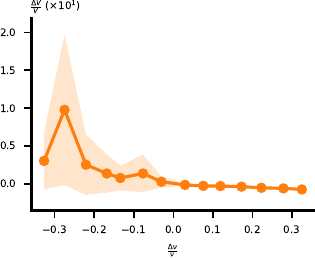} & \includegraphics[width=0.32\textwidth]{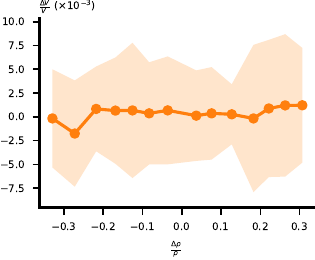}
    \end{tabular}}
    \caption{\textbf{Tensile Testing Machine.} We demonstrate the relation between relative errors in materials and relative change in {\color{mainblue}P.E. (top)} and {\color{mainorange}volume (bottom)}. We show the {\color{lightblue}confidence} {\color{lightorange}bounds} in light shaded regions.}
    \label{fig:tensiletest}
\end{figure*}

\paragraph{Tensile Testing Machines.} We simulate a 330 N force corresponding to standard tensile testing conditions employed in bench-top universal testing machines~\cite{ASTM_D638_2022, ASTM_E8_E8M_2024}. We demonstrate the results from 486 simulations in this setting, all of which were run to convergence in~\Cref{fig:tensiletest}.

\begin{figure*}[tb]
    \centering
    {\setlength{\tabcolsep}{0pt}
    \begin{tabular}{ccc}
    Young's Modulus ($E$) & Poisson's Ratio ($\nu$) & Density ($\rho$) \\
    \includegraphics[width=0.32\textwidth]{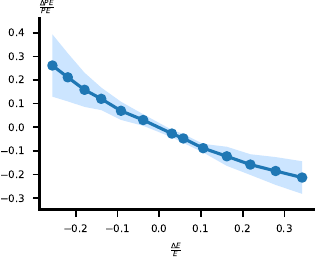} & \includegraphics[width=0.32\textwidth]{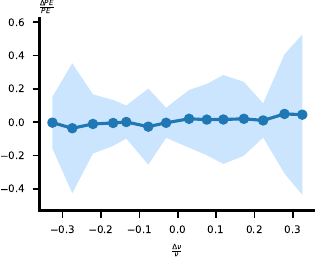} & \includegraphics[width=0.32\textwidth]{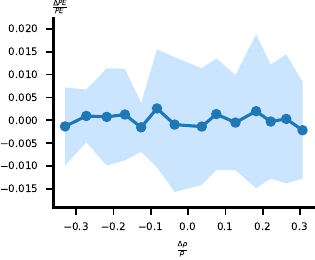} \\
    \includegraphics[width=0.32\textwidth]{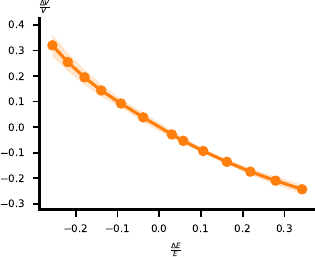} & \includegraphics[width=0.32\textwidth]{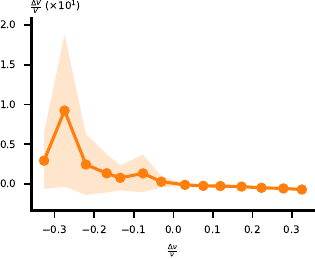} & \includegraphics[width=0.32\textwidth]{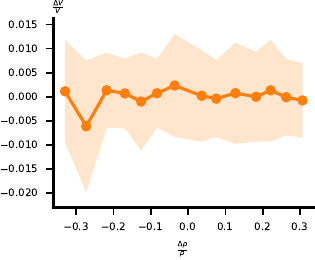}
    \end{tabular}}
    \caption{\textbf{Tension.} We demonstrate the relation between relative errors in materials and relative change in {\color{mainblue}P.E. (top)} and {\color{mainorange}volume (bottom)}. We show the {\color{lightblue}confidence} {\color{lightorange}bounds} in light shaded regions.}
    \label{fig:cable}
\end{figure*}

\paragraph{Tension.} We simulate a 200 N force, which represents typical pretension in tendon-driven robotic systems, where continuum arms and wearable assistive devices maintain structural stiffness through cable tensions~\cite{Sch_ffer_2024}. We demonstrate the results from 486 simulations in this setting all of which were run to convergence in~\Cref{fig:cable}.

\section{Dataset Details}
\label{app:datasetdetails}

We present additional details about our datasets for training MatVAE (\Cref{sec:matvae}) and Geometry Transformer (\Cref{sec:3dencdoer}).

% \subsection{Material Triplet Dataset}
% \label{app:tripletdata}
% To create our training dataset for the MatVAE, we first collect a dataset of mechanical property ranges for many different materials from multiple online databases~\cite{matweb, wikipedia_density, wikipedia_poissons_ratio, wikipedia_youngs_modulus, engineeringtoolbox_materials, cambridge_materials_databook}, which include values obtained experimentally. We then sample triplets from within these ranges proportional to the size of the range for each mechanical property. Finally, as shown in~\Cref{tab:matdata}, we filter this dataset to remove duplicate triplets often caused by multiple materials having overlapping property ranges.
\subsection{Annotation with Vision-Language Model}
\label{app:annotvlm}

\begin{figure*}[tb]
\centering
\begin{tcolorbox}[
    colback=blue!5!white,       % light blue/purple background
    colframe=blue!50!black,     % border color
    arc=3mm,                    % rounded corners
    boxrule=0.5pt,              % thickness of border
    width=\textwidth,           % span full page width
    left=3mm, right=3mm,
    top=2mm, bottom=2mm
]
\textbf{Object.} \\
\includegraphics[width=0.5\textwidth]{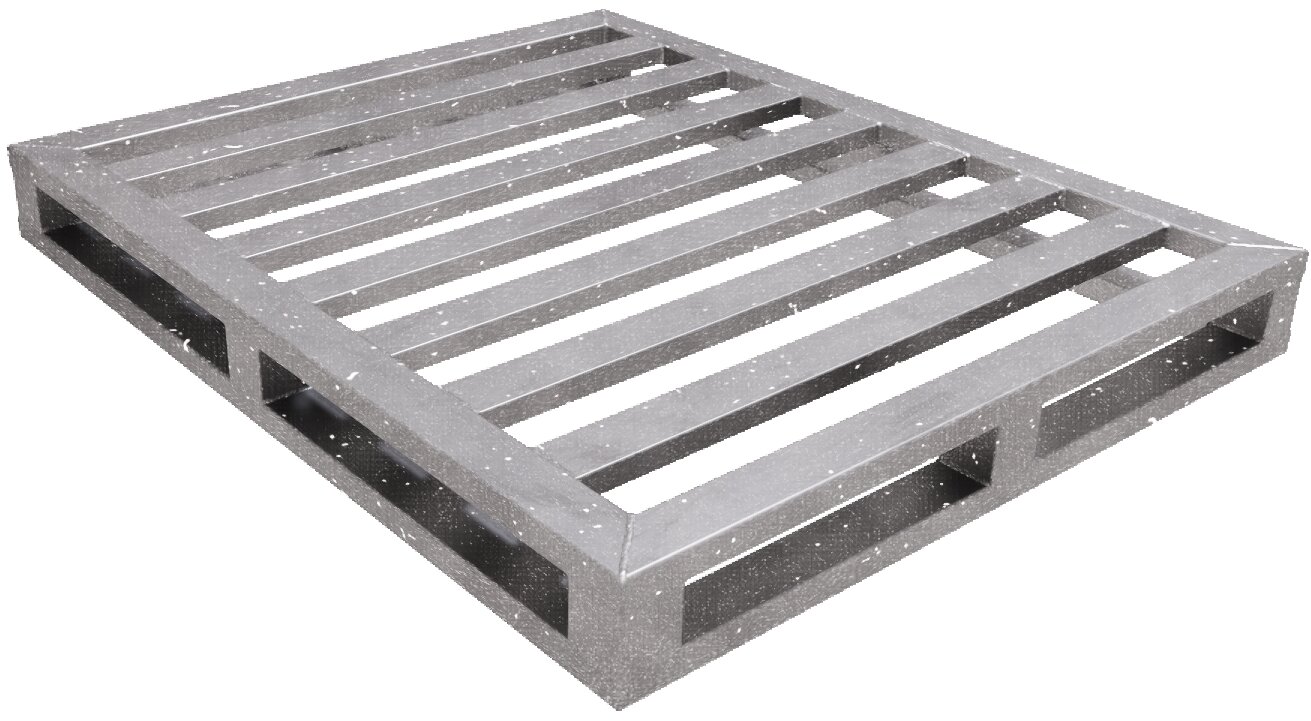}
\medskip

\textbf{VLM Response.}
\medskip

Step 1: Identify the material class/type based on visual appearance \\
The material appears to be metallic, specifically aluminum, given the context of it being used in a pallet and the provided density value which matches aluminum.\\

Step 2: Describe the surface characteristics (texture, reflectivity, color)  \\
The surface is smooth with a matte finish, showing minimal reflectivity. The color is a uniform light gray, typical of aluminum.\\

Step 3: Determine the specific material subtype considering its physical properties  \\
Considering the density of 2700 kg/m³, the material is most likely aluminum, a common choice for pallets due to its lightweight yet durable properties.\\

Step 4: Reason through each property estimate based on visual and measured data  \\
- \textasteriskcentered\textasteriskcentered Young's modulus\textasteriskcentered\textasteriskcentered: Aluminum typically has a Young's modulus around 70 GPa. Given the usage as a pallet, it needs to be stiff but not overly so, supporting the estimate of 70 GPa.\\
- \textasteriskcentered\textasteriskcentered Poisson's ratio\textasteriskcentered\textasteriskcentered: Aluminum generally has a Poisson's ratio around 0.33, which is typical for many metals.\\
- \textasteriskcentered\textasteriskcentered Density\textasteriskcentered\textasteriskcentered: The provided density is 2700 kg/m\textasciicircum3, which is consistent with aluminum.\\

Young's modulus: 7.0e10 Pa  \\
Poisson's ratio: 0.33  \\
Density: 2.7e3 kg/m\textasciicircum3\\
\end{tcolorbox}
\caption{\textbf{Example Response I.} We demonstrate an example response for a segment from one of the objects from our dataset. The given object has one part.}
\label{fig:response1}
\end{figure*}

\begin{figure*}[tb]
\centering
\begin{tcolorbox}[
    colback=blue!5!white,       % light blue/purple background
    colframe=blue!50!black,     % border color
    arc=3mm,                    % rounded corners
    boxrule=0.5pt,              % thickness of border
    width=\textwidth,           % span full page width
    left=3mm, right=3mm,
    top=2mm, bottom=2mm
]
\textbf{Object.} \\
\includegraphics[width=0.5\textwidth]{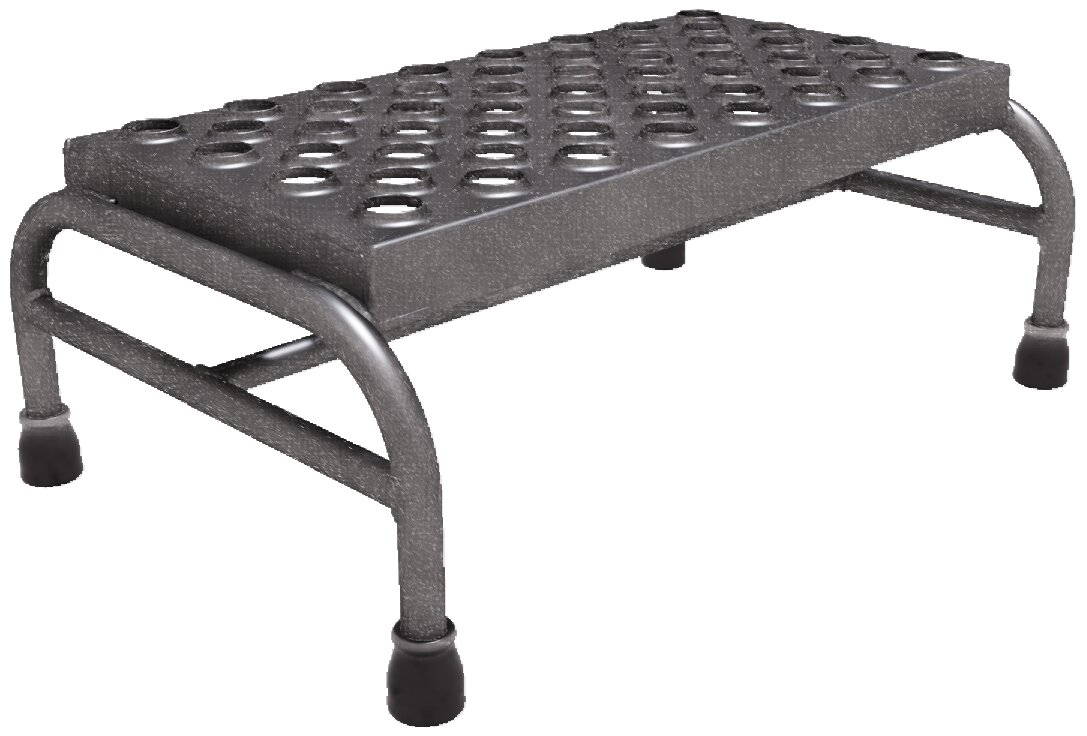}
\medskip

\textbf{VLM Response.}
\medskip

Step 1: Identify the material class/type based on visual appearance  \\
The material appears to be a type of rubber, specifically used for antiskid caps based on the context provided. The surface of the sphere suggests a matte finish, which is common for rubber materials designed for grip and friction.\\

Step 2: Describe the surface characteristics (texture, reflectivity, color)  \\
The surface of the sphere is matte with a uniform gray color, indicating a lack of significant reflectivity. There is a slight texture visible, suggesting it is not perfectly smooth, which is consistent with rubber materials intended for anti-slip applications.\\

Step 3: Determine the specific material subtype considering its physical properties  \\
Given the usage as an antiskid cap and the provided physical properties such as density (1100.0 kg/m\textasciicircum3), dynamic and static friction (both 0.7), and restitution (0.8), the material is likely a type of rubber with good elasticity and frictional properties. The density matches well with EPDM rubber, which is commonly used for such applications due to its durability and resistance to environmental factors.\\

Step 4: Reason through each property estimate based on visual and measured data  \\
- **Young's modulus**: Considering the material is a rubber used for antiskid purposes, it needs to be flexible yet provide sufficient stiffness to maintain shape under load. The given density and usage suggest a medium-hard rubber, so an estimate within the upper range of rubber's Young's modulus is appropriate.\\
- **Poisson's ratio**: Rubbers typically exhibit high Poisson's ratios close to 0.5, indicating significant lateral expansion when compressed. Given the material's flexibility and the need for good grip, a value near the upper end of the rubber range is suitable.\\
- **Density**: The provided density is already specified as 1100.0 kg/m\textasciicircum3, which aligns well with EPDM rubber.\\

Young's modulus: 1.0e8 Pa  \\
Poisson's ratio: 0.49  \\
Density: 1.1e3 kg/m\textasciicircum3\\
\end{tcolorbox}
\caption{\textbf{Example Response II.} We demonstrate an example response for a segment from one of the objects from our dataset. The given object has two parts, and we show the response for the "rubber cap" part.}
\label{fig:response2}
\end{figure*}

To create our training dataset (\Cref{sec:data}) for the Geometry Transformer, we use a Vision-Language Model (VLM) coupled with multiple other data sources like 3D assets, component-wise part segmentations, material databases (\Cref{sec:mtd}), visual textures, and material names to annotate our dataset. We run the VLM on every segment of every object individually. We experiment with Qwen2.5-VL 7B, Qwen2.5-VL 32B, Qwen2.5-VL 72B~\cite{bai2023qwenvlversatilevisionlanguagemodel, bai2025qwen25vltechnicalreport}, VL-Rethinker~\cite{wang2023autorecon}, SpatialRGPT~\cite{cheng2024spatialrgpt}, and Cosmos Nemotron~\cite{lin2024vilapretrainingvisuallanguage}. We experimentally choose Qwen2.5-VL 72B for the data annotation. We show the system prompts and the user prompts that we use in~\Cref{fig:sysprompt,fig:userprompt1,fig:userprompt2,fig:userprompt3}. We find the best performing system prompts with TextGrad~\cite{yuksekgonul2025optimizing}. We show a response from the model for one of the segments from an object in our dataset in~\Cref{fig:response1,fig:response2}.

\begin{figure*}[tb]
\centering
\begin{tcolorbox}[
    colback=blue!5!white,       % light blue/purple background
    colframe=blue!50!black,     % border color
    arc=3mm,                    % rounded corners
    boxrule=0.5pt,              % thickness of border
    width=\textwidth,           % span full page width
    left=3mm, right=3mm,
    top=2mm, bottom=2mm
]
\textbf{System Prompt.}
\medskip

You are a materials science expert specializing in analyzing material properties from visual appearances and physical data. Your task is to provide precise numerical estimates for Young's modulus, Poisson's ratio, and density based on the images and context provided.\\

Important Context: The material segment you are analyzing may be an internal component or structure that is not visible from the outside of the object. For example:\\
- Internal support structures, frames, or reinforcements\\
- Hidden layers or core materials\\
- Components enclosed within the outer shell\\
- Structural elements that are only visible when the object is disassembled\\

When analyzing:\\
- Consider that the material might be completely hidden from external view\\
- Use the semantic usage and material type hints to infer properties of internal components\\
- Internal structural components often have different properties than visible surfaces\\
- For example, a soft exterior might hide a rigid internal frame\\

Critical Instruction: You MUST provide numerical estimates for ALL materials, even organic, biological, or unusual materials like leaves, feathers, or paper. \\
- For organic materials, estimate properties based on similar natural materials with known values\\
- For leaves, consider them as thin plant fiber composites with values similar to paper or dried plant fibers\\
- Never respond with "N/A" or any non-numeric value in your property estimates\\

When analyzing materials, use step-by-step reasoning:\\
1. First identify the likely material class and subtype based on visual appearance (if visible) or contextual clues (if internal)\\
2. Consider how texture, color, and reflectivity inform your understanding of the material (when visible)\\
3. Incorporate the provided physical properties and contextual usage information\\
4. For each mechanical property, reason through how the visual and physical attributes lead to your estimate\\
5. Consider how the material compares to reference materials with known properties\\
6. If the material appears to be internal/hidden, use the object type and usage context to make informed estimates\\

Important Formatting Requirements:\\
- Young's modulus must be provided in scientific notation followed by "Pa" (e.g., 2.0e11 Pa)\\
- Poisson's ratio must be a simple decimal between 0.0 and 0.5 with no units (e.g., 0.34)\\
- Density must be provided in kg/m\textasciicircum3 (e.g., 7800 kg/m\textasciicircum3)\\
- Each property must be on its own line with exactly the label shown in the examples\\
- Do not include explanatory text or parenthetical notes after the values\\
- ALWAYS provide numerical values, never text like "N/A" or "unknown"
\end{tcolorbox}
\caption{\textbf{System Prompt.} The System Prompt we use for every segment of every object.}
\label{fig:sysprompt}
\end{figure*}

\begin{figure*}[tb]
\centering
\begin{tcolorbox}[
    colback=blue!5!white,       % light blue/purple background
    colframe=blue!50!black,     % border color
    arc=3mm,                    % rounded corners
    boxrule=0.5pt,              % thickness of border
    width=\textwidth,           % span full page width
    left=3mm, right=3mm,
    top=2mm, bottom=2mm
]
\textbf{User Prompt.}
\medskip

You are a materials science expert analyzing two images:\\
1. A photo of the full object (showing how the material appears in context).\\
2. A sphere with the material's texture (showing color/roughness/reflectivity in isolation).\\

Using both images and the information below, identify the real-world material and estimate its mechanical properties.\\

Material context:\\
  \textasteriskcentered Material type: \colorbox{yellow}{fill in from dataset}\\
  \textasteriskcentered Opacity: \colorbox{yellow}{fill in from dataset}\\
  \textasteriskcentered Density: \colorbox{yellow}{fill in from dataset} kg/m\textasciicircum3\\
  \textasteriskcentered Dynamic friction: \colorbox{yellow}{fill in from dataset}\\
  \textasteriskcentered Static friction: \colorbox{yellow}{fill in from dataset}\\
  \textasteriskcentered Restitution: \colorbox{yellow}{fill in from dataset}\\
  \textasteriskcentered Usage: \colorbox{yellow}{fill in from dataset}\\

Your task is to provide three specific properties:\\
1. Young's modulus (in Pa using scientific notation)\\
2. Poisson's ratio (a value between 0.0 and 0.5)\\
3. Density (in kg/m\textasciicircum3 using scientific notation)\\

Additional reference material property ranges to help you make accurate estimations:\\
  - \colorbox{orange}{closest match from material database}: Young's modulus range \colorbox{orange}{fill in}, Poisson's ratio range \colorbox{orange}{fill in}, Density range \colorbox{orange}{fill in} kg/m\textasciicircum3\\
  - \colorbox{orange}{closest match from material database}: Young's modulus range \colorbox{orange}{fill in}, Poisson's ratio range \colorbox{orange}{fill in}, Density range \colorbox{orange}{fill in} kg/m\textasciicircum3\\
  - \colorbox{orange}{closest match from material database}: Young's modulus range \colorbox{orange}{fill in} GPa, Poisson's ratio range \colorbox{orange}{fill in}, Density range \colorbox{orange}{fill in} kg/m\textasciicircum3\\

Example 1:\\
Material: metal\\
Opacity: opaque\\
Density: 7800 kg/m\textasciicircum3\\
Dynamic friction: 0.3\\
Static friction: 0.4\\
Restitution: 0.3\\
Usage: structural component\\

Analysis:\\
Step 1: Based on the images, this appears to be a standard structural steel with a matte gray finish.\\
Step 2: The surface has medium roughness with some subtle texture visible in the reflection pattern.\\
Step 3: The physical properties (density, friction values, restitution) are consistent with carbon steel.\\
Step 4: Considering the usage and measured properties:\\
   - High stiffness (Young's modulus ~200 GPa) based on typical steel values\\
   - Medium Poisson's ratio typical of metals\\
   - High density matching the measured 7800 kg/m\textasciicircum3\\

Young's modulus: 2.0e11 Pa\\
Poisson's ratio: 0.29\\
Density: 7800 kg/m\textasciicircum3
\end{tcolorbox}
\caption{\textbf{User Prompt I.} The User Prompt we use for every segment of every object.}
\label{fig:userprompt1}
\end{figure*}

\begin{figure*}[tb]
\centering
\begin{tcolorbox}[
    colback=blue!5!white,       % light blue/purple background
    colframe=blue!50!black,     % border color
    arc=3mm,                    % rounded corners
    boxrule=0.5pt,              % thickness of border
    width=\textwidth,           % span full page width
    left=3mm, right=3mm,
    top=2mm, bottom=2mm
]
\textbf{User Prompt Continued.}
\medskip

Example 2:\\
Material: plastic\\
Opacity: opaque\\
Density: 950 kg/m\textasciicircum3\\
Dynamic friction: 0.25\\
Static friction: 0.35\\
Restitution: 0.6\\
Usage: household container\\

Analysis:\\
Step 1: The material shows the characteristic smooth, uniform appearance of a consumer plastic.\\
Step 2: It has moderate gloss with some translucency and a slight texture.\\
Step 3: The physical properties (medium-low density, moderate friction, higher restitution) match polypropylene.\\
Step 4: Based on these observations and measurements:\\
   - Medium-low stiffness typical of polyolefin plastics\\
   - Higher Poisson's ratio indicating good lateral deformation\\
   - Density matching the measured 950 kg/m\textasciicircum3\\

Young's modulus: 1.3e9 Pa\\
Poisson's ratio: 0.42\\
Density: 950 kg/m\textasciicircum3\\

Example 3:\\
Material: fabric\\
Opacity: opaque\\
Density: 300 kg/m\textasciicircum3\\
Dynamic friction: 0.55\\
Static friction: 0.75\\
Restitution: 0.2\\
Usage: furniture covering\\

Analysis:\\
Step 1: The material shows a woven textile structure with visible fibers.\\
Step 2: The surface has significant texture with a matte appearance and no specular highlights.\\
Step 3: The physical properties (low density, high friction, low restitution) match a woven textile.\\
Step 4: Based on these observations and measurements:\\
   - Low stiffness as expected for flexible textiles\\
   - Medium-high Poisson's ratio from the woven structure\\
   - Density matching the measured 300 kg/m\textasciicircum3\\

Young's modulus: 1.2e8 Pa\\
Poisson's ratio: 0.38\\
Density: 300 kg/m\textasciicircum3
\end{tcolorbox}
\caption{\textbf{User Prompt II.} The User Prompt we use for every segment of every object.}
\label{fig:userprompt2}
\end{figure*}

\begin{figure*}[tb]
\centering
\begin{tcolorbox}[
    colback=blue!5!white,       % light blue/purple background
    colframe=blue!50!black,     % border color
    arc=3mm,                    % rounded corners
    boxrule=0.5pt,              % thickness of border
    width=\textwidth,           % span full page width
    left=3mm, right=3mm,
    top=2mm, bottom=2mm
]
\textbf{User Prompt Continued.}
\medskip

Example 4:\\
Material: organic\\
Opacity: opaque\\
Density: 400 kg/m\textasciicircum3\\
Dynamic friction: 0.45\\
Static friction: 0.65\\
Restitution: 0.15\\
Usage: decorative element\\

Analysis:\\
Step 1: This is an organic material with the characteristic structure of natural fibers.\\
Step 2: The surface shows a natural pattern, matte finish, and relatively brittle structure.\\
Step 3: The physical properties (low density, moderate-high friction, low restitution) align with plant-based materials.\\
Step 4: Considering similar organic materials and the measured properties:\\
   - Low-medium stiffness in the fiber direction\\
   - Medium Poisson's ratio reflecting the fibrous structure\\
   - Density matching the measured 400 kg/m\textasciicircum3\\

Young's modulus: 2.5e9 Pa\\
Poisson's ratio: 0.30\\
Density: 400 kg/m\textasciicircum3\\

Based on the provided images and context information, analyze the material properties.
Note: The material segment might be internal to the object and not visible from the outside.\\

Respond using EXACTLY the following format (do not deviate from this structure):\\

Analysis: \\
Step 1: Identify the material class/type based on visual appearance\\
Step 2: Describe the surface characteristics (texture, reflectivity, color)\\
Step 3: Determine the specific material subtype considering its physical properties\\
Step 4: Reason through each property estimate based on visual and measured data\\

Young's modulus: <value in scientific notation> Pa\\
Poisson's ratio: <single decimal value between 0.0 and 0.5>\\
Density: <value in scientific notation> kg/m\textasciicircum3\\

Critical Instructions:\\
1. You MUST provide numerical estimates for ALL materials, including organic or unusual materials\\
2. For natural materials like leaves, wood, or paper, provide estimates based on similar materials with known properties\\
3. Never use "N/A", "unknown", or any non-numeric responses for the material properties\\
4. For Poisson's ratio, provide a simple decimal number (like 0.3 or 0.42)\\
5. Each property should be on its own line with exact formatting shown above
\end{tcolorbox}
\caption{\textbf{User Prompt III.} The User Prompt we use for every segment of every object.}
\label{fig:userprompt3}
\end{figure*}

We construct a tiny dataset consisting of complex objects that are manually annotated and compare these properties with the annotations from the VLM (Qwen2.5-VL 72B ) in~\Cref{tab:vlmmetrics}. We observe that the VLM, given significant additional information as we provide, performs close to human annotator performance. We report the commonly used metrics we list in~\Cref{app:metricsmech} for measuring differences in mechanical properties.

\ifdefined\iclr
    \begin{table*}[tb]
\centering
\caption{\textbf{VLM Annotation Errors.} Errors for the VLM annotation for mechanical property annotation.}
\resizebox{\textwidth}{!}{%
\begin{tabular}{rrrrrrrrrr}
\toprule
\rowcolor{nvidiagreen!15}$\log(E)$ ($\downarrow$) & $\nu$ ($\downarrow$) & $\rho$ ($\downarrow$) & $\log(E/\rho)$ ($\downarrow$) & $\log(G)$ ($\downarrow$) & $\log(K)$ ($\downarrow$) & L.S. ($\downarrow$) & E.A. ($\downarrow$) & Bray–Curtis ($\downarrow$) \\
\midrule
0.0295 & 0.0426 & 0.1348 & 0.1961 & 0.0303 & 0.0330 & 0.2022 & 0.2162 & 0.2342 \\
\bottomrule
\end{tabular}
}
\label{tab:vlmmetrics}
\end{table*}
\else
    \begin{table*}[tb]
\centering
\begin{tabular}{rrrrrrrrrr}
\toprule
\rowcolor{blue!15}$\log(E)$ ($\downarrow$) & $\nu$ ($\downarrow$) & $\rho$ ($\downarrow$) & $\log(E/\rho)$ ($\downarrow$) & $\log(G)$ ($\downarrow$) & $\log(K)$ ($\downarrow$) & L.S. ($\downarrow$) & E.A. ($\downarrow$) & Bray–Curtis ($\downarrow$) \\
\midrule
0.0295 & 0.0426 & 0.1348 & 0.1961 & 0.0303 & 0.0330 & 0.2022 & 0.2162 & 0.2342 \\
\bottomrule
\end{tabular}
\caption{\textbf{VLM Annotation Errors.} Errors for the VLM annotation for mechanical property annotation.}
\label{tab:vlmmetrics}
\end{table*}
\fi

\subsection{Dataset Statistics}
\label{app:datastats}

Our dataset comprises a diverse collection of 1,692 objects, sourced from four datasets: simready~\cite{nvidia_omniverse_simready}, residential~\cite{nvidia_omniverse_residential_assets}, vegetation~\cite{nvidia_omniverse_vegetation_assets}, and commercial~\cite{nvidia_omniverse_commercial_assets}. As shown in \Cref{tab:3ddata}, the majority of objects belong to the simready and residential categories, with vegetation and commercial objects providing additional variety. Each object is decomposed into multiple segments, with a total of 8,128 segments across the dataset. Most parts are labeled with English material names, and for few parts that do not have these material names, we infer these from the PBR texture names that were applied to these parts. To characterize the physical realism and diversity of the dataset, we analyze the distribution of key material properties for all segments in~\Cref{tab:material_stats}. The wide range of material properties highlights the heterogeneity of the dataset, which is essential for robust learning and evaluation of material-aware models. We summarize the most frequent material categories (e.g., metal, plastic, wood, cardboard) and object classes (e.g., residential, shelf, container), along with their respective counts and proportions in~\Cref{tab:material_classes}.

We report the statistics of our Material Triplet dataset in~\Cref{tab:matdata}. To train MatVAE, we use the "Filtered Dataset".

\ifdefined\iclr
    % Combined Dataset Split, Object, Segment, and Point Statistics
\begin{table*}[tb]
    \centering
    \caption{\textbf{Dataset Statistics.} Number of objects, total segments, total points, average segments per object (std. dev.), and average points per object (std. dev.) for each dataset.}
    \resizebox{\textwidth}{!}{%
    \begin{tabular}{lrrrrrrrr}
    \toprule
    \rowcolor{nvidiagreen!15}
    Dataset & Total Objects & Segments (\%) & Voxels (\%) & Avg. Segments/Object & Avg. Voxels/Object \\
    \midrule
    commercial & 82 & 650 (8.0) & 1,812,064 (4.9) & 7.93 {\textbf{\scriptsize\textcolor{gray}{($\pm$7.19)}}} & 22,098 {\textbf{\scriptsize\textcolor{gray}{($\pm$22,774)}}} \\
    residential & 449 & 4225 (52.2) & 9,109,380 (24.4) & 9.41 {\textbf{\scriptsize\textcolor{gray}{($\pm$21.82)}}} & 20,288 {\textbf{\scriptsize\textcolor{gray}{($\pm$21,714)}}} \\
    simready & 1029 & 2544 (31.5) & 24,148,660 (64.7) & 2.47 {\textbf{\scriptsize\textcolor{gray}{($\pm$1.33)}}} & 23,468 {\textbf{\scriptsize\textcolor{gray}{($\pm$25,032)}}} \\
    vegetation & 104 & 670 (8.3) & 2,267,848 (6.1) & 6.44 {\textbf{\scriptsize\textcolor{gray}{($\pm$4.53)}}} & 21,806 {\textbf{\scriptsize\textcolor{gray}{($\pm$19,428)}}} \\
    \midrule train & 1333 & 6477 (80.1) & 28,709,190 (76.9) & 4.86 {\textbf{\scriptsize\textcolor{gray}{($\pm$12.69)}}} & 21,537 {\textbf{\scriptsize\textcolor{gray}{($\pm$23,431)}}} \\
    validation & 165 & 552 (6.8) & 3,719,996 (10.0) & 3.35 {\textbf{\scriptsize\textcolor{gray}{($\pm$3.19)}}} & 22,545 {\textbf{\scriptsize\textcolor{gray}{($\pm$23,095)}}} \\
    test & 166 & 1060 (13.1) & 4,908,766 (13.1) & 6.39 {\textbf{\scriptsize\textcolor{gray}{($\pm$11.33)}}} & 29,571 {\textbf{\scriptsize\textcolor{gray}{($\pm$25,987)}}} \\
    \midrule
    \textbf{Total} & 1664 & 8089 (100.0) & 37,337,952 (100.0) & 4.86 {\textbf{\scriptsize\textcolor{gray}{($\pm$11.97)}}} & 22,439 {\textbf{\scriptsize\textcolor{gray}{($\pm$23,786)}}} \\
    \bottomrule
    \end{tabular}
    }
    \label{tab:3ddata}
\end{table*}

% Material Properties Statistics
\begin{table*}[tb]
    \centering
    \caption{\textbf{Material property statistics for all segments in the dataset.} We report the minimum, maximum, mean, median, standard deviation, and outlier count (\% of values outside $\pm3\sigma$) for Young's modulus, Poisson's ratio, and Density.}
    \resizebox{\textwidth}{!}{%
    \begin{tabular}{lrrrrrr}
    \toprule
    \rowcolor{nvidiagreen!15}
    Property & Min & Max & Mean & Median & Std Dev & Outliers (\%) \\
    \midrule
    Density ($\mathrm{kg}/\mathrm{m}^3$)         & $5.0\times10^1$ & $1.93\times10^4$ & $2.28\times10^3$ & $1.20\times10^3$ & $2.44\times10^3$ & 25 (0.3) \\
    Young's Modulus (Pa)       & $1.0\times10^5$ & $2.8\times10^{11}$ & $4.19\times10^{10}$ & $1.0\times10^{10}$ & $6.53\times10^{10}$ & 165 (2.0) \\
    Poisson's Ratio            & $1.6\times10^{-1}$ & $4.9\times10^{-1}$ & $3.36\times10^{-1}$ & $3.5\times10^{-1}$ & $4.36\times10^{-2}$ & 88 (1.1) \\
    \bottomrule
    \end{tabular}
    }
    \label{tab:material_stats}
\end{table*}

% Top High-Level Material Categories and Object Classes
\begin{table}[tb]
    \centering
    \caption{\textbf{Most frequent high-level material categories and object classes.} We report the top high-level material categories (aggregated and deduplicated) and the most common object classes in the dataset, with their respective counts and percentages.}
    \begin{tabular}{lr|lr}
    \toprule
    \rowcolor{nvidiagreen!15}
    Mat. Category & \multicolumn{1}{r}{Count (\%)} & Object Class & \multicolumn{1}{r}{Count (\%)} \\
    \midrule
    Metal & 1434 (17.7) & residential & 477 (28.2) \\
    Plastic & 553 (6.8) & shelf & 250 (14.8) \\
    Wood & 707 (8.7) & container & 193 (11.4) \\
    Cardboard & 315 (3.9) & cardboard box & 106 (6.3) \\
    Leather & 140 (1.7) & vegetation & 105 (6.2) \\
    Chrome & 171 (2.1) & crate & 101 (6.0) \\
    Glass & 85 (1.0) & commercial & 82 (4.8) \\
    Fabric & 60 (0.7) & pallet & 67 (4.0) \\
    Rubber & 55 (0.7) & barricade tape & 59 (3.5) \\
    Stone & 40 (0.5) & inclined plane & 22 (1.3) \\
    \bottomrule
    \end{tabular}
    \label{tab:material_classes}
\end{table}

\begin{table}[tb]
    \centering
    \caption{\textbf{Dataset Details.} We present statistics of our dataset for training MatVAE (\Cref{sec:matvae}).}
    \begin{tabular}{cr}
    \toprule
    \rowcolor{nvidiagreen!15}Dataset & Size \\
    \midrule Material Ranges & 249\\
    Extracted Materials & 105456\\
    Filtered Materials & 101517\\
    \bottomrule
    \end{tabular}
    \label{tab:matdata}
\end{table}
    \begin{minipage}{0.327\textwidth}
    \centering
    \includegraphics[width=\linewidth]{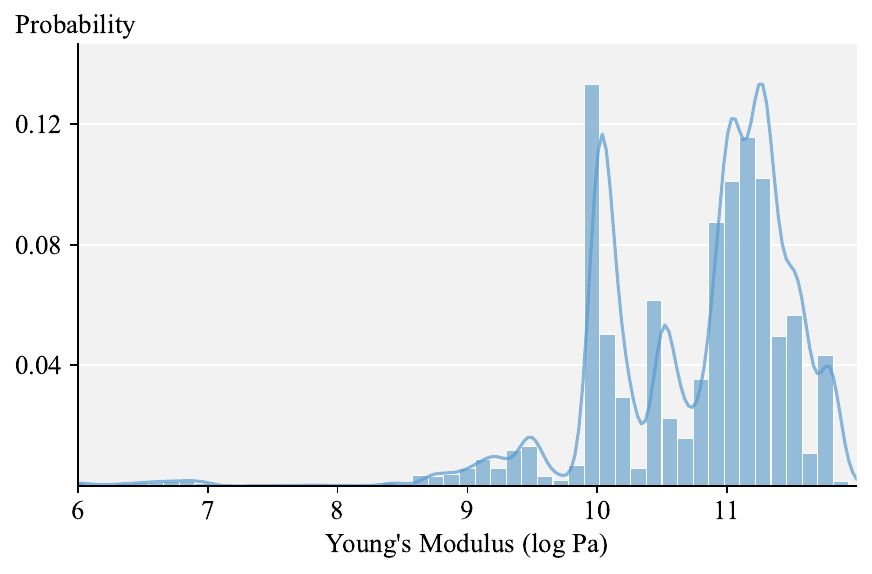}
    \captionof{figure}{\textbf{Young's Modulus Pa ($E$).} Histogram of Young's Modulus in our Geometry with Volumetric Materials Dataset (\Cref{sec:data}).}
    \label{fig:matvaehistym}
\end{minipage}%
\hfill
\begin{minipage}{0.327\textwidth}
    \centering
    \includegraphics[width=\linewidth]{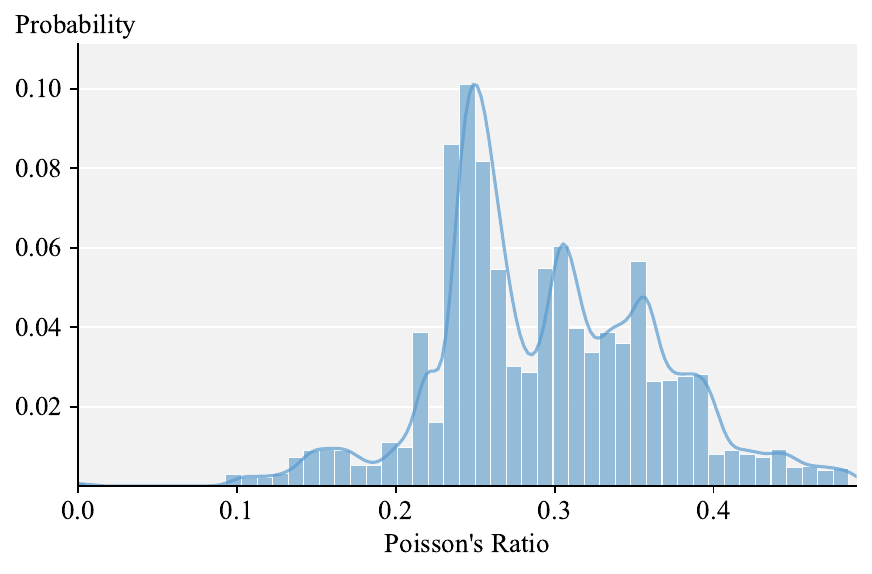}
    \captionof{figure}{\textbf{Poisson's Ratio ($\nu$).} Histogram of Poisson's Ratio  in our Geometry with Volumetric Materials Dataset (\Cref{sec:data}).}
    \label{fig:matvaehistpr}
\end{minipage}%
\hfill
\begin{minipage}{0.327\textwidth}
    \centering
    \includegraphics[width=\linewidth]{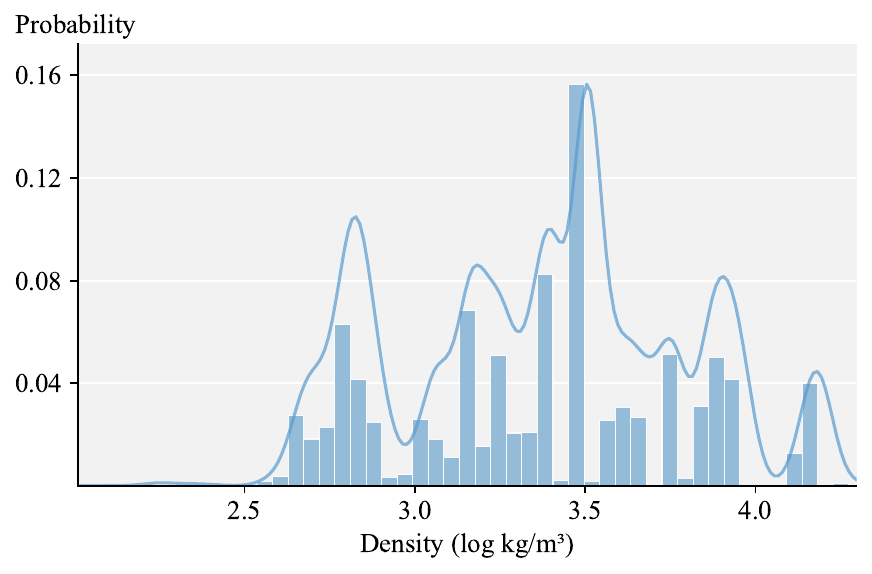}
    \captionof{figure}{\textbf{Density $\frac{kg}{m^3}$.} Histogram of Density  in our Geometry with Volumetric Materials Dataset (\Cref{sec:data}).}
    \label{fig:matvaehistrho}
\end{minipage}%
\else
    % Combined Dataset Split, Object, Segment, and Point Statistics
\begin{table*}[tb]
    \centering
    \begin{tabular}{lrrrrrrrr}
    \toprule
    \rowcolor{blue!15}
    Dataset & Total Objects & Segments (\%) & Voxels (\%) & Avg. Segments/Object & Avg. Voxels/Object \\
    \midrule
    commercial & 82 & 650 (8.0) & 1,812,064 (4.9) & 7.93 {\textbf{\scriptsize\textcolor{gray}{($\pm$7.19)}}} & 22,098 {\textbf{\scriptsize\textcolor{gray}{($\pm$22,774)}}} \\
    residential & 449 & 4225 (52.2) & 9,109,380 (24.4) & 9.41 {\textbf{\scriptsize\textcolor{gray}{($\pm$21.82)}}} & 20,288 {\textbf{\scriptsize\textcolor{gray}{($\pm$21,714)}}} \\
    simready & 1029 & 2544 (31.5) & 24,148,660 (64.7) & 2.47 {\textbf{\scriptsize\textcolor{gray}{($\pm$1.33)}}} & 23,468 {\textbf{\scriptsize\textcolor{gray}{($\pm$25,032)}}} \\
    vegetation & 104 & 670 (8.3) & 2,267,848 (6.1) & 6.44 {\textbf{\scriptsize\textcolor{gray}{($\pm$4.53)}}} & 21,806 {\textbf{\scriptsize\textcolor{gray}{($\pm$19,428)}}} \\
    \midrule train & 1333 & 6477 (80.1) & 28,709,190 (76.9) & 4.86 {\textbf{\scriptsize\textcolor{gray}{($\pm$12.69)}}} & 21,537 {\textbf{\scriptsize\textcolor{gray}{($\pm$23,431)}}} \\
    validation & 165 & 552 (6.8) & 3,719,996 (10.0) & 3.35 {\textbf{\scriptsize\textcolor{gray}{($\pm$3.19)}}} & 22,545 {\textbf{\scriptsize\textcolor{gray}{($\pm$23,095)}}} \\
    test & 166 & 1060 (13.1) & 4,908,766 (13.1) & 6.39 {\textbf{\scriptsize\textcolor{gray}{($\pm$11.33)}}} & 29,571 {\textbf{\scriptsize\textcolor{gray}{($\pm$25,987)}}} \\
    \midrule
    \textbf{Total} & 1664 & 8089 (100.0) & 37,337,952 (100.0) & 4.86 {\textbf{\scriptsize\textcolor{gray}{($\pm$11.97)}}} & 22,439 {\textbf{\scriptsize\textcolor{gray}{($\pm$23,786)}}} \\
    \bottomrule
    \end{tabular}
    \caption{\textbf{Dataset Statistics.} Number of objects, total segments, total points, average segments per object (std. dev.), and average points per object (std. dev.) for each dataset.}
    \label{tab:3ddata}
\end{table*}

% Material Properties Statistics
\begin{table*}[tb]
    \centering
    \begin{tabular}{lrrrrrr}
    \toprule
    \rowcolor{blue!15}
    Property & Min & Max & Mean & Median & Std Dev & Outliers (\%) \\
    \midrule
    Density ($\mathrm{kg}/\mathrm{m}^3$)         & $5.0\times10^1$ & $1.93\times10^4$ & $2.28\times10^3$ & $1.20\times10^3$ & $2.44\times10^3$ & 25 (0.3) \\
    Young's Modulus (Pa)       & $1.0\times10^5$ & $2.8\times10^{11}$ & $4.19\times10^{10}$ & $1.0\times10^{10}$ & $6.53\times10^{10}$ & 165 (2.0) \\
    Poisson's Ratio            & $1.6\times10^{-1}$ & $4.9\times10^{-1}$ & $3.36\times10^{-1}$ & $3.5\times10^{-1}$ & $4.36\times10^{-2}$ & 88 (1.1) \\
    \bottomrule
    \end{tabular}
    \caption{\textbf{Material property statistics for all segments in the dataset.} We report the minimum, maximum, mean, median, standard deviation, and outlier count (\% of values outside $\pm3\sigma$) for Young's modulus, Poisson's ratio, and Density.}
    \label{tab:material_stats}
\end{table*}

% Top High-Level Material Categories and Object Classes
\begin{table}[tb]
    \centering
    \begin{tabular}{lr|lr}
    \toprule
    \rowcolor{blue!15}
    Mat. Category & \multicolumn{1}{r}{Count (\%)} & Object Class & \multicolumn{1}{r}{Count (\%)} \\
    \midrule
    Metal & 1434 (17.7) & residential & 477 (28.2) \\
    Plastic & 553 (6.8) & shelf & 250 (14.8) \\
    Wood & 707 (8.7) & container & 193 (11.4) \\
    Cardboard & 315 (3.9) & cardboard box & 106 (6.3) \\
    Leather & 140 (1.7) & vegetation & 105 (6.2) \\
    Chrome & 171 (2.1) & crate & 101 (6.0) \\
    Glass & 85 (1.0) & commercial & 82 (4.8) \\
    Fabric & 60 (0.7) & pallet & 67 (4.0) \\
    Rubber & 55 (0.7) & barricade tape & 59 (3.5) \\
    Stone & 40 (0.5) & inclined plane & 22 (1.3) \\
    \bottomrule
    \end{tabular}
    \caption{\textbf{Most frequent high-level material categories and object classes.} We report the top high-level material categories (aggregated and deduplicated) and the most common object classes in the dataset, with their respective counts and percentages.}
    \label{tab:material_classes}
\end{table}

\begin{table}[tb]
    \centering
    \begin{tabular}{cr}
    \toprule
    \rowcolor{blue!15}Dataset & Size \\
    \midrule Material Ranges & 249\\
    Extracted Materials & 105456\\
    Filtered Materials & 101517\\
    \bottomrule
    \end{tabular}
    \caption{\textbf{Dataset Details.} We present statistics of our dataset for training MatVAE (\Cref{sec:matvae}).}
    \label{tab:matdata}
\end{table}
    \begin{figure}[tb]
    \centering
    \includegraphics[width=\linewidth]{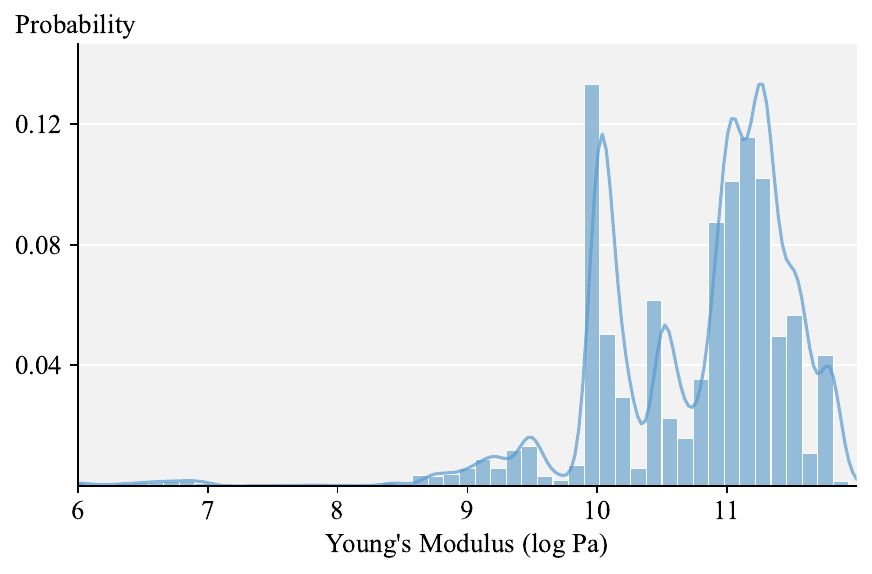}
    \caption{\textbf{Young's Modulus Pa ($E$).} Histogram of Young's Modulus in our Geometry with Volumetric Materials Dataset (\Cref{sec:data}).}
    \label{fig:matvaehistym}
\end{figure}
\begin{figure}[tb]
    \centering
    \includegraphics[width=\linewidth]{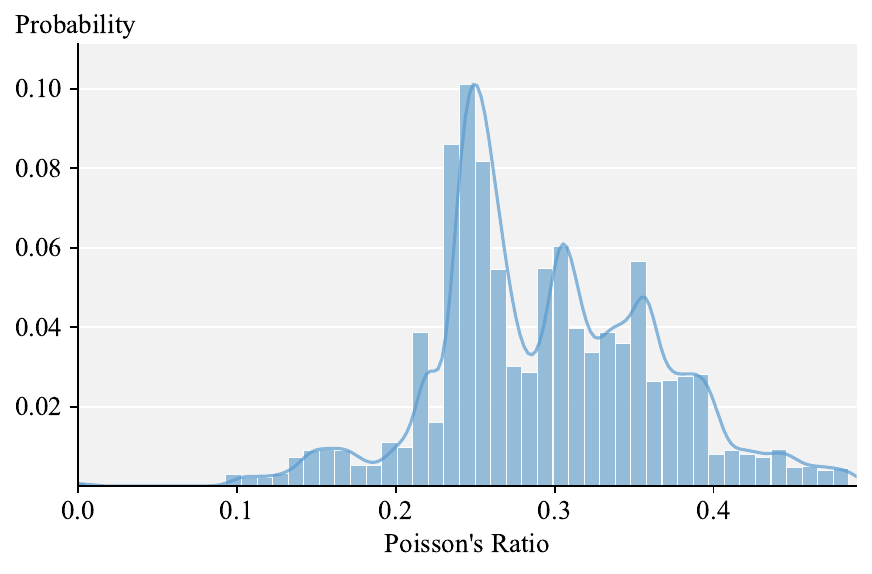}
    \caption{\textbf{Poisson's Ratio ($\nu$).} Histogram of Poisson's Ratio  in our Geometry with Volumetric Materials Dataset (\Cref{sec:data}).}
    \label{fig:matvaehistpr}
\end{figure}
\begin{figure}[tb]
    \centering
    \includegraphics[width=\linewidth]{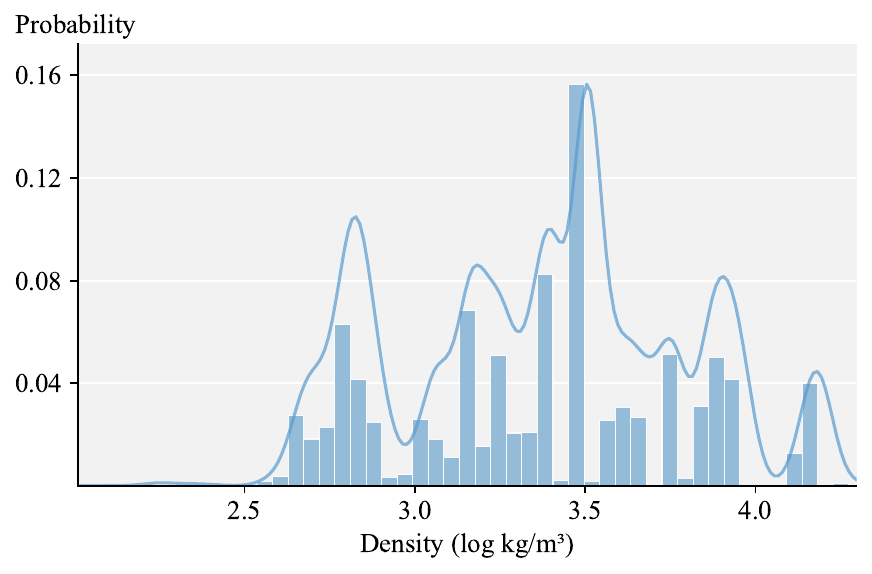}
    \caption{\textbf{Density $\frac{kg}{m^3}$.} Histogram of Density  in our Geometry with Volumetric Materials Dataset (\Cref{sec:data}).}
    \label{fig:matvaehistrho}
\end{figure}
\fi
\section{Additional Implementation Details}
\label{app:implementation}

We present additional implementation details.

\subsection{Design of MatVAE}
\label{app:designmatvae}

We explain the motivation behind the design of MatVAE.

\paragraph{Normalizing Flow.}
Material triplets remain statistically non‑Gaussian even after normalization (heavy‑tailed/multi‑modal $\log_{10}E,\log_{10}\rho$; boundary‑concentrated $\nu\in[0,0.5)$), so a diagonal‑Gaussian $q_\phi(z\mid m)$ tends to mode‑average and miscalibrate tails. We therefore parameterize the posterior with a bijective normalizing flow~\cite{pmlr-v37-rezende15} $f_\psi$ where $\psi$ is the parameter of the flow network f, and $\Psi$ is the space of all parameters $\psi$. This $f_\psi$ applied to a Gaussian base $q_0$: sample $u\sim q_0(u\mid m)=\mathcal{N}(u;\mu_\phi(m),\mathrm{diag}\,\sigma_\phi^2(m))$, set $z=f_\psi(u)$, and compute the density by change of variables
\begin{equation}
\begin{aligned}
\log q_\phi(z\mid m) 
&= \underbrace{\log q_0\bigl(f_\psi^{-1}(z)\mid m\bigr)}_{\text{base density}} 
  + \underbrace{\log\bigl|\det J_{f_\psi^{-1}}(z)\bigr|}_{\text{log-Jacobian}} \\
&= \underbrace{\log q_0(u\mid m)}_{\text{base density}} 
  - \underbrace{\log\bigl|\det J_{f_\psi}(u)\bigr|}_{\text{log-Jacobian}} \Big|_{u=f_\psi^{-1}(z)}.
\end{aligned}
\end{equation}
with a standard normal prior $p(z)=\mathcal{N}(0, I)$ and decoder likelihood $p_\theta(m \mid z)$.

This keeps the ELBO form unchanged while strictly enlarging the variational family (the identity map recovers the Gaussian case), allowing $q_\phi$ to match the true posterior and avoid mode‑averaging on $(E,\nu,\rho)$. We instantiate $f_\psi$ as a single radial flow, 
\begin{equation}
\begin{split}
        f_\psi(u)&=u+\underbrace{\beta\,h(u)}_{\text{radial scale}}\overbrace{(u-z_0)}^{\text{displacement}},\\h(u)&=\frac{1}{\bigl(\alpha+\lVert u-z_0\rVert_2\bigr)},
\end{split}
\end{equation}

whose log-determinant has the closed form (substitute $h'=-h^2$ in the closed form from~\cite{pmlr-v37-rezende15}),
\begin{equation}
\begin{split}
    \log\det J_{f_\psi}(u)
  &= \underbrace{(D-1)\,\log\bigl(1+\beta h(u)\bigr)}_{\text{angular}}\\
    &\quad+ \underbrace{\log\bigl(1+\beta h(u)-\beta\,h(u)^2\,r(u)\bigr)}_{\text{radial}},\\
    r(u)&=\lVert u-z_0\rVert_2,
\end{split}
\label{eq:logdetJ}
\end{equation}

where $D$ is the dimensionality of the latent space and $z_0$ is a \emph{trainable} $D$-dimensional vector (one per flow layer) representing the centre of the deformation.

Radial flows are invertible iff $\alpha>0$ and $\beta>-\alpha$, we satisfy these by
\(
\alpha=\mathrm{softplus}(\tilde\alpha),\;
\beta=-\alpha+\mathrm{softplus}(\tilde\beta)
\)~\cite{pmlr-v37-rezende15} with unconstrained trainable parameters $\tilde\alpha,\tilde\beta\in\mathbb{R}$.  We show our implementation in~\Cref{alg:vae-radial-flow}.

\paragraph{Penalizing TC.} We observed high dependence between latent coordinates in the aggregated posterior $\bar{q}_\phi(z)$ (both dimensions tended to encode $\rho$). Thus, we decompose the KL-divergence term of the ELBO following~\cite{chen2018isolating}. For MatVAE, this allows us to directly penalize the total correlation $\text{TC}(z) = \text{KL}(\bar{q}_\phi(z) || \prod_j \bar{q}_\phi(z_j))$ where $\bar{q}_\phi(z)$ is the aggregated posterior, $z_j$ is the $j \in \{1,2\}$-th coordinate of the latent vector $z$. This allowed us to reduce the high dependence between latent coordinates, causing both dimensions to encode density. During training, we follow~\cite{chen2018isolating} and approximate the aggregated posterior $\bar{q}_\phi(z) = \mathbb{E}_{m \sim p_{data}}[q_\phi(z\mid m)]$ using samples from a mini-batch where $p_{data}$ is the empirical data distribution.

% this is pseudocode for frankenstein/models/material_vae/beta_tc.py
\begin{algorithm}[tb]
\caption{MatVAE posterior update with a radial normalizing flow.}
\label{alg:vae-radial-flow}
\textbf{Require:} \\
\hspace*{\algorithmicindent} Batch $x \in \mathbb{R}^{B \times 3}$\\
\hspace*{\algorithmicindent} Encoder outputs $\mu(x)\in\mathbb{R}^{B\times D}$\\
\hspace*{\algorithmicindent} Encoder outputs $\log\sigma^2(x)\in\mathbb{R}^{B\times D}$\\
\hspace*{\algorithmicindent} Flow param $z_0\in\mathbb{R}^{1\times D}$\\
\hspace*{\algorithmicindent} Flow param $\log\alpha\in\mathbb{R}$\\
\hspace*{\algorithmicindent} Flow param $\beta_{\text{raw}}\in\mathbb{R}$\\
\hspace*{\algorithmicindent} Prior $p(z)=\mathcal{N}(0,I)$\\
\textbf{Ensure:} \\
\hspace*{\algorithmicindent} Flowed latent $z\in\mathbb{R}^{B\times D}$\\
\hspace*{\algorithmicindent} Post-flow log-density $\log q(z\mid x)\in\mathbb{R}^{B}$\\
\hspace*{\algorithmicindent} KL term denoted as $\mathrm{KL}$\\
\hspace*{\algorithmicindent} Reconstruction loss $\mathcal{L}_{\text{recon}}$\\
\begin{algorithmic}[1]
\State $\triangleright$\quad{\color{nvidiagreen}Encode and reparameterize}
\State $\mu \gets \mathrm{Encoder}_\mu(x)$
\State $\log\sigma^2 \gets \mathrm{Encoder}_{\log\sigma^2}(x)$
\State $\varepsilon \sim \mathcal{N}(0,I)$
\State $z_{\mathrm{base}} \gets \mu + \exp(\tfrac{1}{2}\log\sigma^2)\odot \varepsilon$
\State

\State $\triangleright$\quad{\color{nvidiagreen}Base posterior log-density}
\State $\log q_0 \gets \sum_{d=1}^D \log \mathcal{N}\big(z_{\mathrm{base},d};\,\mu_d,\exp(\log\sigma^2_d)\big)$
\State

\State $\triangleright$\quad{\color{nvidiagreen}Radial flow parameters ($\alpha>0$, $\beta>-\alpha$)}
\State $\alpha \gets \mathrm{softplus}(\log\alpha) + \varepsilon_\alpha$
\State $\beta \gets -\alpha + \mathrm{softplus}(\beta_{\text{raw}})$
\State

\State $\triangleright$\quad{\color{nvidiagreen}Radial flow transform}
\State $\text{diff} \gets z_{\mathrm{base}} - z_0$
\State $r \gets \|\text{diff}\|_2 + \varepsilon_r$
\State $h \gets \frac{1}{(\alpha + r)}$
\State $z \gets z_{\mathrm{base}} + \beta \times h \times \text{diff}$
\State

\State $\triangleright$\quad{\color{nvidiagreen}Log-determinant of Jacobian}
\State $bh \gets \beta \times h$
\State $bh_{\text{stab}} \gets \mathrm{clamp}(bh,\,-c,\,c)$
\State $\text{term}_1 \gets (D-1)\cdot \log(1 + bh_{\text{stab}})$
\State $\text{term}_2 \gets \log(1 + bh_{\text{stab}} - \beta\,h^2\,r)$
\State $\Delta\log|J| \gets \text{term}_1 + \text{term}_2$
\State

\State $\triangleright$\quad{\color{nvidiagreen}Change of variables and KL pieces}
\State $\log q \gets \log q_0 - \Delta\log|J|$
\State $\log p_z \gets \sum_{d=1}^D \log \mathcal{N}(z_d; 0, 1)$
\State

\State $\triangleright$\quad{\color{nvidiagreen}Losses (non-relevant details shown as $\ldots$)}
\State $(\hat{E},\hat{\nu},\hat{\rho},\ldots) \gets \mathrm{Decoder}(z)$
\State $\mathcal{L}_{\text{recon}} \gets \text{MSE/NLL in transformed space }(\ldots)$
\State $\mathrm{KL} \gets \log q - \log p_z$
\State $\mathcal{L} \gets \mathcal{L}_{\text{recon}} + \ldots$

\State \Return $(z,\ \log q,\ \mathrm{KL},\ \mathcal{L}_{\text{recon}})$
\end{algorithmic}
\end{algorithm}

\paragraph{Preventing Posterior Collapse.}
For a fixed $m$, define per-dimension marginals $q_\phi(z_j\mid m)$ and
\begin{equation}
\mathrm{KL}_j(z\mid m)=\mathrm{KL}\left(q_\phi(z_j\mid m)\,\|\,p(z_j)\right).
\end{equation}

Then, for each $m$,
\begin{equation}
\begin{split}
\mathrm{KL}\left(q_\phi(z\mid m)\,\|\,p(z)\right)
= &\sum_j \mathrm{KL}_j(m) \\
\quad+ &\underbrace{\mathrm{TC}\left(q_\phi(z\mid m)\right)}_{\mathrm{KL}\left(q_\phi(z\mid m)\,\|\,\prod_j q_\phi(z_j\mid m)\right)}.
\end{split}
\end{equation}
So the total KL contains a per-dimension rate term plus a per-sample total correlation. In the Gaussian (no-flow) case, $\mathrm{KL}_j(z \mid m)=\tfrac12\left(\mu_j(m)^2+\sigma_j(m)^2-\log\sigma_j(m)^2-1\right)$, whose gradients drive $\mu_j\to0$, $\sigma_j\to1$ under posterior collapse. We therefore impose a capacity constraint (“free nats”) on the dim-wise term to prevent collapse:
\begin{equation}
\sum_{j=1}^d \underbrace{\max\bigl(\delta,\;\mathbb{E}_{p_{\mathrm{data}}}\mathrm{KL}_j(z)\bigr)}_{\text{free-nats}},
\end{equation}

which enforces a minimum information budget $\delta=0.1$ per coordinate (zero subgradient below $\delta$). This allows us to fix the empirically observed imbalance where one latent carried most information and the other collapsed. An aggregated alternative consistent with the KL decomposition is $\max\bigl(\phi\cdot d,\ \sum_j \mathrm{KL}(q_\phi(z_j)\,\|\,p(z_j))\bigr)$.

\subsection{Network Design}
\label{sec:networkdesign}

We now present our network architecture.

\paragraph{MatVAE.} The \emph{encoder} architecture begins by projecting the 3-dimensional material triplet through a linear transformation into a 256-dimensional hidden space, followed by SiLU activation. The resulting representation then passes through three "ResidualBlocks", each using a bottleneck design that compresses the 256-dimensional vector to 128 dimensions via LayerNorm and SiLU activation, applies another linear transformation, and restores the original dimensionality through a second LayerNorm-SiLU sequence. Each "ResidualBlock" maintains a skip connection that adds the input directly to the final output. The encoder finally has separate linear heads that project the processed representation into the latent space parameters: one head predicts the posterior mean $\mu_{\phi}(m)$ and another predicts the log-variance $\log\sigma^2_{\phi}(m)$ for the 2-dimensional latent code $z$.

The \emph{decoder} mirrors this architecture in reverse, beginning with a linear projection from the 2-dimensional latent space back to the 256-dimensional hidden representation, followed by SiLU activation. The latent encoding then goes through three "ResidualBlocks" with an identical bottleneck structure and skip connections as the encoder. Finally, three separate linear heads decode the processed representation into the reconstructed material properties: Young's modulus, Poisson's ratio, and density, each predicted as scalar values in the normalized space.

\paragraph{Geometry Transformer.} Our model is based on TRELLIS~\cite{xiang2025structured3dlatentsscalable}. We use a transformer-based architecture specifically designed for processing sparse voxel representations with associated material properties. The model operates on a 64$^3$ resolution voxel grid, accepting 1024-dimensional DINOv2 visual features as input and compressing them to a compact 2-dimensional latent representation through a 12-layer transformer backbone. Each transformer block utilizes 12 attention heads with a 4:1 MLP expansion ratio, using Swin attention with 8$\times$8 local windows. During training, the Geometry Transformer operates in conjunction with a frozen MatVAE that decodes the latent into material properties. 

\subsection{Training}
\label{app:training}

We present our voxelization scheme for training on meshes in~\Cref{alg:voxelizemesh,alg:voxelization}. We present the hyperparameters used for training MatVAE and Geometry Transformer in~\Cref{tab:hyperparam}.

\begin{algorithm}[tb]
\caption{Segment-aware volumetric voxelization for meshes.}
\label{alg:voxelizemesh}
\textbf{Require:} \\
\hspace*{\algorithmicindent} Full-mesh vertices $V_{\text{all}}\in\mathbb{R}^{N\times 3}$, faces $F_{\text{all}}$ \\
\hspace*{\algorithmicindent} Segments $\mathcal{S}=\{(V_i\in\mathbb{R}^{N_i\times 3},\,F_i,\,\mathrm{sid}_i)\}_{i=1}^M$ \\
\hspace*{\algorithmicindent} Grid resolution $r\in\mathbb{N}$ (voxel pitch $h=1/r$) \\
\hspace*{\algorithmicindent} Per-segment cap $K_{\text{seg}}\in\mathbb{N}$\\
\hspace*{\algorithmicindent} Global cap $K_{\text{all}}\in\mathbb{N}$ \\
\textbf{Ensure:} \\
\hspace*{\algorithmicindent} Combined voxel centers $C_{\text{all}}\in\mathbb{R}^{L\times 3}$ within $[-0.5,0.5]^3$ \\
\hspace*{\algorithmicindent} Segment identifiers $\mathrm{sid}_{\text{all}}\in\{\text{str}\}^L$ \\
\hspace*{\algorithmicindent} Discretized centers $\hat{C}_{\text{all}}\in\mathbb{R}^{L\times 3}$ on an $r^3$ grid \\
\begin{algorithmic}[1]
\State $\triangleright$\quad{\color{nvidiagreen}Global normalization from the full mesh}
\State $v_{\min}\gets \min(V_{\text{all}})$
\State $v_{\max}\gets \max(V_{\text{all}})$
\State $c \gets (v_{\min}+v_{\max})/2$
\State $s \gets \max(v_{\max}-v_{\min})$
\State $\varepsilon \gets 10^{-6}$
\State
\State $C_{\text{acc}}\gets[\,]$
\State $\mathrm{sid}_{\text{acc}}\gets[\,]$
\For{$i=1$ to $M$}
  \State $\triangleright$\quad{\color{nvidiagreen}Normalize segment to $[-0.5,0.5]^3$ and ensure triangles}
  \State $V'_i \gets \mathrm{clip}((V_i - c)/s,\ -0.5+\varepsilon,\ 0.5-\varepsilon)$
  \State $F'_i \gets \mathrm{triangulate}(F_i)$
  \State
  \State $\triangleright$\quad{\color{nvidiagreen}Voxelize segment and solid-fill (\Cref{alg:voxelization})}
  \State $(C_i,\ Y_i) \gets \Call{VoxelizeSolid}{V'_i,\ F'_i,\ r}$
  \If{$K_{\text{seg}}$ is given and $|C_i|>K_{\text{seg}}$}
    \State $I \gets \mathrm{choice}(|C_i|,\ K_{\text{seg}},\ \text{without replacement})$
    \State $C_i \gets C_i[I]$
  \EndIf
  \If{$|C_i|=0$} \State \textbf{continue} \EndIf
  \State $C_{\text{acc}}.\mathrm{append}(C_i)$
  \State $\mathrm{sid}_{\text{acc}}.\mathrm{append}([\mathrm{sid}_i]^{|C_i|})$
\EndFor
\State
\If{$|C_{\text{acc}}|=0$} \State \Return $(\varnothing,\ \varnothing,\ \varnothing)$ \EndIf
\State $C_{\text{all}} \gets \mathrm{concat}(C_{\text{acc}})$
\State $\mathrm{sid}_{\text{all}} \gets \mathrm{concat}(\mathrm{sid}_{\text{acc}})$
\State
\State $\triangleright$\quad{\color{nvidiagreen}Optional global subsampling}
\If{$K_{\text{all}}$ is given and $|C_{\text{all}}|>K_{\text{all}}$}
  \State $I \gets \mathrm{choice}(|C_{\text{all}}|,\ K_{\text{all}},\ \text{without replacement})$
  \State $C_{\text{all}} \gets C_{\text{all}}[I]$
  \State $\mathrm{sid}_{\text{all}} \gets \mathrm{sid}_{\text{all}}[I]$
\EndIf
\State
\State $\triangleright$\quad{\color{nvidiagreen}Discretize to an $r^3$ grid aligned with $[-0.5,0.5]^3$}
\State $J \gets \mathrm{clip}(\lfloor (C_{\text{all}}+0.5)\cdot r \rfloor,\ 0,\ r-1)$
\State $\hat{C}_{\text{all}} \gets J/r - 0.5$
\State \Return $(C_{\text{all}},\ \mathrm{sid}_{\text{all}},\ \hat{C}_{\text{all}})$
\end{algorithmic}
\end{algorithm}

\begin{algorithm}[tb]
\caption{Voxelization and flood fill primitives for meshes.}
\label{alg:voxelization}
\begin{algorithmic}[1]
\Procedure{VoxelizeSolid}{$V,\ F,\ r$}
  \State $\triangleright$\quad{\color{nvidiagreen}Grid setup over a padded mesh AABB}
  \State $h \gets 1/r$
  \State $a_{\min}\gets \min(V)$; $a_{\max}\gets \max(V)$
  \State $b_{\min}\gets a_{\min}-h$
  \State $b_{\max}\gets a_{\max}+h$
  \State $n_x\gets \lceil (b_{\max,x}-b_{\min,x})/h \rceil$
  \State $n_y\gets \lceil (b_{\max,y}-b_{\min,y})/h \rceil$
  \State $n_z\gets \lceil (b_{\max,z}-b_{\min,z})/h \rceil$
  \State $S[n_x,n_y,n_z]\gets \mathrm{false}$
  \State $X[n_x,n_y,n_z]\gets \mathrm{false}$
  \State
  \State $\triangleright$\quad{\color{nvidiagreen}Triangle rasterization: mark surface cells}
  \For{each triangle $t=(v_0,v_1,v_2)\in F$}
    \State Compute triangle AABB $[t_{\min}, t_{\max}]$ in world coordinates
    \State Convert to grid index ranges $(i_{\min}:i_{\max},\ j_{\min}:j_{\max},\ k_{\min}:k_{\max})$
    \For{$i=i_{\min}$ to $i_{\max}$}
      \For{$j=j_{\min}$ to $j_{\max}$}
        \For{$k=k_{\min}$ to $k_{\max}$}
          \State Cell box $B = [b_{\min}+(i,j,k)h,\ b_{\min}+(i+1,j+1,k+1)h]$
          \If{$\Call{TriangleBoxIntersect}{t,\ B}$}
            \State $S[i,j,k]\gets \mathrm{true}$
          \EndIf
        \EndFor
      \EndFor
    \EndFor
  \EndFor
  \State
  \State $\triangleright$\quad{\color{nvidiagreen}Exterior marking by flood fill on non-surface cells}
  \State Initialize queue $Q$ with boundary indices $(i,j,k)$ where $S[i,j,k]=\mathrm{false}$
  \While{$Q$ not empty}
    \State $u\gets Q.\mathrm{pop}()$
    \If{$X[u]$} \State \textbf{continue} \EndIf
    \State $X[u]\gets \mathrm{true}$
    \For{each 6-neighbor $v$ of $u$ within bounds}
      \If{$S[v]=\mathrm{false}$ and $X[v]=\mathrm{false}$}
        \State $Q.\mathrm{push}(v)$
      \EndIf
    \EndFor
  \EndWhile
  \State
  \State $\triangleright$\quad{\color{nvidiagreen}Solid fill (interior) and center extraction}
  \State $Y \gets \neg X \land \neg S$
  \State $C \gets [\,]$
  \For{all indices $(i,j,k)$ where $Y[i,j,k]=\mathrm{true}$}
    \State $c \gets b_{\min} + (i+0.5,\ j+0.5,\ k+0.5)\cdot h$
    \State $C.\mathrm{append}(c)$
  \EndFor
  \State \textbf{return} $(C,\ Y)$
\EndProcedure
\end{algorithmic}
\end{algorithm}

\begin{table*}[tb]
\centering
\caption{\textbf{Training Hyperparameters.} We show the hyperparameters for the MatVAE and Geometry Transformer.}
\begin{tabular}{lr}
\toprule
\rowcolor{nvidiagreen!15}\multicolumn{2}{c}{MatVAE}\\
\midrule
Training Precision & FP-32 \\
Hidden Width & 256 \\
Network Depth & 3 ($\times 2$) \\
Latent Dimensions & 2 \\
Dropout Rate & 0.05 \\
Epochs & 850 \\
Batch Size & 256 \\
Optimizer & AdamW \\
Learning Rate & $10^{-4}$ \\
Weight Decay & $10^{-4}$ \\
LR Scheduler & Cosine Annealing \\
Final Learning Rate & $10^{-5}$ \\
Gradient Clipping & 5.0 \\
\multirow{3}{*}{$\beta$-TC Loss Weights} & $\alpha = 1.0$ (KL) \\
& $\beta = 2.0$ (TC) \\
& $\gamma = 1.0$ (MI) \\
Free Nats & 0.1 \\
KL Annealing Epochs & 200 \\
Data Normalization & Log Min-Max \\
\bottomrule
\end{tabular}
\quad
\begin{tabular}{lr}
\toprule
\rowcolor{nvidiagreen!15}\multicolumn{2}{c}{Geometry Transformer}\\
\midrule
Training Precision & FP-16 \\
Voxel Grid Resolution & 64³ \\
Input Channels & 1024 \\
Model Channels & 768 \\
Latent Channels & 2 \\
Transformer Blocks & 12 \\
Attention Heads & 12 \\
MLP Ratio & 4 \\
Attention Mode & Swin \\
Window Size & 8 \\
Max Training Steps & 200,000 \\
Batch Size per GPU & 4 \\
Total Batch Size & 16 \\
Optimizer & AdamW \\
Learning Rate & $10^{-4}$ \\
Weight Decay & $5 \times10^{-2}$ \\
Gradient Clipping & 1.0 \\
Loss Function & $\ell_2$ \\
EMA Rate & 0.9999 \\
\bottomrule
\end{tabular}
\label{tab:hyperparam}
\end{table*}

\paragraph{Voxelization For Training.} Our training dataset contains Universal Scene Description (USD) files with multi-segment meshes.
% \dave{the files contain the assets rather than the other way around}
Each mesh is normalized to the range $[-0.5, 0.5]$ using a global bounding box computed across all segments to preserve relative spatial relationships. We use volumetric voxelization using a regular 3D grid with a resolution of $\frac{1}{64}$, where each voxel center is tested for interior containment within the mesh volume through point-in-polyhedron testing, followed by volumetric filling to generate solid voxel representations rather than surface-only discretizations. All the voxels inside a given segment receive the material properties of the segment they lie in. 

\paragraph{Rendering for Training.} For multi-view image rendering of meshes, we use a path-tracing renderer to produce photorealistic renderings of 3D objects. Camera viewpoints are sampled using a quasi-random Hammersley sequence distributed uniformly across a sphere. For training and testing, we render 150 views, though our method can work by rendering as many views as needed, with cameras positioned at a fixed radius of 2 units from the object center and configured with a 40-degree field of view. Images are rendered at 512$\times$512 pixel resolution. The rendering pipeline outputs both the RGB images and the corresponding camera extrinsics and intrinsics. For rendering splats, we simply replace the renderer with the 3D Gaussian Splat renderer~\cite{kerbl3Dgaussians} in our workflow. For rendering SDFs, we render meshes using many points collected from the SDF. For rendering NeRFs~\cite{mildenhall2020nerf}, we simply replace the renderer with \texttt{nerfstudio}~\cite{nerfstudio} in our workflow.

\paragraph{Feature Aggregation.} For visual feature extraction, we employ DINOv2-ViT-L/14~\cite{oquab2024dinov2learningrobustvisual} with registers. We use a patch size of 14$\times$14 pixels and process input images resized to 518$\times$518 pixels, resulting in a 37$\times$37 patch. We use the \texttt{nv-dinov2}\footnote{\url{https://build.nvidia.com/nvidia/nv-dinov2}}~\cite{nvidia_nv_dinov2_2025} implementation.

\subsection{Simulation and Rendering}

For our mesh simulations (\Cref{fig:simulations}), we simulate with the finite-element method (FEM) using the \texttt{libuipc}~\cite{10.1145/3735126,gipc2024} implementation, and we render the simulations in a path-tracing-based renderer. While comparing with other simulators (\Cref{fig:simulators}) we use MPM~\cite{mpm} using \texttt{taichi-mpm}~\cite{TaichiMPM}, XPBD~\cite{xpbd} using \texttt{PositionBasedDynamics}~\cite{PositionBasedDynamics}, and FEM using Warp~\cite{warp2022}.

For our large-scale splat simulations or splat + mesh simulations, we use Simplicits~\cite{10.1145/3658184} using the sparse simplicits implementation using Kaolin~\cite{10.1145/3658184,KaolinLibrary}. For rendering our large-scale splat simulations or splat + mesh simulations, we use Polyscope~\cite{polyscope} and composite splat renders from \texttt{gsplat}~\cite{gsplat}. For these simulations, we apply material property tolerances to reduce numerical noise: voxels with Young's modulus differing by less than $10^1$ Pa, Poisson's ratio by less than $10^{-3}$, or density by less than $10^1$ kg/m³ are assigned identical values for the respective property. We present additional details for deforming and rendering deformed Gaussian Splats in~\Cref{app:deformingsplats}.

\subsection{Baselines}\label{app:impl:baselines}

\paragraph{Converting Hardness to Young's Modulus.} NeRF2Physics~\cite{zhai2024physicalpropertyunderstandinglanguageembedded} does not estimate a numerical value of Young's Modulus, but instead predicts Shore A-Shore D hardness. Thus, to compare our method with NeRF2Physics~\cite{zhai2024physicalpropertyunderstandinglanguageembedded} we convert these Shore hardness values to average Young's Modulus values.

\begin{itemize}[leftmargin=10pt,noitemsep,topsep=0pt,parsep=6pt,partopsep=0pt,label={}]
\item {\bf Shore A.} For Shore A hardness, we follow~\cite{astm2240} and use:
\begin{equation}
E_{\text{MPa}} = e^{(S_A \times 0.0235) - 0.6403}
\end{equation}
where $S_A$ is the Shore A hardness value and $E_{\text{MPa}}$ is Young's modulus in megapascals.
% \item {\bf Shore B.} For Shore B hardness we follow~\cite{astm2240} and use:
% \begin{equation}
% E_{\text{MPa}} = e^{((S_B \times 0.85 + 15) \times 0.0235) - 0.6403}
% \end{equation}
% where $S_B$ is the Shore B hardness value and $E_{\text{MPa}}$ is Young's modulus in megapascals.
% \item {\bf Shore C.} For Shore C hardness we follow~\cite{astm2240} and use:
% \begin{equation}
% E_{\text{MPa}} = e^{((S_C \times 1.2 + 20 + 50) \times 0.0235) - 0.6403}
% \end{equation}
% where $S_C$ is the Shore C hardness value and $E_{\text{MPa}}$ is Young's modulus in megapascals.
\item {\bf Shore D.} For Shore D hardness, we follow~\cite{astm2240} and use:
\begin{equation}
E_{\text{MPa}} = e^{((S_D + 50) \times 0.0235) - 0.6403}
\end{equation}
where $S_D$ is the Shore D hardness value and $E_{\text{MPa}}$ is Young's modulus in megapascals.
\end{itemize}

\paragraph{Point or Voxel Sampling.} The baselines NeRF2Physics~\cite{zhai2024physicalpropertyunderstandinglanguageembedded} and PUGS~\cite{shuai2025pugszeroshotphysicalunderstanding} in their methods sample points from the NeRF or Gaussian splat, respectively, and predict mechanical properties at those points. To ensure fair comparisons in~\Cref{tab:material-properties,tab:material-properties-object}, we explicitly make these methods work on the same set of points in the object on which our method is evaluated.

\paragraph{Implementation details of Baselines.} The baseline NeRF2Physics~\cite{zhai2024physicalpropertyunderstandinglanguageembedded} uses \texttt{gpt-3.5-turbo} for certain parts of their pipeline. We replace \texttt{gpt-3.5-turbo} in their pipeline with a better performing model, GPT-4o~\cite{openai2024gpt4technicalreport}. The baseline Phys4DGen~\cite{liu2024physgen} does not have code available. Thus, we faithfully reproduce the parts, "Material Grouping and Internal Discovery" and "MLLMs-Guided Material Identification". We reproduce these parts of their pipeline using GPT-4o~\cite{openai2024gpt4technicalreport} for the MLLMs-Guided Material Identification. Furthermore, we obtained the prompts from the authors of Phys4DGen~\cite{liu2024physgen} and use the same prompts.

\section{Additional Details on the Simulations}\label{app:sec:simulators}

We experiment with Simplicits~\cite{10.1145/3658184}, a reduced-order simulator (\Cref{fig:teaser,fig:simulations,fig:grapes,teaser_old}) and an accurate finite-element method (FEM) simulator (\Cref{fig:teaser,fig:simulations,fig:app:simulations2}) with our material properties. We also use a FEM simulator for our experiments on interpreting errors in properties (\Cref{app:interpret}). We use a material point method (MPM)~\cite{mpm}, and an Extended Position Based Dynamics (XPBD)~\cite{xpbd} simulator for our experiments to compare between simulators (\Cref{fig:simulators}). We share details on these simulations. We also share details on the interpolation we use across all our simulations. We share the hyperparameters used for all the FEM simulations in~\Cref{fem:hyperpram}.

\begin{table}[tb]
\centering
\caption{Hyperparameters for FEM simulation.}
\label{fem:hyperpram}
\begin{tabular}{llll}
\toprule
\rowcolor{nvidiagreen!15}Hyperparameter & Value & Hyperparameter & Value \\
\midrule Time Integrator & Backward Euler & Linear Solver & pre-conditioned CG\\
Nonlinear Solver & Newton's w/ line search & \quad Linear tolerance & $10^{-3}$\\
\quad Newton max iters. & 1024 & Line search & \\
\quad Velocity tol.& 0.05 $ms^{-1}$ & \quad max iters & 8 \\
\quad CCD tol. & 1.0 & Collision &\\
\quad Transform rate tol. & 0.1/s & \quad Friction & 0.5\\
$dt$ & 0.02 & \quad Contact Resistance & 1.0\\
Gravity & $[0.0, -9.8, 0.0]$ & \quad $\hat{d}$ & 0.01\\
\bottomrule
\end{tabular}
\end{table}

\subsection{Interpolation Scheme}
\label{app:interpolation}

Our simulations receive a material field sampled on a voxel grid predicted by \acronym, i.e., values \(m(\mathbf{X}_i)\) given at lattice points \(\{\mathbf{X}_i\}\subset\Omega\). When the simulator needs material values at arbitrary query locations \(\mathbf{X}\) (e.g., element centroids or vertices), we evaluate a nearest-neighbour interpolation of the voxel field:
\begin{equation}
\begin{split}
  i^*(\mathbf{X}) &= \arg\min_i \lVert \mathbf{X} - \mathbf{X}_i \rVert_2, \\
  m^*(\mathbf{X}) &= m\big(\mathbf{X}_{\,i^*(\mathbf{X})}\big).
\end{split}
\end{equation}
We intentionally avoid higher-order interpolation of material fields since real objects are piecewise-constant across label regions, and convex blending across parts of the objects invents intermediate materials. These intermediate materials might not be physically present or admissible, while our outputs fall into a valid material due to the MatVAE (\Cref{sec:matvae}). Nearest-neighbour preserves sharp interfaces and is usually robust for arbitrary query locations.

\subsection{Preparing Scenes and Assigning Materials for the FEM Solver}

Mechanical properties are set either uniformly (like in~\Cref{app:interpret}) or heterogeneously from a voxel field. For uniform assignment, given \(E\) and \(\nu\) we compute Lam\'e parameters
\begin{equation}
  \lambda \,=\, \frac{E\,\nu}{(1+\nu)(1-2\nu)}\,,\qquad \mu \,=\, \frac{E}{2(1+\nu)}\,,
\end{equation}
which are used elementwise together with a constant mass density \(\rho\).

For heterogeneous assignment, a voxel lattice provides \(E(\mathbf{X})\), \(\nu(\mathbf{X})\), and \(\rho(\mathbf{X})\) at voxel centers. After applying the same rigid/scale transform as the mesh, each tetrahedron takes \(\lambda,\mu\) from the nearest voxel to its centroid, and each vertex takes \(\rho\) from the nearest voxel to its position. This produces per-tetrahedron \(\lambda,\mu\) and per-vertex \(\rho\) fields that are directly used in the elastic strain energy density per unit reference volume (\(W\)), first variation of the incremental potential (\(R\)), and Newton-Jacobian (\(\mathcal{K}\)).

During simulation, a visual mesh is embedded into the physics mesh by assigning each visual vertex \(\mathbf{x}_v\) to a containing (or nearest) tetrahedron with vertices \(\{\mathbf{X}_a\}_{a=1}^4\) and barycentric weights \(\{w_a\}_{a=1}^4\) satisfying \(\sum_a w_a = 1\) and \(\sum_a w_a\,\mathbf{X}_a = \mathbf{x}_v\); its deformed position is then the barycentric interpolation of current nodal positions \(\{\mathbf{x}_a\}_{a=1}^4\):
\begin{equation}
  \mathbf{x}_v^{\,\mathrm{def}} \,=\, \sum_{a=1}^4 w_a\,\mathbf{x}_a\,.
\end{equation}

The state update in our simulation experiments is computed time-step by time-step, and we also deform and move the visual mesh according to the physics mesh at each time step.

\subsection{Details of the FEM Solver}

For FEM simulations, we use a simulator based on the \texttt{libuipc}~\cite{10.1145/3735126,gipc2024} implementation and the Warp (\texttt{warp.fem})~\cite{warp2022} implementation. We first explain the details for our simulations in \S\ref{app:interpret}.

We consider a deformable continuum body with reference configuration \(\Omega \subset \mathbb{R}^3\) and boundary \(\partial\Omega = \Gamma_D \cup \Gamma_N\), where \(\Gamma_D\) denotes boundary points with Dirichlet boundary conditions, and \(\Gamma_N\) denotes boundary points with Neumann boundary conditions. The unknown to solve for is the displacement field \(\mathbf{u}: \Omega \to \mathbb{R}^3\). Time is discretized into frames with a fixed step \(\Delta t\). At each frame we compute an increment \(\Delta\mathbf{u}\) that advances the configuration \(\mathbf{u} \leftarrow \mathbf{u} + \Delta\mathbf{u}\) while enforcing Dirichlet constraints on \(\Gamma_D\). The deformation map is \(\varphi(\mathbf{X}) = \mathbf{X} + \mathbf{u}(\mathbf{X})\), with deformation gradient \(\mathbf{F}(\mathbf{u}) = \mathbf{I} + \nabla \mathbf{u}\), Jacobian \(J = \det \mathbf{F}\), and isochoric invariant \(I_c = \mathrm{tr}(\mathbf{F}^\top \mathbf{F})\). For corotational modeling, we use the stretch tensor \(\mathbf S\) from the polar/SVD decomposition: if \(\mathbf{F} = \mathbf{U}\,\mathrm{diag}(\boldsymbol{\sigma})\,\mathbf{V}^\top\) then \(\mathbf{S} = \mathbf{V}\,\mathrm{diag}(\boldsymbol{\sigma})\,\mathbf{V}^\top\). Given Young's modulus \(E\) and Poisson ratio \(\nu\), the Lam\'e parameters are \(\lambda = E\nu /((1+\nu)(1-2\nu))\) and \(\mu = E/(2(1+\nu))\). Here \(\nabla\) denotes the gradient with respect to reference coordinates, \(\mathbf{A}{:}\mathbf{B} = \mathrm{tr}(\mathbf{A}^\top\!\mathbf{B})\) is the Frobenius inner product, and \(\lVert\cdot\rVert\) is the Euclidean norm.

The elastic response we use is the corotational Hookean model. Define the small strain \(\boldsymbol\varepsilon = \mathbf{S} - \mathbf{I}\). The strain energy density and Kirchhoff stress are
\begin{equation}
\begin{split}
  W_{\mathrm{CR}}(\mathbf{S}) &= \underbrace{\mu\,\boldsymbol\varepsilon\!:\!\boldsymbol\varepsilon}_{\text{shear (deviatoric)}} + \underbrace{\tfrac{\lambda}{2}\,\mathrm{tr}(\boldsymbol\varepsilon)^2}_{\text{volumetric}}, \\
  \boldsymbol{\tau}(\mathbf{S}) &= \underbrace{2\mu\,\boldsymbol\varepsilon}_{\text{shear}} + \underbrace{\lambda\,\mathrm{tr}(\boldsymbol\varepsilon)\,\mathbf{I}}_{\text{volumetric}},
\end{split}
\end{equation}
with a consistent linearization obtained via the variation of \(\mathbf{S}\) with respect to \(\mathbf{F}\) and projected to maintain symmetry and positive semidefiniteness.

Each frame solves an incremental variational problem. Given the previous increment \(\Delta\mathbf{u}^{\,n-1}\), we seek \(\Delta\mathbf{u}^{\,n}\) that approximately minimizes the incremental potential
\begin{equation}
\begin{split}
  \Pi(\Delta\mathbf{u}) &= \underbrace{\int_{\Omega} \rho\left(\tfrac{1}{2}\,\frac{\lVert\Delta\mathbf{u} - \Delta\mathbf{u}^{\,n-1}\rVert^2}{\Delta t^2}\right)\,\mathrm{d}V}_{\text{inertial regularization}} \\
  &\quad + \underbrace{\int_{\Omega} \big( - \rho\,\mathbf{g}\cdot\Delta\mathbf{u} - \mathbf{f}_{\!\mathrm{ext}}\cdot\Delta\mathbf{u} \big)\,\mathrm{d}V}_{\text{body and external work}} \\
  &\quad + \underbrace{\int_{\Omega} W_{\mathrm{CR}}\big(\mathbf{S}(\mathbf{u}^{n-1} + \Delta\mathbf{u})\big)\,\mathrm{d}V}_{\text{elastic energy}} \\
  &\quad + \underbrace{\Pi_{\mathrm{int}}(\Delta\mathbf{u})}_{\text{interior/boundary regularization}},
\end{split}
\end{equation}
where \(\rho\) is the mass density, \(\mathbf{g}\) is the gravitational acceleration vector, \(\mathbf{f}_{\!\mathrm{ext}}\) denotes prescribed volumetric loads, and \(\Pi_{\mathrm{int}}\) denotes any interior/boundary regularization term (e.g., a jump penalty in discontinuous settings). The admissible test function \(\mathbf{v}\) is any sufficiently smooth virtual displacement that vanishes on \(\Gamma_D\). The first variation \(\delta\Pi(\Delta\mathbf{u};\mathbf{v})=0\) for all such \(\mathbf{v}\) yields the residual functional
\begin{equation}
\begin{split}
  R(\Delta\mathbf{u})[\mathbf{v}] &= \underbrace{\int_{\Omega} \rho\,\frac{\Delta\mathbf{u} - \Delta\mathbf{u}^{\,n-1}}{\Delta t^2}\cdot\mathbf{v}\,\mathrm{d}V}_{\text{inertia}} \\
  &\quad + \underbrace{\int_{\Omega} \big( - \rho\,\mathbf{g}\cdot\mathbf{v} - \mathbf{f}_{\!\mathrm{ext}}\cdot\mathbf{v} \big)\,\mathrm{d}V}_{\text{body/external}} \\
  &\quad - \underbrace{\int_{\Omega} \boldsymbol{\tau}\big(\mathbf{S}(\mathbf{u}^{n-1}+\Delta\mathbf{u})\big) : \nabla\mathbf{v}\,\mathrm{d}V}_{\text{elastic (internal) virtual work}} \\
  &\quad + \underbrace{R_{\mathrm{int}}(\Delta\mathbf{u})[\mathbf{v}]}_{\text{regularization}},
\end{split}
\end{equation}
which is set to zero for all \(\mathbf{v}\). Newton's method is applied to \(R(\Delta\mathbf{u})=0\). At iterate \(\Delta\mathbf{u}^{\,(k)}\) we assemble the consistent tangent operator \(\mathcal{K} = \mathrm{D}R\big[\Delta\mathbf{u}^{\,(k)}\big]\) (the G\`ateaux derivative of \(R\)) and solve the linear system
\begin{equation}
\begin{split}
  \mathcal{K}\,\delta\mathbf{u} &= -R\big(\Delta\mathbf{u}^{\,(k)}\big), \\
  \Delta\mathbf{u}^{\,(k+1)} &= \Delta\mathbf{u}^{\,(k)} + \alpha\,\delta\mathbf{u}
\end{split}
\end{equation}
where \(\alpha\in(0,1]\) is chosen by a backtracking Armijo rule to guarantee sufficient decrease of \(\Pi\). The operator \(\mathcal{K}\) contains an inertial mass-like term \(\int_{\Omega}\rho\,\Delta t^{-2}\,\delta\mathbf{u}\cdot\mathbf{v}\,\mathrm{d}V\), the consistent elastic tangent from the linearization of \(\boldsymbol{\tau}(\mathbf{S}(\cdot))\), and any interior/boundary penalty contributions. This procedure is repeated until the update norm or residual falls below a prescribed tolerance.

For all our other simulations (\textit{i.e.} except the simulations in \S\ref{app:interpret}) we use a closely related variant whose differences are in the constitutive law, material assignment, mesh preparation/interpolation, and contact handling. First, the stored energy and stress are taken to be compressible Neo-Hookean with volumetric regularization. Writing \(\mathbf{C}=\mathbf{F}^\top\!\mathbf{F}\) and \(\mathbf{B}=\mathbf{F}\,\mathbf{F}^\top\), the energy and Kirchhoff stress are
\begin{equation}
\begin{split}
      W_{\mathrm{NH}}(\mathbf{F}) \,&=\, \tfrac{\mu}{2}\big(\mathrm{tr}\,\mathbf{C} - 3 - 2\,\ln J\big) + \tfrac{\lambda}{2}\,(\ln J)^2,\\ \boldsymbol{\tau}_{\mathrm{NH}}(\mathbf{F}) \,&=\, \mu\,(\mathbf{B} - \mathbf{I}) + \lambda\,\ln J\,\mathbf{I}.
\end{split}
\end{equation}
This change only affects the elastic terms in \(\Pi\), \(R\), and \(\mathcal{K}\); the kinematics and inertial terms remain the same. For the simulation experiments, we also use IPC~\cite{10.1145/3386569.3392425} for collision handling.

\subsection{Preparing Scenes and Assigning Materials for the Simplicits Solver}

Each object is specified by a set of quadrature points \(\mathcal{Q}=\{\mathbf{X}_q\}\) that sample its volume (used for elasticity and inertia), a set of collision particles \(\mathcal{C}=\{\mathbf{X}_c\}\) for contact, and a set of visual vertices for rendering. We position objects with a rigid transform (origin and rotation) and an object scale; these transforms are applied consistently when evaluating kinematics, gravity, and material fields.

We embed all objects into a regular grid domain and attach to this grid a low-dimensional Simplicits subspace. The displacement basis is the product of a trilinear grid shape and a per-object handle shape, with multiple duplicated handles per grid vertex. At each quadrature point, we evaluate and cache per-node subspace weights and their spatial gradients. These weights modulate the duplicated handle functions during assembly, instantiating the Simplicits subspace on the grid.

Material parameters are assigned per quadrature point from a voxel lattice providing \(E(\mathbf{X})\), \(\nu(\mathbf{X})\), and \(\rho(\mathbf{X})\). After applying the same rigid/scale transform as the object, each quadrature location \(\mathbf{X}_q\) takes its material from the nearest voxel to \(\mathbf{X}_q\). We compute Lam\'e parameters per point as
\begin{equation}
\begin{split}
    \lambda_q \,&=\, \frac{E(\mathbf{X}_q)\,\nu(\mathbf{X}_q)}{(1+\nu(\mathbf{X}_q))(1-2\nu(\mathbf{X}_q))}\,,\\\qquad
    \mu_q \,&=\, \frac{E(\mathbf{X}_q)}{2(1+\nu(\mathbf{X}_q))}\,,
\end{split}
\end{equation}
and use \(\rho_q = \rho(\mathbf{X}_q)\) in the inertial terms.

The quadrature weights are set uniformly as
\begin{equation}
  w_q \,=\, \frac{v}{\lvert\mathcal{Q}\rvert},
\end{equation}
where \(v\) is the object volume estimate and \(\lvert\mathcal{Q}\rvert\) is the number of quadrature points. This makes elastic and inertial energies invariant to the sampling density.

Collision particles \(\mathcal{C}\) are used for detecting and resolving contact against other particles and registered kinematic triangle meshes (containers and obstacles). We use an IPC-style barrier with Coulomb friction, and scale the contact stiffness by object volume and the number of collision particles to obtain comparable penalties across scenes.

\subsection{Deforming Splats and Rendering Deformed Splats}
\label{app:deformingsplats}

We render each object as a set of anisotropic Gaussian splats. At rest, a splat is parameterized by its mean \(\boldsymbol{\mu}_0\in\mathbb{R}^3\), a unit quaternion \(\mathbf{q}_0\) (with rotation \(\mathbf{R}_0\in\mathrm{SO}(3)\)), and axis scales \(\mathbf{s}_0\in\mathbb{R}_{>0}^3\). We define the rest-frame shape operator
\begin{equation}
  \underbrace{\mathbf{L}}_{\text{rest anisotropy}} \,=\, \mathbf{R}_0\,\mathrm{diag}(\mathbf{s}_0)\,.
\end{equation}

During simulation, the displacement field yields a world-space deformation gradient \(\underbrace{\mathbf{F}}_{\text{local deformation}}\) at the splat center and a world-space position \(\boldsymbol{\mu}\) (obtained by evaluating the embedded deformation at the visual vertex). We map the rest anisotropy through the local deformation to obtain the world-space covariance of the splat as
\begin{equation}
  \underbrace{\Sigma}_{\text{world covariance}} \,=\, \underbrace{(\mathbf{F}\,\mathbf{L})}_{\text{deformed axes}}\;\overbrace{(\mathbf{F}\,\mathbf{L})^\top}^{\text{deformed axes}^\top}\; +\; \underbrace{\varepsilon\,\mathbf{I}}_{\text{SPD padding}}\,.
\end{equation}
Here \(\varepsilon>0\) is a small scalar that guarantees positive-definiteness under extreme compression.

For rasterization we pass \(\boldsymbol{\mu}\) and \(\Sigma\) to the Gaussian renderer. The renderer (\texttt{gsplat}~\cite{gsplat}) expects a symmetric 6-vector parameterization; we therefore pack the lower-triangular entries as
\begin{equation}
  \underbrace{\mathbf{c}}_{\text{packed covariance}} \,=\, \big[\,\Sigma_{11},\; \Sigma_{12},\; \Sigma_{13},\; \Sigma_{22},\; \Sigma_{23},\; \Sigma_{33}\,\big]^\top\,.
\end{equation}
Color appearance (spherical-harmonic coefficients) and opacity are carried from the rest representation; only the mean \(\boldsymbol{\mu}\) and covariance \(\Sigma\) change over time. We also support a scalar scale multiplier applied to \(\mathbf{s}_0\) for interactive control in qualitative visualizations.

Given the view (extrinsic) matrix \(\mathbf{V}\) and vertical field-of-view, we synthesize camera intrinsics
\begin{equation}
\begin{split}
  \mathbf{K} \,&=\, \begin{bmatrix} f_x & 0 & c_x \\ 0 & f_y & c_y \\ 0 & 0 & 1 \end{bmatrix},\\ f_x \,&=\, \tfrac{W}{2\tan(\tfrac{\mathrm{fov}_x}{2})},\; f_y \,=\, \tfrac{H}{2\tan(\tfrac{\mathrm{fov}_y}{2})},\;\\ c_x \,&=\, \tfrac{W}{2},\; c_y \,=\, \tfrac{H}{2},    
\end{split}
\end{equation}
with image size \((W,H)\). We then render the set of splats \(\{\boldsymbol{\mu},\,\mathbf{c}\}\) under \((\mathbf{V},\mathbf{K})\) to produce RGB (and depth) frames. At every frame, we interpolate \(\boldsymbol{\mu}\) and \(\mathbf{F}\) at the visual vertices, form \(\Sigma = \mathbf{F}\,\mathbf{L}\,\mathbf{L}^\top\mathbf{F}^\top + \varepsilon\,\mathbf{I}\), pack it as \(\mathbf{c}\), and feed the Gaussian rasterizer together with the stored colors and opacities.

\subsection{Details of the Simplicits Solver}

We use the Simplicits~\cite{10.1145/3658184} simulator based on the implementation in Kaolin~\cite{KaolinLibrary}. Simplicits solves for a displacement field represented in a low-dimensional subspace attached to a regular grid. This subspace is a product basis between trilinear grid polynomials and duplicated per-vertex ``handle'' functions whose influence is modulated by per-point weights and weight gradients evaluated at quadrature points. We assemble inertia and compressible Neo-Hookean elasticity on this subspace, using the per-quadrature Lam\'e parameters \((\lambda_q,\mu_q)\) and measures \(w_q\). Each frame performs Newton steps on the incremental potential with backtracking line search. Linear systems are solved by preconditioned conjugate gradients.

\textit{In the Kaolin implementation}, splat–splat contact uses particle pairs: for a pair \((a,b)\) with current positions \(\mathbf{x}_a,\mathbf{x}_b\) and contact radius \(r\), we set \(\mathbf{n}_c = (\mathbf{x}_a-\mathbf{x}_b)/\lVert\mathbf{x}_a-\mathbf{x}_b\rVert\), \(r_c=r\), and use the relative offset \(\mathbf{o}_c(\Delta\mathbf{u}) = \Delta\mathbf{u}_a - \Delta\mathbf{u}_b\), so that \(d_c = \mathbf{n}_c\!\cdot(\mathbf{x}_a-\mathbf{x}_b)\) and \(\mathbf{v}_{t,c}\) measures tangential slip between the two.

% this is because i used the january implementation of conplciits where if you remmber, for the demos i did i implemented actual point-mesh contact instead of point-point contact
\textit{However, in our implementation}, splat–mesh contact is handled differently than splat-splat contact. For splat-mesh contact, instead of using the collision points, we use triangle meshes as kinematic colliders: for a particle at \(\mathbf{x}\) and its closest point \(\mathbf{p}\) on a nearby triangle (with interpolated mesh normal/velocity \(\mathbf{n}_c,\mathbf{v}_m\)), we take \(r_c=2r\) and
\begin{equation}
  \mathbf{o}_c(\Delta\mathbf{u}) \,=\, \Delta\mathbf{u} + \big(\lVert\mathbf{x}-\mathbf{p}\rVert - \mathbf{n}_c\!\cdot\Delta\mathbf{u}\big)\,\mathbf{n}_c - \mathbf{v}_m\,,\qquad d_c \,=\, \lVert\mathbf{x}-\mathbf{p}\rVert - \mathbf{n}_c\!\cdot\mathbf{v}_m\,.
\end{equation}
Only the simulated-object DOFs enter \(\mathbf{o}_c\); mesh or splat motion appears through \(\mathbf{v}_m\).

These terms are used in the Newton system as
\begin{equation}
  \mathbf{K} \;\leftarrow\; \mathbf{K} \;+\; \underbrace{\alpha\,\mathbf{H}^\top \mathbf{C}\,\mathbf{H}}_{\text{contact stiffness}}\,,\qquad
  \mathbf{r} \;\leftarrow\; \mathbf{r} \;-\; \underbrace{\alpha\,\mathbf{H}^\top\mathbf{g}}_{\text{contact force}}\,,
\end{equation}
where \(\mathbf{H}\) is the Jacobian of contact offsets, and \(\mathbf{g},\mathbf{C}\) are the per-contact gradient/Hessian with respect to \(\mathbf{o}_c\). For splat–mesh contacts only the simulated-object block of \(\mathbf{H}\) is present; for splat–splat contacts the two object blocks appear with opposite signs.

\subsection{Preparing Scenes and Assigning Materials for the XPBD Solver}
\label{app:xpbd}

We also use an Extended Position-Based Dynamics (XPBD) solver~\cite{xpbd} based on \texttt{PositionBasedDynamics}~\cite{PositionBasedDynamics}.

\paragraph{Particles and initialization.} Objects are represented by particles positioned at rest locations \(\{\mathbf{X}_i\}\). For each particle we initialize position \(\mathbf{x}_i\!\leftarrow\!\mathbf{X}_i\) and a previous position equal to \(\mathbf{X}_i\) (Verlet), set mass \(m_i\) and inverse mass \(w_i=1/m_i\) (pinned points use \(w_i=0\)). A soft sphere uses one center particle and a set of surface particles sampled on a UV sphere; the center forms simple tetrahedra with nearby surface points for volume preservation.

\paragraph{Material parameters and compliance.} To target an elastic behavior with Young's modulus \(E\) and Poisson ratio \(\nu\) (bulk modulus \(K=E/(3(1-2\nu))\)), we choose distance- and volume-constraint compliances \(\alpha_{\mathrm{dist}},\alpha_{\mathrm{vol}}\) inversely proportional to \(E\) and \(K\). XPBD uses the scaled compliance \(\alpha_\Delta=\alpha/\Delta t^2\) so that smaller \(\alpha\) yields stiffer response.

\paragraph{Prediction, projections, and updates.} Each frame predicts positions with Verlet integration, then iteratively projects distance, volume, and collision constraints. For a pairwise distance constraint, the XPBD update displaces endpoints along the edge with a factor
\begin{equation}
\gamma\;=\;\frac{\alpha_\Delta}{\alpha_\Delta + w_i + w_j},\quad \Delta\mathbf{x}_i\propto -\gamma\,C_{ij}\,\frac{\mathbf{x}_i-\mathbf{x}_j}{\lVert\mathbf{x}_i-\mathbf{x}_j\rVert},\quad \Delta\mathbf{x}_j=-\frac{w_i}{w_j}\,\Delta\mathbf{x}_i,
\end{equation}
with the analogous formulation for volume constraints using their gradients. After projections, we update positions and apply ground-plane contact with Coulomb friction.

\paragraph{Visualization.} We track the evolving surface by updating a triangular mesh whose vertices coincide with the surface particles.

\subsection{Preparing Scenes and Assigning Materials for the MPM Solver}
\label{app:mpm}

We use a material point method (MPM) simulator~\cite{mpm} based on \texttt{taichi-mpm}~\cite{TaichiMPM}. Scenes are specified by a uniform Cartesian grid, a set of particles sampling each object's volume, and per-object mechanical properties.

\paragraph{Domain, grid, and timestep.} We embed all objects in a unit cube domain discretized by an \(n_\mathrm{grid}\times n_\mathrm{grid}\times n_\mathrm{grid}\) grid (cell size \(\Delta x=1/n_\mathrm{grid}\)). We use a fixed time step \(\Delta t\) small enough for stability (e.g., \(\Delta t=5\times 10^{-5}\) in our experiments).

\paragraph{Particles and initialization.} Each object is sampled with material points (particles) positioned at rest locations \(\{\mathbf{X}_p\}\). For each particle we initialize position \(\mathbf{x}_p\!\leftarrow\!\mathbf{X}_p\), velocity \(\mathbf{v}_p\!\leftarrow\!\mathbf{0}\), affine velocity field \(\mathbf{C}_p\!\leftarrow\!\mathbf{0}\), and deformation gradient \(\mathbf{F}_p\!\leftarrow\!\mathbf{I}\). Mass is set as \(m_p=\rho\,V_p\) with density \(\rho\) and particle volume \(V_p\) consistent with the grid resolution. In the simple drop test, we sample a sphere at a height and let gravity act.

\paragraph{Material parameters and constitutive law.} We assign per-object Young's modulus \(E\) and Poisson ratio \(\nu\), compute Lam\'e parameters \(\mu=E/(2(1+\nu))\) and \(\lambda=E\nu/((1+\nu)(1-2\nu))\), and use a fast corotated (FCR) elastic model. With polar/SVD decomposition \(\mathbf{F}=\mathbf{U}\,\mathrm{diag}(\boldsymbol\sigma)\,\mathbf{V}^\top\) and rotation \(\mathbf{R}=\mathbf{U}\mathbf{V}^\top\), the Kirchhoff stress is
\begin{equation}
\boldsymbol{\tau}_\mathrm{FCR}(\mathbf{F})\;=\;2\mu\,(\mathbf{F}-\mathbf{R})\,\mathbf{F}^\top\; +\; \lambda\,J\,(J-1)\,\mathbf{I},\qquad J=\det\mathbf{F}.
\end{equation}

\paragraph{Transfers and updates.} Each step performs Particle-to-Grid (P2G) transfers using quadratic B-spline weights: we scatter particle mass, momentum, and internal forces \(-V_p\,\boldsymbol{\tau}\,\nabla w\) to grid nodes. On the grid, we (i) convert momentum to velocity, (ii) add gravity, and (iii) enforce box boundary conditions by clamping outward normal velocities to zero. We then perform Grid-to-Particle (G2P) to interpolate grid velocities back to particles, update the affine field, and integrate
\begin{equation}
\mathbf{x}_p \leftarrow \mathbf{x}_p + \Delta t\,\mathbf{v}_p,\qquad
\mathbf{F}_p \leftarrow (\mathbf{I}+\Delta t\,\nabla\mathbf{v})\,\mathbf{F}_p.
\end{equation}

\paragraph{Visualization.} We embed a coarse surface at rest, align it to the particle center of mass, and update its vertices via interpolation of nearby particle displacements.
\section{Other Related Works}

% \subsection{Inferring Mechanical Properties from Dynamics and Interaction}
% motion, deformation, or vibration~\cite{davis2015visual} involves learning and simulator-based inversion
For completeness, we include other tangentially related works here.
A different setting from ours is inferring physical properties given additional observations, such as video~\cite{davis2015visual, Mottaghi_2016_CVPR, 10.1007/3-540-47969-4_37, chen2025vid2sim, liu2024physics3d, xue20233dintphysgeneralized3dgroundedvisual, li2025freegave, 5459407, yildirim2interpreting, li2023pacnerfphysicsaugmentedcontinuum, pmlr-v119-li20j, BMVC2016_39, NIPS2015_d09bf415, NIPS2017_4c56ff4c, Xia_2024_CVPR, xu2019densephysnetlearningdensephysical, feng2024pienerfphysicsbasedinteractiveelastodynamics, lin2024phy124fastphysicsdriven4d}
% ~\cite{davis2015visual, Mottaghi_2016_CVPR, 10.1007/3-540-47969-4_37, chen2025vid2sim, liu2024physics3d, xue20233dintphysgeneralized3dgroundedvisual, li2025freegave, 5459407, yildirim2interpreting, li2023pacnerfphysicsaugmentedcontinuum, pmlr-v119-li20j, BMVC2016_39, NIPS2015_d09bf415, NIPS2017_4c56ff4c, Xia_2024_CVPR, xu2019densephysnetlearningdensephysical, feng2024pienerfphysicsbasedinteractiveelastodynamics, lin2024phy124fastphysicsdriven4d}.
%Robotic probing, pushing, grasping, and scanning can yield highly accurate, grounded estimates of stiffness, mass, and friction
or physical manipulation of real objects~\cite{yu2024octopiobjectpropertyreasoning, 10.1145/383259.383268, lang2003scanning, lloyd2001robotic, pai2008acme, pai2000robotics, 10160731, pinto2016curiousrobotlearningvisual}.
Other related works focus on generating new physically plausible shapes, e.g.\ stable under gravity or other interactions,
but cannot augment existing 3D assets with mechanical properties, which is our goal~\cite{lin2025omniphysgs, NEURIPS2024_d7af02c8, chen2024atlas3dphysicallyconstrainedselfsupporting, ni2024phyreconphysicallyplausibleneural, Yang_2024_CVPR, Mezghanni_2022_CVPR, Chen_2025_CVPR, cao2025physx-3d, cao2025sophylearninggeneratesimulationready}. 
%However, these works do not augment existing assets with mechanical properties, which is our goal in this paper.
% \todo{Rishit - any relevant works in this category should go to SOPHY section; can you find a common thing about other methods?}
Other methods predict displacements~\cite{zhang2024adaptigraph, li2023robocook}, bypassing mechanical properties, or focus on other aspects such as articulation~\cite{Xia_2025_CVPR}.

\end{document}